\RequirePackage{fix-cm}
\documentclass{svjour3}                     
\smartqed  
\usepackage{graphicx}
\usepackage{morefloats}
\usepackage{subfig}

\begin{document}

\title{A Semi-automated Statistical Algorithm for Object Separation
}


\author{Madhur Srivastava \and Satish K. Singh \and Prasanta K. Panigrahi 
}


\institute{
}

\date{}

\maketitle

\begin{abstract}
We explicate a semi-automated statistical algorithm for object identification and segregation in both gray scale and color images. The algorithm makes optimal use of the observation that definite objects in an image are typically represented by pixel values having narrow Gaussian distributions about characteristic mean values. Furthermore, for visually distinct objects, the corresponding Gaussian distributions have negligible overlap with each other and hence the Mahalanobis distance between these distributions are large. These statistical facts enable one to sub-divide images into multiple thresholds of variable sizes, each segregating similar objects. The procedure incorporates the sensitivity of human eye to the gray pixel values into the variable threshold size, while mapping the Gaussian distributions into localized $\delta-$functions, for object separation. The effectiveness of this recursive statistical algorithm is demonstrated using a wide variety of images.
\end{abstract}

\section{Introduction}
\label{intro}
With the development of image based systems, the need for object extraction from images has increased exponentially in different fields. One of the precursors for object separation is image segmentation, which divides an image into different regions having similar or same features. The utility and effectiveness of segmentation arise from the following two statistical observations about images. Firstly, the objects in an image are generally characterized by pixel values having Gaussian distributions about their characteristic mean values. Secondly, these distributions have narrow overlap with each other for visually distinct objects. These two statistical features can be utilized to segregate different distributions from each other which can lead to effective image separation. Evidently, image segmentation is a complex process, involving analysis of color, shape, texture and/or motion of different objects in an image \cite{1}. It is a key prerequisite for image analysis, ranging from feature extraction to object recognition. Hence, the efficacy of object separation and recognition algorithms crucially depend upon the quality of image segmentation. Over segmentation results in separating similar regions into two or more, while under segmentation merges separate regions into one, resulting in failure of image analysis \cite{2}. Most of the work on segmentation is focused on monochrome (gray scale) images$-$segmenting an image into regions based on intensity levels \cite{3}. 

As is well known, segmentation techniques can be broadly classified into five classes based on pixel values, edge information, regional homogenity, physical principles and hybrid methods \cite{4}. Pixel based approaches operate on intensity values of individual pixels and their correlation with immediate neighbourhood, while edge based ones through different approaches search for discontinuities in local regions to form distinct categories. Region based techniques take into consideration the homogeneity of regions for image segmentation. Physics based methods eliminate the effect of shadows, shadings and highlights, focusing on the boundaries of objects. Hybrid techniques incorporate the advantages of different methods for optimal effect.

Image extraction methods are of two types: supervised and unsupervised. In each of this category, one can have two further subdivisions: one based on a single copy of the image and the other having access to multiple images of the same object. In all these approaches, one can search for single or multiple objects at a given time. Borenstein and Ullman determined the object and background by overlapping the automatically extracted fragments \cite{5}. This approach is vulnerable to wrong assumption of background objects as foreground, as individual fragments are considered independent of each other. Winn and Jojic assumed that within a class, object shape and color distribution remain consistent \cite{6}. Hence, to get consistent segmentation, all images were combined. Huang et al. combined Markov Random Field (MRF) and a deformable model for image segmentation \cite{7}. They used brief propagation in the MRF part, while variational approaches were used to estimate the deformable contour. In their method, Rother et al. \cite{8} only needed two images, to segment the common parts subject to the condition that they have similar features like color, shape or/and texture. Liu et al. \cite{9} used Hybrid Graph Model (HGM) for object segmentation. They considered class specific information and local texture/color similarities by taking both symmetric and asymmetric relationships among the samples. Tu et al. \cite{10} generated the parsing graph and reconfigured it with the help of Markov chain jumps. This approach included both generative and discriminative methods. Leibe et al. \cite{11} used a two stage approach, 'Codebook of Local Appearance and Implicit Shape Model', to find the shapes and appearances of images consistent with each other. Yu and Shi optimized joint grouping criterion by using low$-$level pixel and high$-$level patch grouping processes \cite{12}.

For effective separation of objects, one requires suitable thresholding techniques for segregating the different objects in an image into non$-$overlapping sets. Thresholding is broadly divided into two categories: bilevel and multilevel techniques. In bilevel thresholding, a complete image is divided into two regions with the help of a single threshold; the two regions have mutually exclusive values of 0 and 1. In multilevel thresholding, an image is characterized into multiple  distinct regions with the help of different features and properties of image \cite{13}. More specifically, thresholding methods in general can be categorized into six groups, based on histogram shape, clustering properties, image entropy, object properties, spatial extremes and local attributes \cite{14}. Histogram shape based techniques analyse and threshold images with the help of peaks, valleys and curvatures. In clustering methods, foreground and backgrounds are formed by separating the samples of gray level images into distinct clusters. Entropy methods use variation between the background and the foreground to distinguish the two. Approaches using object attribute search for the similarity between original and thresholded images. Spatial feature based methods use probability distribution and correlation between pixels. Local approaches use local threshold values of each pixel in a given sub-band for categorising different objects. 

Among all the above mentioned techniques, the clustering method is chosen and the thresholding based on the statistics of histogram is proposed in the approach used in this paper. Considering the fact that the constituent objects forming an image have no or minimal overlap of intensity values in the histogram, the paper applies a histogram based thresholding technique to divide the histogram in such a way that the object required to be separated, is in the non-overlapping segment. Moreover, the aim of the paper is to provide flexibility to the user to pick any distinct object, which can be attained by changing the parameters that varies segment sizes. The other methods might result in effective segmentation, but the flexibility to extract a distinct object of choice by varying segmentation parameters at user's end may not be possible. Furthermore, overall optimal segmentation is goal of other techniques, whereas histogram thresholding techniques can be utilized for optimizing segmentation for a particular object of the image.

Amongst all, Otsu's method is the most widely used in bilevel thresholding \cite{15}. Kittler and Illingworth proposed a minimum error thresholding method, showing that mixture distribution of the object and background pixels can characterize an image \cite{16}\cite{17}. Ridler and Calvard used an iterative scheme to get the thresholds by taking the average of the foreground and the background class means \cite{18}. This concept is based on two class Gaussian mixture models. Yanni and Horne followed a similar concept to find the thresholds by taking the midpoint of the two peaks \cite{19}. Reddi et al. extended Otsu's method of maximizing between$-$class variance recursively to obtain multiple thresholds \cite{20}. Liu and Li generalized Otsu's algorithm into two dimensions \cite{21}. Llyod initially assumed equal variance Gaussian density functions for foreground and background and then used an iterative search to minimize the total misclassification error \cite{22}. Jawahar et al. proposed a fuzzy thresholding technique, which is based on iterative optimization, resulting in the minimization of a suitable cost function \cite{23}. Sen and Pal \cite{24} proposed two algorithms, one for bilevel and the other for multilevel thresholding. In bilevel thresholding, they divided the histogram into three regions, in which the bright and the dark regions have predefined seed values, while the undefined region pixels are merged with the other two regions. On the other hand, in multilevel thresholding, tree structured technique is used to extend the concept of bilevel thresholding. Gao et al. employed particle swam optimization technique, instead of population$-$based stochastic optimization methods, to obtain the thresholds \cite{25}. Lopes et al. overcome the limitations for finding fuzziness measure of initial subsets for high contrast images in histogram based technique, by proposing an automated algorithm \cite{26}. Hemchander et al. developed a local binarization method which maintains image continuity rigorously \cite{27}.

In gray identifiable objects, images are typically characterized by pixel values having Gaussian distributions about characteristic intensity values. The visually distinguishable objects usually have a narrow overlap of their respective distributions, implying a large Mahalanobis distance between their respective distributions. These statistical features can be optimally utilized to separate the distributions of multiple objects from each other. In this paper, a clustering based method is used to separate out the overlapping Gaussian distributions of the foreground and the background pixels with the help of weighted mean and standard deviation of the image histogram. We explicate a semi-automated algorithm, which employs a recursive statistical multi-thresholding approach, naturally separating the object based pixel distributions of the foreground into different classes. The algorithm follows variable block size segmentation of the weighted histogram of the image, based on the approach of separating the overlapping Gaussians into completely non$-$overlapping $\delta-$functions.

Taking into account complexity of images, two algorithms are used; the first one does a finer division of pixel values, away from black and white regions, while the second does the same in the black and white regions. After multilevel thresholding, image segmentation is carried out by separating out each domain into independent images. Subsequently, bilevel thresholding is used to isolate desired objects from the background. The proposed approach defines a framework which enables object separation in color images belonging to various fields. It provides three variable parameters to identify and separate the object: number of thresholds, combination of segmented regions, and skewness. This paper, through a variety of images, shows that the two methodologies can successfully separate the desired object from the rest of the image. Hence, it can be applied to the different set of images used in distinct fields.

The following section describes the approach in steps. Section 3 illustrates the methodology used for segmentation in detail. In section 4, experimental results are provided with inferences. We conclude in section 5, with further prospects for the proposed approach and directions for future research.

\section{Approach}
\label{sec:1}

From a physical perspective, images can be classified into two categories. In the first case, the object dominates the background, while in the second case the background is dominant. We adopt two different methodologies, one for each category of images. Methodology-I is preferred when the intensity values of the object to be extracted have sensitive boundary or boundaries with the rest of the image towards the weighted mean of the histogram. On the other hand, methodology-II is adopted when boundary separating object and others is away from the weighted mean of histogram. In nutshell, the application of methodology-I or methodology-II depends on whether the object lies in the complex foreground or the complex background, respectively. 

In a number of cases, for color images, gray level segmentation algorithms are extended to all the three channels, without considering the correlation amongst them. In our approach, thresholding is performed on the luminance (intensity) component of Y (luminance) I (hue) and Q (saturation) color space, after converting the R (red) G (green) B (blue) color space into the YIQ domain. Hence, only the gray level information of the color image is used in the proposed monochrome image segmentation algorithm. The reason for using this approach is that the gray scale component of a color image captures sufficient information to effectively segment the color image with either of the two segmentation methodologies. Moreover, the segmentation methodologies can be applied to I (hue) and Q (saturation), if needed. In other words, if an object in the color image can be extracted from only its gray scale component, then it is futile to segment other color channels. It also reduces the processing time for segmentation and object separation. However, in the rare case where two distinct object possess same luminance value and only one of them is needed to be extracted, the other color channels can be segmented to separate them. The reason for choosing YIQ color space, is due to the fact that this space has less correlation among its constituent components \cite{28}. It is also worth noting that there are only a limited number of segmentation algorithms for color images. They lack the efficacy of gray scale segmentation algorithms for monochrome images \cite{29}.

After segmentation of the Y channel, the desired object to be extracted, can be identified by choosing one or more combination of segmented values, $S_1,\, S_2,\,S_3,\,\dots,\,$ $S_{n+1}$, where $n$ is the number of thresholds. These $S_i's$ represent different regions of the image having same, similar or nearby pixel values. Let the region/s to be separated is/are denoted by $\zeta$, where $\zeta  \in \left\{S_1,\, S_2,\,S_3,\,\dots,\,S_{m+1}\right\}$, $m \leq n$. After segmentation, the object separation is carried out through the following process. The binarization of three channels is done representing the object to be separated by 1 and the rest by 0.  

The binarized $Y$ channel components are represented as,

$$
Y_B = 
\left\{
\begin{array}{lr}
1 & Y(x,y) \in \zeta\\
0 & otherwise,
\end{array}
\right.
$$\\
Here $x \in [1,M]$ and $y \in [1,N]$, where $M$ and $N$ are matrix dimensions of the image under consideration. The binarization of $I$ and $Q$ components are done in correspondence with the binarized $Y_B$ component:
\begin{equation}
I_B = Q_B = Y_B .
\end{equation}
Let the object to be separated $O_S$, be represented by three channels, $Y_S,\,I_S$ and $Q_S$. The three channels for the separated objects can then be represented by,
\begin{equation}
Y_S = Y_B \otimes Y  ,
\end{equation}
\begin{equation}
I_S = I_B \otimes I
\end{equation}
and
\begin{equation}
Q_S = Q_B \otimes Q ,
\end{equation}\\
where $\otimes$ denotes pixel-by-pixel multiplication. 

In order to isolate the objects in the original image, corresponding to the above segmentation, the original pixel values of $Y$, $I$ and $Q$ are respectively retained where the pixel value of the binarized $Y_B$, $I_B$ and $Q_B$ matrices have unit entries. The rest of the values are either replaced with 0 or 255, depending on the user or characteristics of the object. If the object is of darker shade, it is better to replace the background pixels by 255 to distinctly separate the two. In case of object having a lighter shade, replacing the background with 0 is more suitable.

The approach illustrated above can be summarized in the following steps:\\
\\
$Step\, 1$: Input the image $F$ and convert it into YIQ color space ($F_{YIQ}$).\\
$Step\, 2$: Segment $F_Y$ using the methodology I or II depending on the image.\\
$Step\, 3$: Binarize $F_Y$, $F_I$ and $F_Q$ on the segmented values required to separate the object.\\
$Step\, 4$: Extract the original pixels of $F_Y$, $F_I$ and $F_Q$ with the help of binarized $F_Y$, $F_I$ and $F_Q$.\\
$Step\, 5$: Convert $F_{YIQ}$ (extracted) after concatenating $F_Y$, $F_I$ and $F_Q$ to RGB color space $F_{RGB}$.\\

\section{Segmentation Methodologies}
\label{sec:2}
As mentioned earlier, we adopt two methodologies on the basis of the types of images under study. Both the methodologies use histogram based thresholding techniques for segmentation of gray scale images. The two prime instruments, used in selecting the thresholds, are weighted mean and standard deviation:
\begin{equation}
\mu = \frac{\displaystyle\sum_{i=a}^b \eta{(i)}\beta{(i)}}{\displaystyle\sum_{i=a}^b \eta{(i)}}
\end{equation}
and
\begin{equation}
\sigma = \sqrt{\frac{\displaystyle\sum_{i=a}^b \eta{(i)}[\beta{(i)}-\mu]^2}{\displaystyle\sum_{i=a}^b \eta{(i)}}} ,
\end{equation}\\
where $\mu$ = weighted mean of histogram (gray scale image),\\
           $\beta$ = different pixel values in a gray scale image,\\
             $\eta$ = frequency of each pixel value occurring in a particular image,\\
             $\sigma$ = standard deviation of image histogram,\\
             $a$ = minimum pixel value that is to be considered for segmentation, and\\
             $b$ = maximum pixel value that is to be considered for segmentation.\\
\\
Here, $[a, b]$ represent the range of different segmented regions in the histogram. It is to be noted that for the uint8 gray level images, $\beta$ will range from $[0, 255]$. Both the methodologies adopt the concept of vanishing variance of histogram, in complimentary regions. The variable thresholds are chosen, based on mean ($\mu$), standard deviation ($\sigma$) and skewness of the image histogram:
\begin{equation}
\tau = \mu\pm\kappa\sigma .
\end{equation}\\
with the parameter $\kappa$ taking into account the skewness of the image [30].

\subsection{Methodology-I}
\label{sec:3}
This methodology is used for images, where the foreground is dominant over the background. In these type of images, the focus is on separating the background from the objects, as it will be distinct, having pixel values at the two ends of histogram. This methodology is also favorable for simple background at the ends of the histogram, having regions of pixel values, which are well separated.

Methodology-I is Arora et al.'s algorithm \cite{30} with active usage of $\kappa$ parameter. Unlike Arora  et al., where the value of $\kappa_1=\kappa_2=\kappa=1$, in methodology-$I$, $\kappa_1$ and $\kappa_2$ are independent of each other. In other words, methodology-$I$ incorporates the suggestion of \cite{30} on using $\kappa_1$ and $\kappa_2$, in place of $\kappa$, and varying $\kappa_1$ and $\kappa_2$ to obtain optimal or desired segmentation. The main objective of this paper is object separation through image segmentation. Thus, methodology-$I$ applies Arora et al. algorithm by varying $\kappa_1$ and $\kappa_2$ actively for obtaining appropriate thresholds to carry out segmentation, and finally extracting the desired object.

The proposed procedure takes into account the fact that important information lie towards the weighted mean of the histogram. Furthermore, it also incorporates the fact that human eye is more sensitive to the gray pixel values as compared to black and white ones. Hence, the aim is to have wider segment size at the ends of histogram of a gray scale image \cite{30}. This will separate the background from the foreground. In some images, a very small part of the background may have pixel values near to the foreground or object pixel values. Therefore, the segmented region in the histogram becomes finer as it reaches its weighted mean, to separate out these background pixel values from the object.

The thresholds for the histogram are defined by:
\begin{equation}
T_i=\left\{(a,b):\mu-\kappa_1\sigma, a \in T_{i-1}, b \in T_{j+1}\right\},
\end{equation}
\begin{equation}
T_j=\left\{(a,b):\mu+\kappa_2\sigma, a \in T_{i-1}, b \in T_{j+1}\right\},
\end{equation}
and
\begin{equation}
T_{\frac{N+1}{2}}=\left\{(a,b):\mu_{\frac{N+1}{2}}, a \in T_{i-1}, b \in T_{j+1}\right\},
\end{equation}\\
where $i \in [1,\frac{N-1}{2}]$, $j \in [(\frac{N+1}{2}),N]$.
\\
\\
$N$ is the number of thresholds, which is always odd. It is to be noted that initial values of $a$ and $b$, or $T_0$ and $T_{N+1}$ are 0 and 255 repectively, because they include the complete image. The segmented values are the weighted mean of the histogram between the thresholds, and are defined as, 
\begin{equation}
\mu_S = \left\{(a,b):\mu_i, a = T_i, b = T_{i+1}\right\}.
\end{equation}\\
Therefore, the segmented $F_Y$ defined as $F_{YS}$ can be written as:
\begin{equation}
F_{Y_S}=\left\{(a,b):F_Y(a,b)=\mu_i, F_Y(a,b) \in [T_i,T_{i+1}], i \in [0,N]\right\}.
\end{equation}

\subsection{Methodology-II}
\label{sec:4}

The images for which methodology II is applied have dominant backgrounds. This methodology is found to be best suited for complex background. It is due to the fact that the object has to be separated out from the background, which itself is complex.

As the background is complex, the methodology incorporates the fact that important information lies at the end of the histogram plot. This approach segments the background into different regions to extract the desired objects. Hence, the methodology has finer segment size at the ends of histogram, while having wider segment size at the weighted mean. The wider segment size at the weighted mean of histogram is in conformity of not segmenting the object but the background.

The thresholds for the histogram are defined as:
\begin{equation}
T_{i}=\left\{(a,b):\mu-\kappa_1\sigma, a = 0, b = T_{i+1}\right\}
\end{equation}
and
\begin{equation}
T_j=\left\{(a,b):\mu+\kappa_2\sigma, a = T_{j-1}, b = 255\right\}.
\end{equation}\\

The initial values of $T_{i+1}$ and $T_{j-1}$ are: 
\begin{equation}
T_{\frac{N+1}{2}} = T_{i+1} = T_{j-1} = \left\{(a,b):\mu_{\frac{N+1}{2}}, a = T_{i-1}, b = 255\right\},
\end{equation}\\
where $i \in [0,\frac{N-1}{2}]$, $j \in [(\frac{N-1}{2}),N+1]$.
\\
\\
$N$ is the number of thresholds. The segmented values are the weighted means of the histogram between the thresholds:
\begin{equation}
\mu_S = \left\{(a,b):\mu_i, a = T_i, b = T_{i+1}\right\}.
\end{equation}\\
Therefore, segmented $F_Y \equiv F_{Y_S}$ is given by:
\begin{equation}
F_{Y_S}=\left\{(a,b):F_Y(a,b)=\mu_i, F_Y(a,b) \in [T_i,T_{i+1}], i \in [0,N]\right\}.
\end{equation}
For the purpose of illustration, we explicitly demonstrate the histograms of images, in which the two methodologies are successfully applied at threshold level 5 for different sets of $\kappa_1$ and $\kappa_2$ values. The image histograms in $Fig.\,1$ and $Fig.\,2$ clearly demostrate the non$-$overlapping Gaussian distribution corresponding to different objects. The corresponding impulse functions extracted through the present procedure given in $Fig.\,1$ and $Fig.\,2$ show the mapping of Gaussian distributions into respective $\delta-$functions.

\begin{figure*}[ht]
\begin{center}
\includegraphics[width=2.70cm]{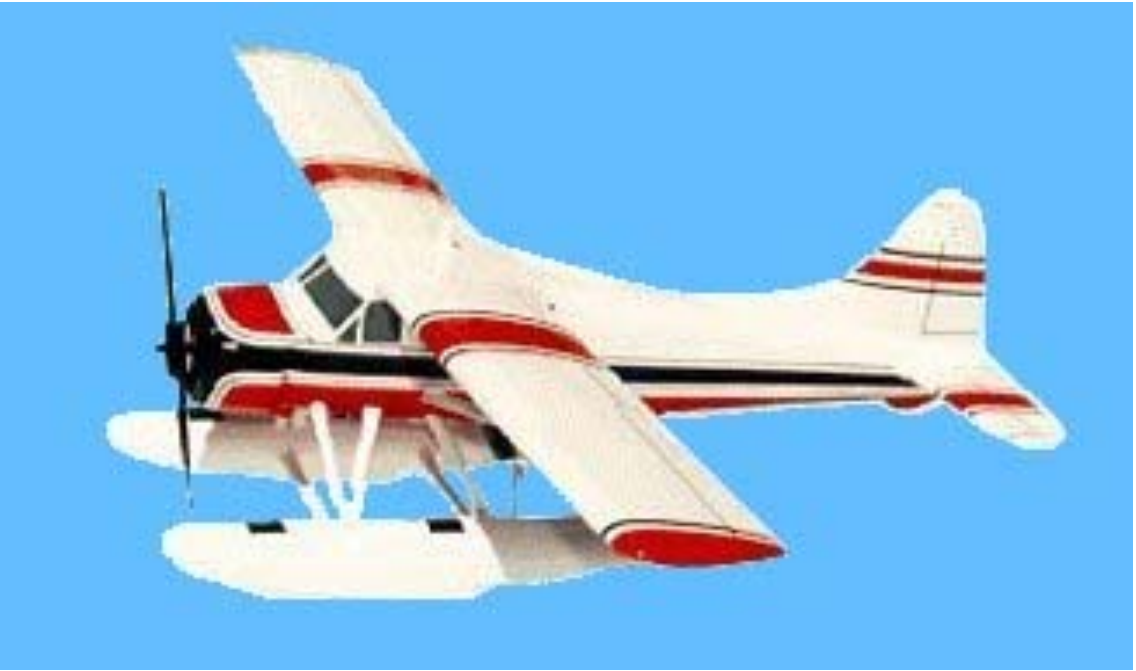}
\includegraphics[width=2.70cm]{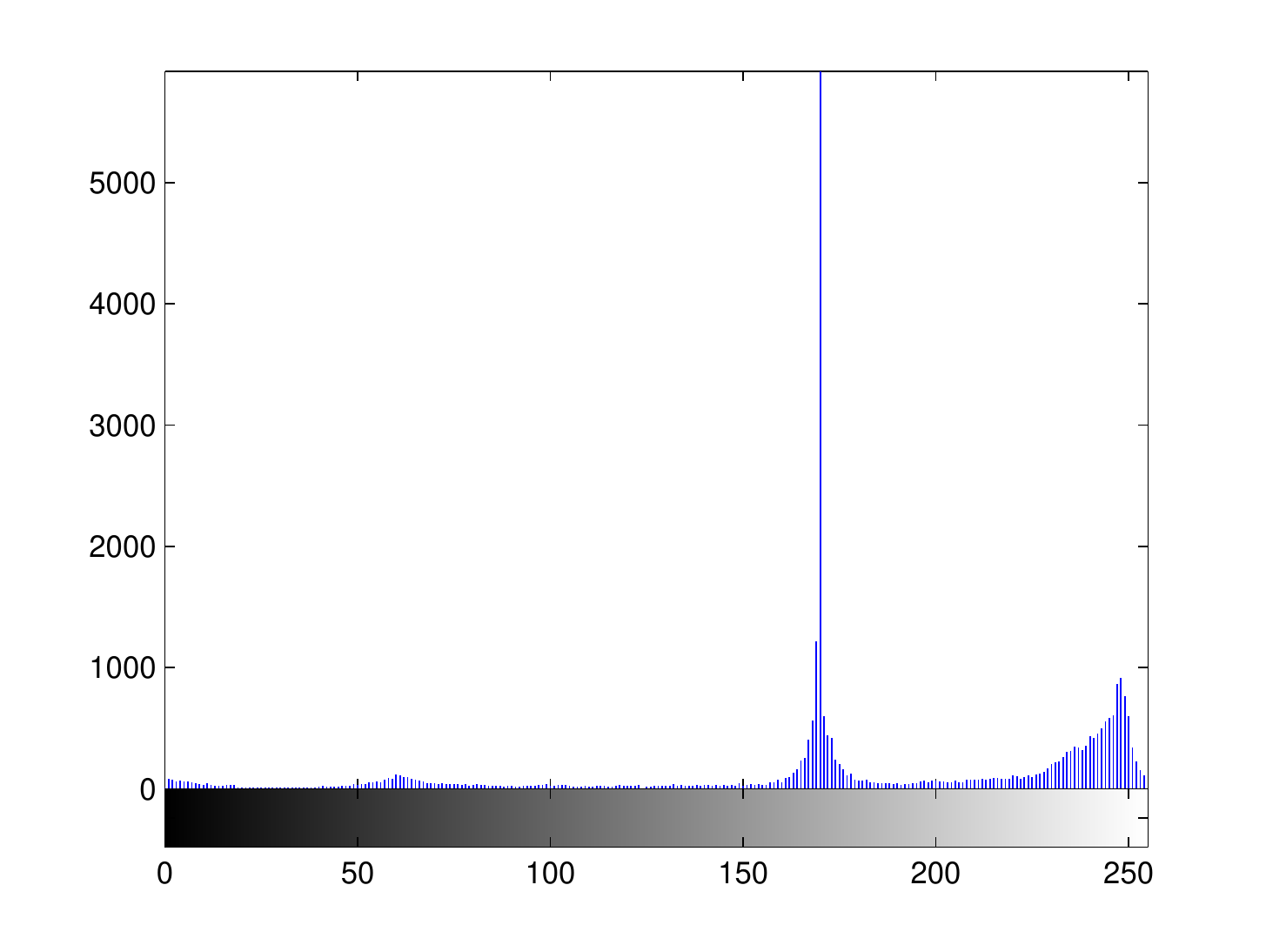}\\
\includegraphics[width=2.70cm]{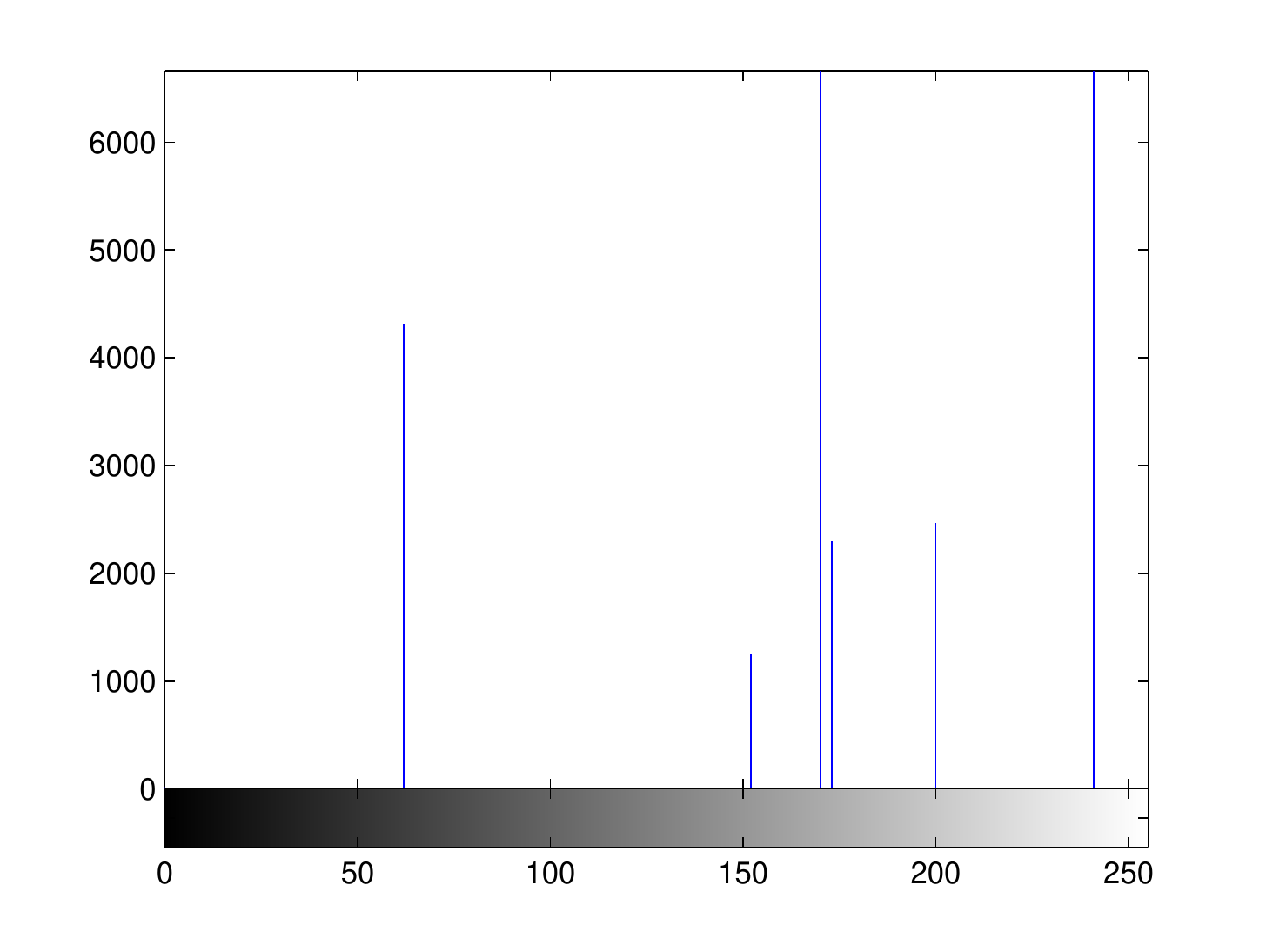}
\includegraphics[width=2.70cm]{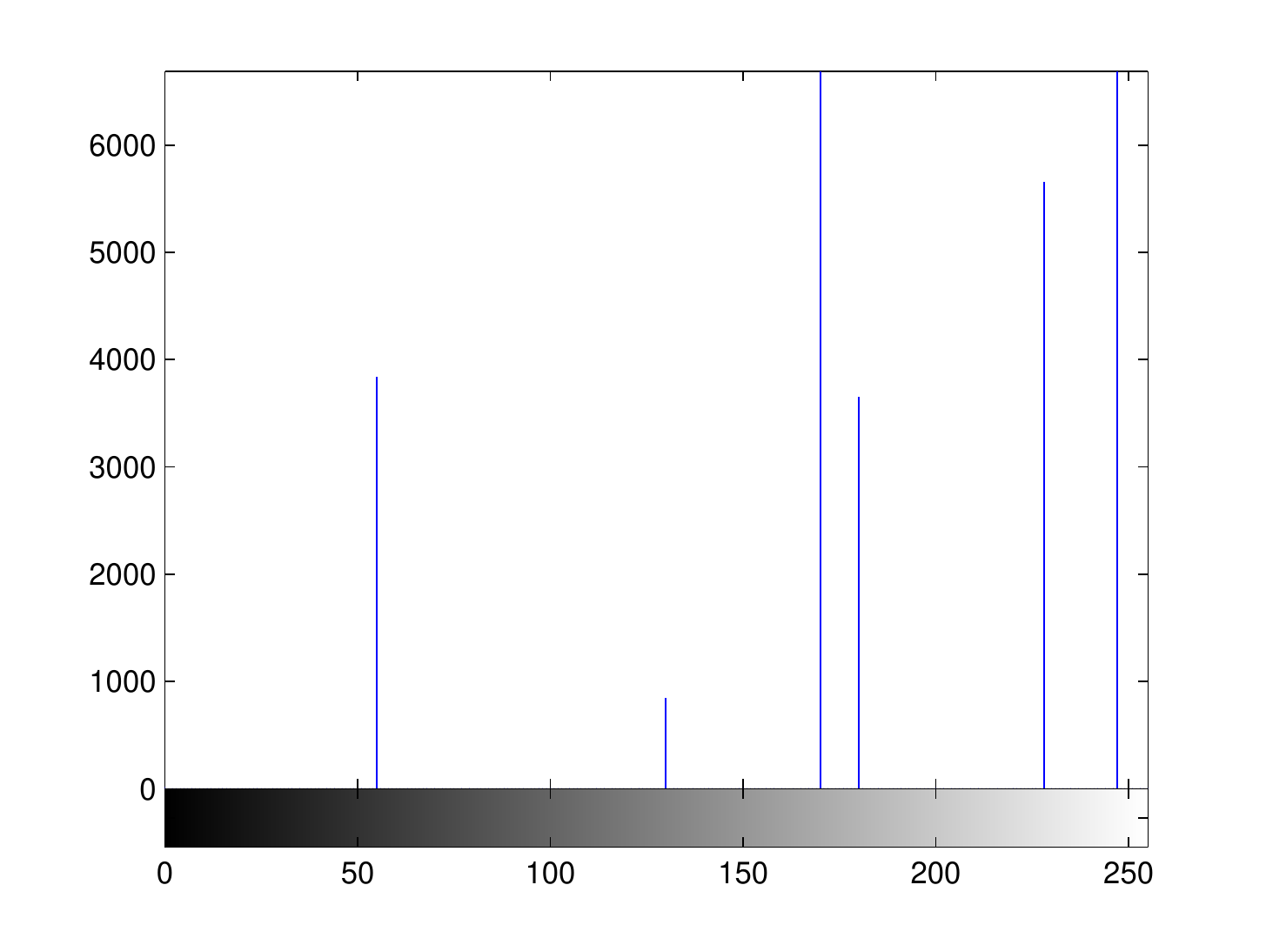}
\includegraphics[width=2.70cm]{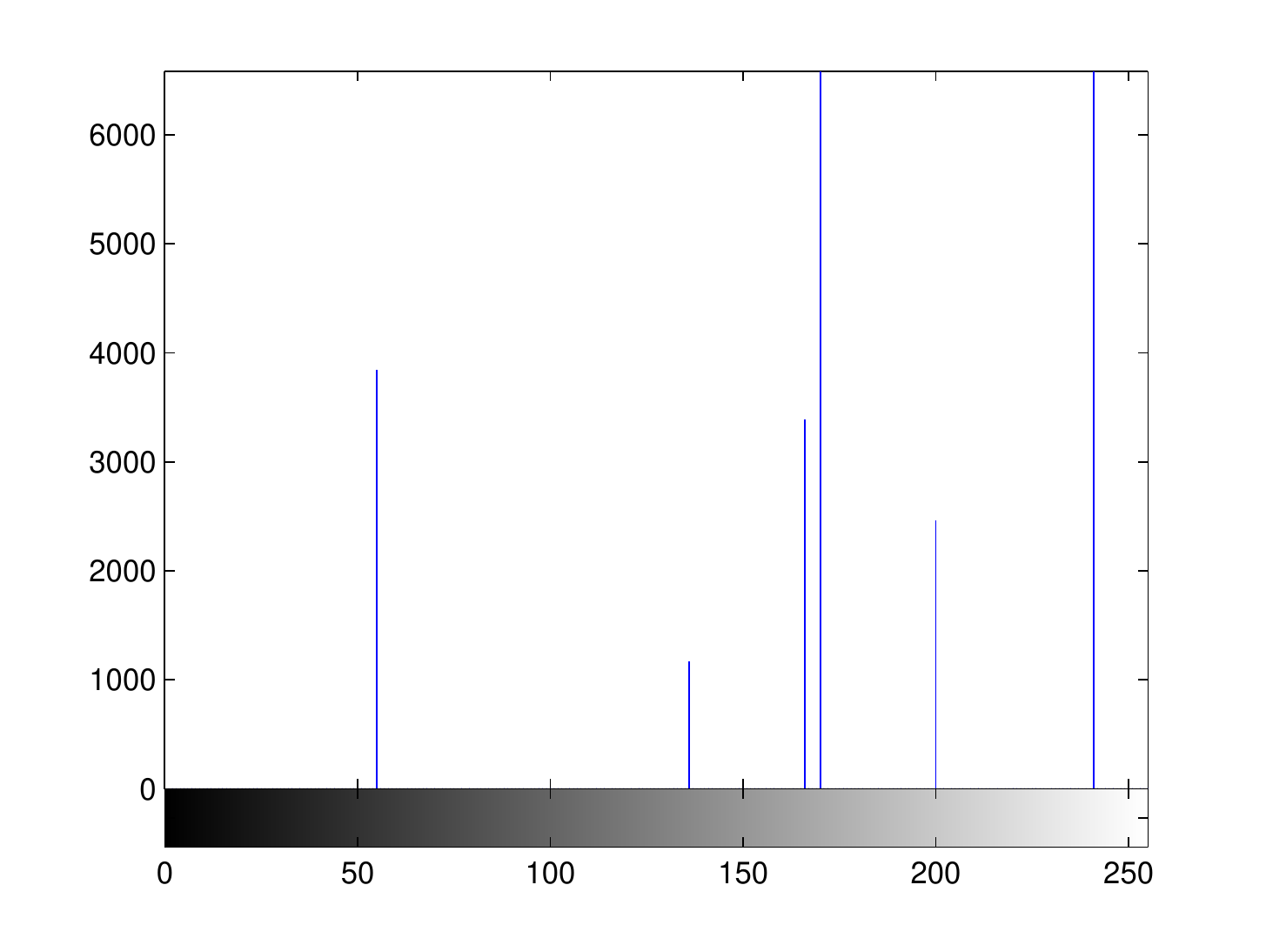}
\includegraphics[width=2.70cm]{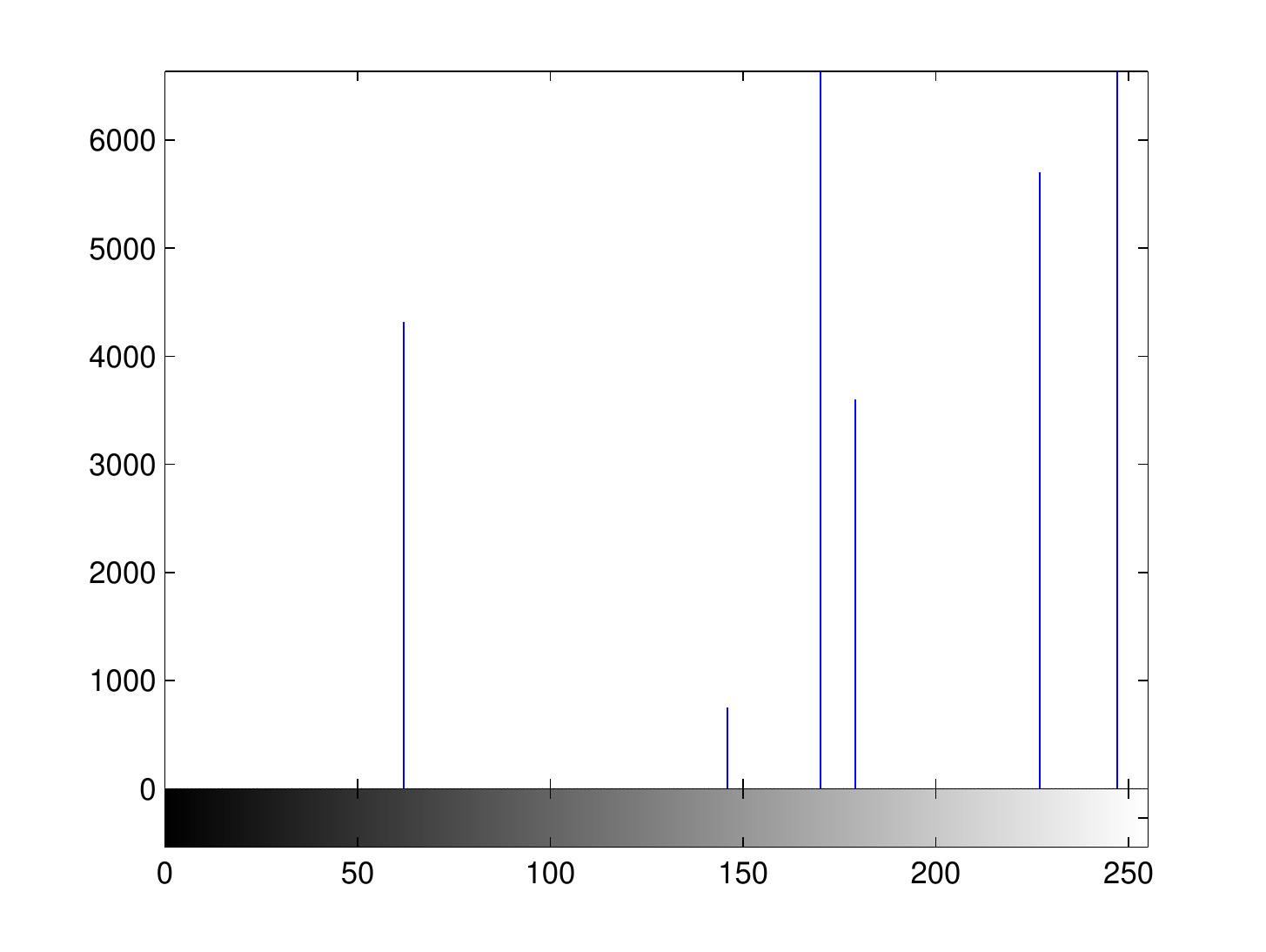}\\
\caption{Clockwise: Original image of Aeroplane; Histogram plot of Gray image of Aeroplane; Histogram plot of Segmented image using methodology-I at $\kappa_1=1\, and\, \kappa_2=1$; $\kappa_1=1.5\, and\, \kappa_2=1.5$; $\kappa_1=1.5\, and\, \kappa_2=1$; $\kappa_1=1\, and\, \kappa_2=1.5$, repectively. The normal distributions corresponding to different objects have been mapped to localized, well separated $\delta-$functions.}
\label{fig:1}
\end{center}
\end{figure*}

\begin{figure*}[ht]
\begin{center}
\includegraphics[width=2.70cm]{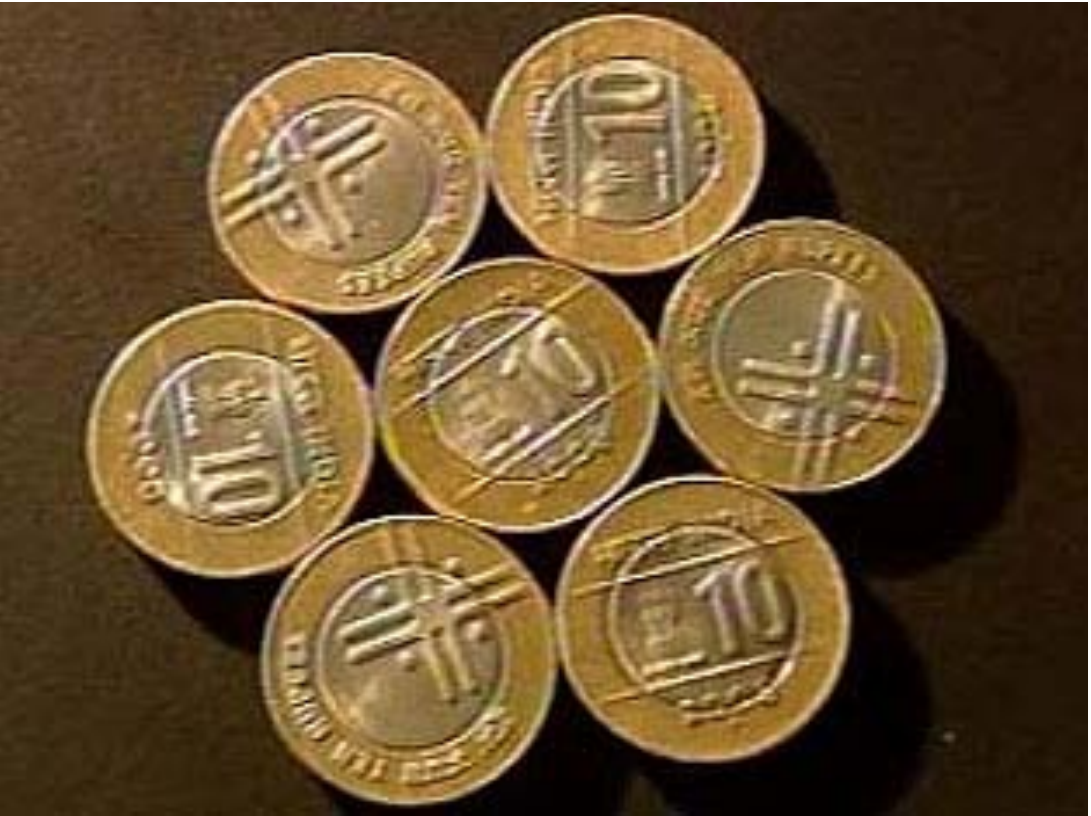}
\includegraphics[width=2.70cm]{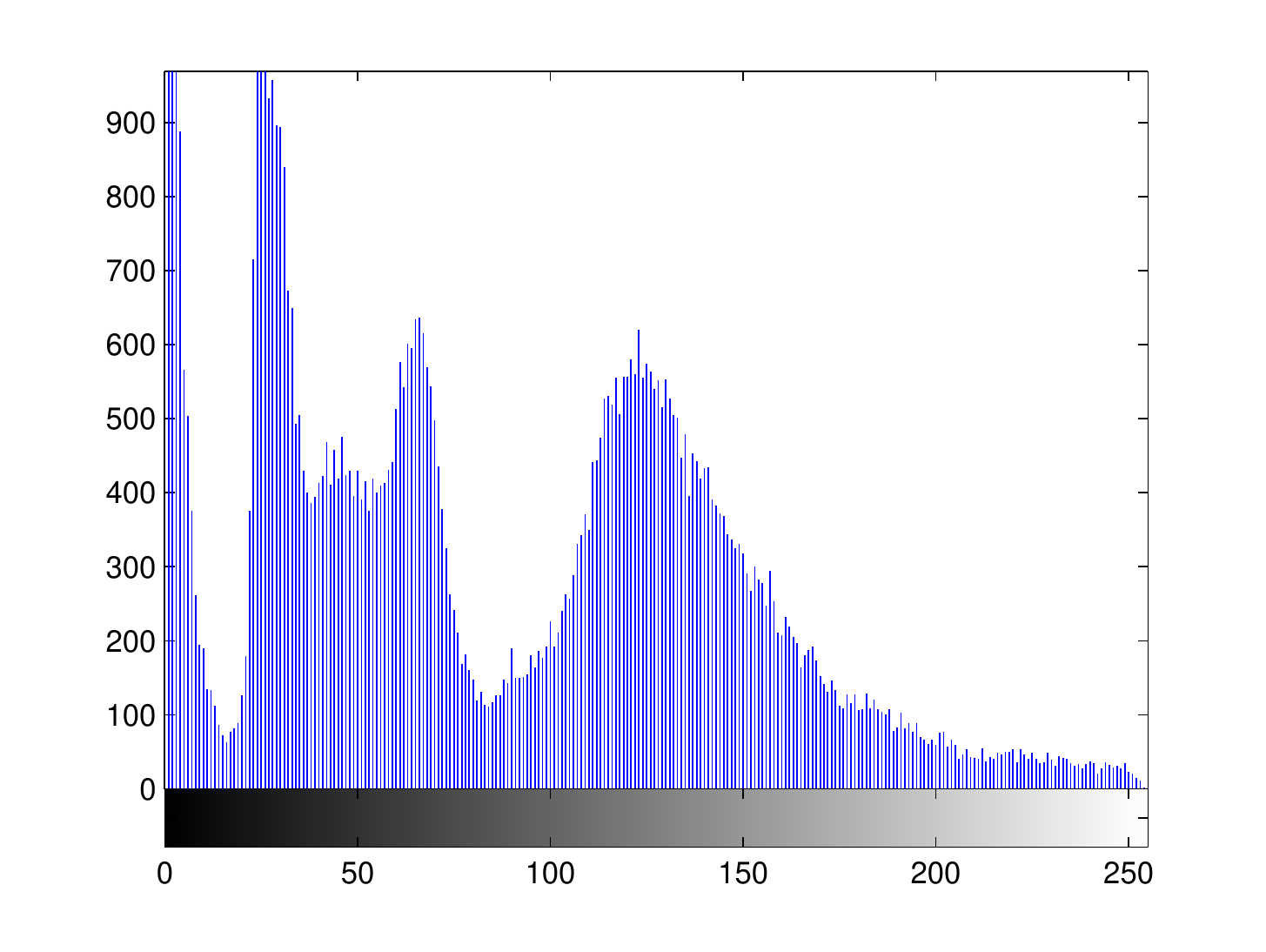}\\
\includegraphics[width=2.70cm]{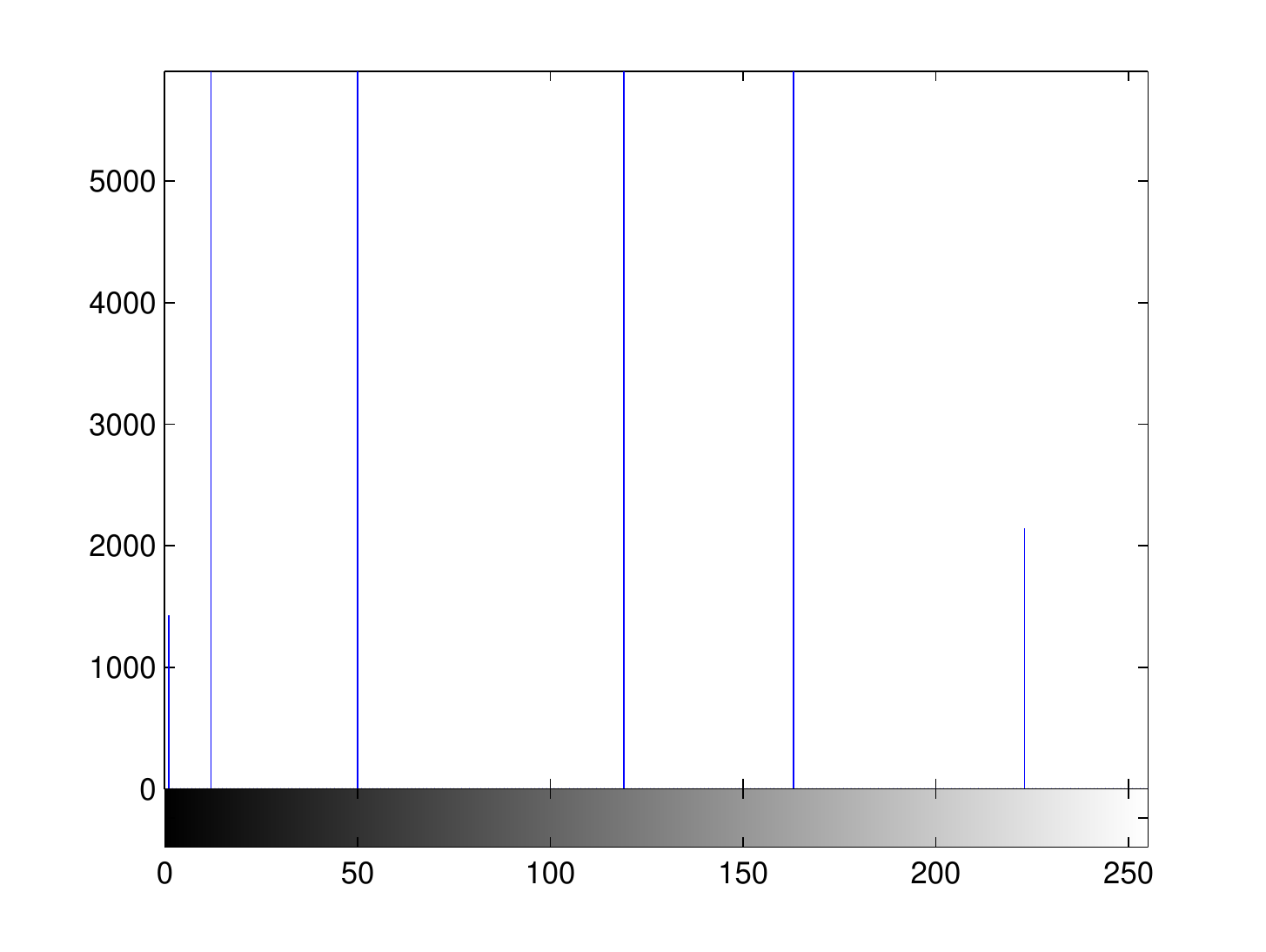}
\includegraphics[width=2.70cm]{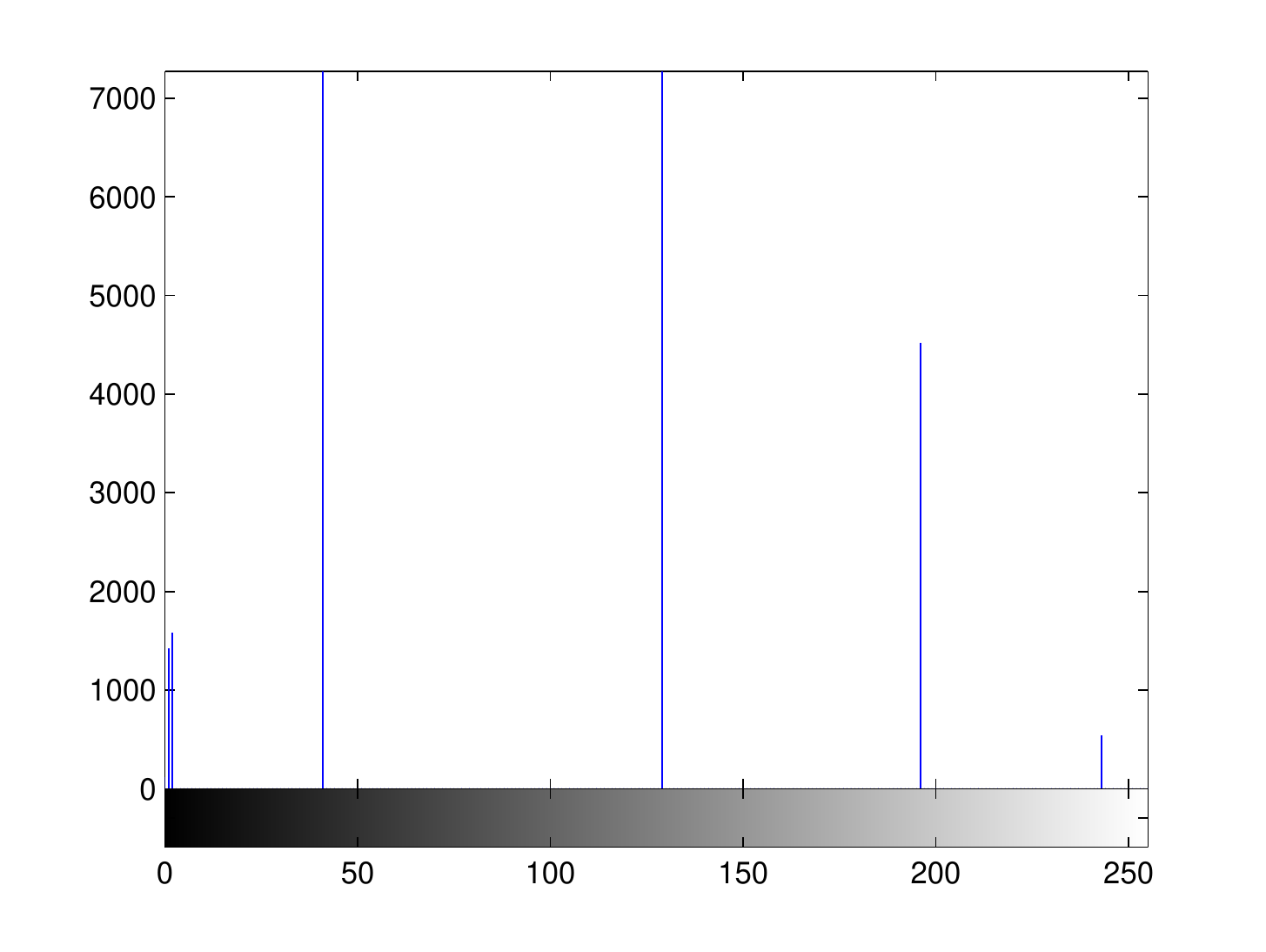}
\includegraphics[width=2.70cm]{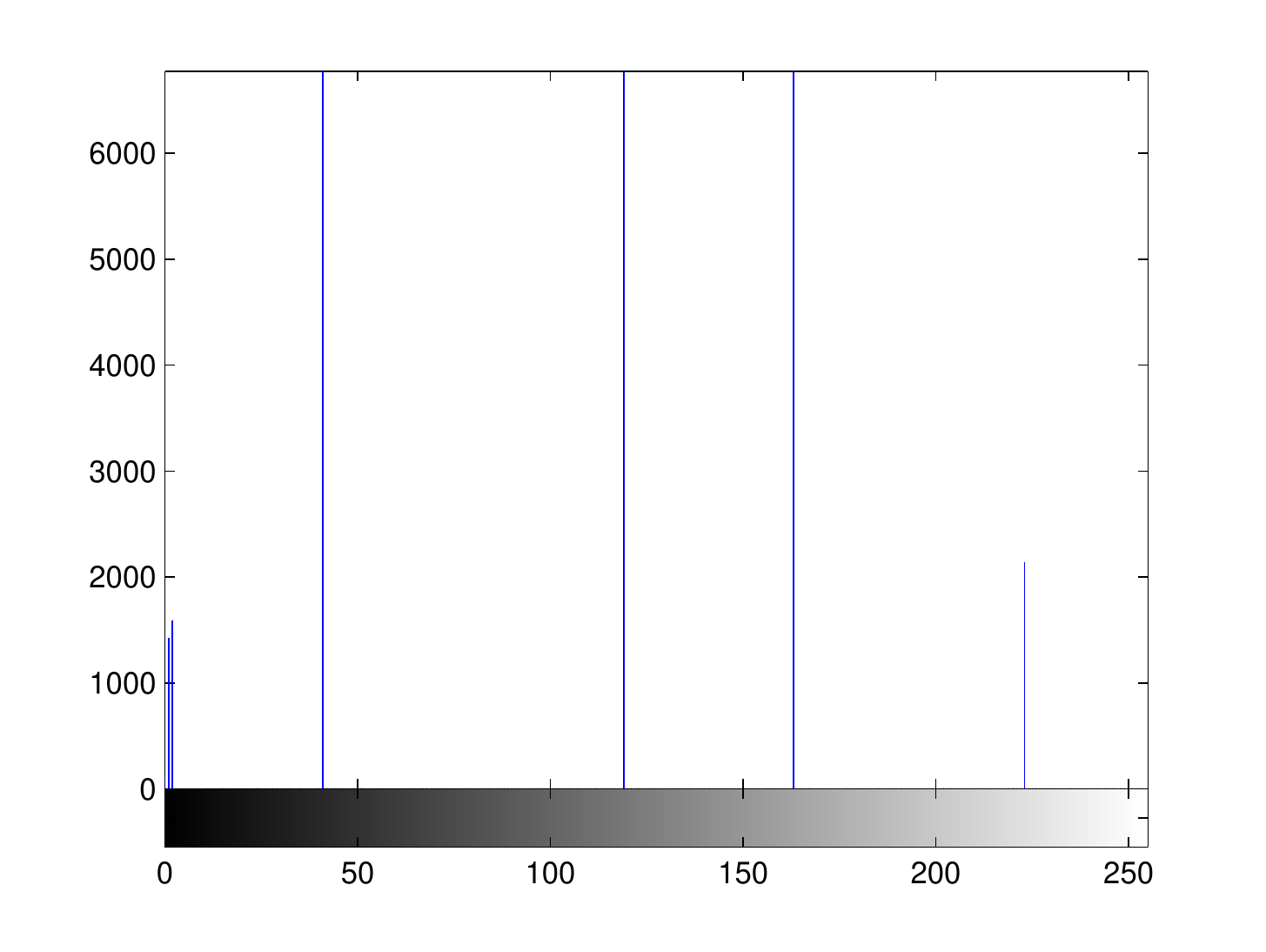}
\includegraphics[width=2.70cm]{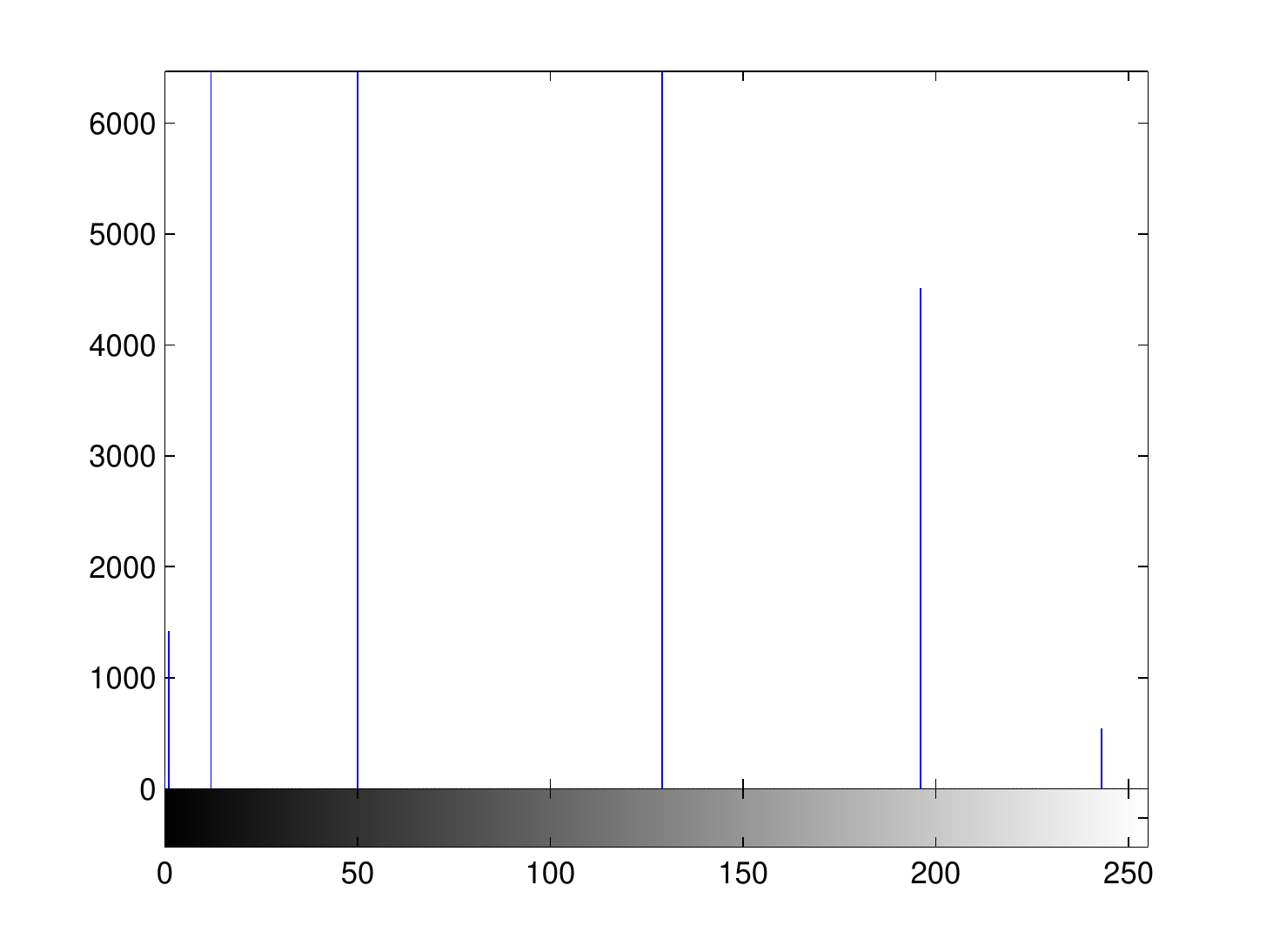}\\
\caption{Clockwise: Original image of Coin; Histogram plot of Gray image of Coin; Histogram plot of Segmented image using methodology-II at $\kappa_1=1\, and\, \kappa_2=1$; $\kappa_1=1.5\, and\, \kappa_2=1.5$; $\kappa_1=1.5\, and\, \kappa_2=1$; $\kappa_1=1\, and\, \kappa_2=1.5$, repectively. The normal distributions corresponding to different objects have been mapped to localized, well separated $\delta-$functions.}
\label{fig:1}
\end{center}
\end{figure*}

\section{Experimental Results and Discussion}

The presented algorithms have been tested for different classes of images. The experimental results section is divided into three parts followed by comments. First, the quantitative analysis of segmentation algorithms and Otsu's multithreshold algorithm \cite{15}\cite{20} is provided. After that, the detailed results of object separation using and comparing methodology-I, methodology-II, and Otsu's multithreshold algorithm, are illustrated. At last, object separation results with varying $\kappa_1$ and $\kappa_2$ are displayed with observations. In the end, comments are provided regarding algorithms and its usage.

\subsection{Quantitative Analysis of Segmentation Algorithm}

To assess the quality of the segmentation methodologies and compare them with Otsu's multithreshold algorithm objectively, Mean Structural Similarity Index Measure (MSSIM) \cite{31} is used in this paper. MSSIM assesses the quality of segmented image by calculating and comparing the structural information degradation w.r.t the original image. This evaluation parameter is considered because every object in an image has a structural form, easily perceived by human eye. Thus, the segmentation algorithm retaining the maximum structural features of the original image is most suitable for extracting object segments and hence, performing object separation. The formulae for finding MSSIM are defined as, 

\begin{equation}
SSIM(p,q)=\frac{(2\mu_p\mu_q + K_1)(2\sigma_{pq} + K_2)}{(\mu_p^2 + \mu_q^2 + K_1)(\sigma_p^2 + \sigma_q^2 + K_2)}
\end{equation}

\begin{equation}
MSSIM(I,\tilde{I})=\frac{1}{M} \sum_{i=1}^{M} SSIM(p_i,q_i)
\end{equation}

where $\mu$ is the mean; $\sigma$ is the standard deviation; $p$ and $q$ are the window sizes of the original and the reconstructed images which are typically $8 \times 8$; $K_1$ and $K_2$ are the constants having values $0.01$ and $0.03$, respectively; $M$ is the total number of windows. It has be noted that MSSIM value ranges from 0 to 1 representing no similarity and complete similarity with the original image, respectively.

The results in $Table\,1$ show the MSSIM values of each segmentation algorithm i.e. methodology-I, methodology-II and Otsu's multithreshold algorithm, at different number of segments. It can be seen from $Table\,1$ that MSSIM of Otsu's method is substantially less in comparison to the two proposed methodologies in the paper, with minimum and maximum MSSIM ranging in 0.30's and 0.80's, respectively, for the given test images. The superiority of the two algorithms described in this paper can be seen from the fact that no MSSIM value at any number of segments and for any test images, lies below 0.84. This superiority can be attributed to the efficacy of the proposed algorithms to effectively separate the non-overlapping or minimally overlapping distributions. As mentioned earlier, images consist of objects with distinct set of intensity values, which results in non-overlapping or minimally overlapping distributions in the histogram of image. Hence, the structure of the original image is maximally preserved in the segmented image through proposed methodologies, while also retaining the objects' structure at certain number of segments. On the other hand, as evident from MSSIM values from Otsu's method of maximizing the class variance, separate the distributions with some part of other distributions resulting in the distortion of structural information.

\begin{table*}[ht]
\caption{MSSIM of various test images at different number of segments of the image using methodology-I, methodology-II and Otsu's method.}
\label{tab:3}
\begin{center}
\begin{tabular}{ccccc}
\hline
Image Name & Number of Segments & Methodology-I & Methodology-II & Otsu's Method\\
\hline
Airplane & 2 & 0.944439048 & 0.944439048 & 0.861760028\\
	& 4	&	0.983242212 &	0.987593927 &	0.600859466\\
& 6	&	0.98856355 &	0.989160305 &	0.606888357\\
& 8	&	0.988563281 &	0.989174075 &	0.604777984\\
& 10	&	0.98867102 &	0.989174962 &	0.608014416\\
\\
Eagle & 2	&	0.953403014 &	0.953403014 &	0.322423915\\
& 4 &		0.966730016 &	0.956287227 &	0.355894963\\
& 6 &		0.986707214 &	0.95634721 &	0.358507741\\
& 8 &		0.996196072 &	0.956353393 &	0.361168687\\
& 10 &		0.997061958 &	0.956353449 &	0.363427451\\
\\
House & 2	&	0.950822379 &	0.950822379 &	0.73208051\\
& 4 &		0.96826398 &	0.976776477 &	0.659689631\\
& 6 &		0.977492518 &	0.99198246 &	0.672277065\\
& 8 &		0.977490385 &	0.992307364 &	0.673129246\\
& 10 &		0.977491222 &	0.992308977 &	0.673960416\\
\\
Coin & 2	&	0.846094607 &	0.846094607 &	0.627948851\\
& 4 &		0.944789498 &	0.947160725 &	0.69072605\\
& 6 &		0.955724741 &	0.972553444 &	0.715542787\\
& 8 &		0.959571777 &	0.973566561 &	0.716621746\\
& 10 &		0.959615586 &	0.973582226 &	0.715071021\\
\\
Coins & 2 &		0.851321383 &	0.851321383 &	0.43831022\\
& 4 &		0.970455106 &	0.963198732 &	0.532771674\\
& 6 & 		0.96952448 &	0.974975678 &	0.524782516\\
& 8 &		0.977635895 &	0.975190193 &	0.534911366\\
& 10 &		0.97908687 &	0.975205358 &	0.538537487\\

\hline
\end{tabular}
\end{center}
\end{table*}

Between the two methodologies, it can be noted that a particular methodology dominates the other, in terms of the MSSIM value for all segmentations with number of segments greater than two. When there are just two segments, both the methodologies would result in the same segmented regions because the formulae for calculating thresholds is same for both methodologies. The difference lies in the part of the histogram chosen for applying the algorithm recursively. The choice of methodology to be used for an image can somewhat be indicated by the MSSIM value in $Table\,1$. The gradient of the two MSSIM values is an important factor in objectively choosing the appropriate methodology for segmenting an image. However, the absolute MSSIM value or gradient is only suggestive. For example, in images 'Airplane', 'Eagle', and 'Coins' (see $Fig.\,3-4,7$), methodology-I is used through subjective observations. The MSSIMs of 'Eagle' and 'Coins' have higher values than the other methodology, but not with a substantial gradient. However, MSSIMs of 'Airplane' suggest the usage of methodology-II, again with small gradient. The gradient is too small to indicate a preference towards a methodology. Hence, subject evaluation by the user is the most reliable method. On the other hand, methodology-II is used in images 'House' and 'Coin' (see $Fig.\,5-6$). In this case, the MSSIMs and their gradients provide stronger inclination towards methodology-II. However, it is still advisable to trust a subjective evaluation by the user, especially if MSSIM difference is not sharp or substantial, because the definition of the object in an image is highly dependent on the user and the application.

At last, MSSIM does not provide any information about the appropriate threshold required to separate the object completely. This is because MSSIM evaluates the structure of the image, not explicitly the structure of the object to be separated. Thus, the MSSIM increases with the increase in the number of segments. Although, the proposed methodologies are designed to overcome the over-segmentation issue i.e. even after over segmentation, the desired object can be extracted, it would be futile in terms of time and processing to choose a higher number of thresholds or segments than the requirement. In addition, one cannot determine through the MSSIM, the minimum threshold value at which the problem of under segmentation is overcome for the desired object to be separated.

\subsection{Object Separation Results}

As mentioned in the previous section, MSSIM cannot be used for choosing appropriate thresholds. Also, the type of methodology to be used cannot be reliably determined by MSSIM where gradient between the methodologies are low. Therefore, following method is suggested for finding the right methodology, the required threshold, and the extracting object for a class of images.
\\
$Step\, 1$: Segment the image with the number of thresholds starting from 1 and increased by 2 at each iteration.\\
$Step\, 2$: Compare the segmented image of methodology-I and methodology-II, subjectively. If the object to be extracted exclusively lies or seems to lie in one methodology, then choose that methodology.\\
$Step\, 3$: Repeat steps 1-2 till the desired object seems to lie in a segment value or combination of segmented values which represents the desired object.\\

\begin{figure*}[hb]
\begin{center}
\includegraphics[width=1.80cm]{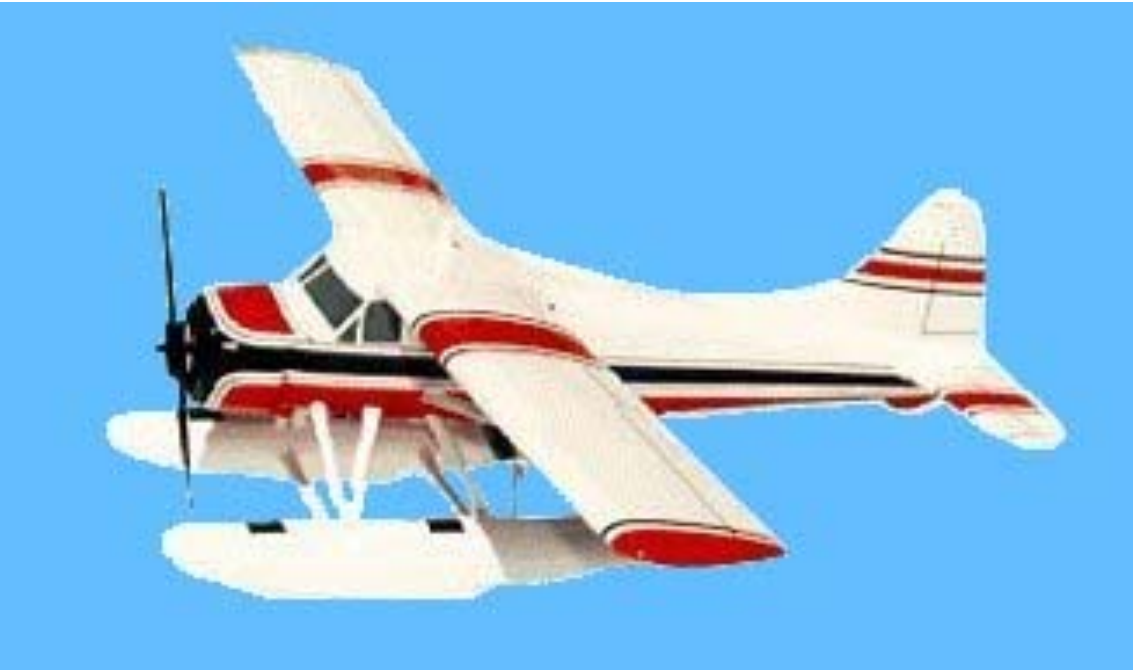}
\includegraphics[width=1.80cm]{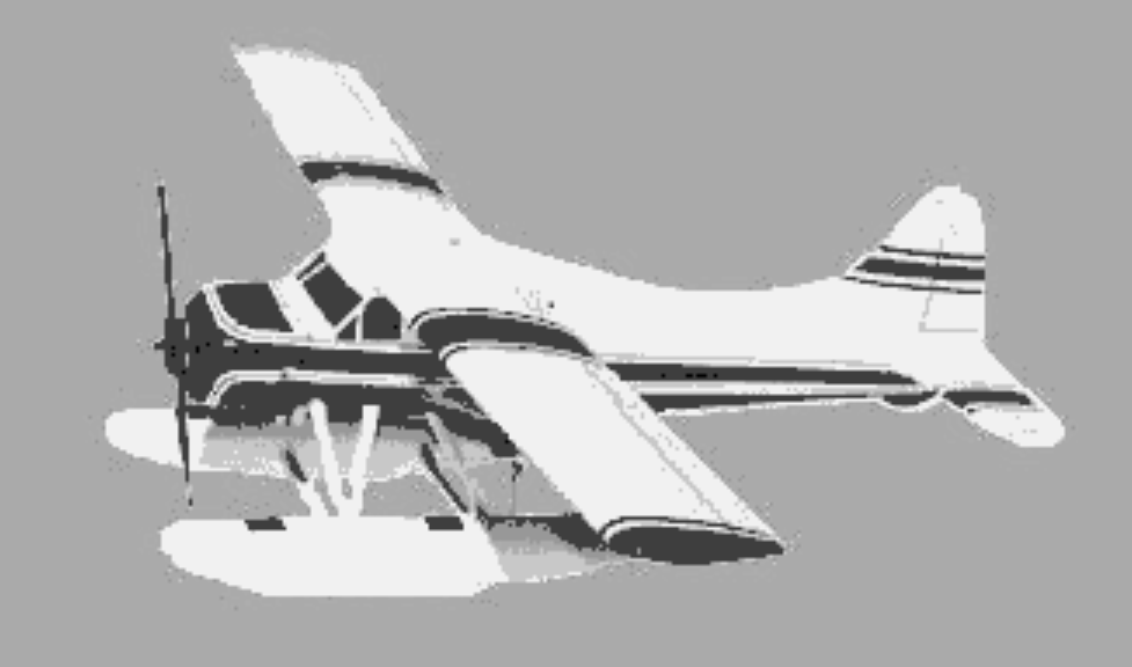}
\includegraphics[width=1.80cm]{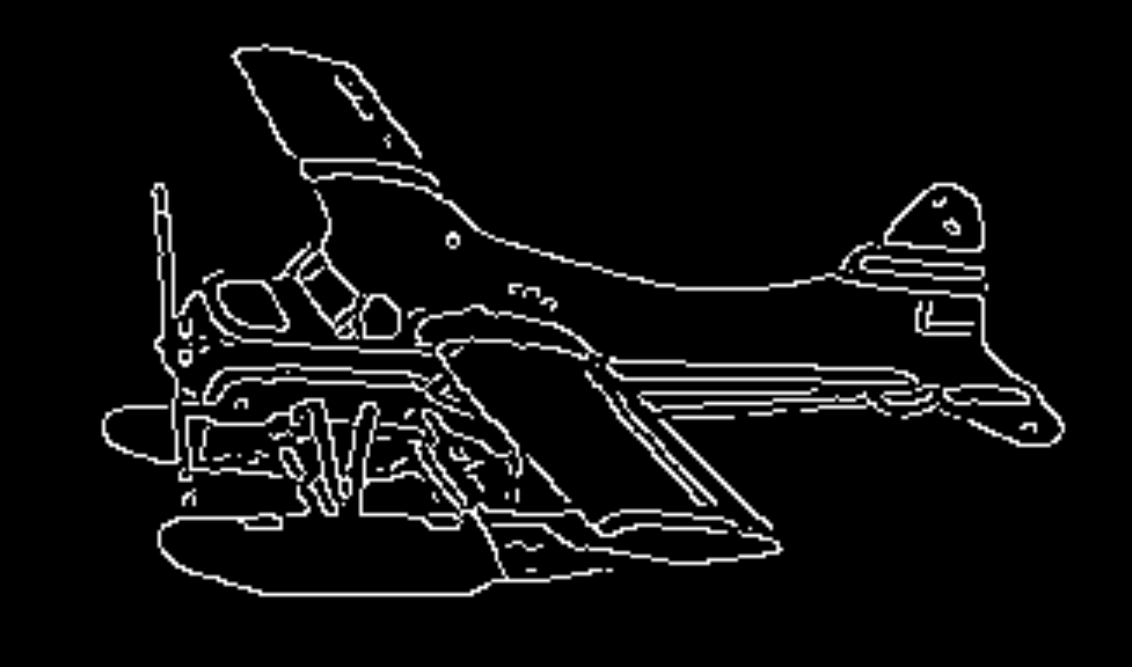}
\includegraphics[width=1.80cm]{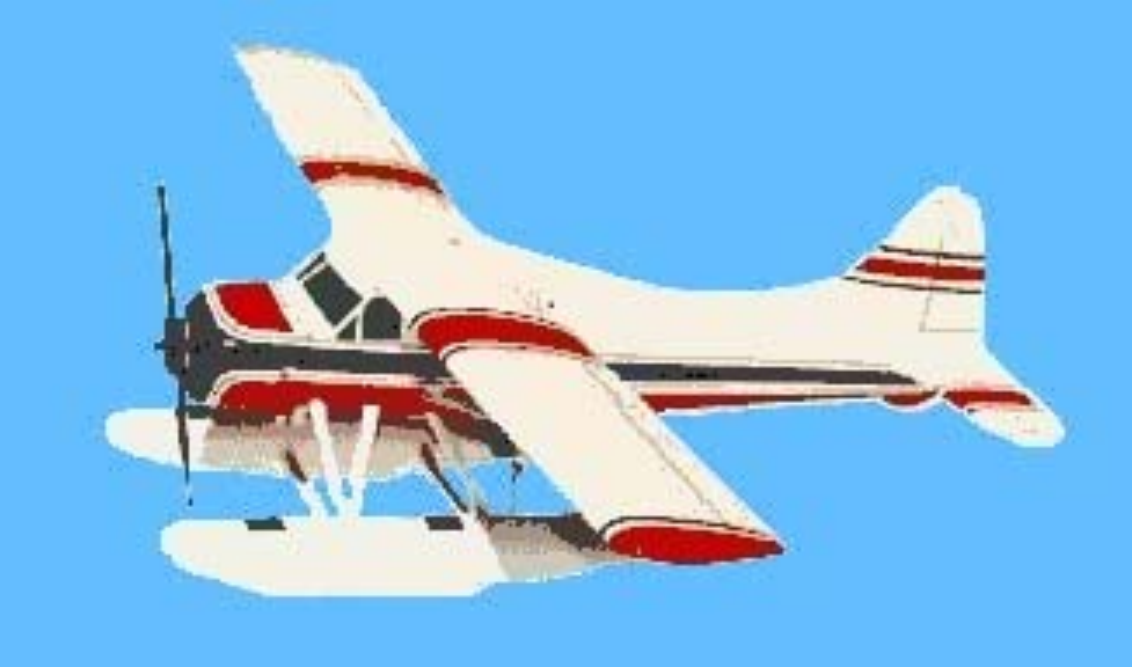}
\includegraphics[width=1.80cm]{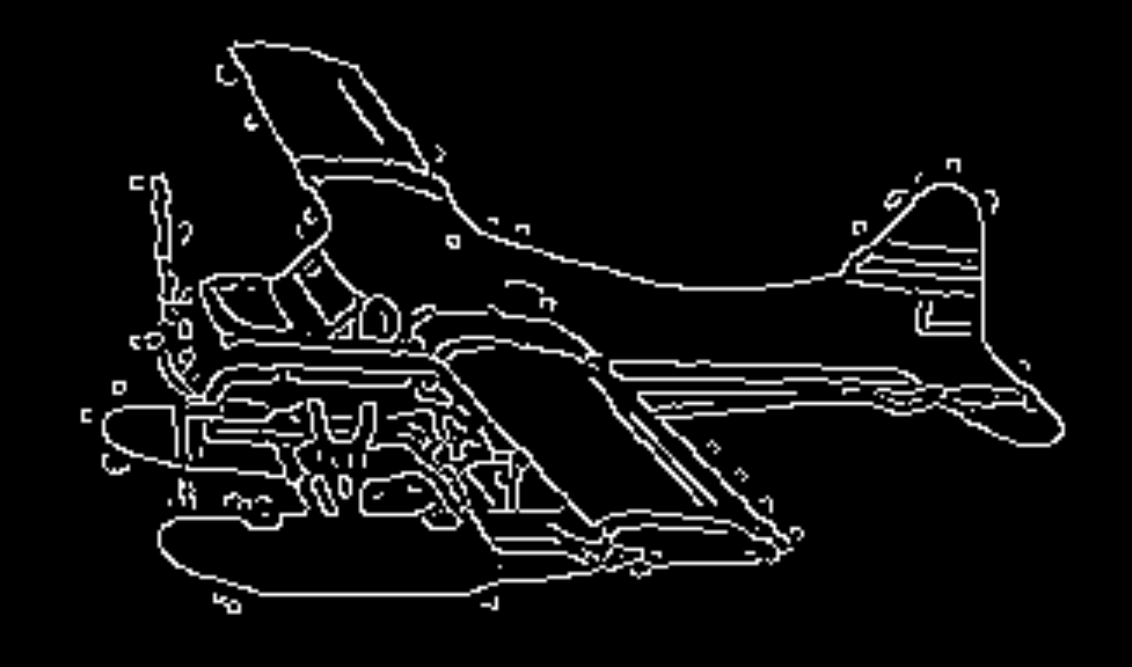}
\includegraphics[width=1.80cm]{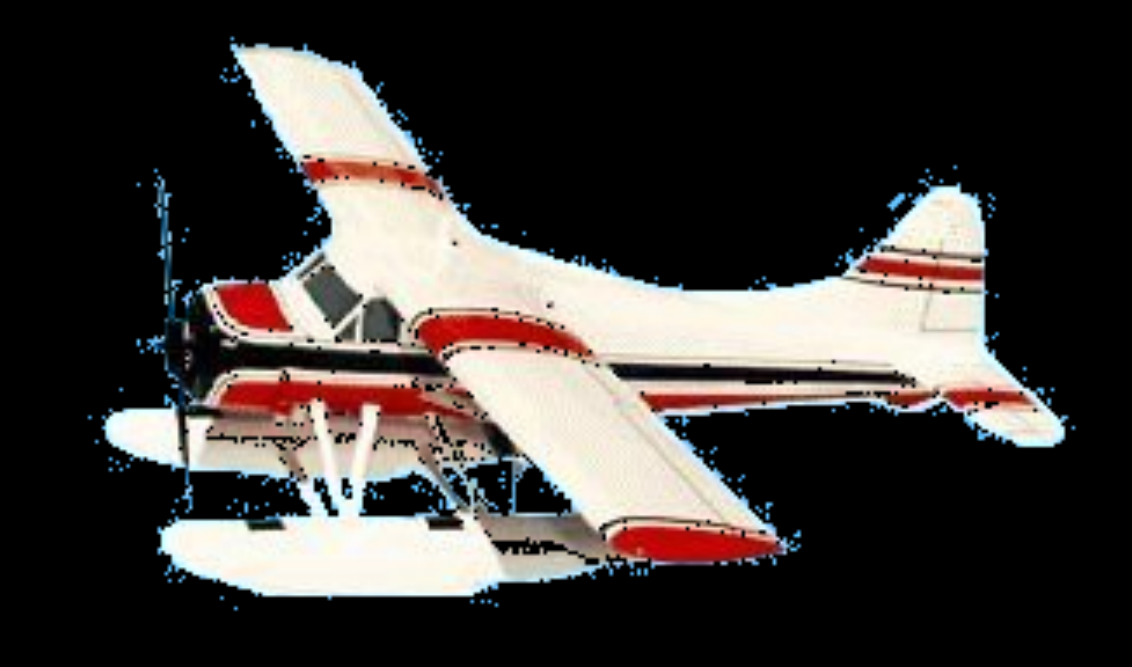}\\

\includegraphics[width=1.80cm]{aeroplane.pdf}
\includegraphics[width=1.80cm]{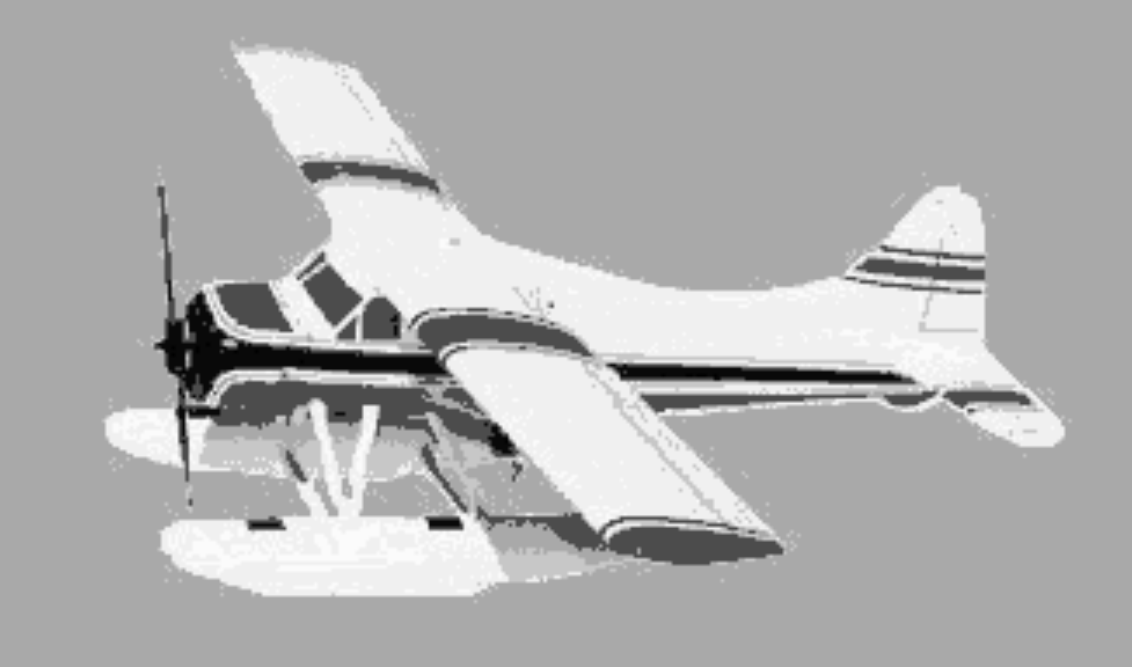}
\includegraphics[width=1.80cm]{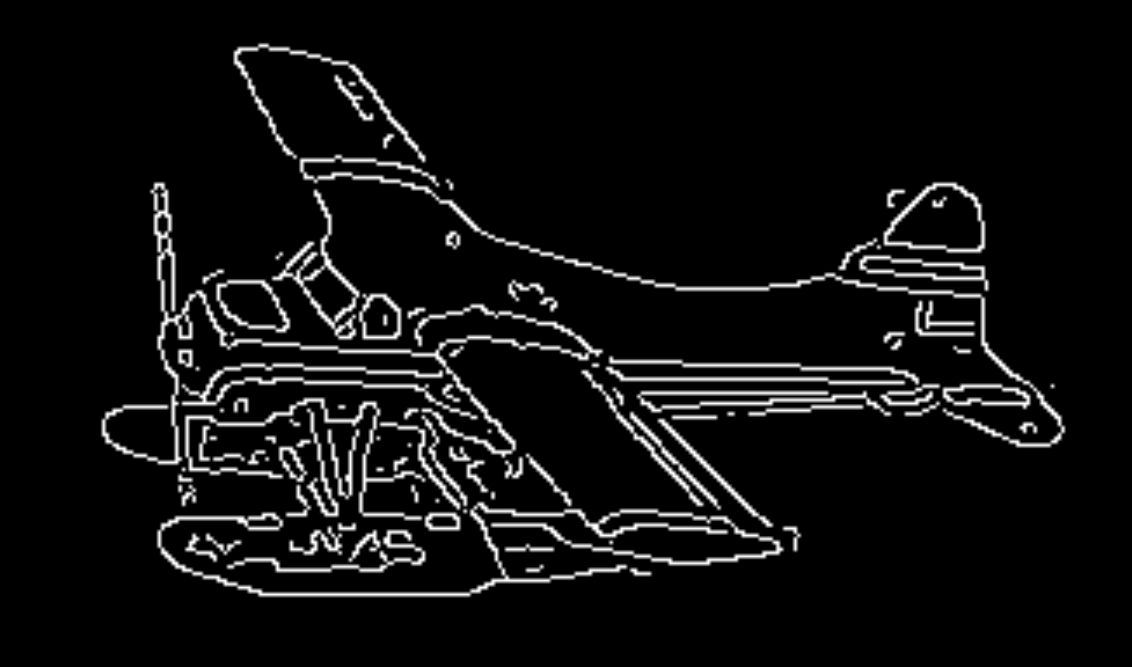}
\includegraphics[width=1.80cm]{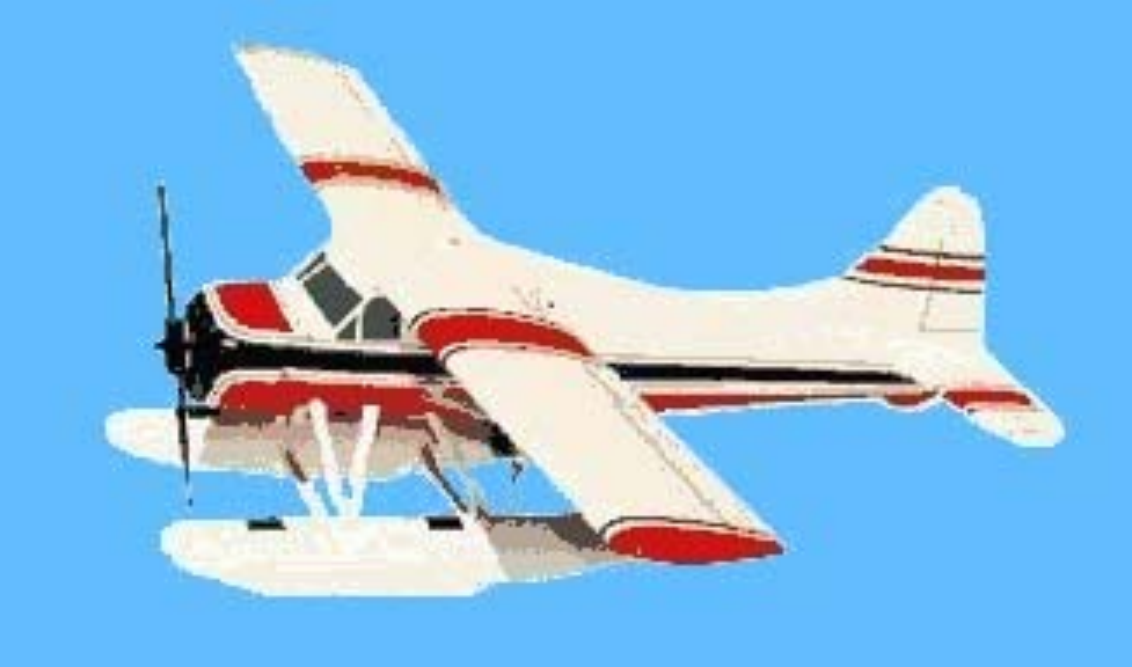}
\includegraphics[width=1.80cm]{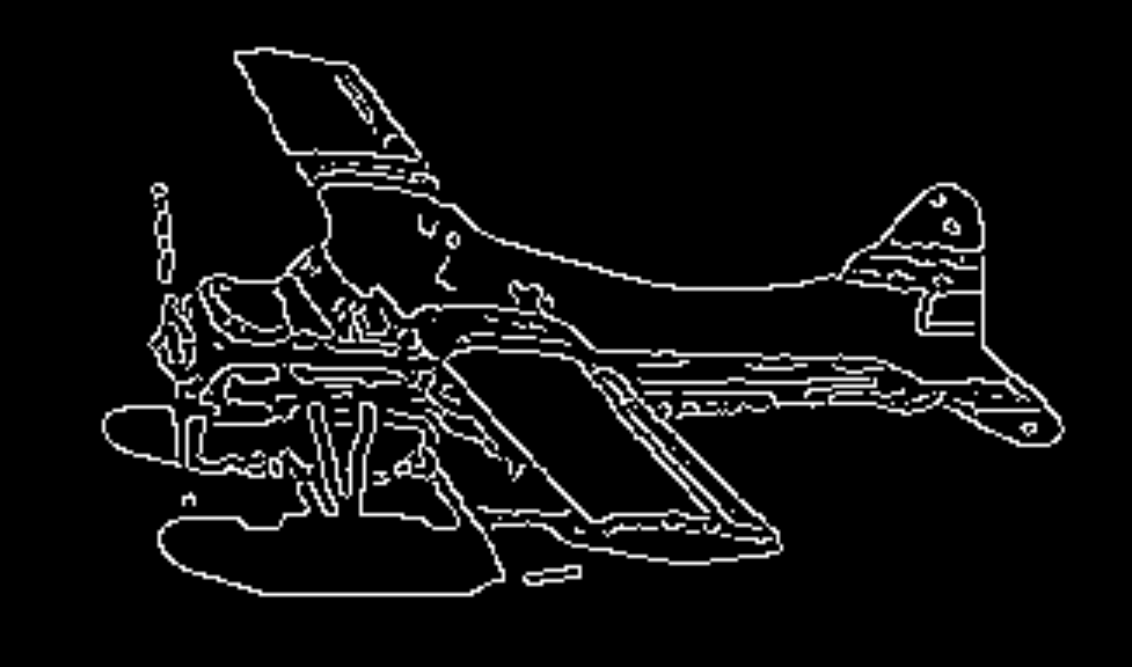}
\includegraphics[width=1.80cm]{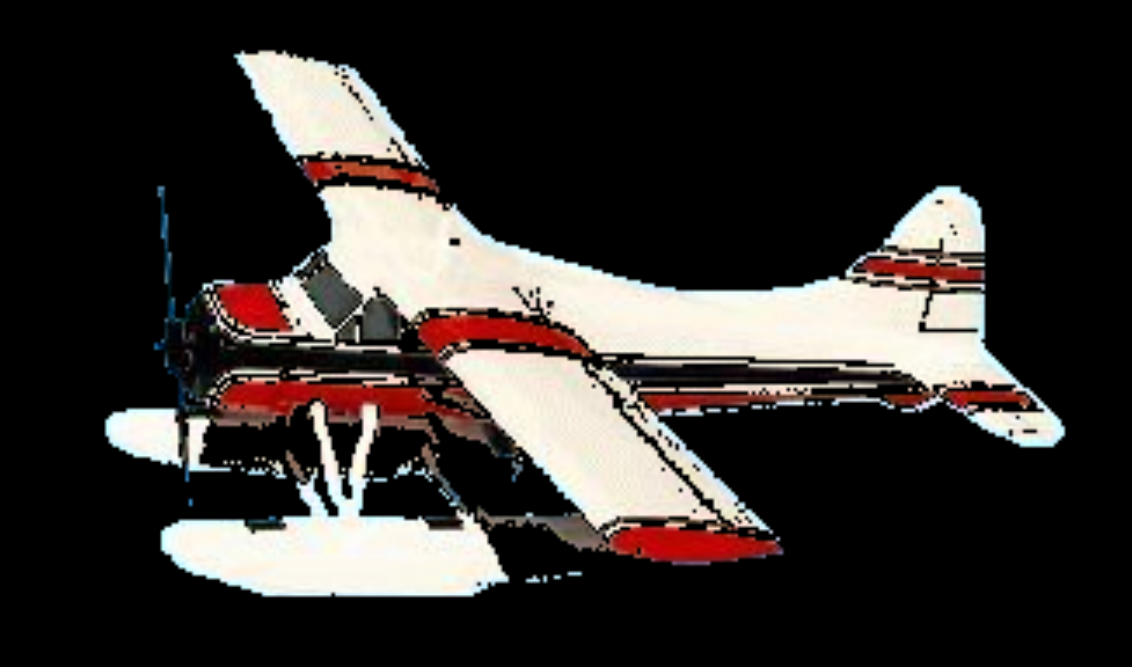}\\

\includegraphics[width=1.80cm]{aeroplane.pdf}
\includegraphics[width=1.80cm]{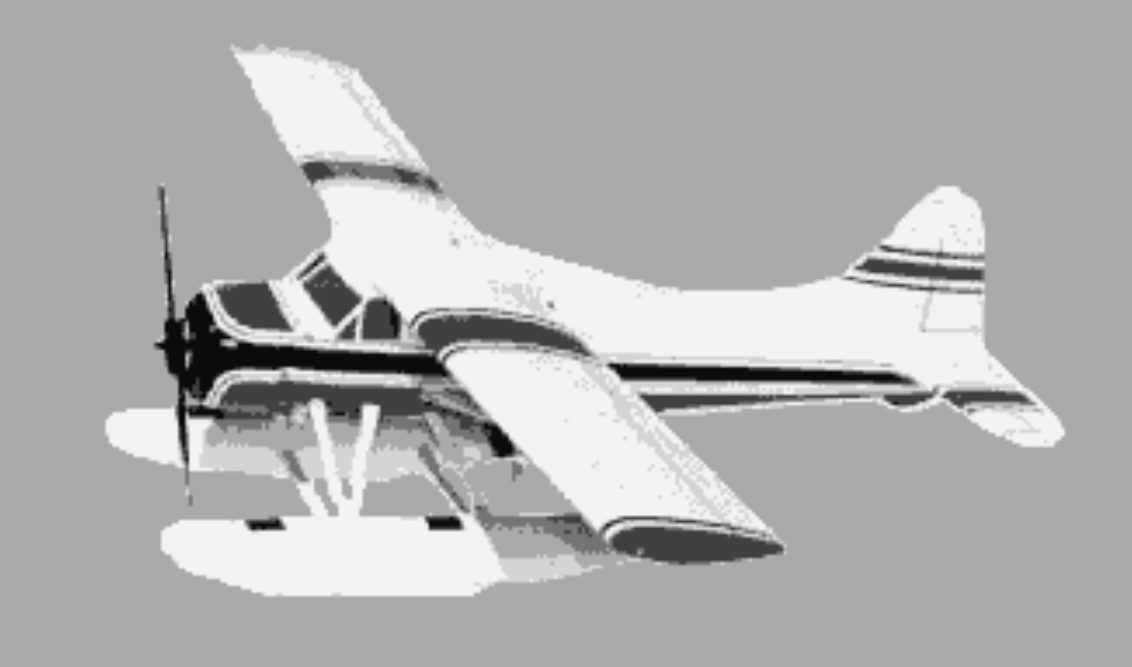}
\includegraphics[width=1.80cm]{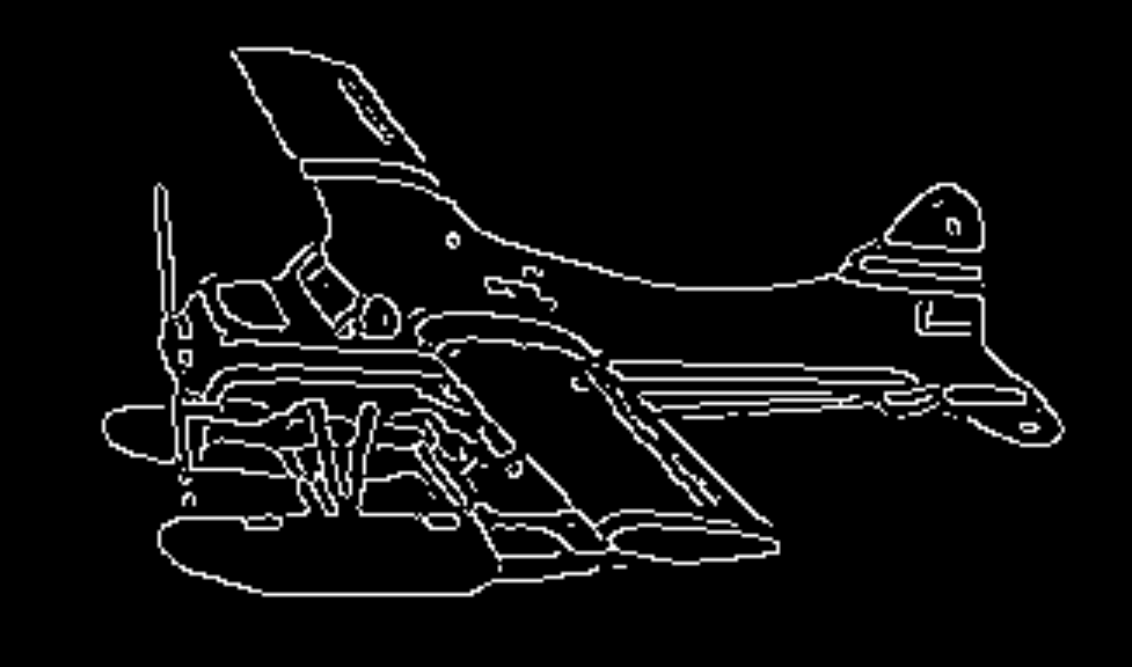}
\includegraphics[width=1.80cm]{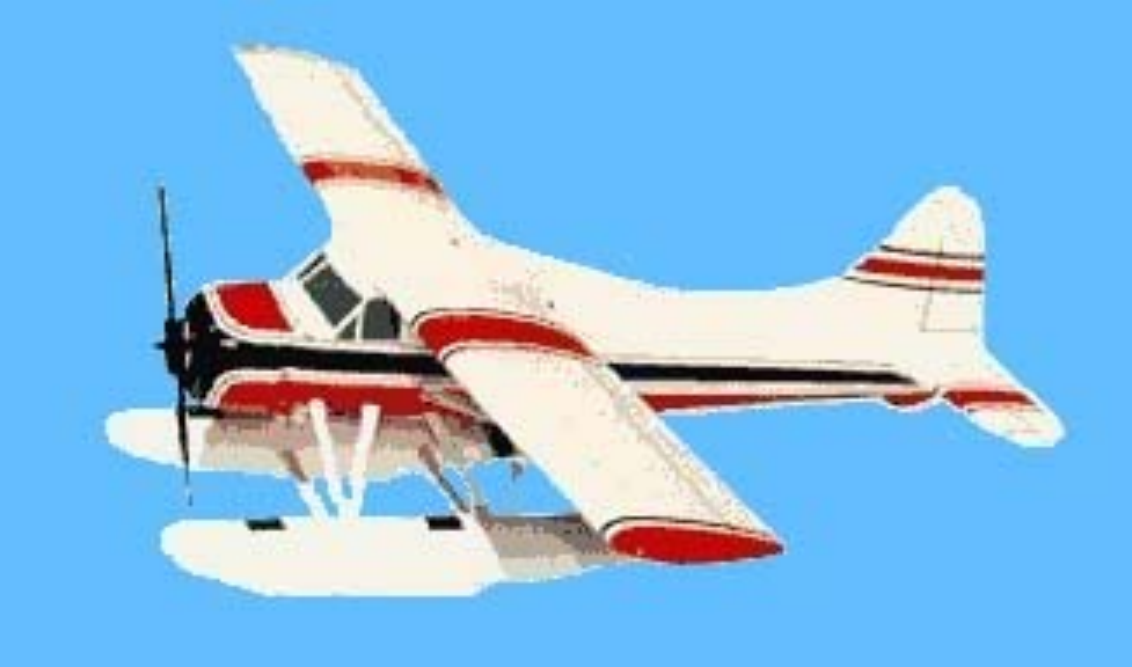}
\includegraphics[width=1.80cm]{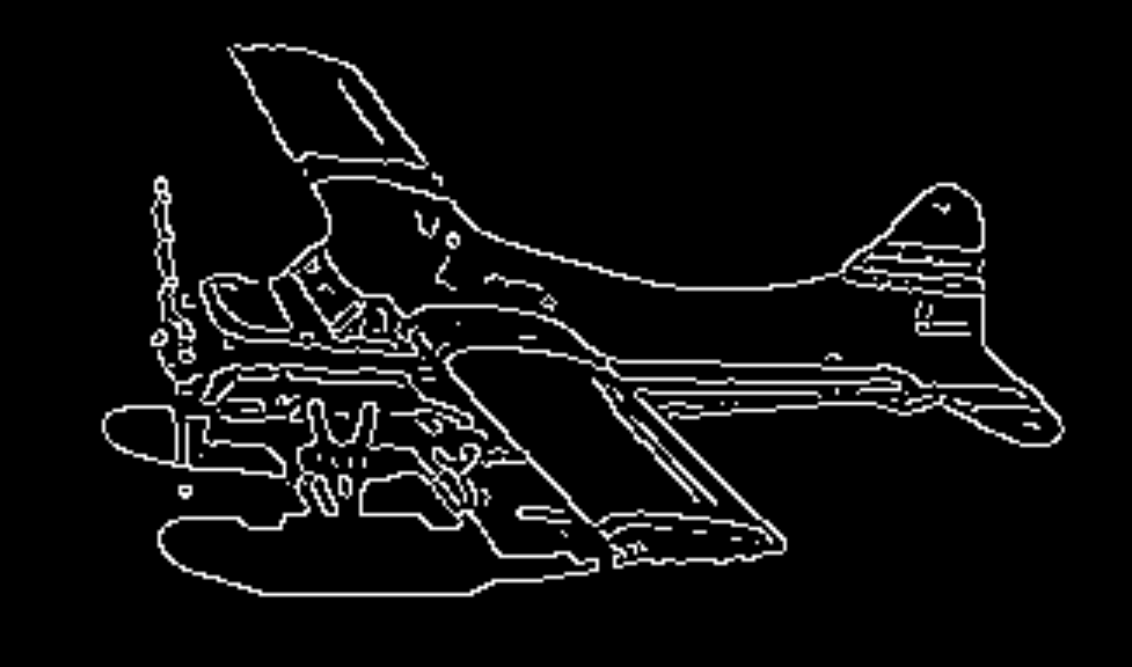}
\includegraphics[width=1.80cm]{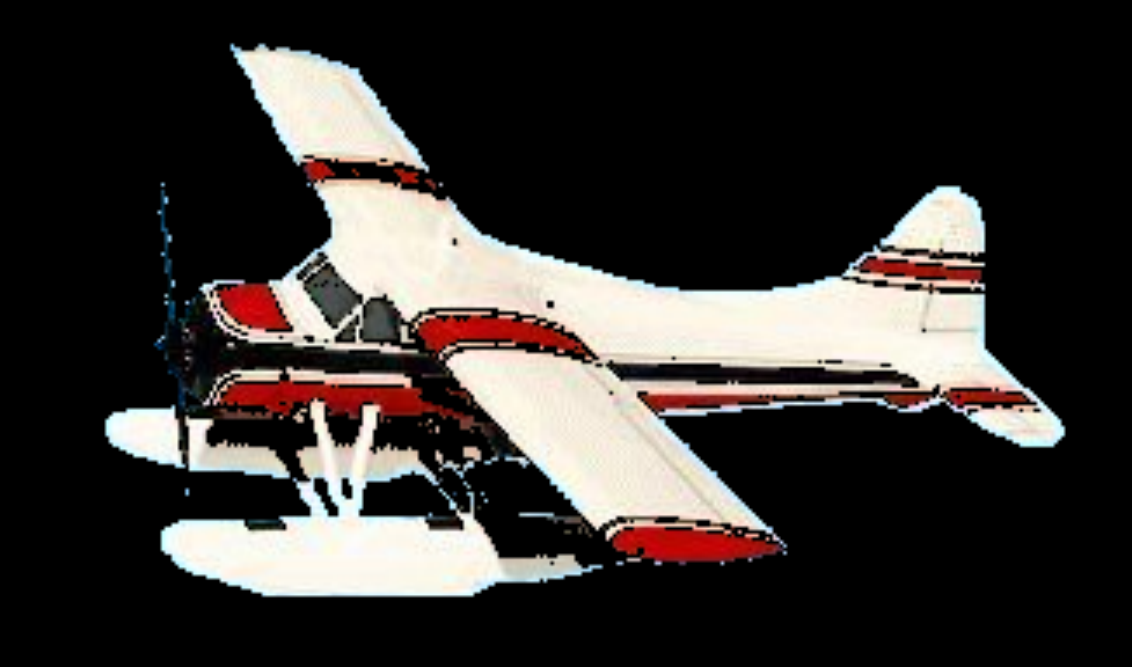}\\

\caption{Simulated output of Airplane. Row 1 - Methodology-I, Row 2 - Methodology-II, Row 3 - Otsu's method. Column 1 - Orginal Image, Column 2 - Segmented Y component, Column 3 - Canny edge detection of segmented Y component, Column 4 - Segmented color image, Column 5 - Canny edge detection of extracted Y component, Column 6 - Extracted image.}
\label{fig:2}
\end{center}
\end{figure*}

\begin{figure*}[ht]
\begin{center}
\includegraphics[width=1.80cm]{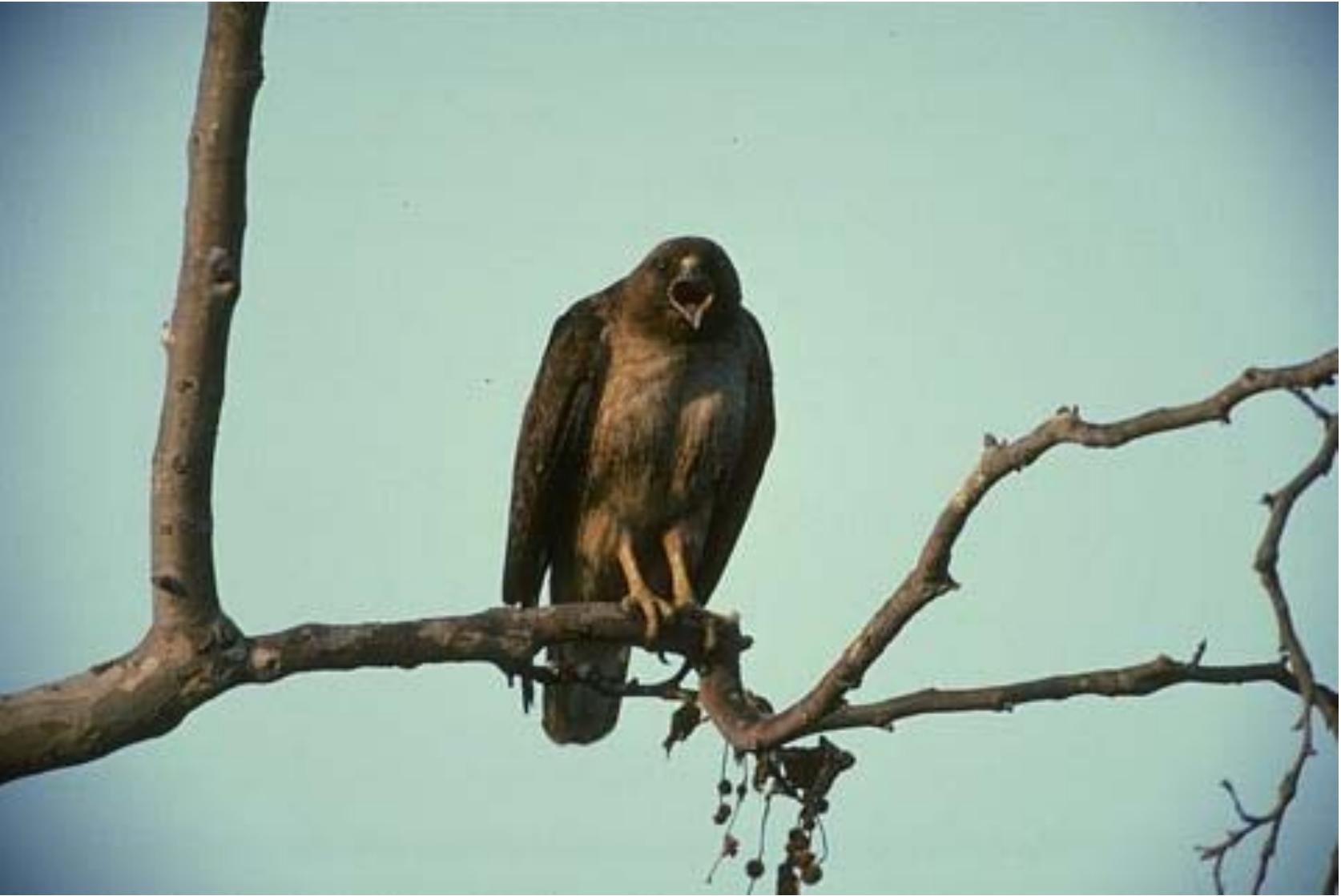}
\includegraphics[width=1.80cm]{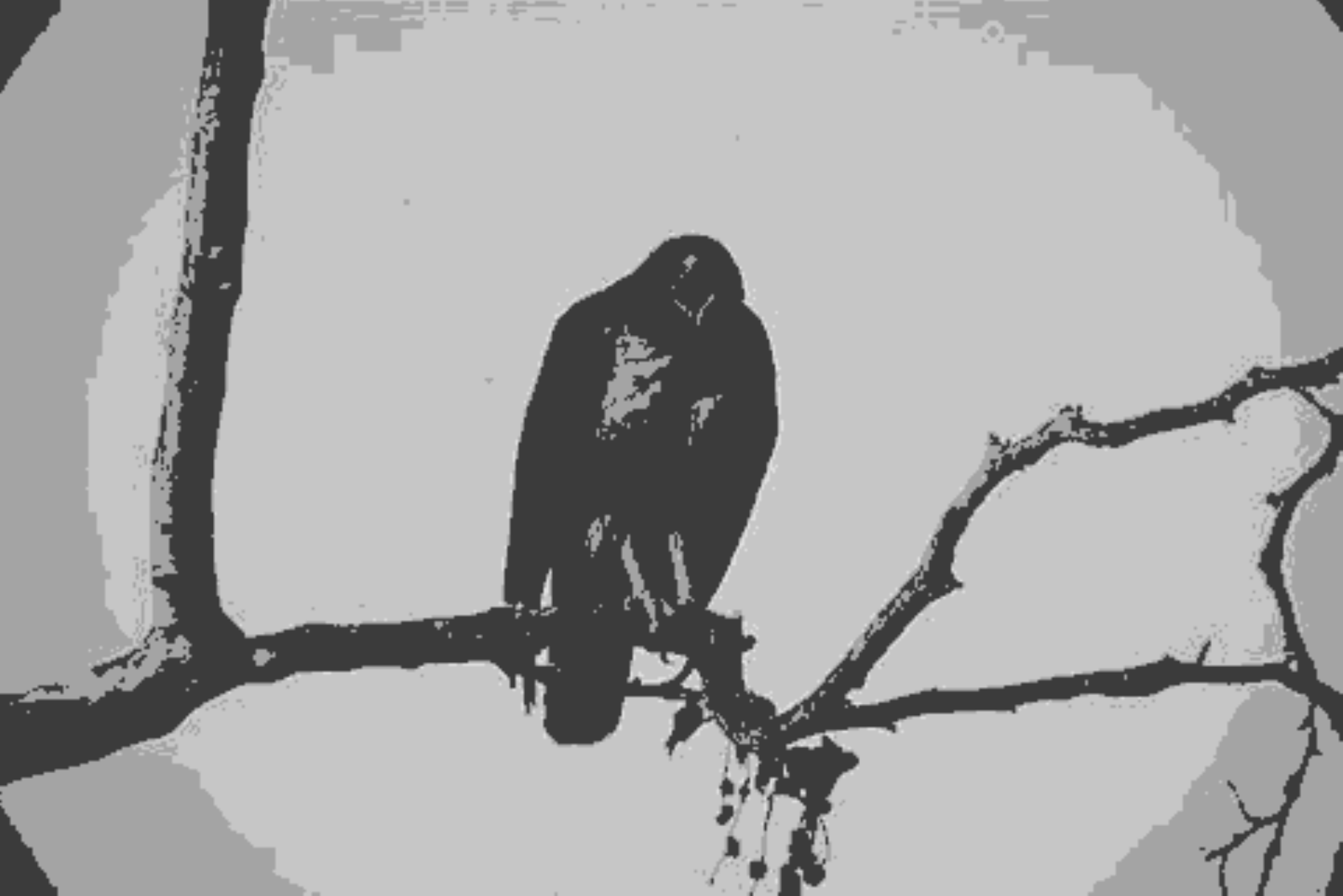}
\includegraphics[width=1.80cm]{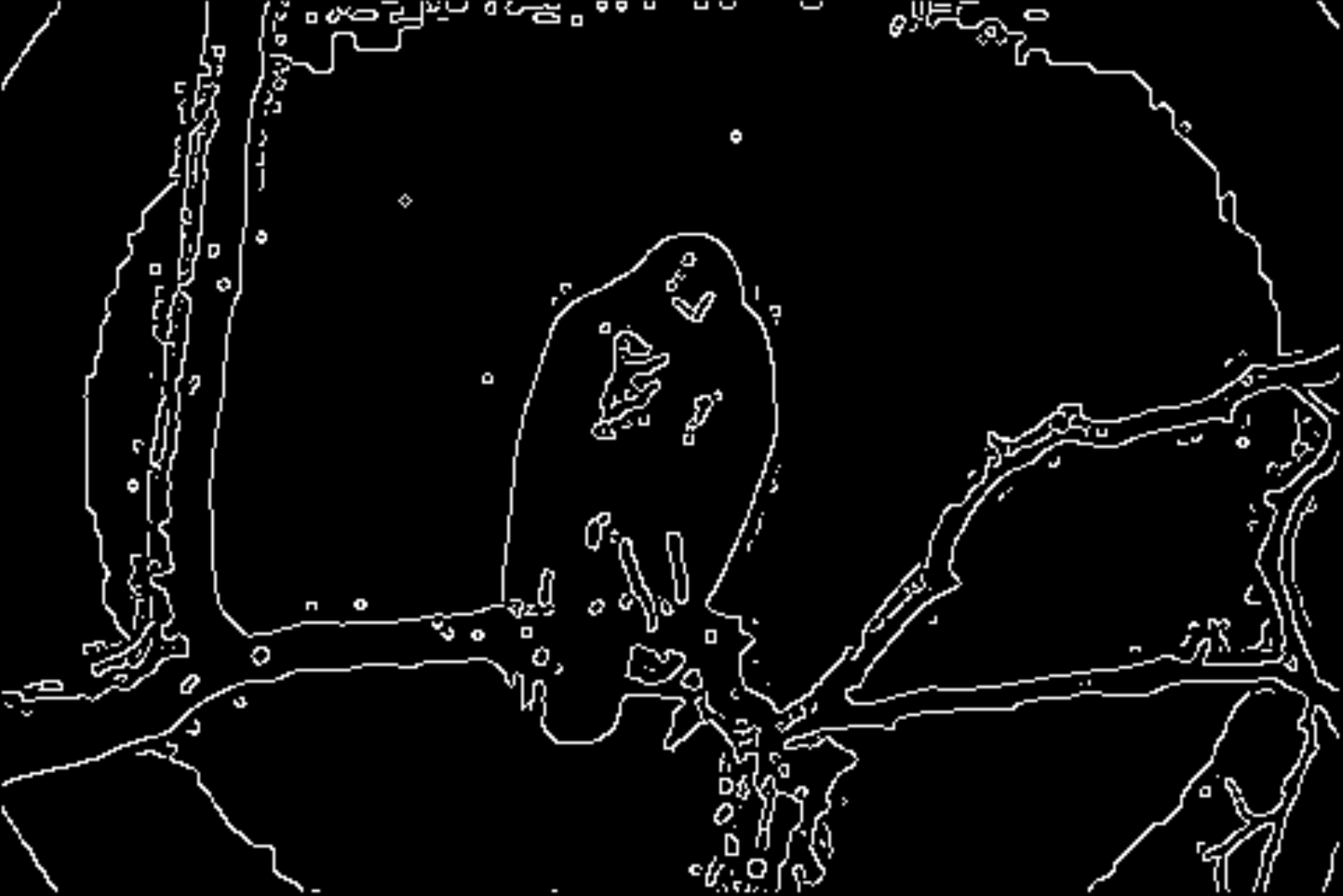}
\includegraphics[width=1.80cm]{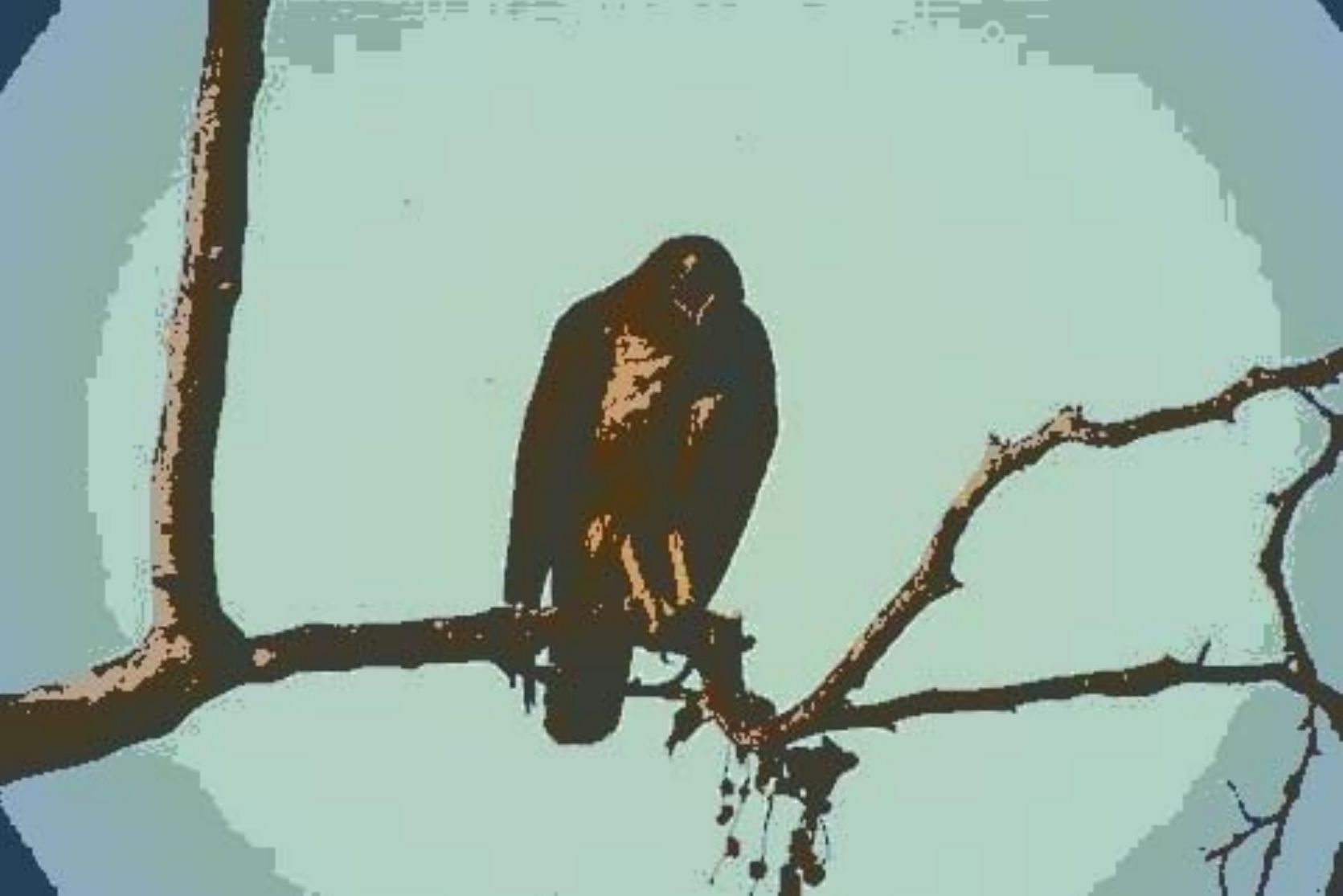}
\includegraphics[width=1.80cm]{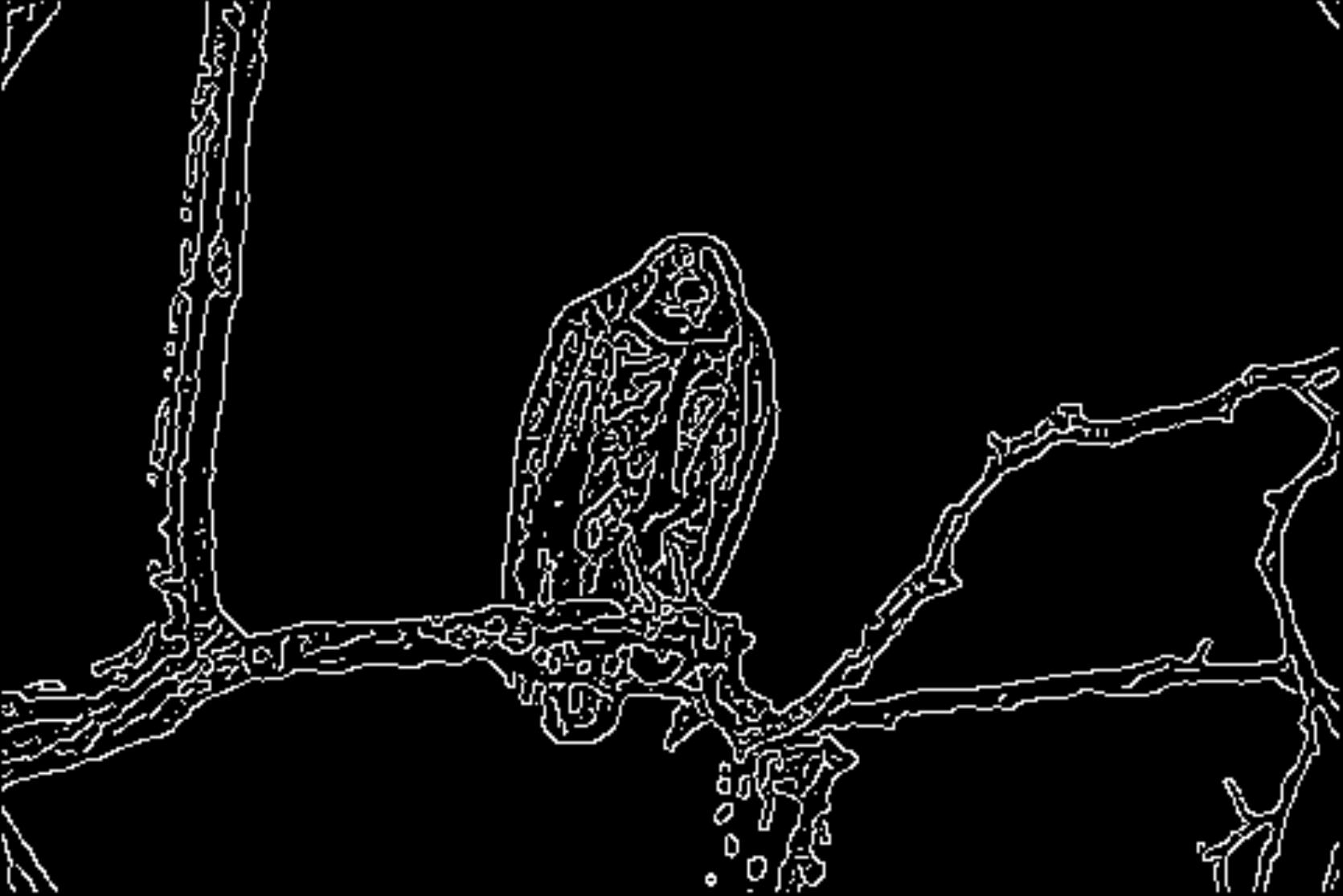}
\includegraphics[width=1.80cm]{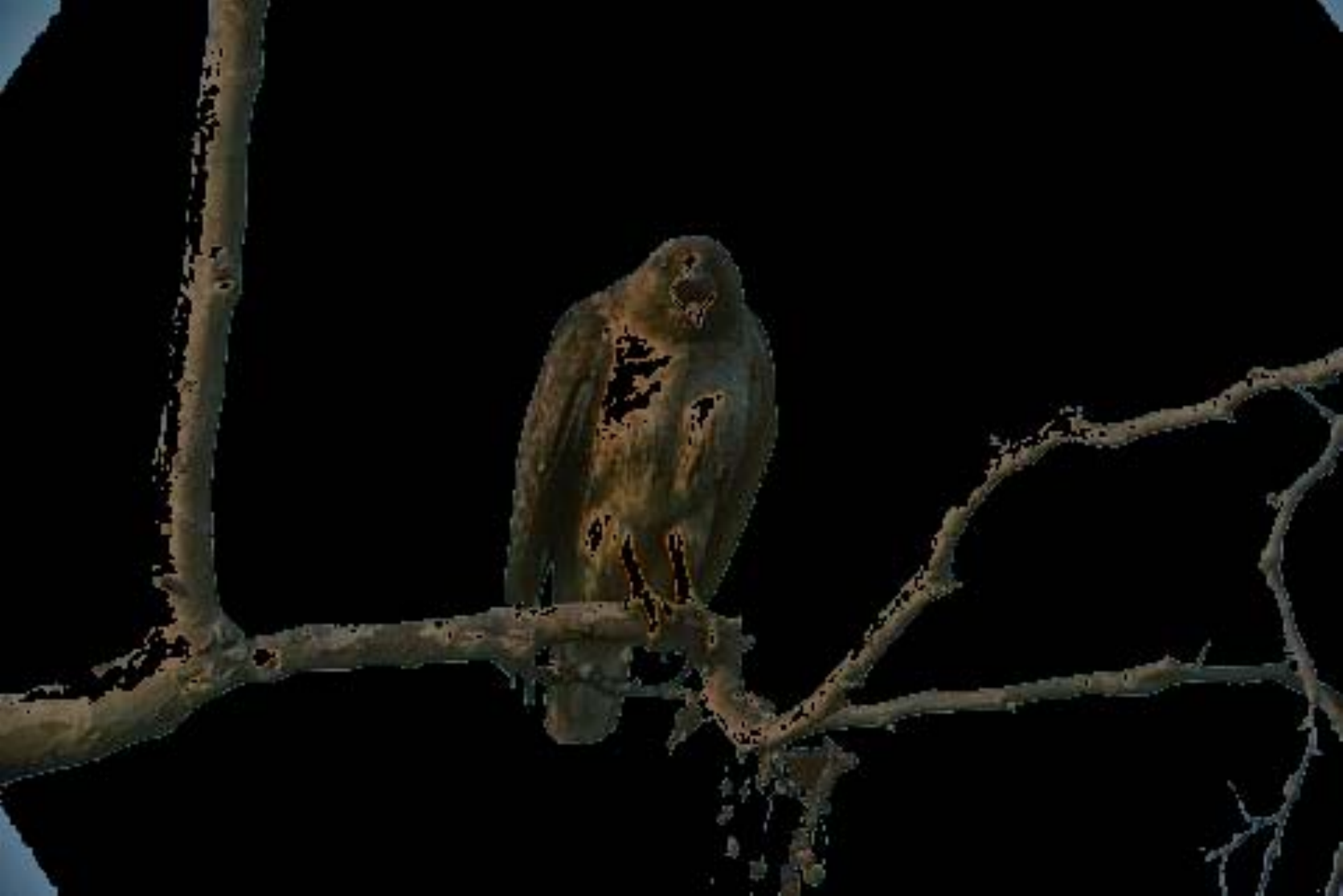}\\

\includegraphics[width=1.80cm]{Eagle.pdf}
\includegraphics[width=1.80cm]{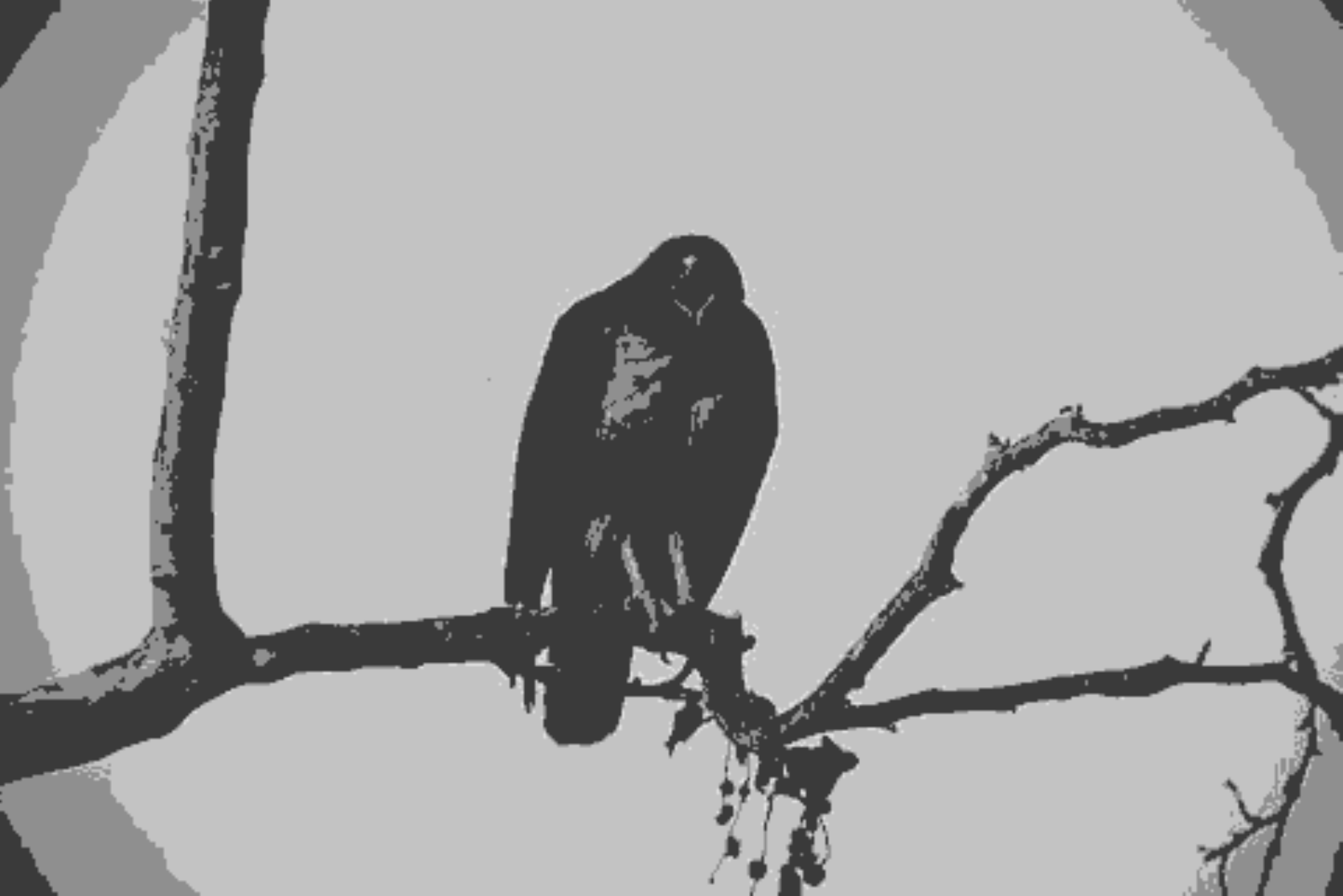}
\includegraphics[width=1.80cm]{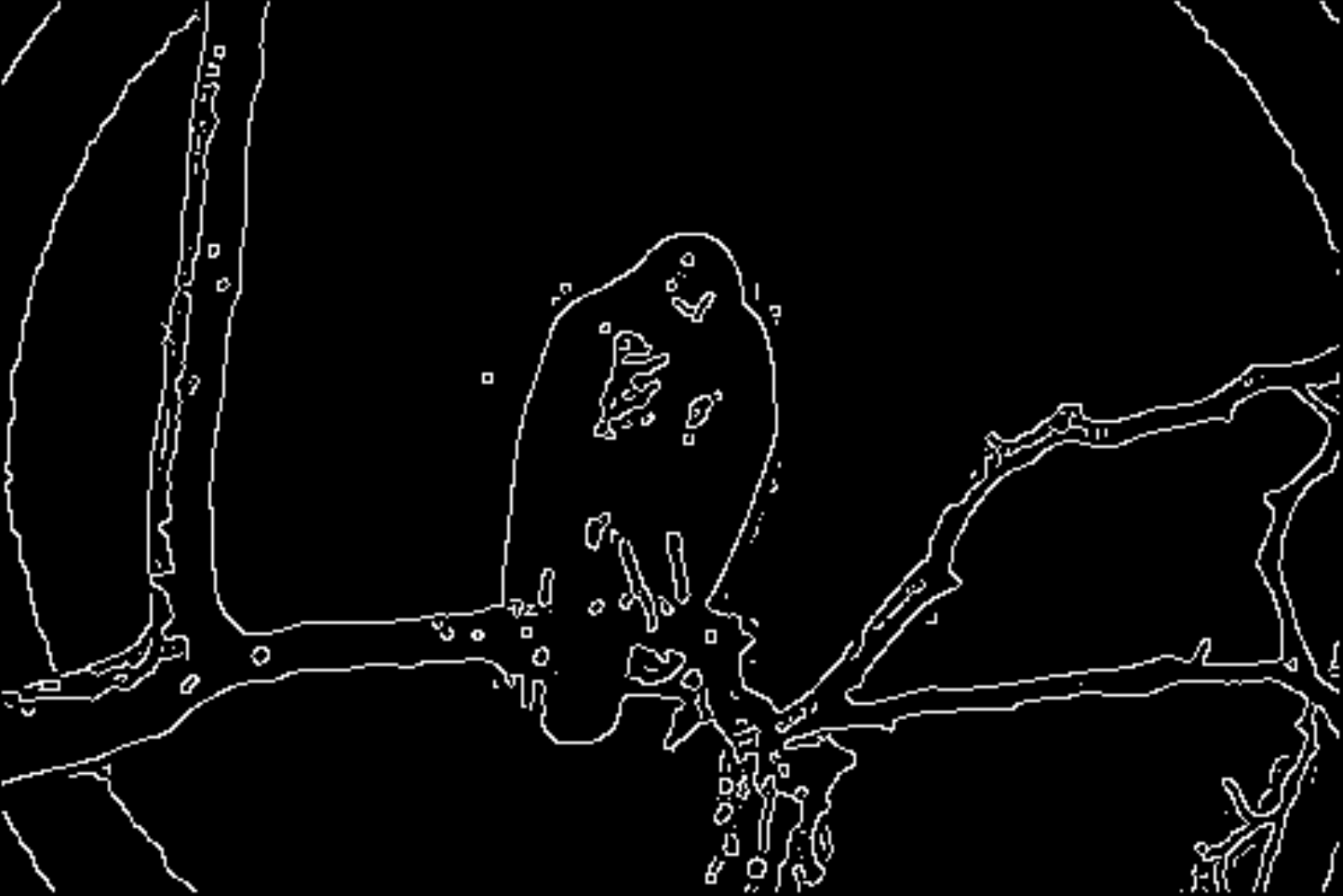}
\includegraphics[width=1.80cm]{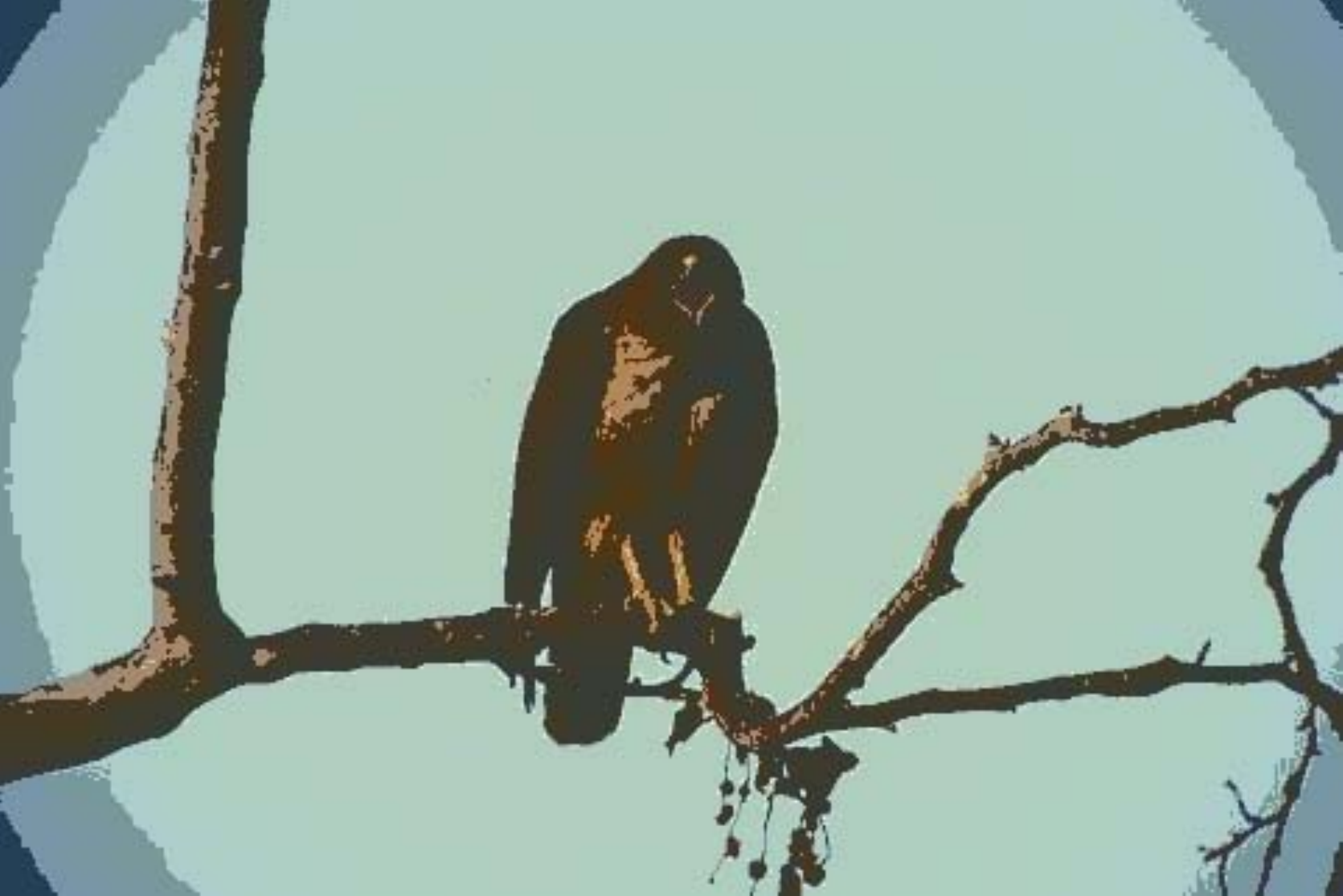}
\includegraphics[width=1.80cm]{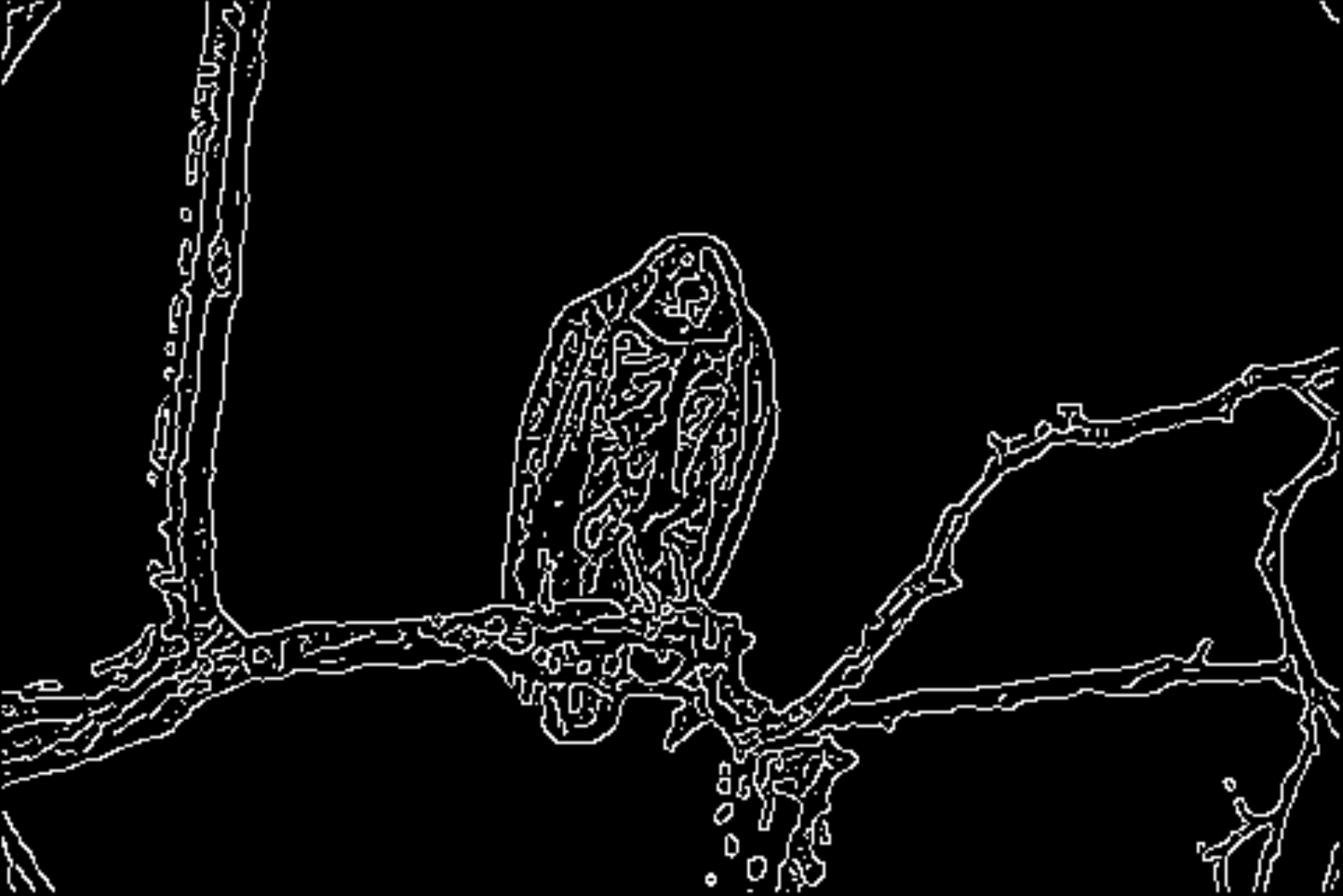}
\includegraphics[width=1.80cm]{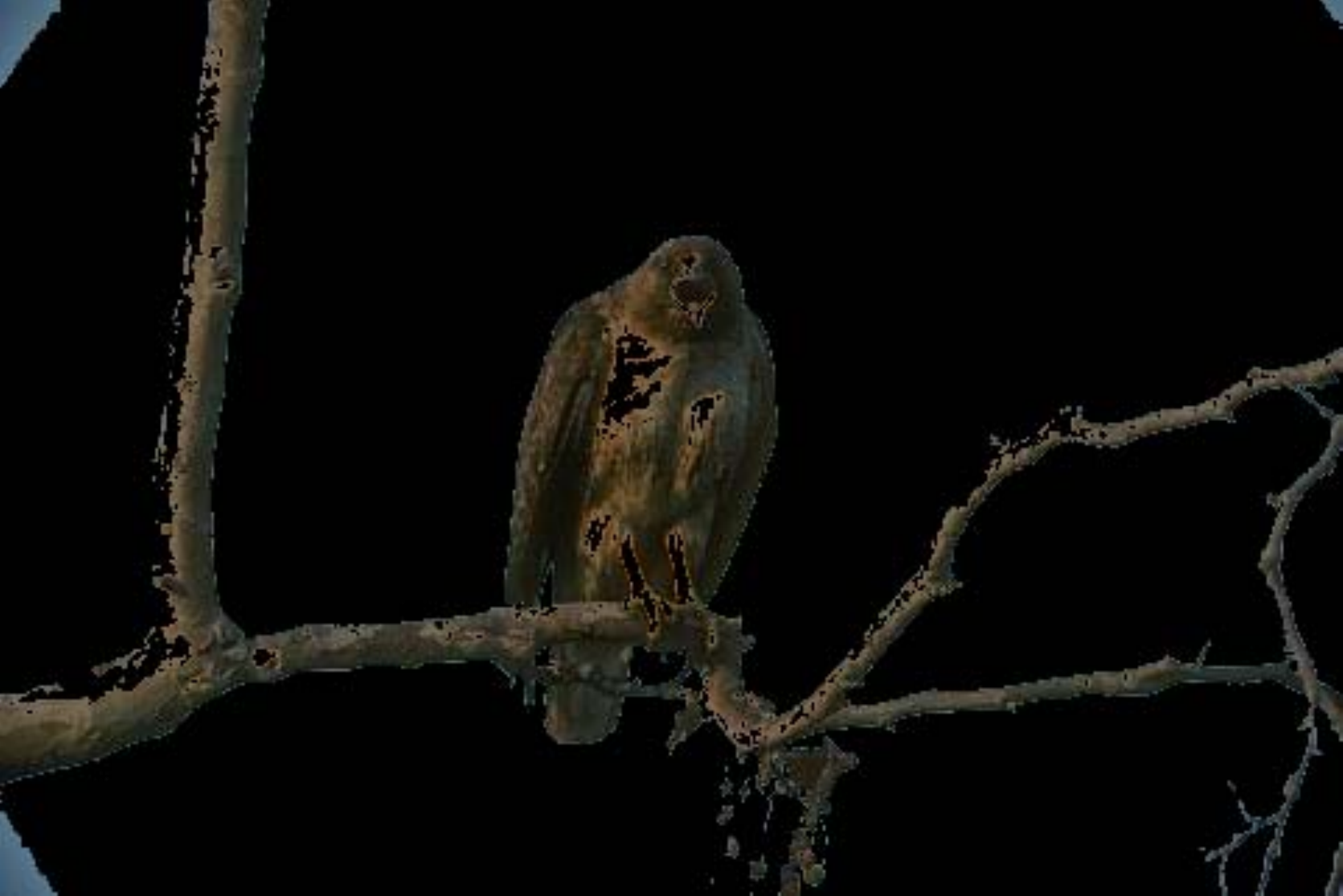}\\

\includegraphics[width=1.80cm]{Eagle.pdf}
\includegraphics[width=1.80cm]{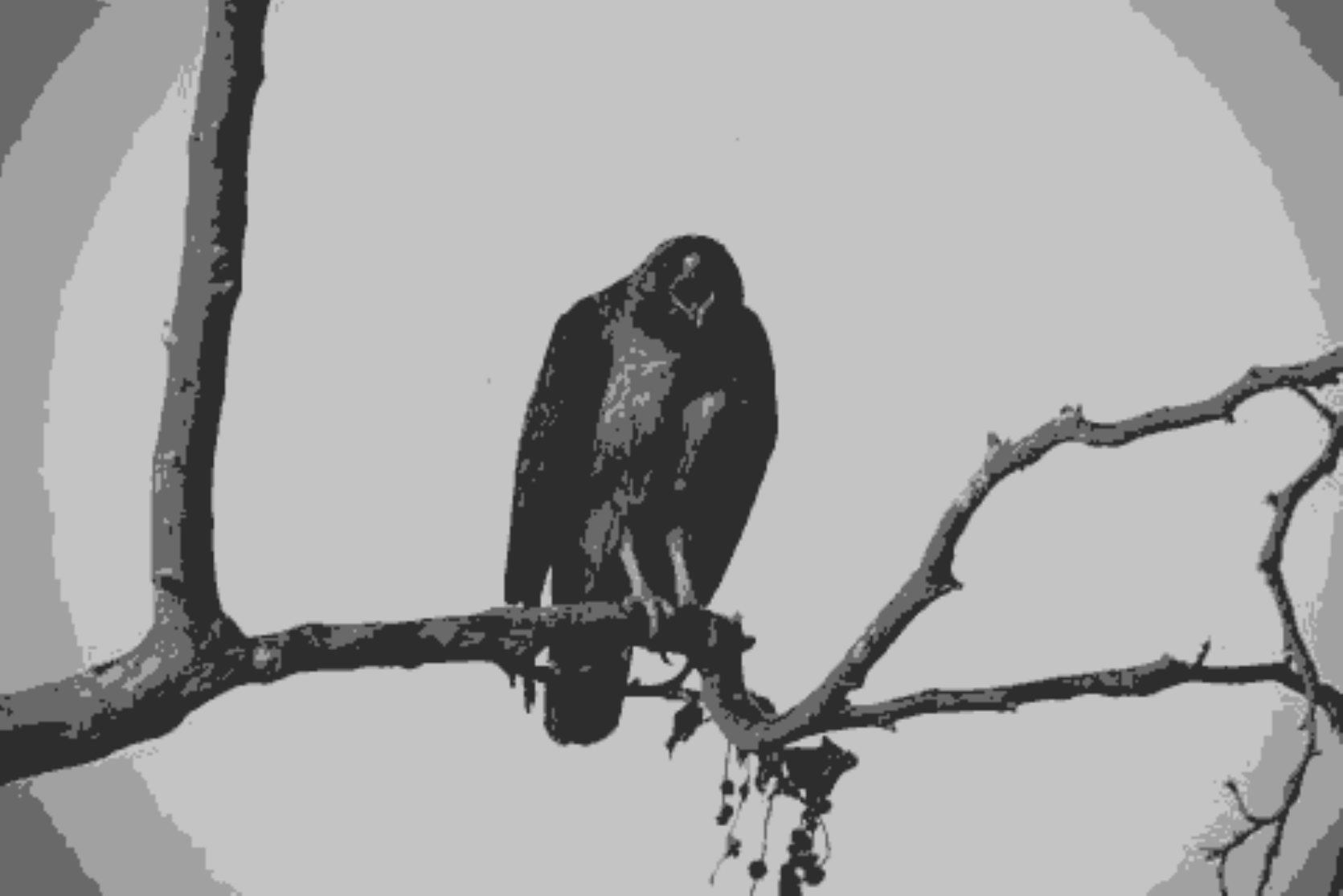}
\includegraphics[width=1.80cm]{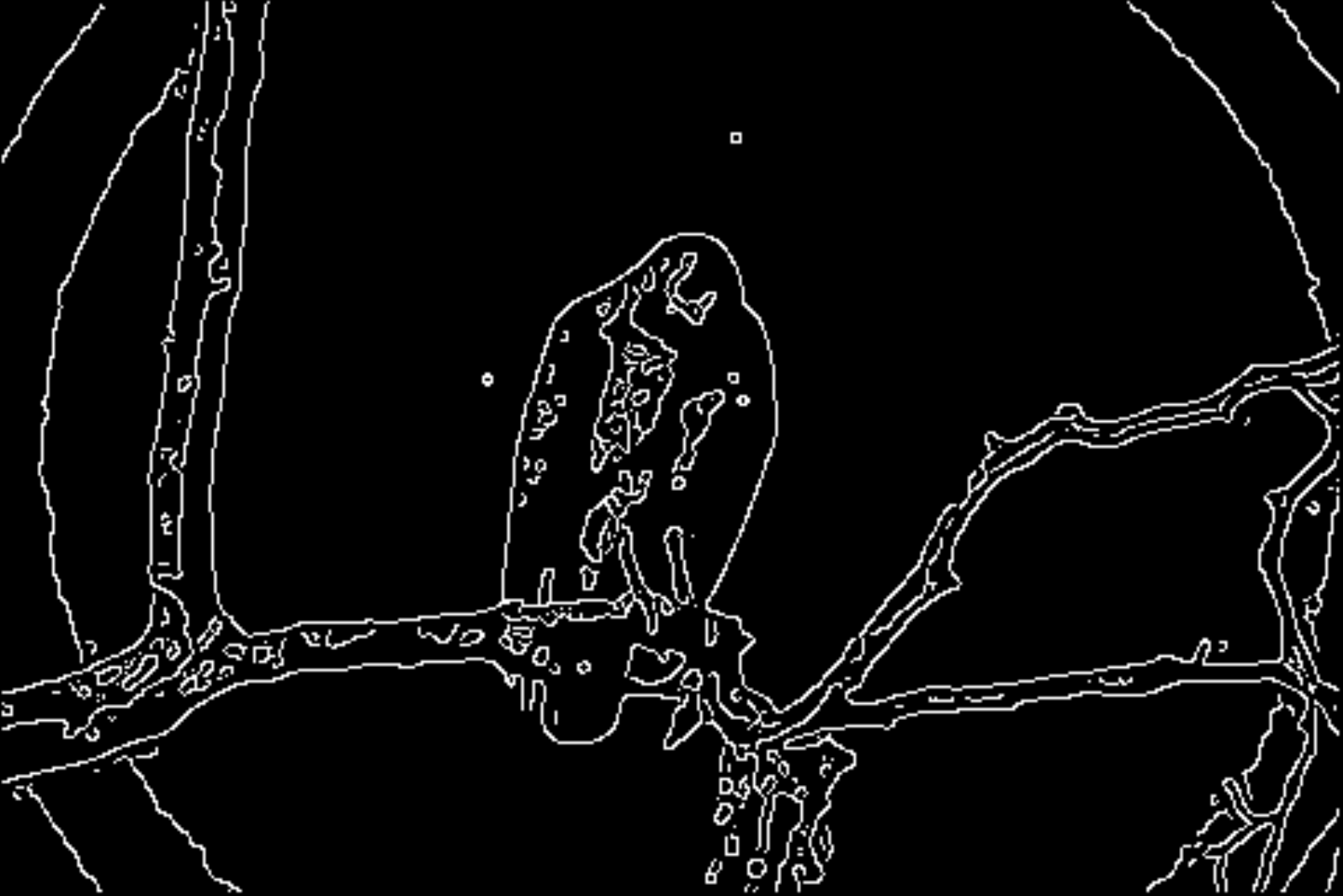}
\includegraphics[width=1.80cm]{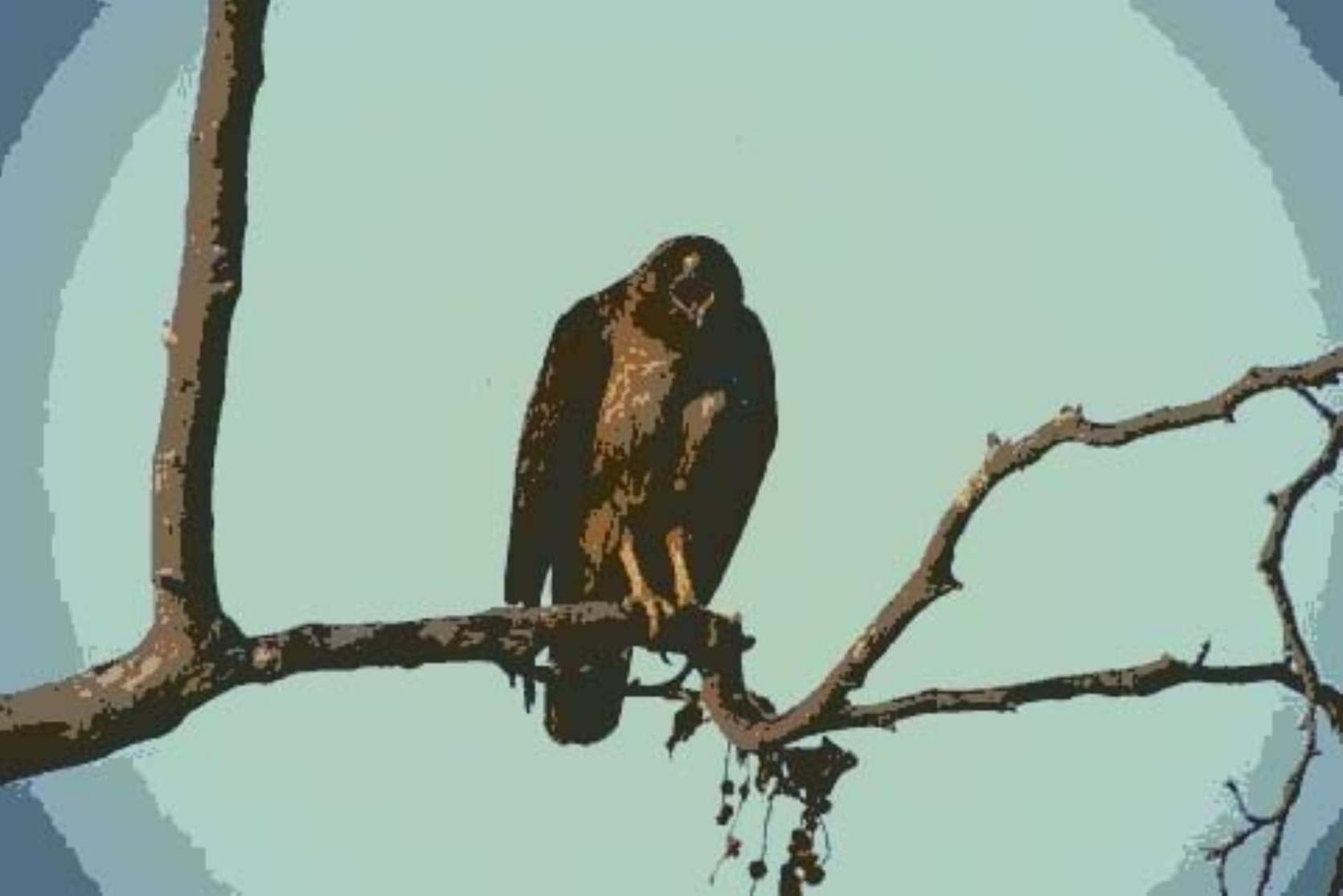}
\includegraphics[width=1.80cm]{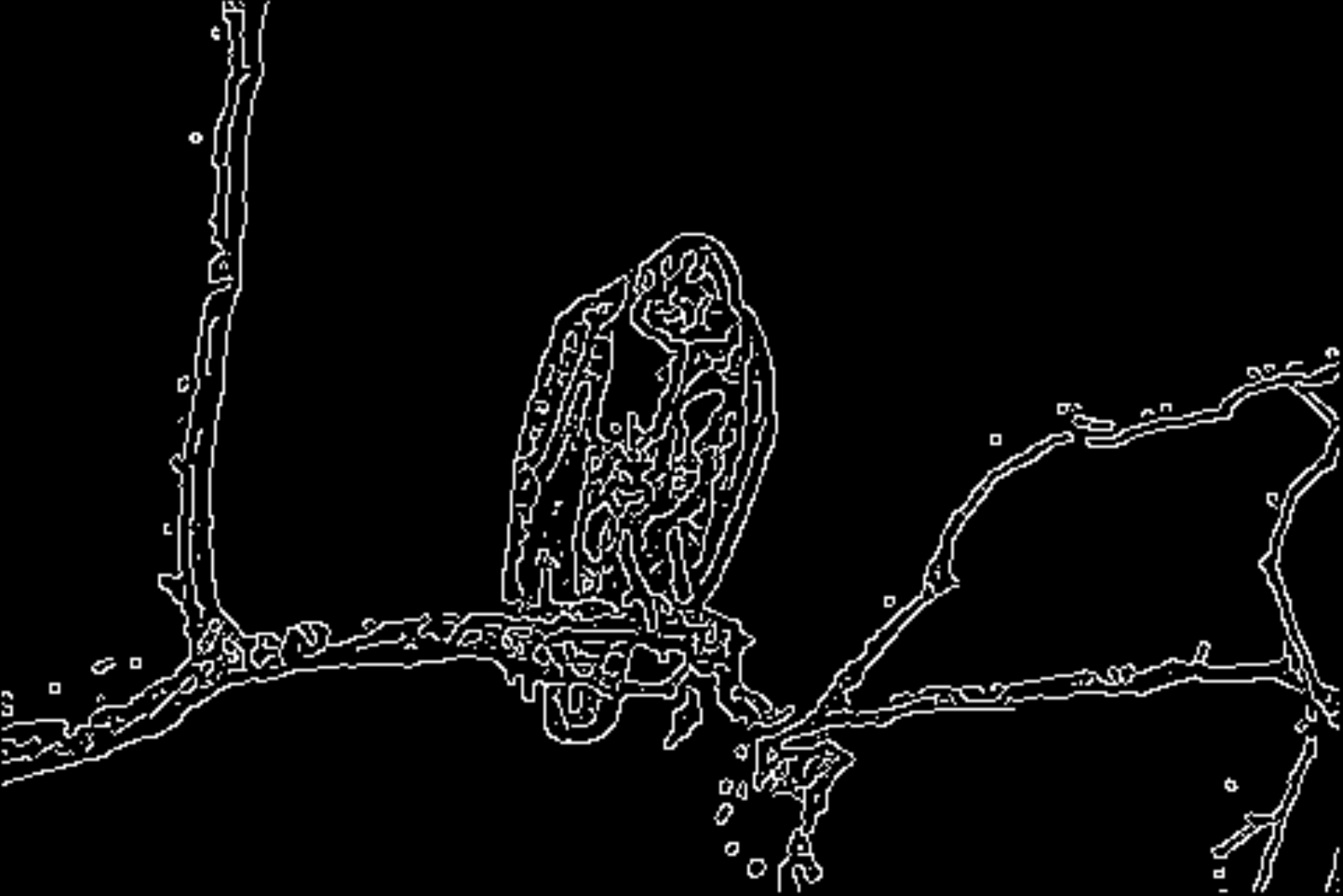}
\includegraphics[width=1.80cm]{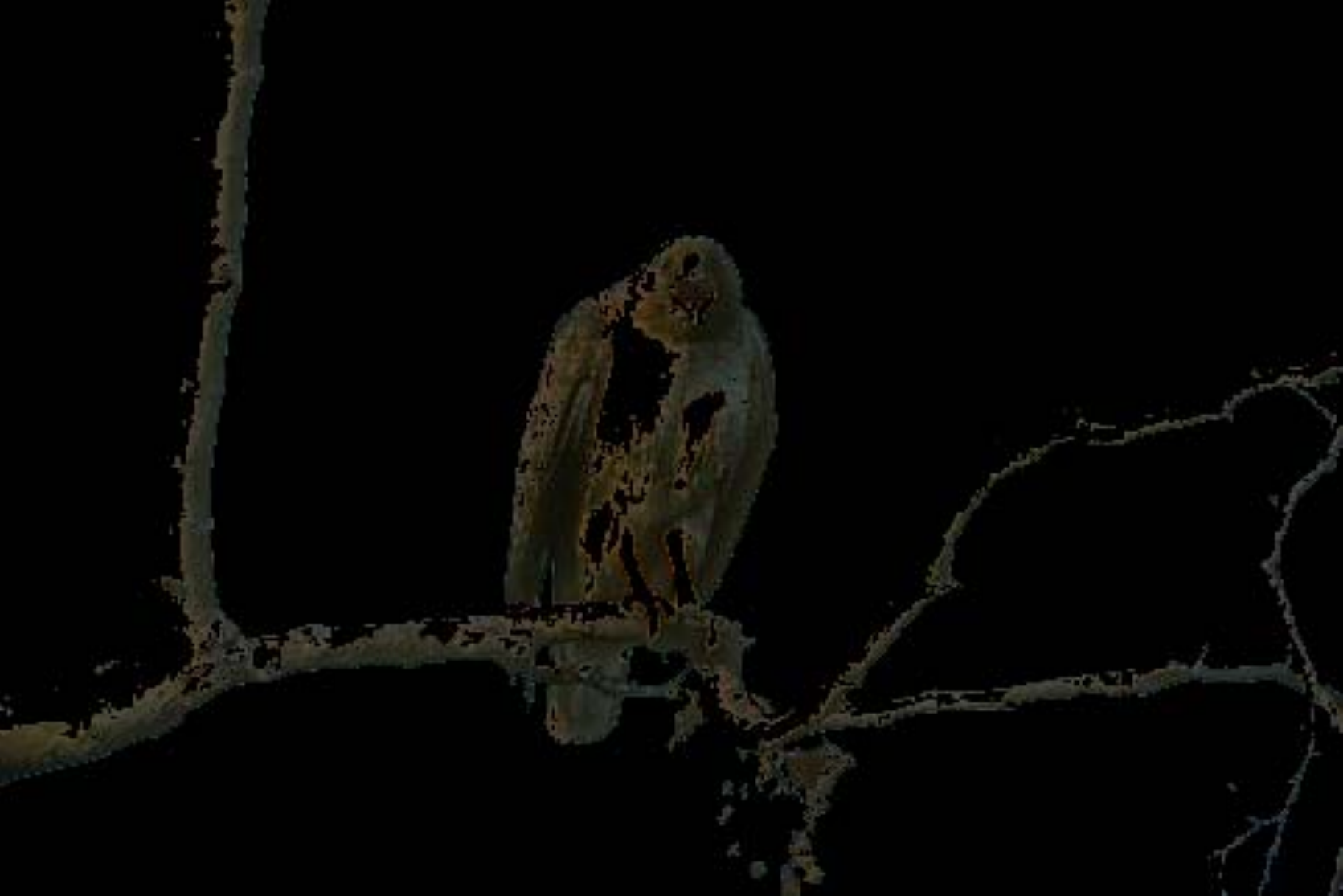}\\
\caption{Simulated output of Eagle. Row 1 - Methodology-I, Row 2 - Methodology-II, Row 3 - Otsu's method. Column 1 - Orginal Image, Column 2 - Segmented Y component, Column 3 - Canny edge detection of segmented Y component, Column 4 - Segmented color image, Column 5 - Canny edge detection of extracted Y component, Column 6 - Extracted image.}

\label{fig:3}
\end{center}
\end{figure*}

\begin{figure*}[ht]
\begin{center}

\includegraphics[width=1.80cm]{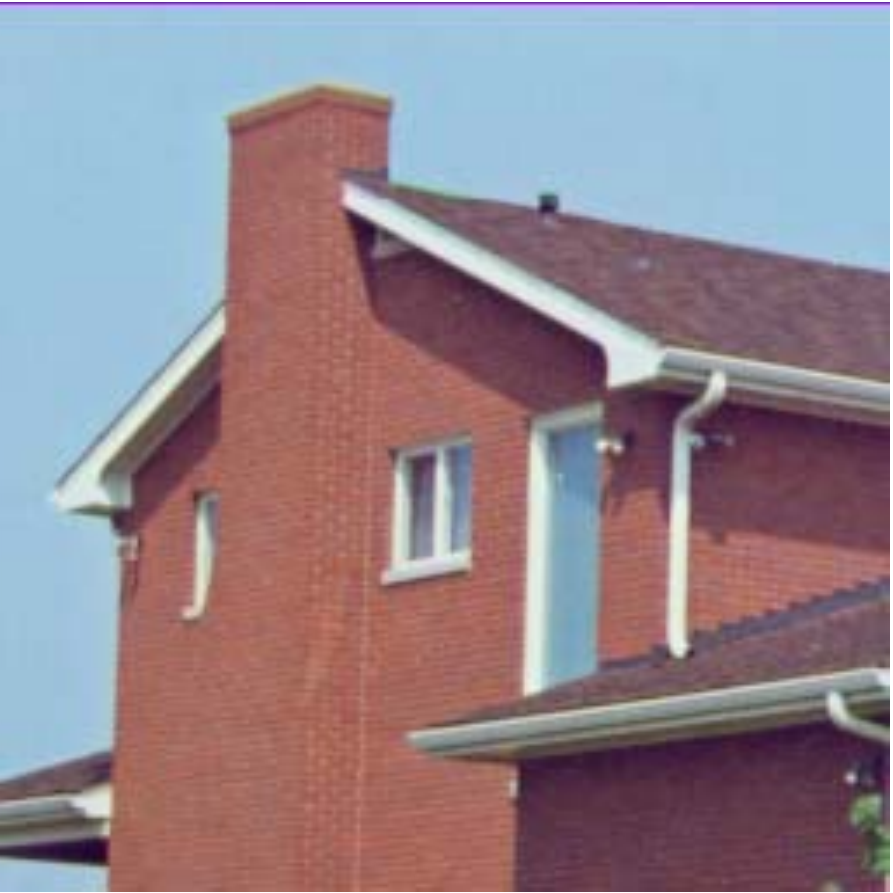}
\includegraphics[width=1.80cm]{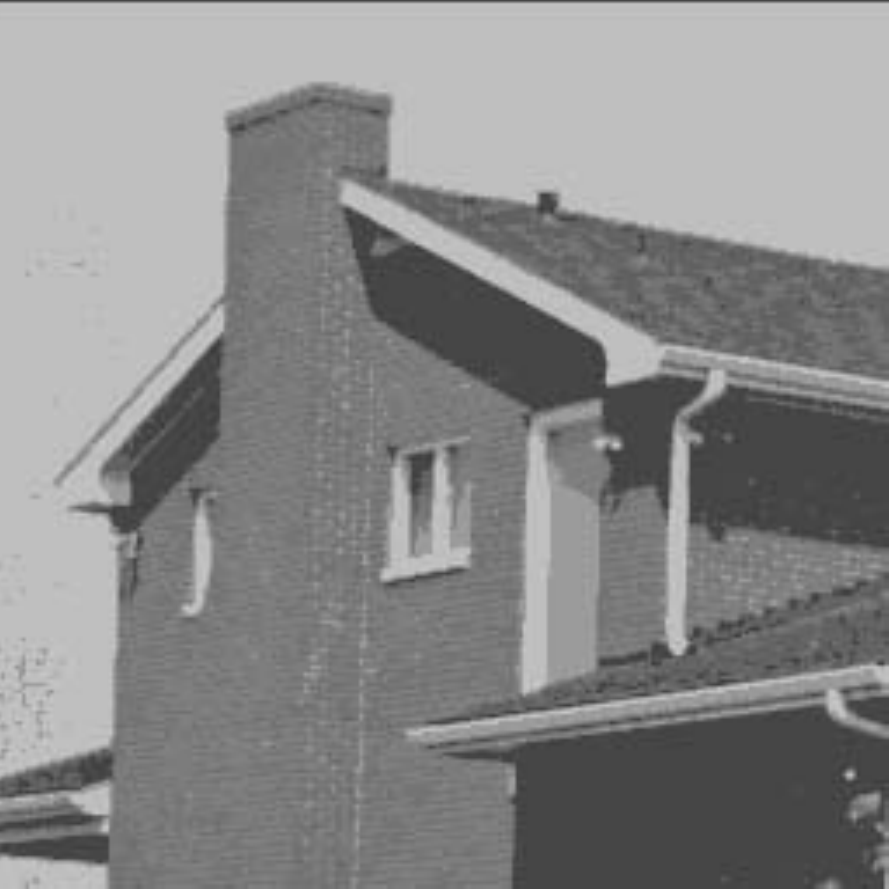}
\includegraphics[width=1.80cm]{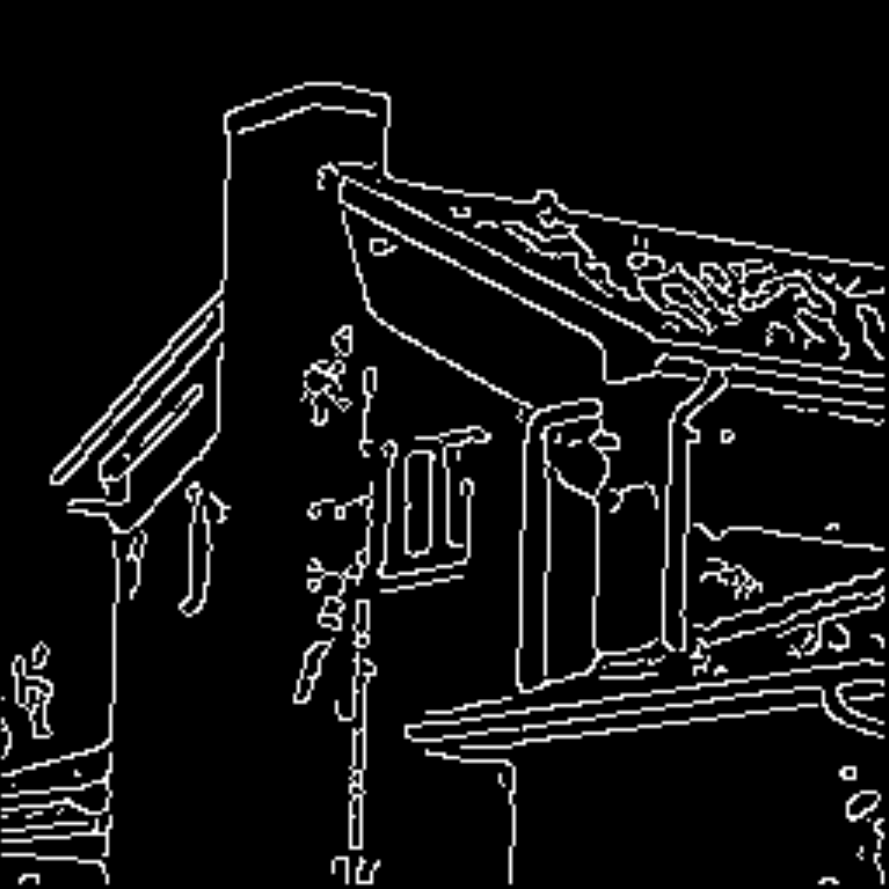}
\includegraphics[width=1.80cm]{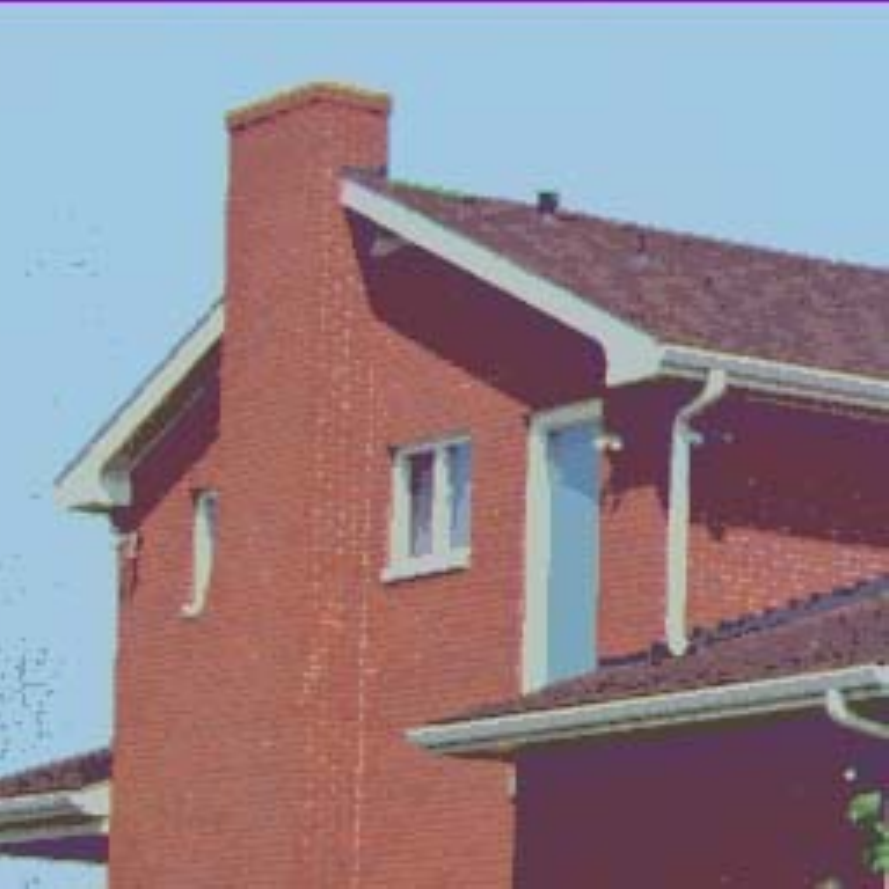}
\includegraphics[width=1.80cm]{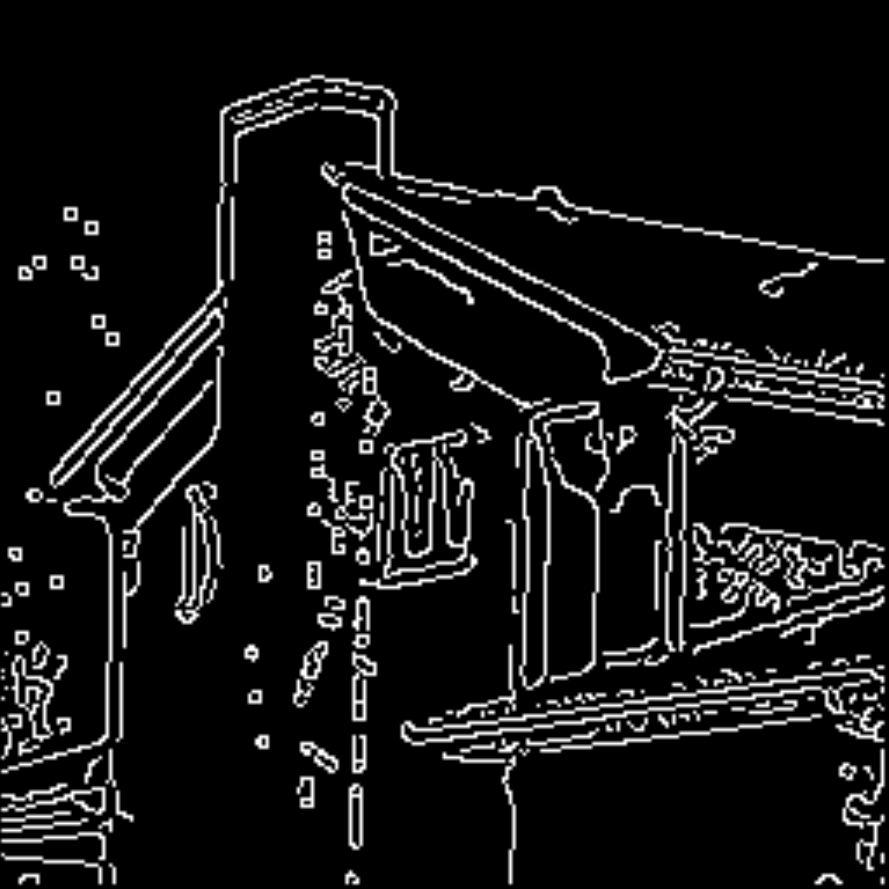}
\includegraphics[width=1.80cm]{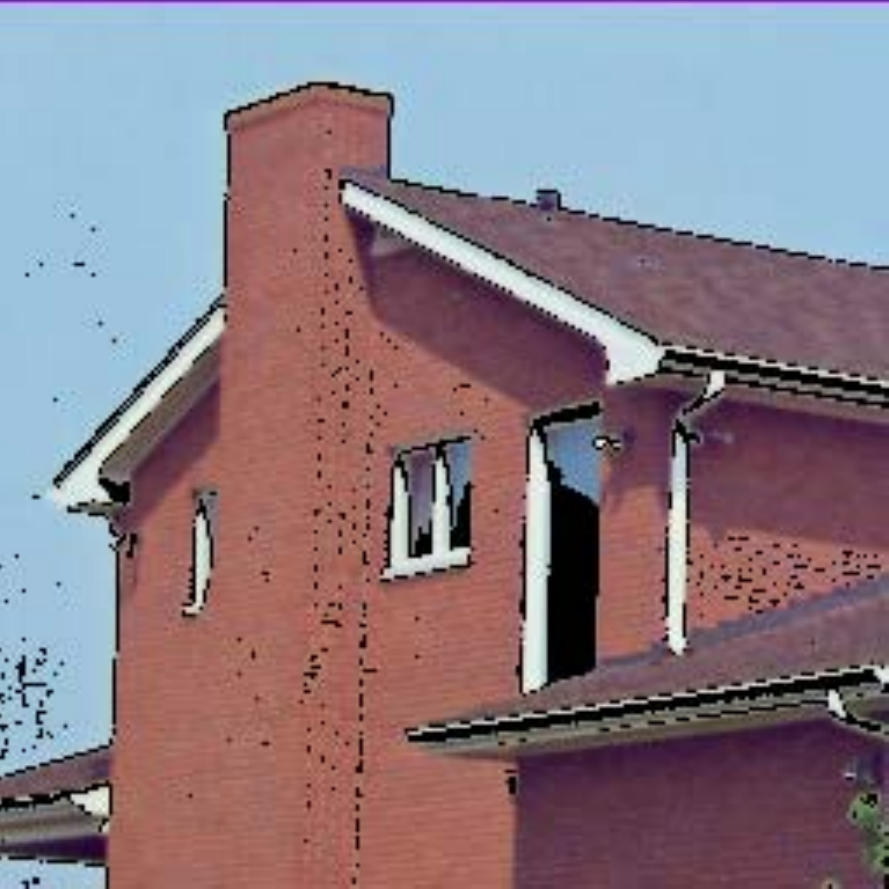}\\

\includegraphics[width=1.80cm]{Houses.pdf}
\includegraphics[width=1.80cm]{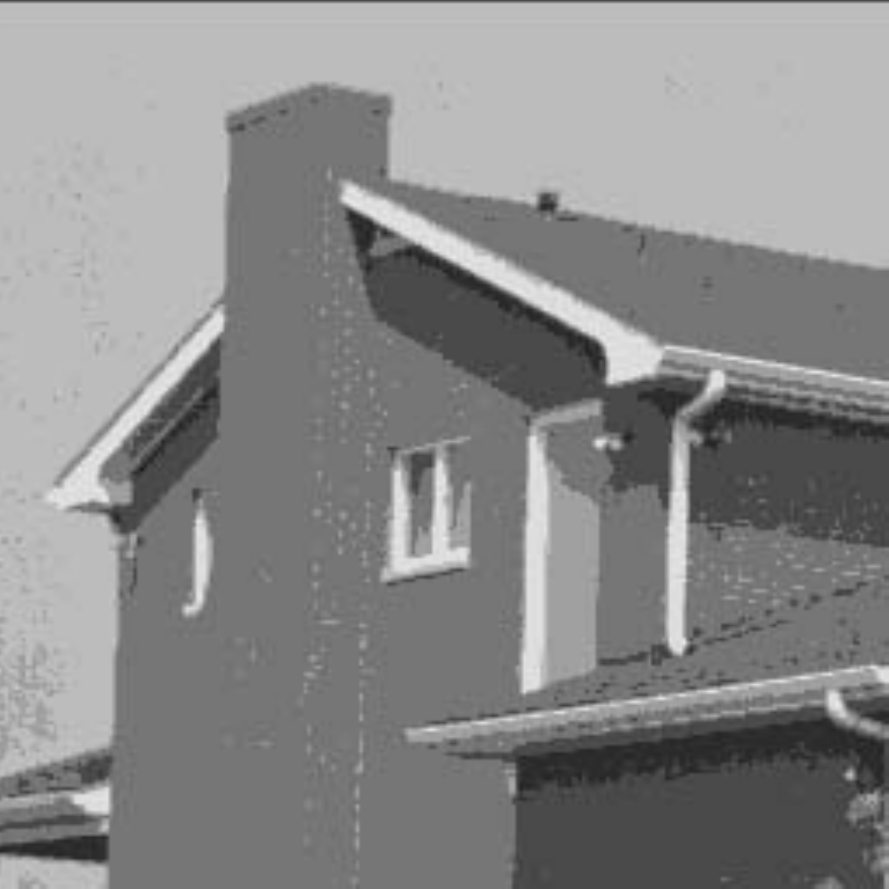}
\includegraphics[width=1.80cm]{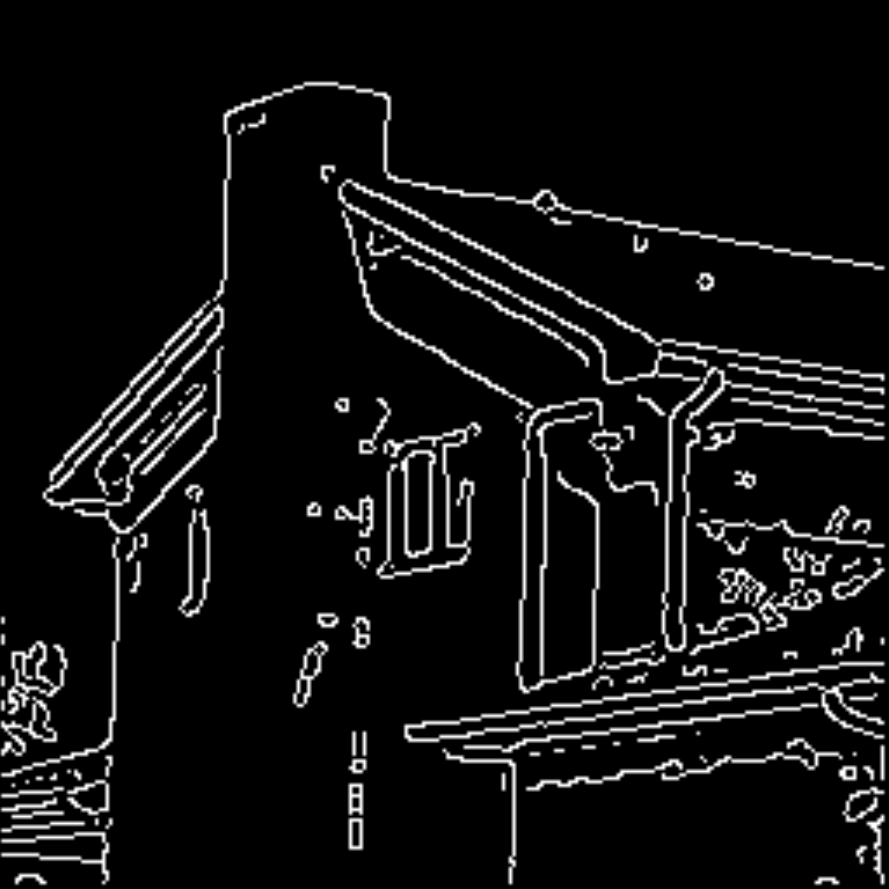}
\includegraphics[width=1.80cm]{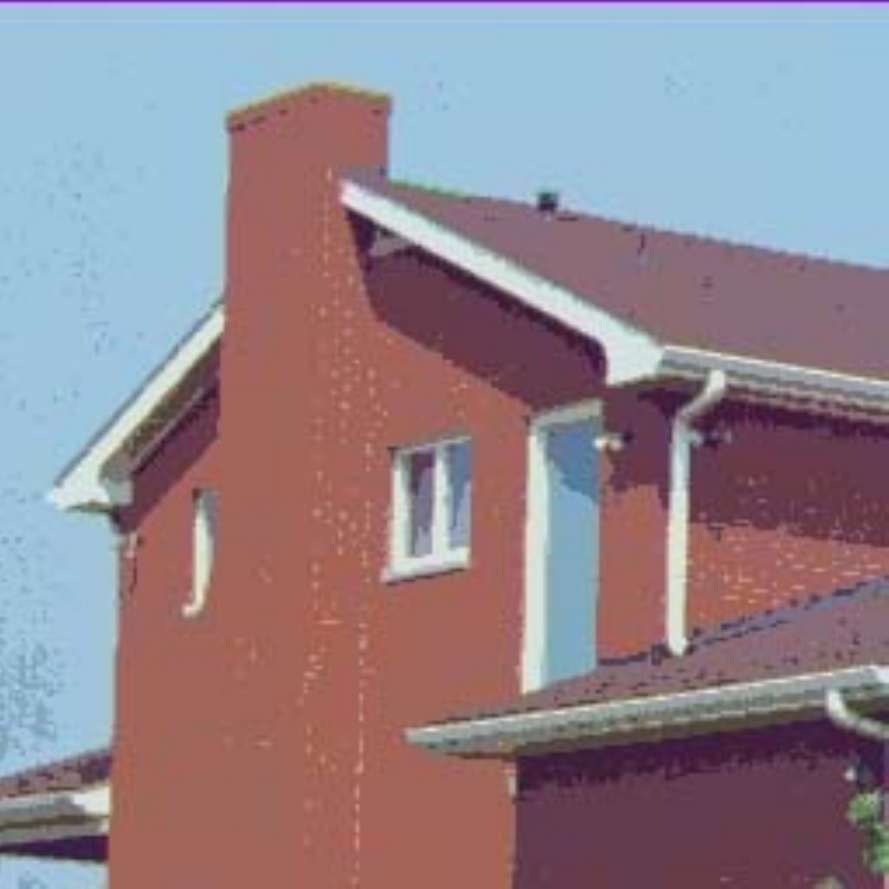}
\includegraphics[width=1.80cm]{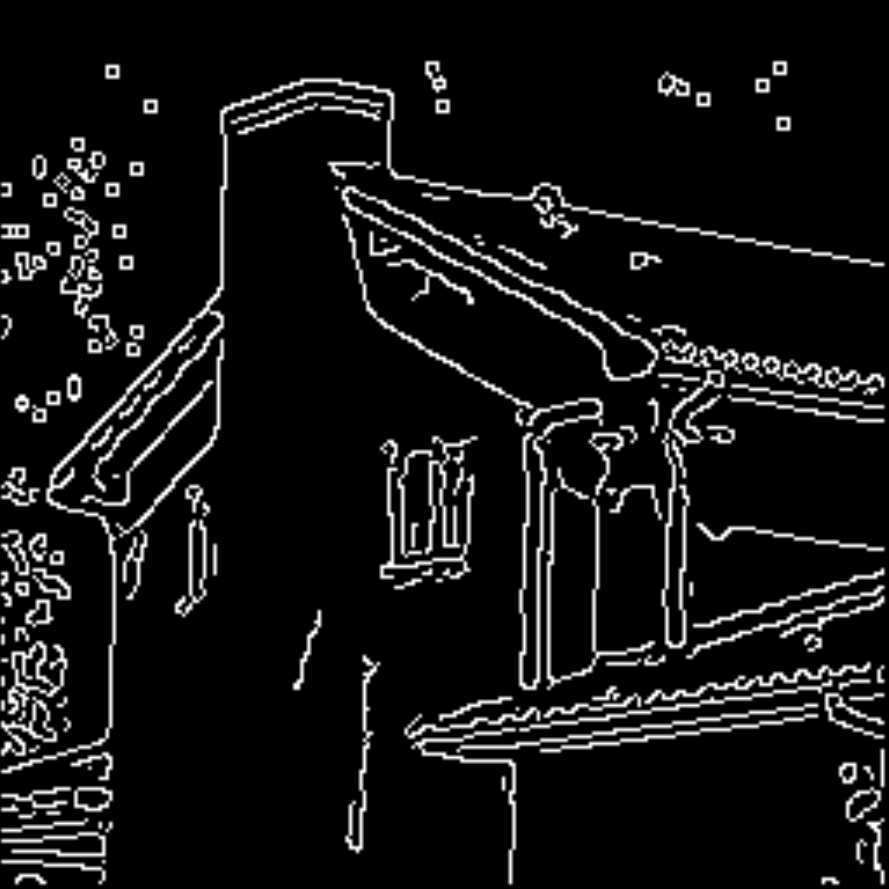}
\includegraphics[width=1.80cm]{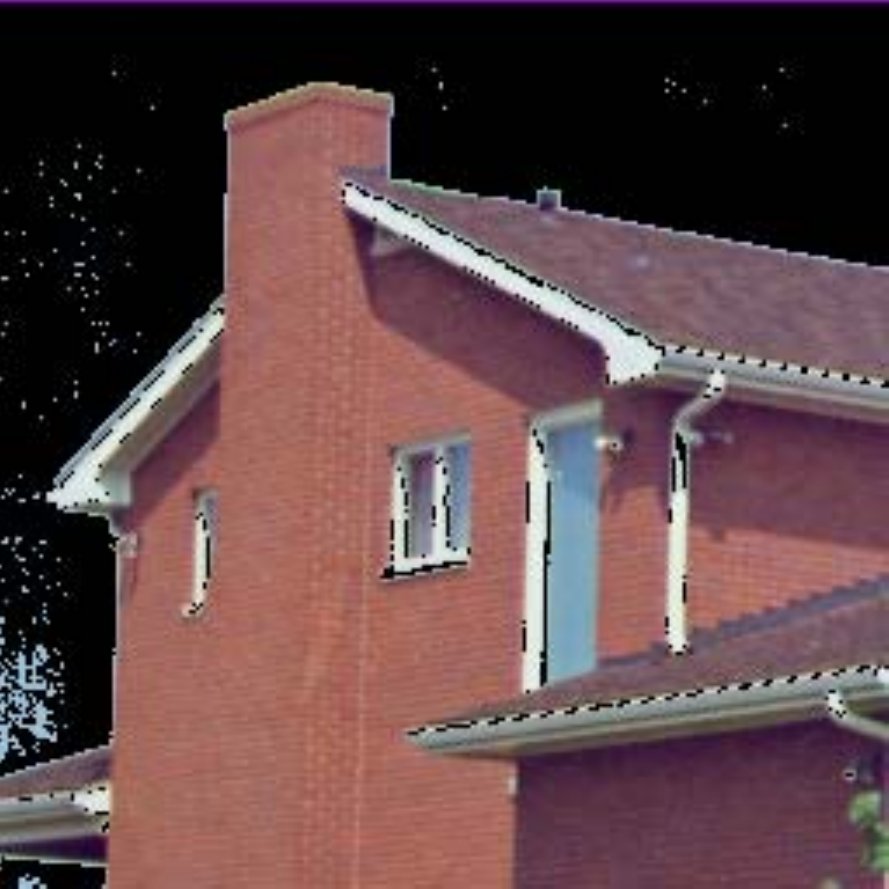}\\

\includegraphics[width=1.80cm]{Houses.pdf}
\includegraphics[width=1.80cm]{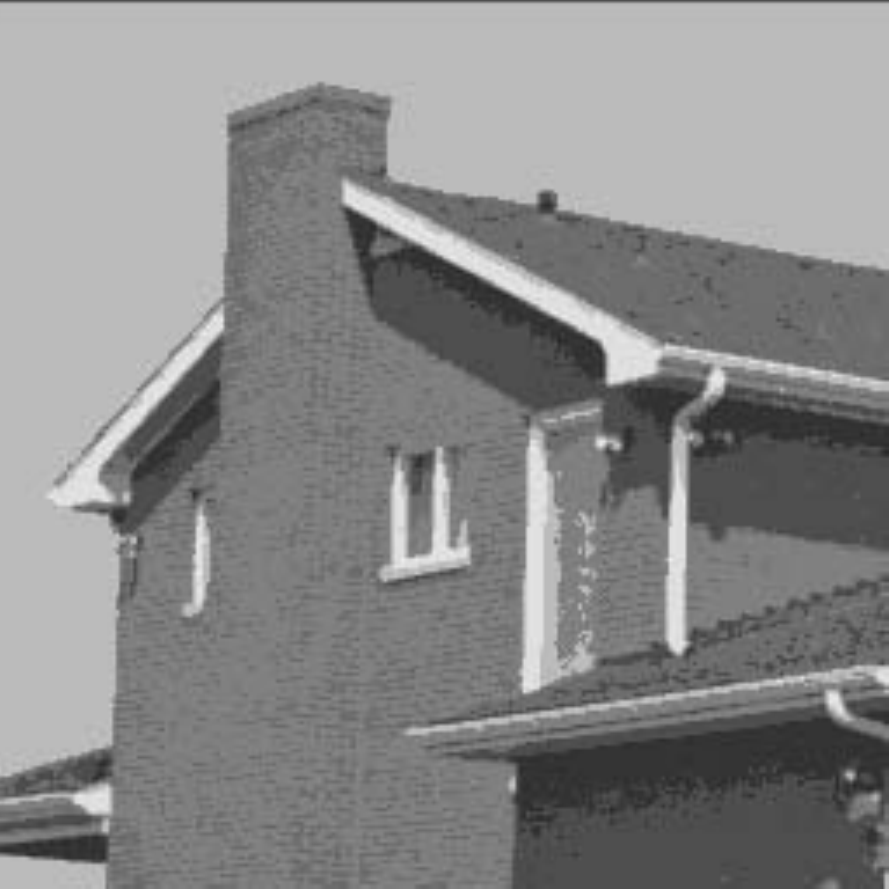}
\includegraphics[width=1.80cm]{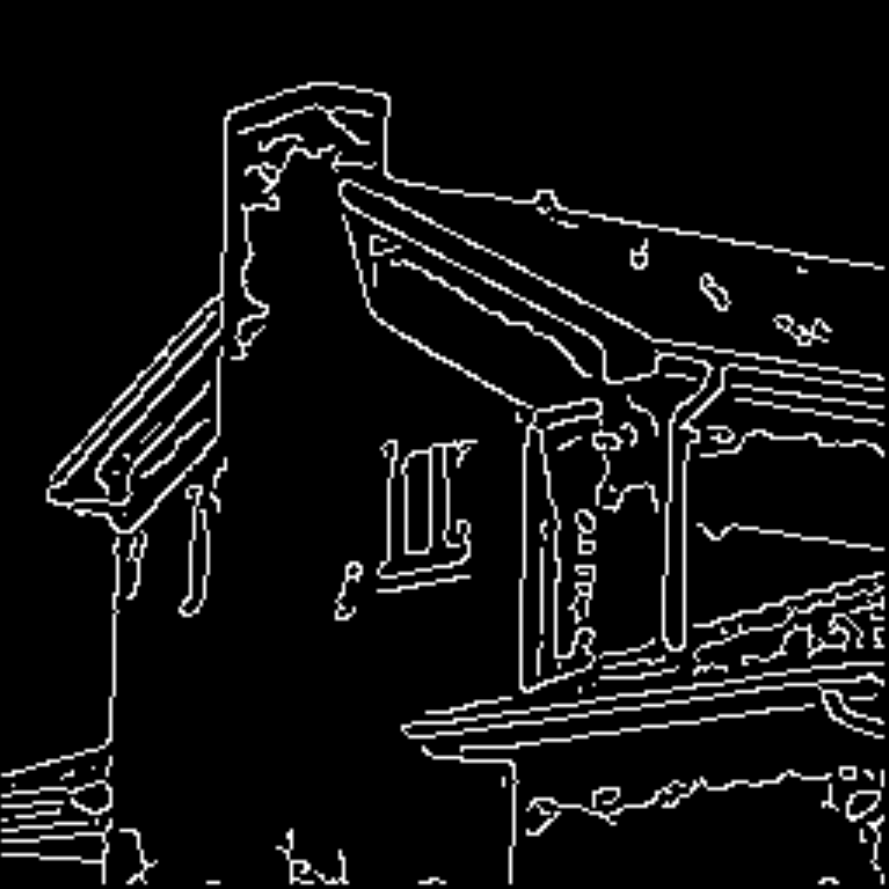}
\includegraphics[width=1.80cm]{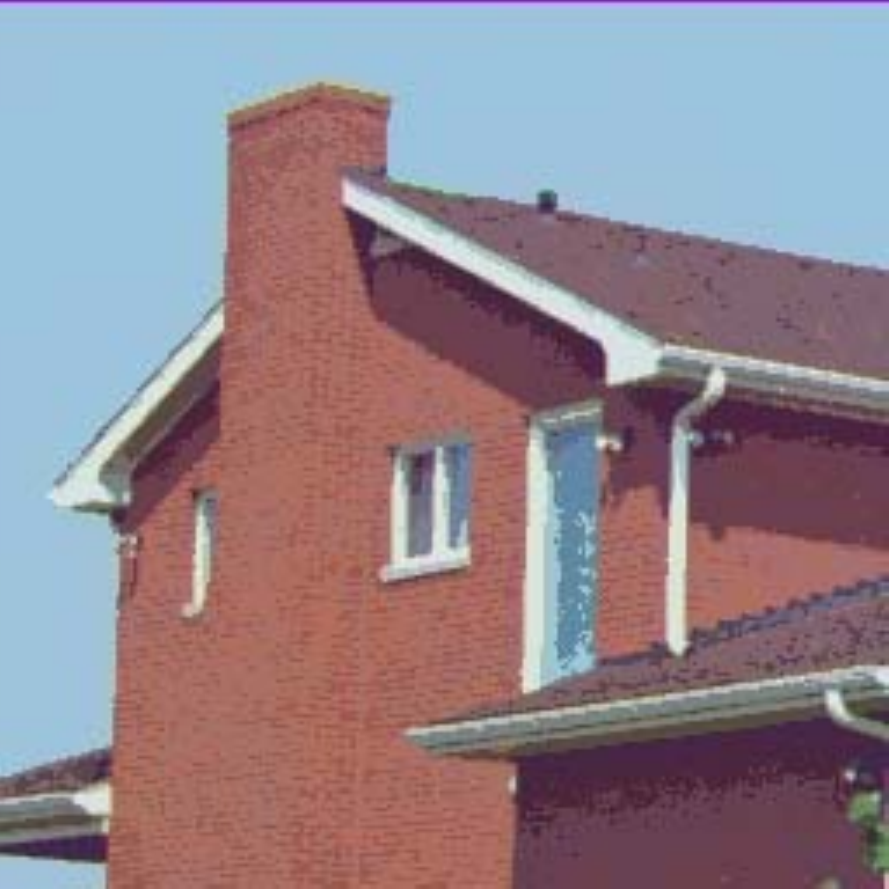}
\includegraphics[width=1.80cm]{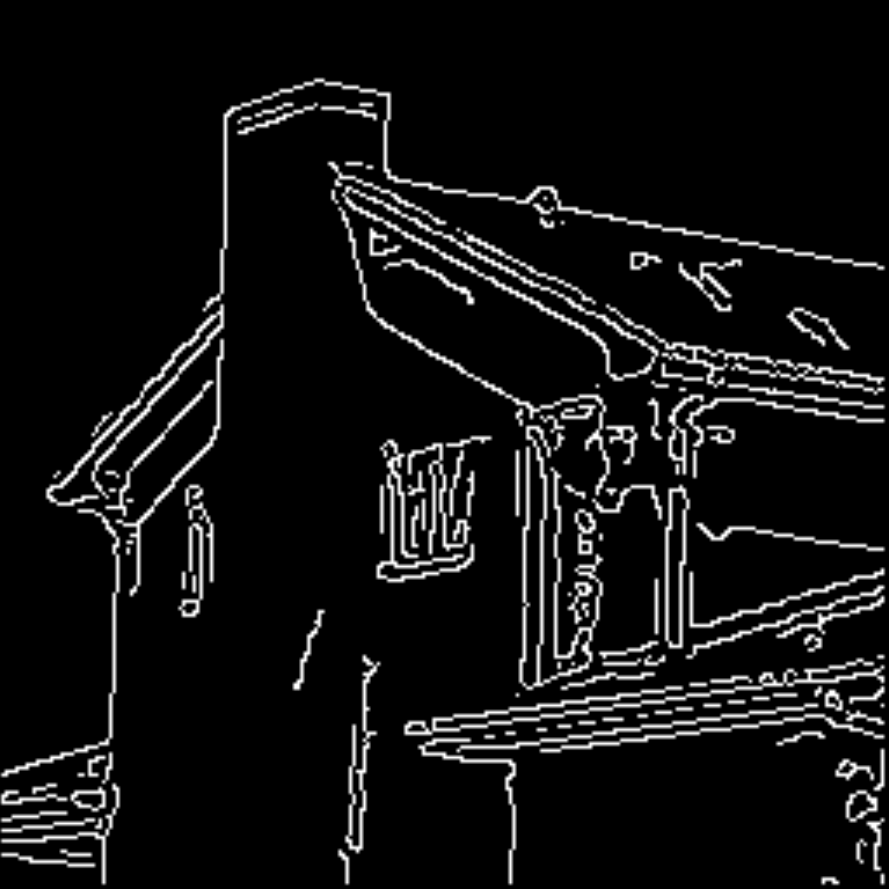}
\includegraphics[width=1.80cm]{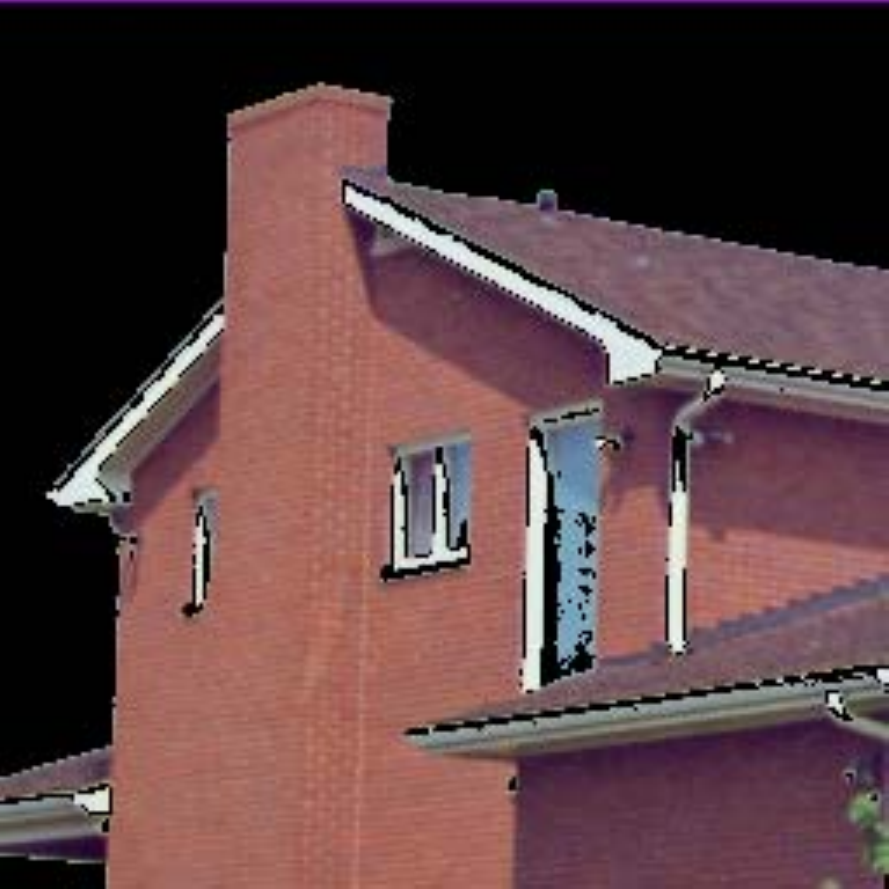}\\
\caption{Simulated output of House. Row 1 - Methodology-I, Row 2 - Methodology-II, Row 3 - Otsu's method. Column 1 - Orginal Image, Column 2 - Segmented Y component, Column 3 - Canny edge detection of segmented Y component, Column 4 - Segmented color image, Column 5 - Canny edge detection of extracted Y component, Column 6 - Extracted image.}

\label{fig:4}
\end{center}
\end{figure*}

\begin{figure*}[h]
\begin{center}

\includegraphics[width=1.80cm]{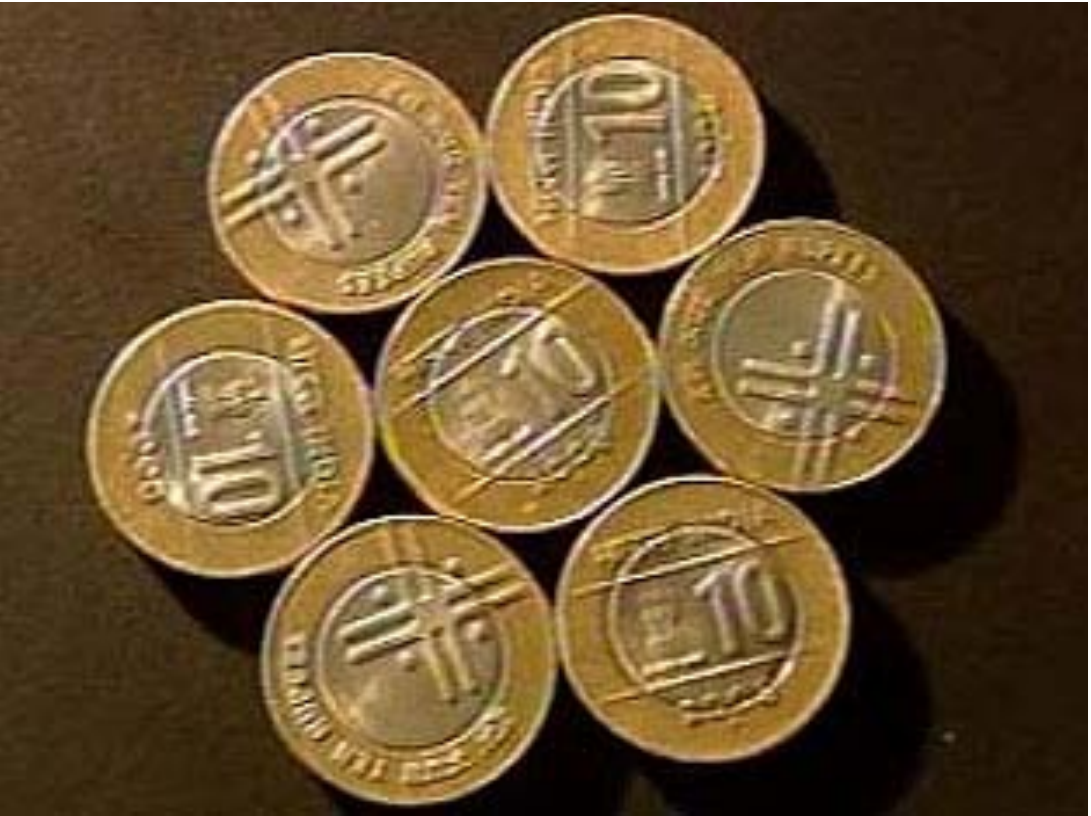}
\includegraphics[width=1.80cm]{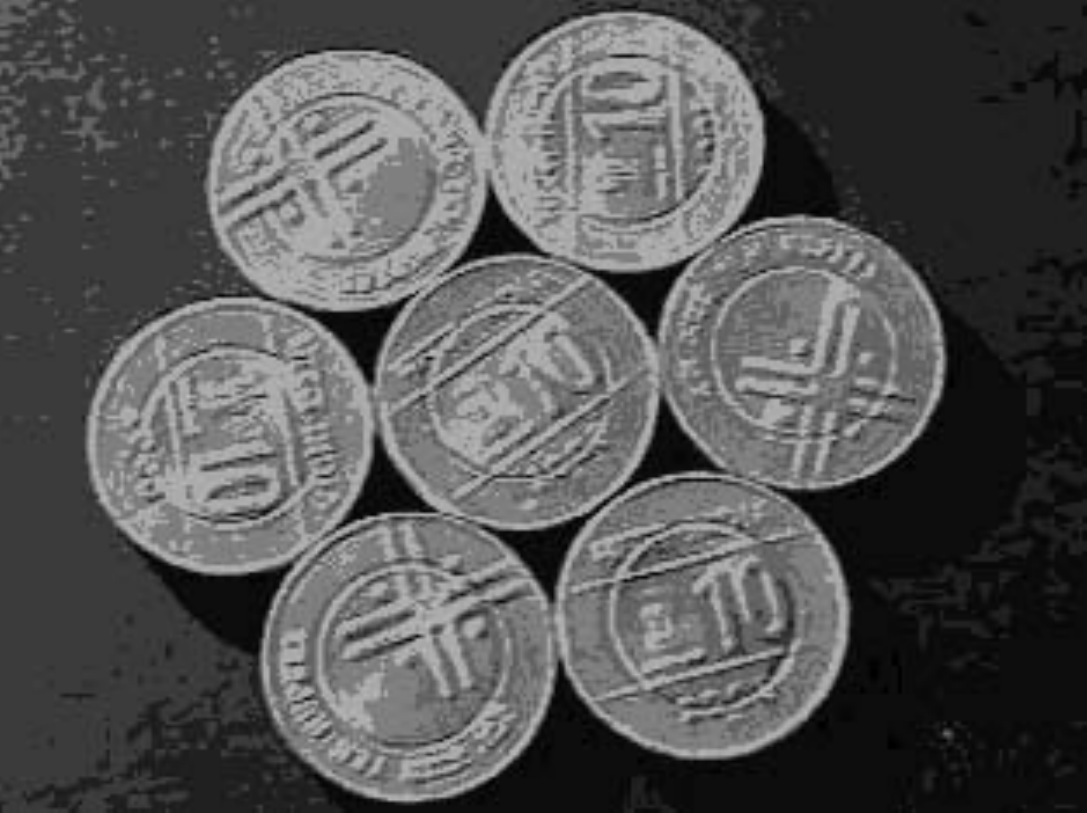}
\includegraphics[width=1.80cm]{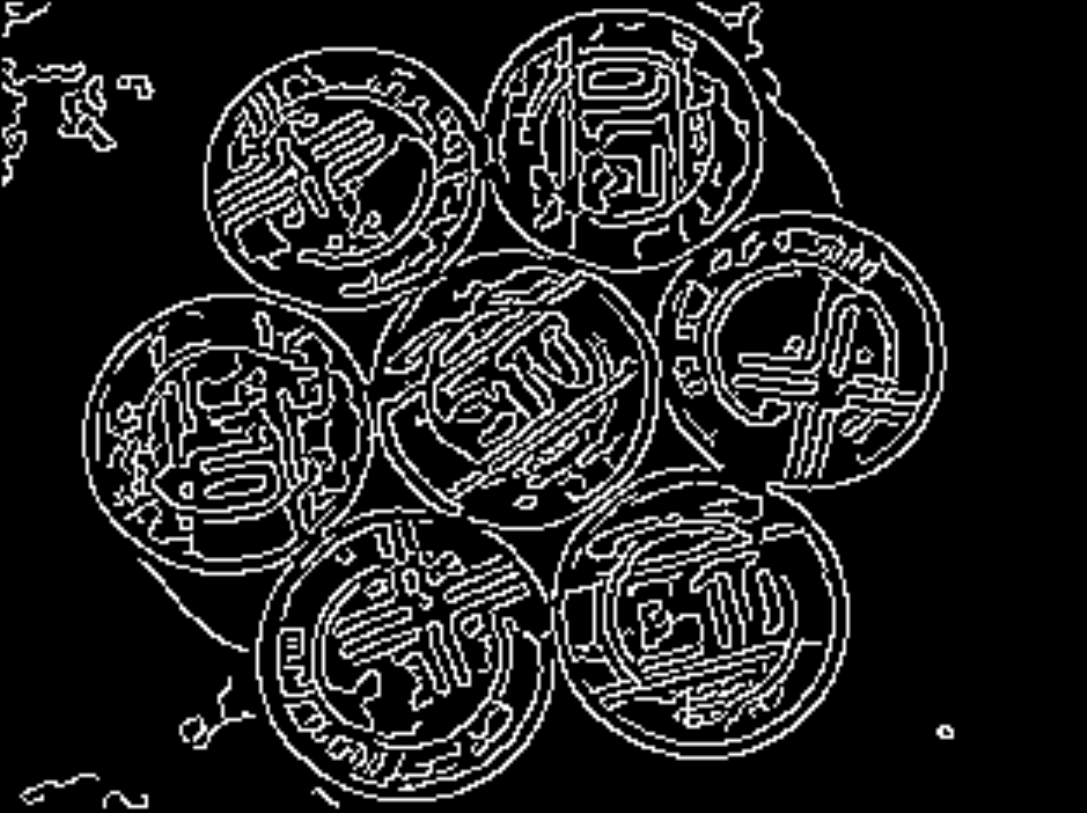}
\includegraphics[width=1.80cm]{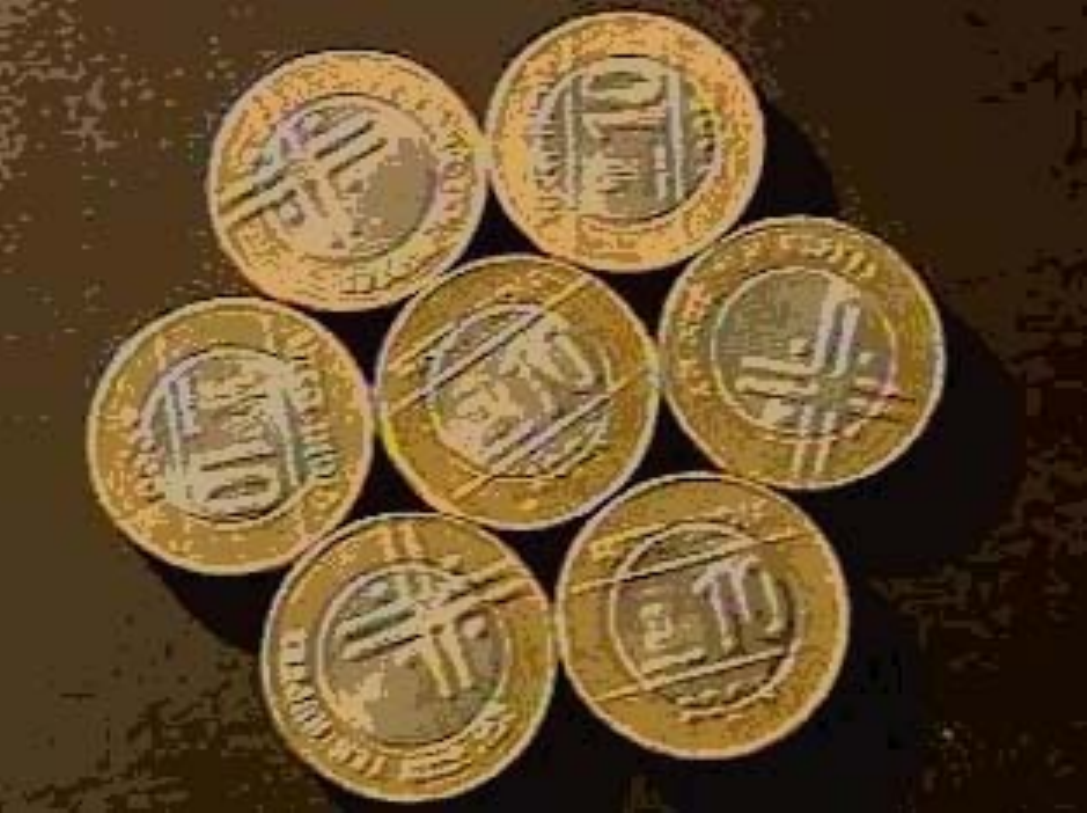}
\includegraphics[width=1.80cm]{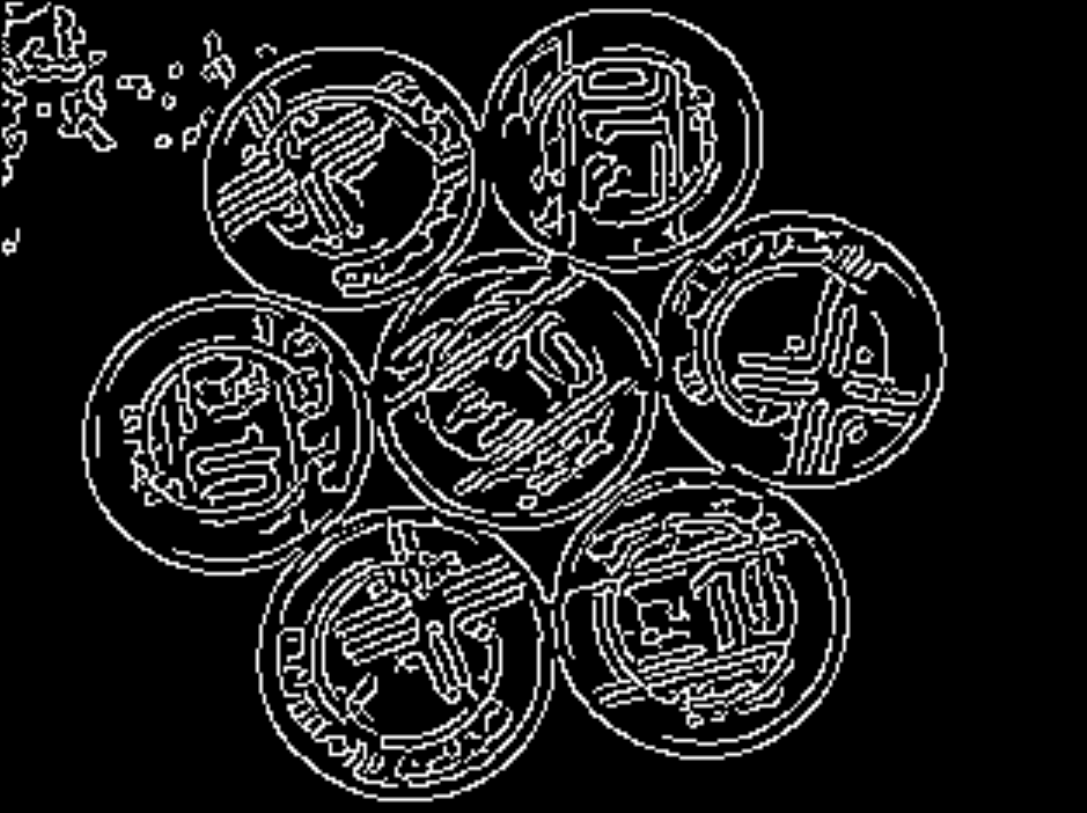}
\includegraphics[width=1.80cm]{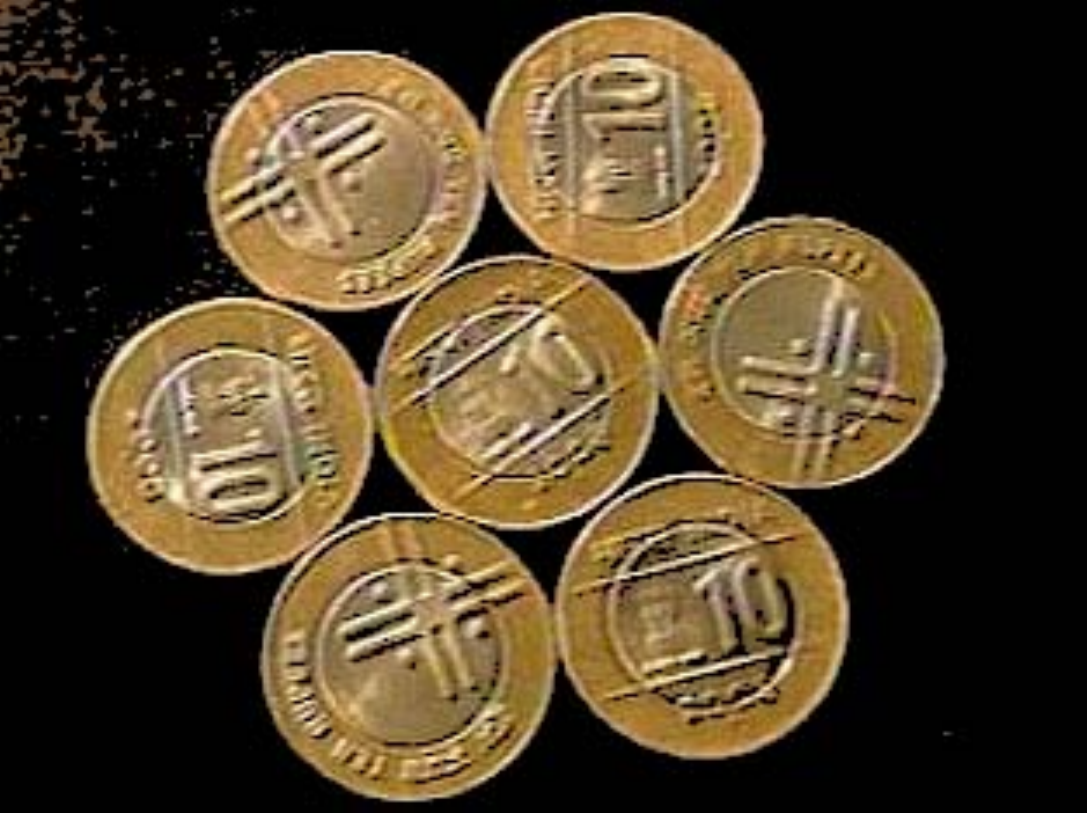}\\

\includegraphics[width=1.80cm]{Coin.pdf}
\includegraphics[width=1.80cm]{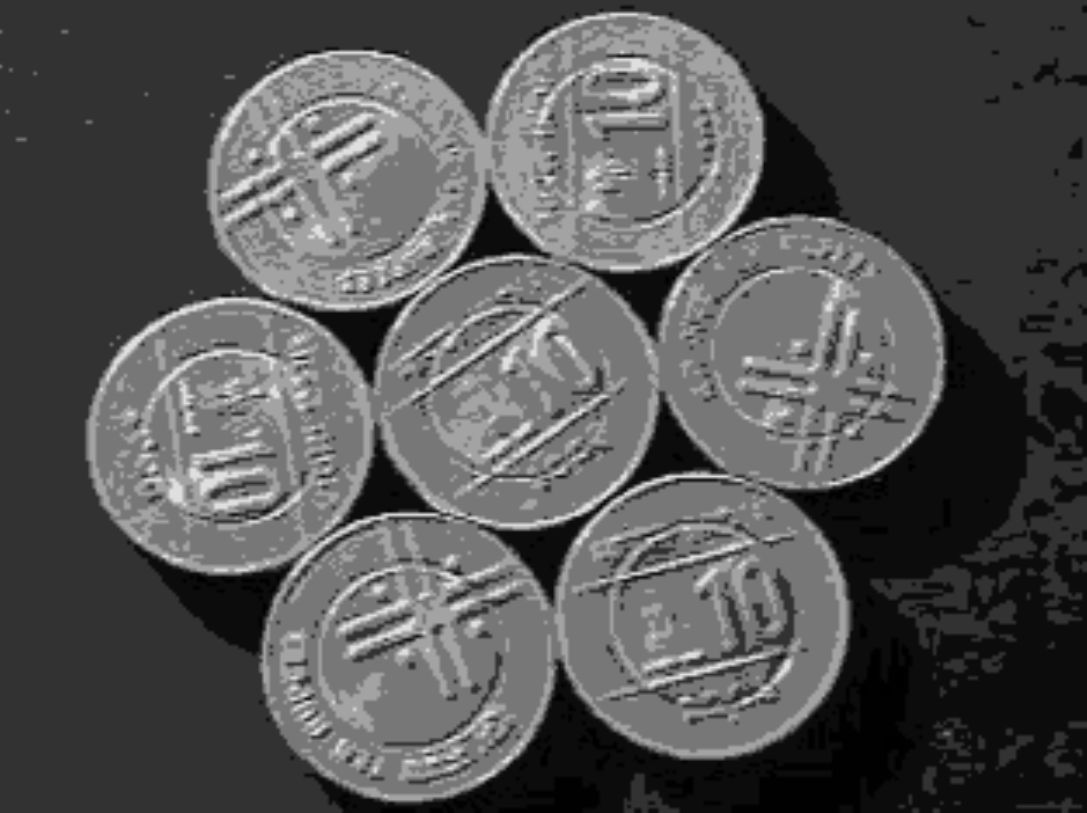}
\includegraphics[width=1.80cm]{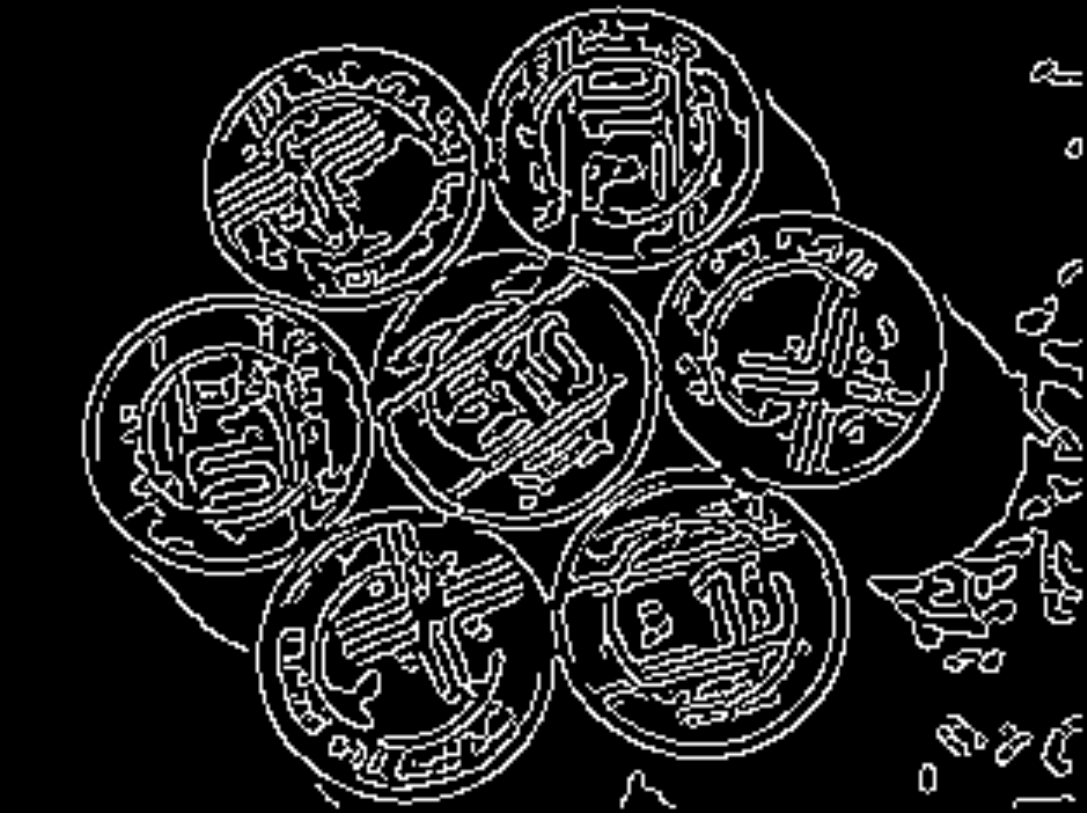}
\includegraphics[width=1.80cm]{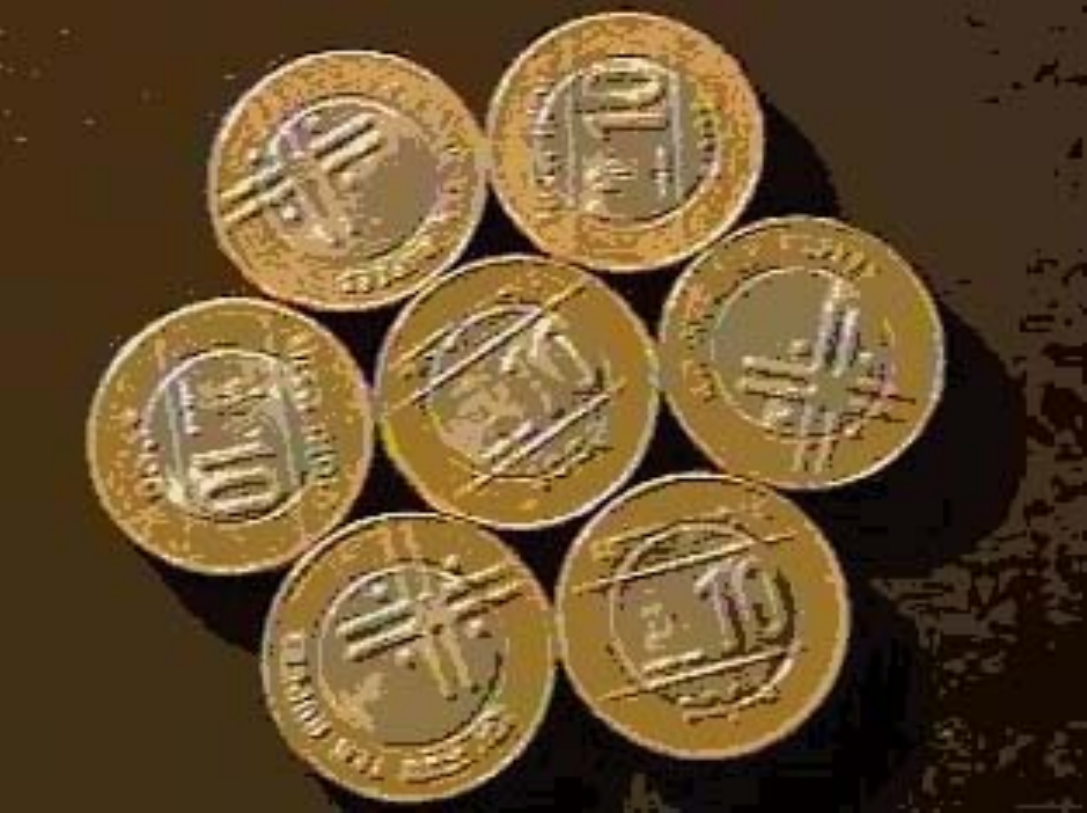}
\includegraphics[width=1.80cm]{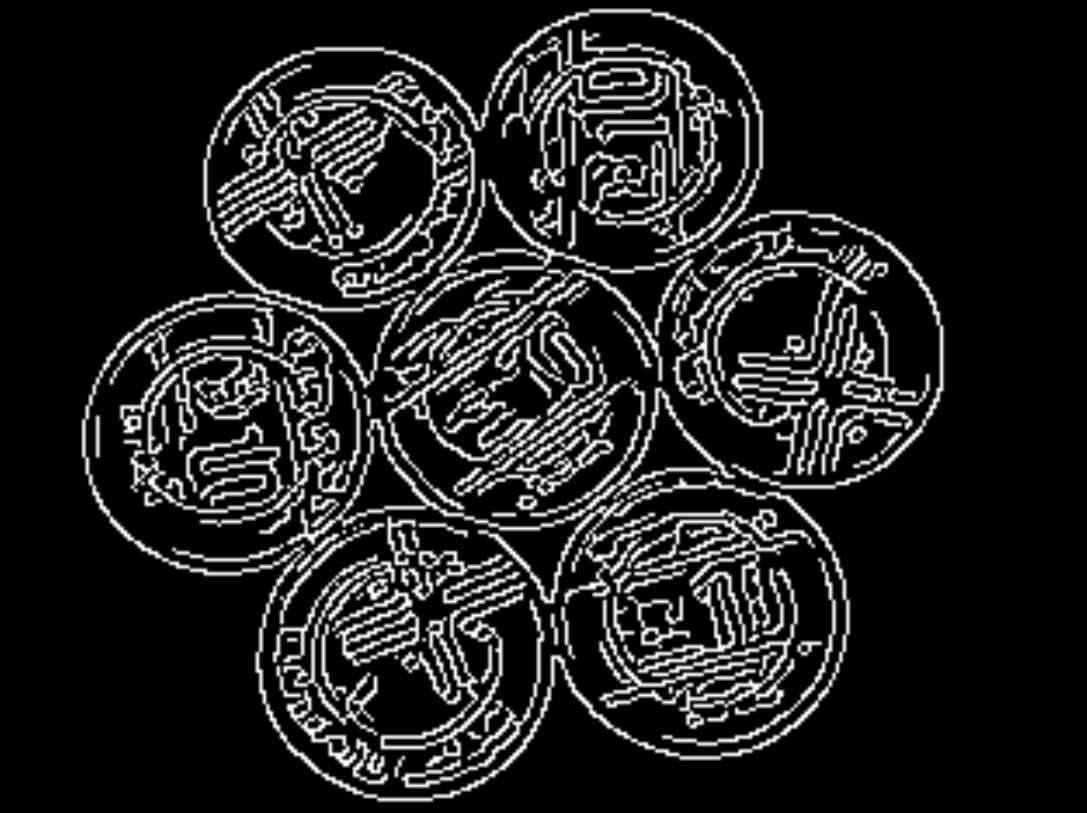}
\includegraphics[width=1.80cm]{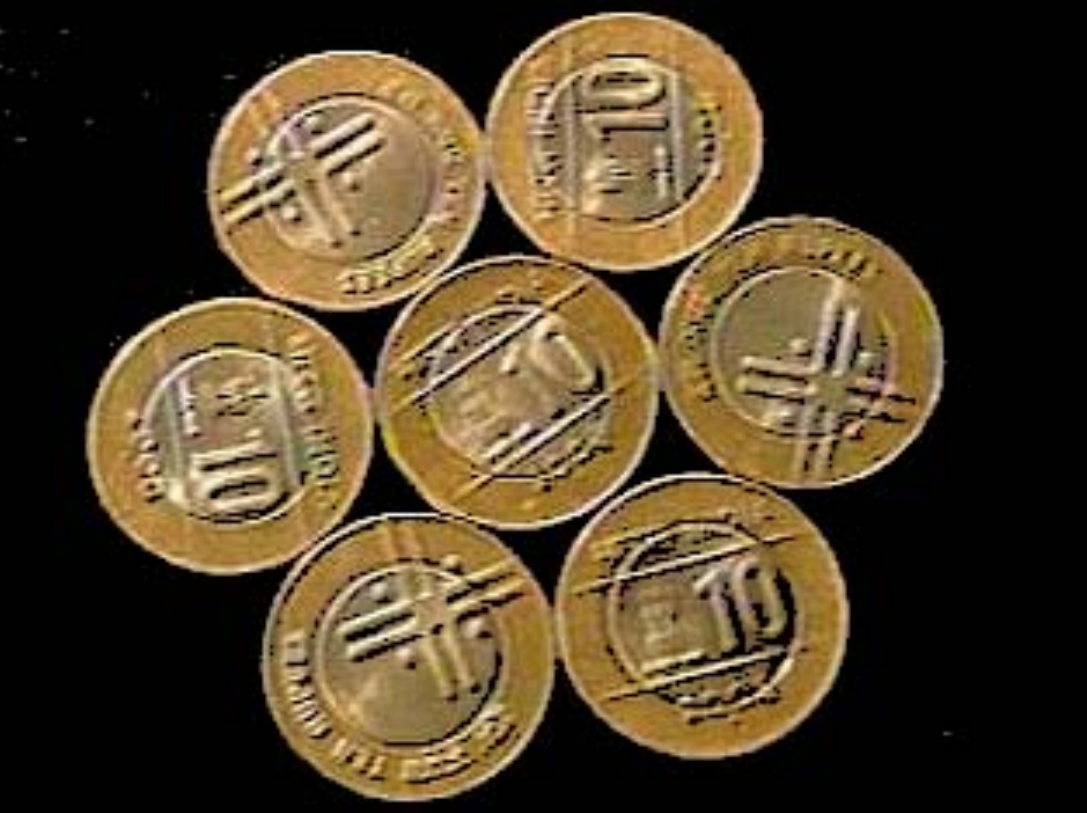}\\

\includegraphics[width=1.80cm]{Coin.pdf}
\includegraphics[width=1.80cm]{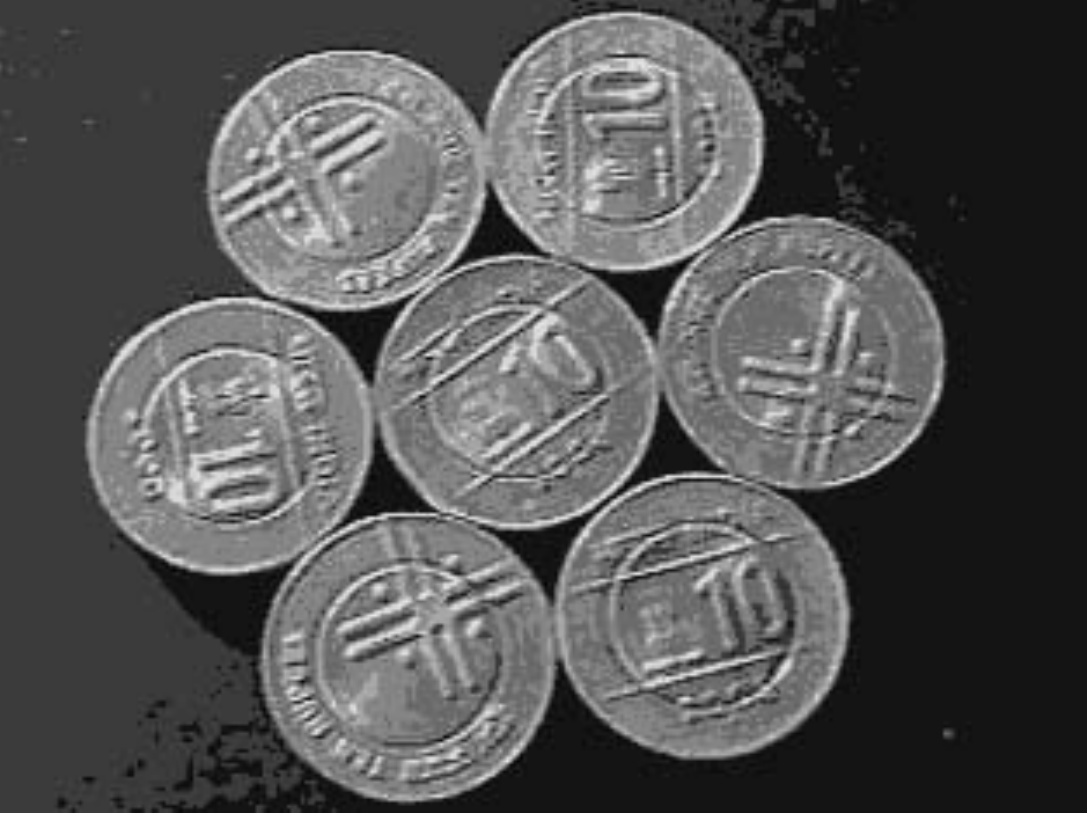}
\includegraphics[width=1.80cm]{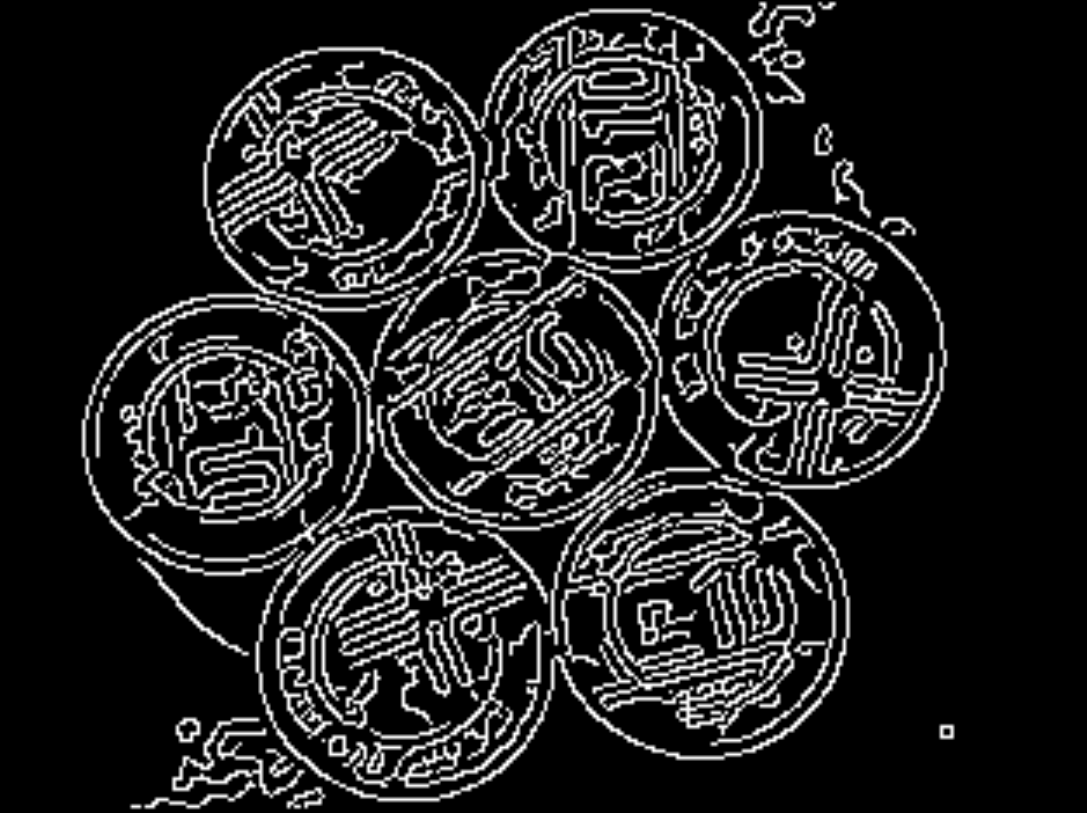}
\includegraphics[width=1.80cm]{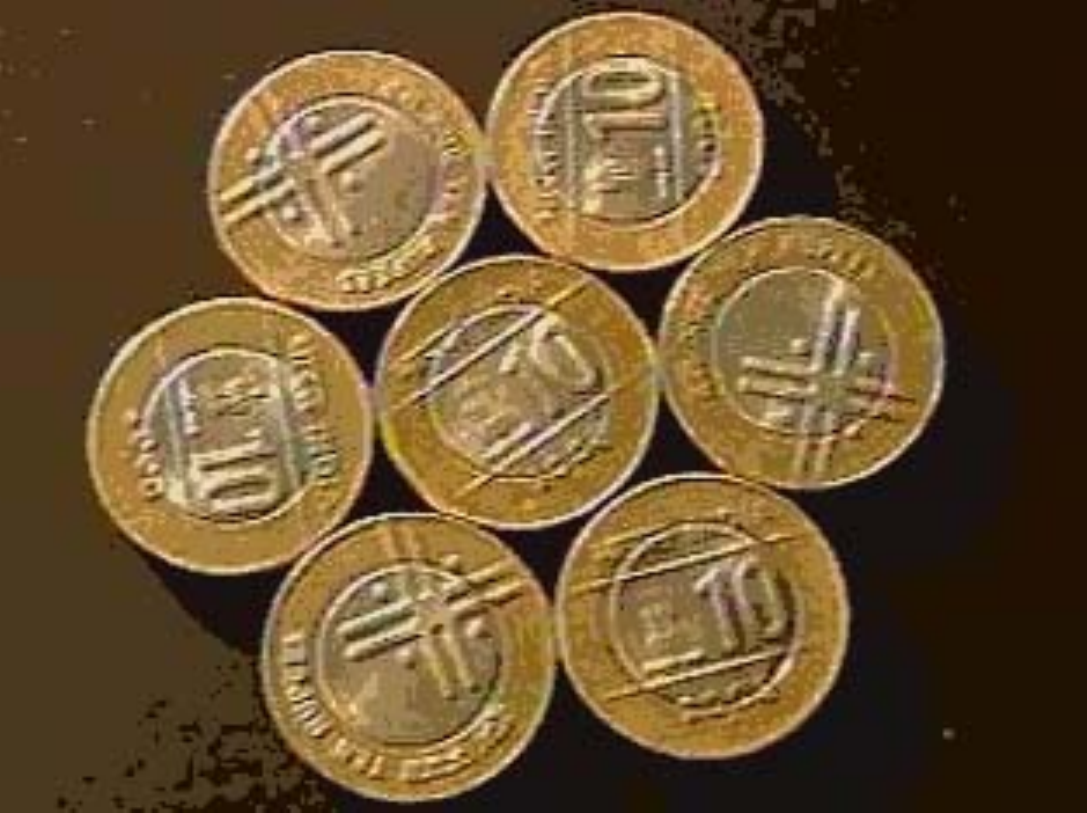}
\includegraphics[width=1.80cm]{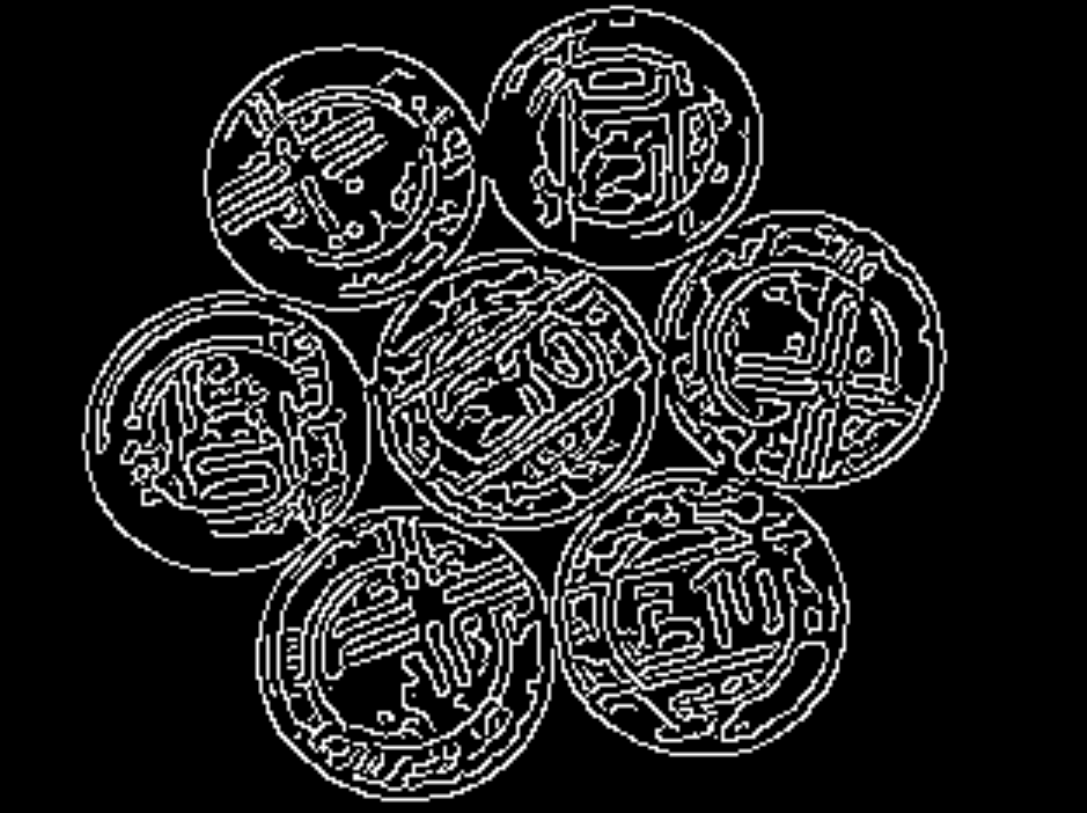}
\includegraphics[width=1.80cm]{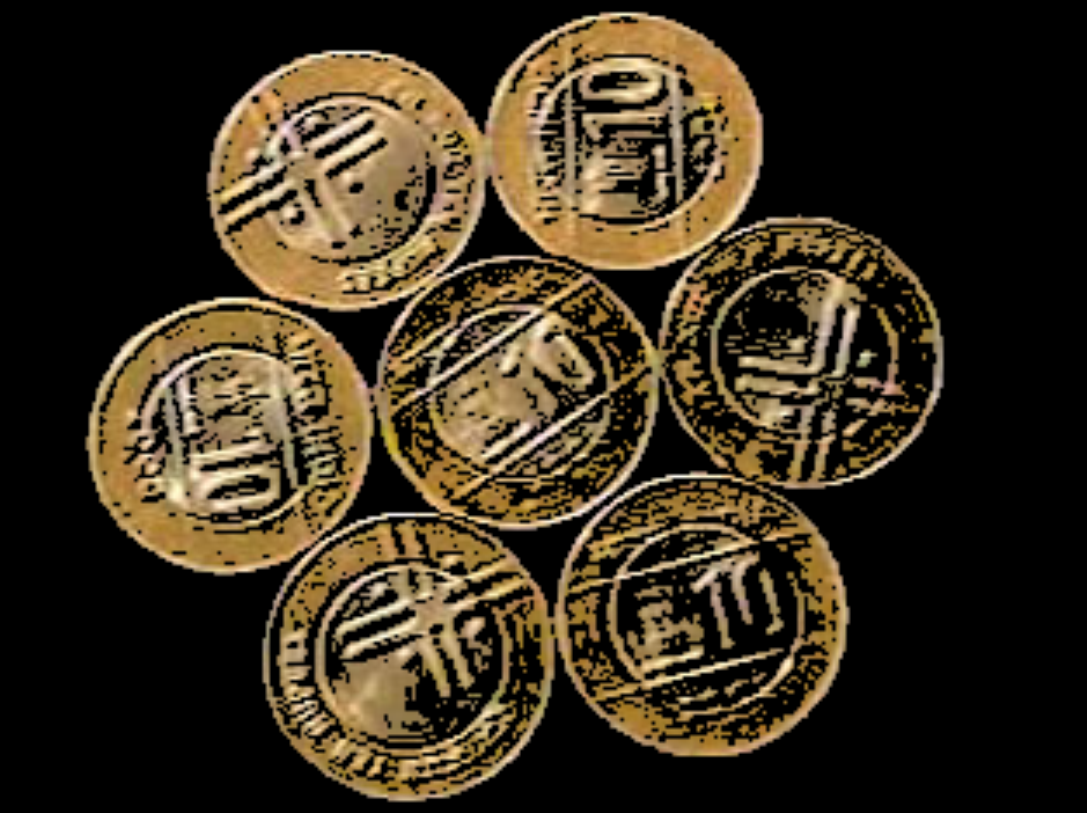}\\
\caption{Simulated output of Coin. Row 1 - Methodology-I, Row 2 - Methodology-II, Row 3 - Otsu's method. Column 1 - Orginal Image, Column 2 - Segmented Y component, Column 3 - Canny edge detection of segmented Y component, Column 4 - Segmented color image, Column 5 - Canny edge detection of extracted Y component, Column 6 - Extracted image.}

\label{fig:5}
\end{center}
\end{figure*}

\begin{figure*}[h]
\begin{center}

\includegraphics[width=1.80cm]{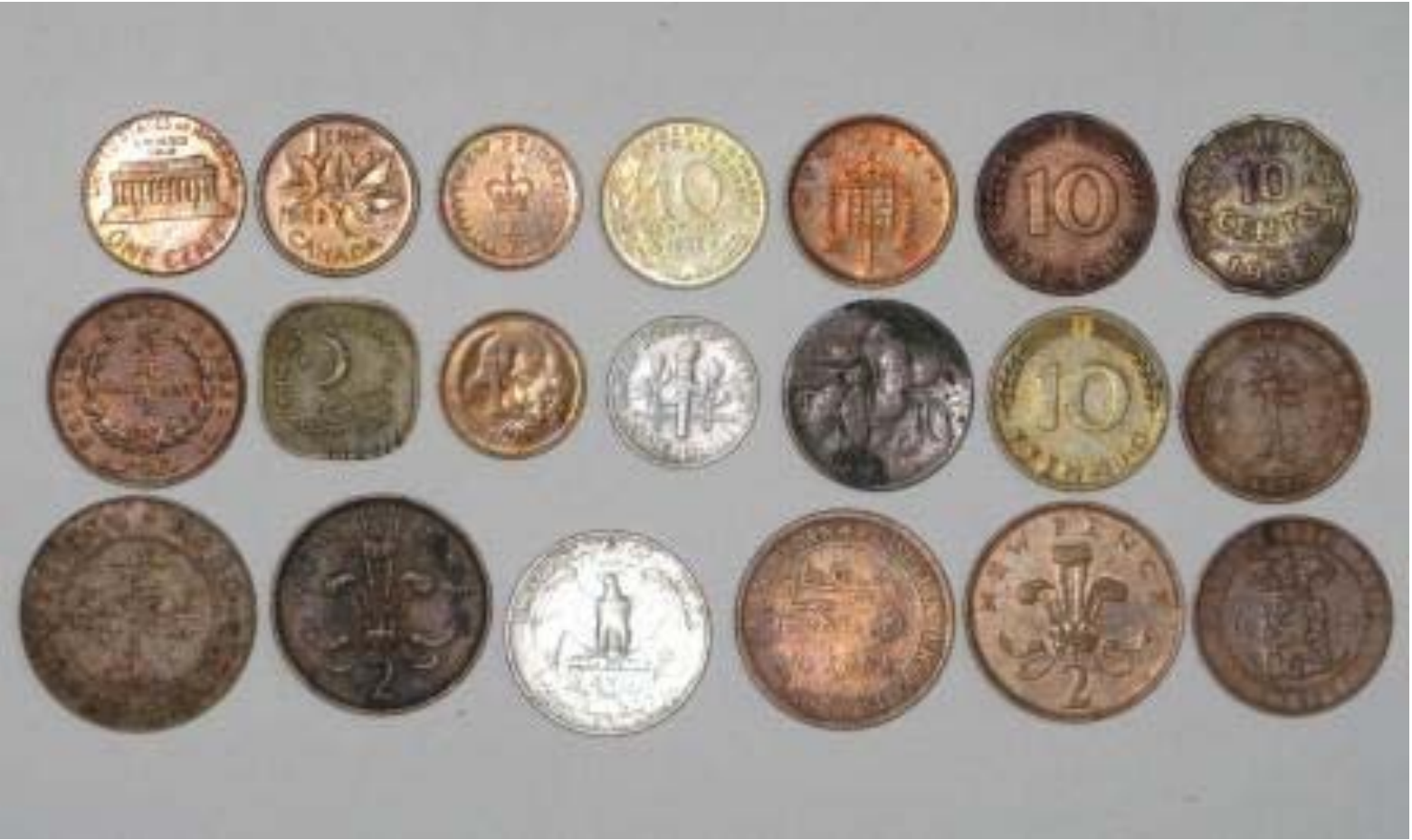}
\includegraphics[width=1.80cm]{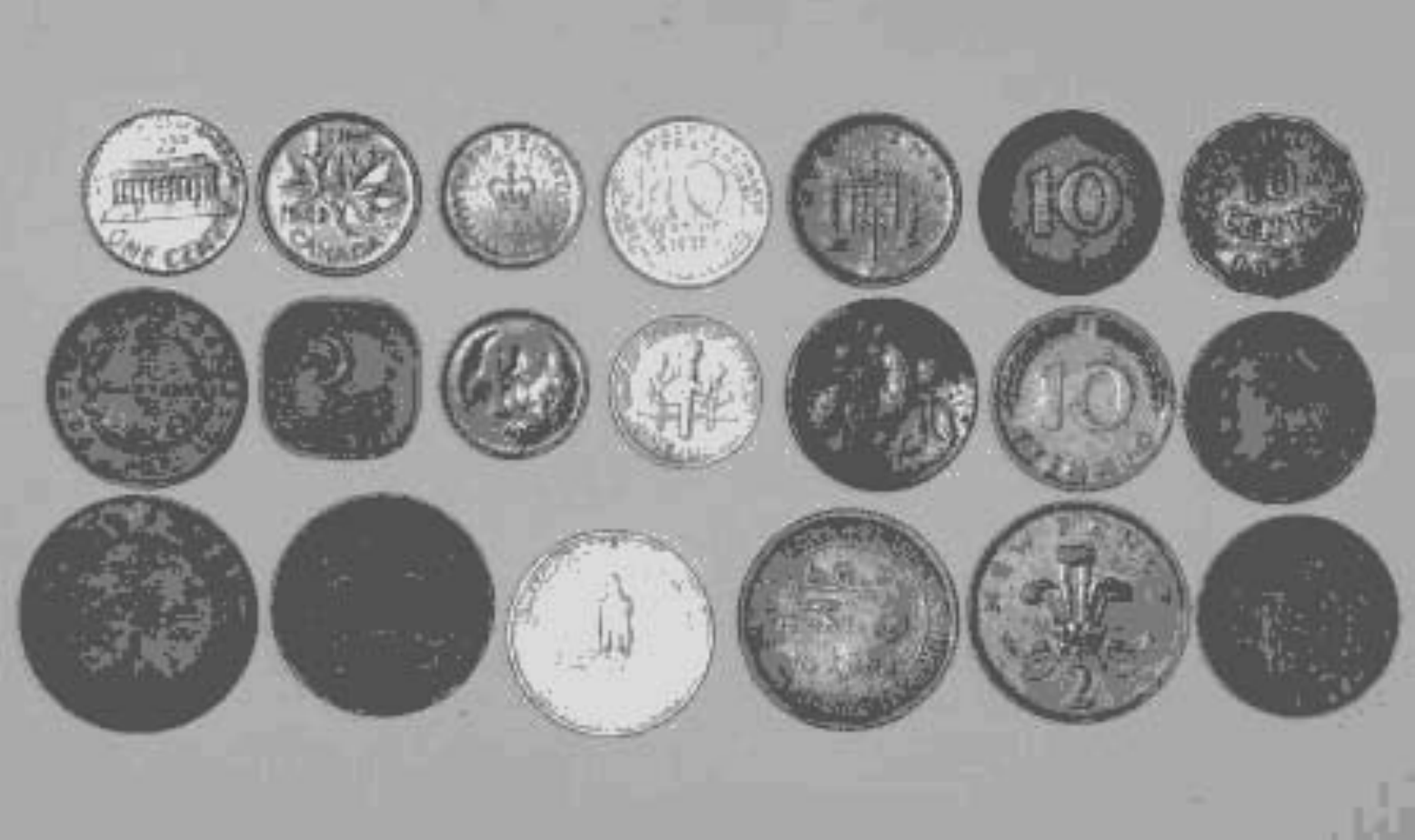}
\includegraphics[width=1.80cm]{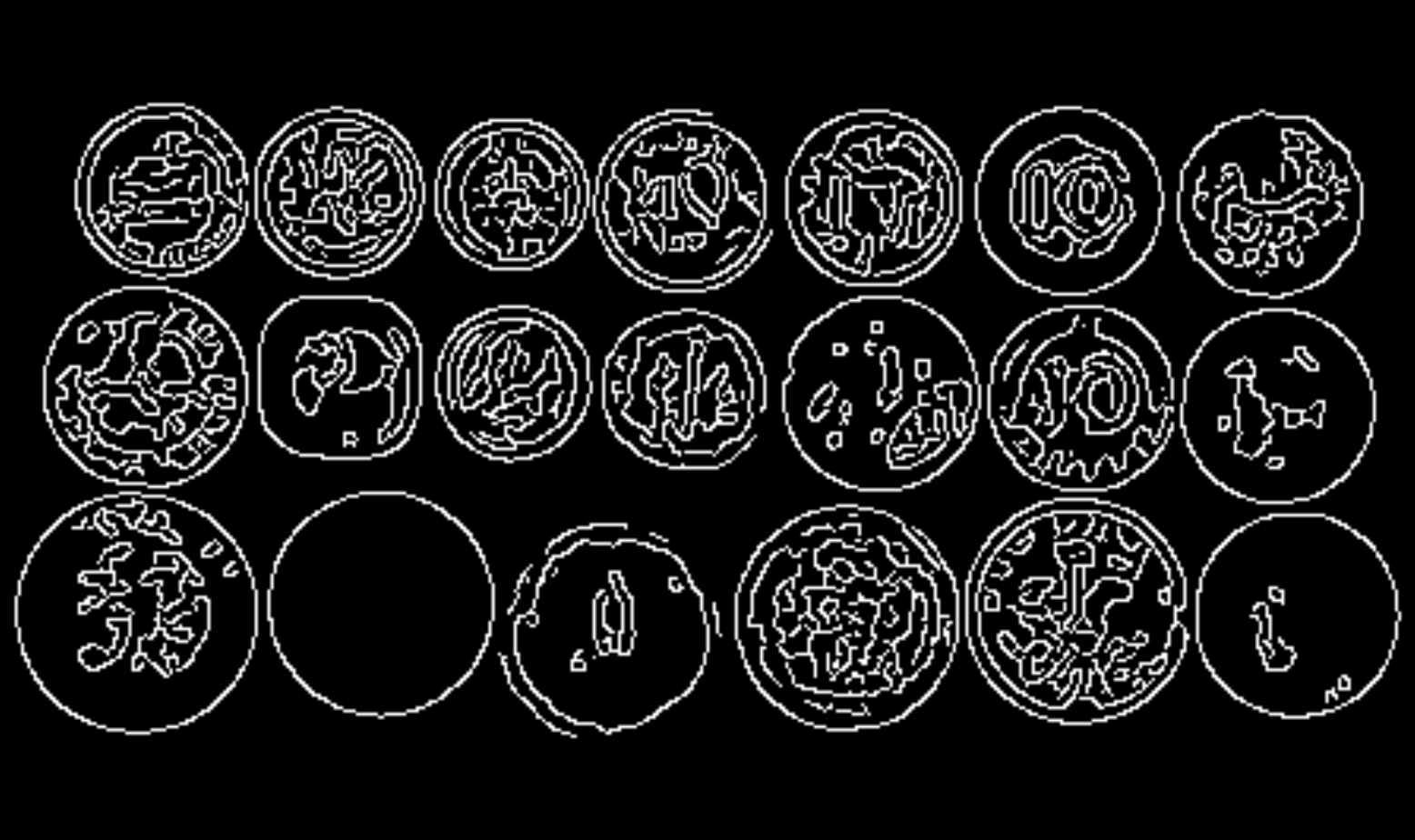}
\includegraphics[width=1.80cm]{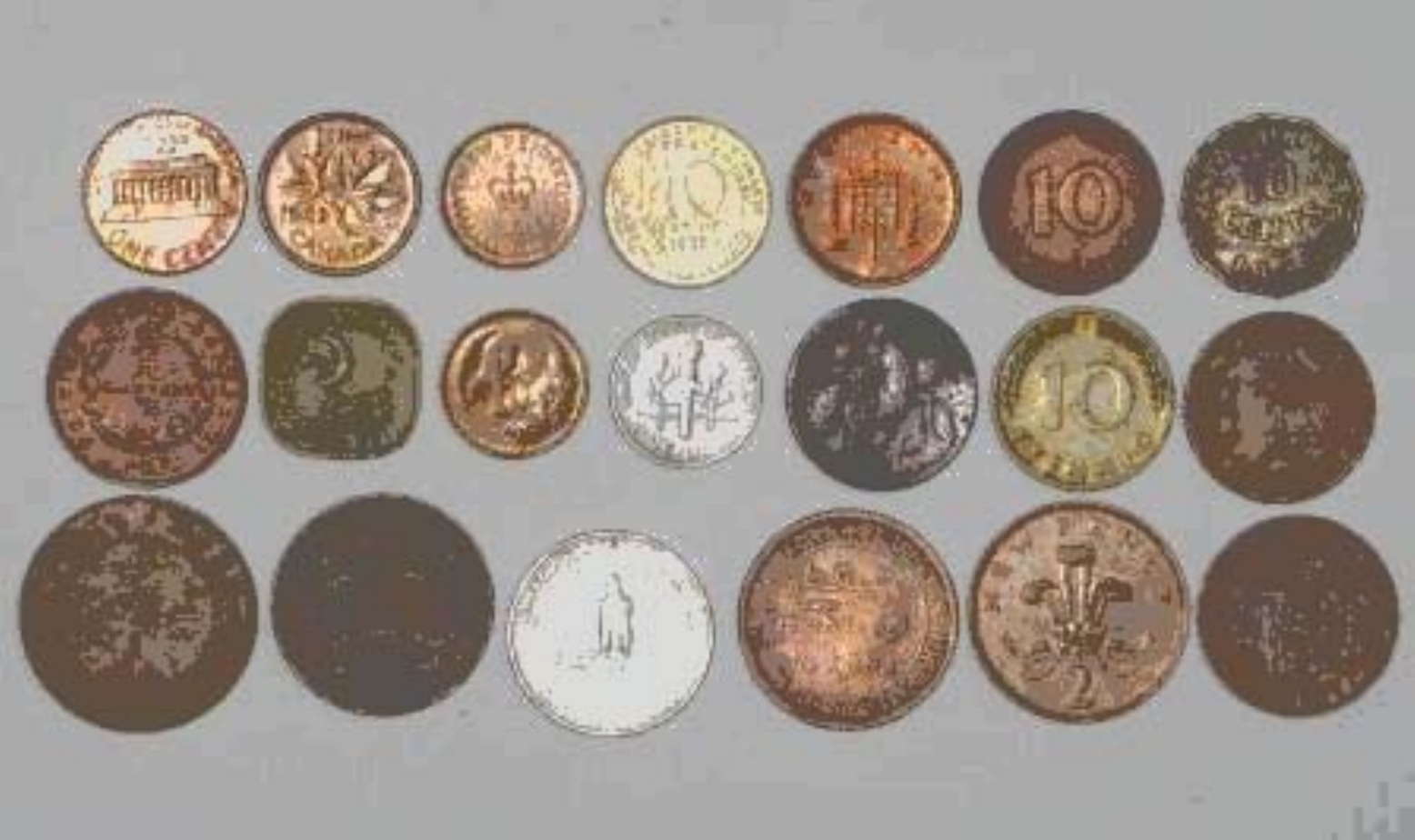}
\includegraphics[width=1.80cm]{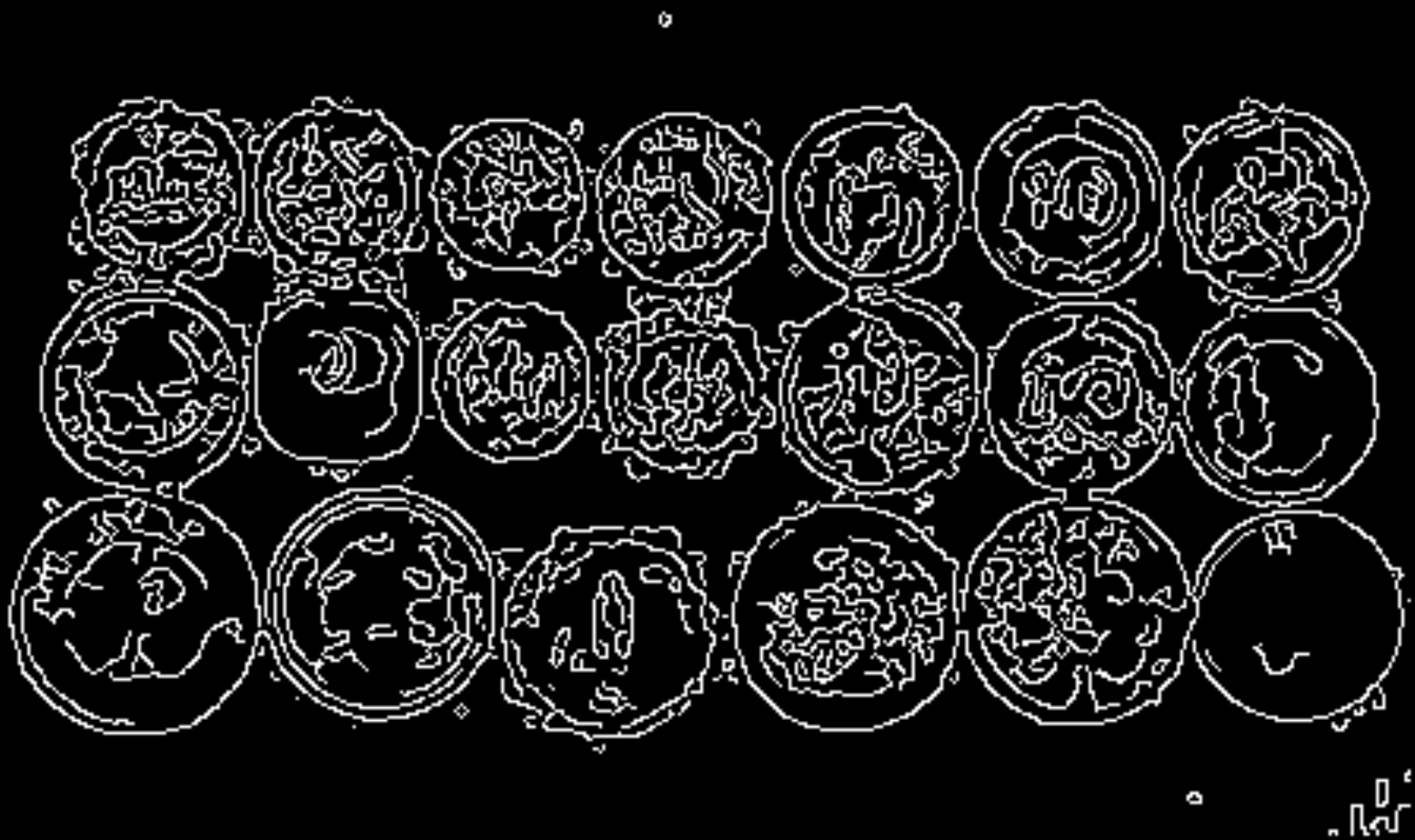}
\includegraphics[width=1.80cm]{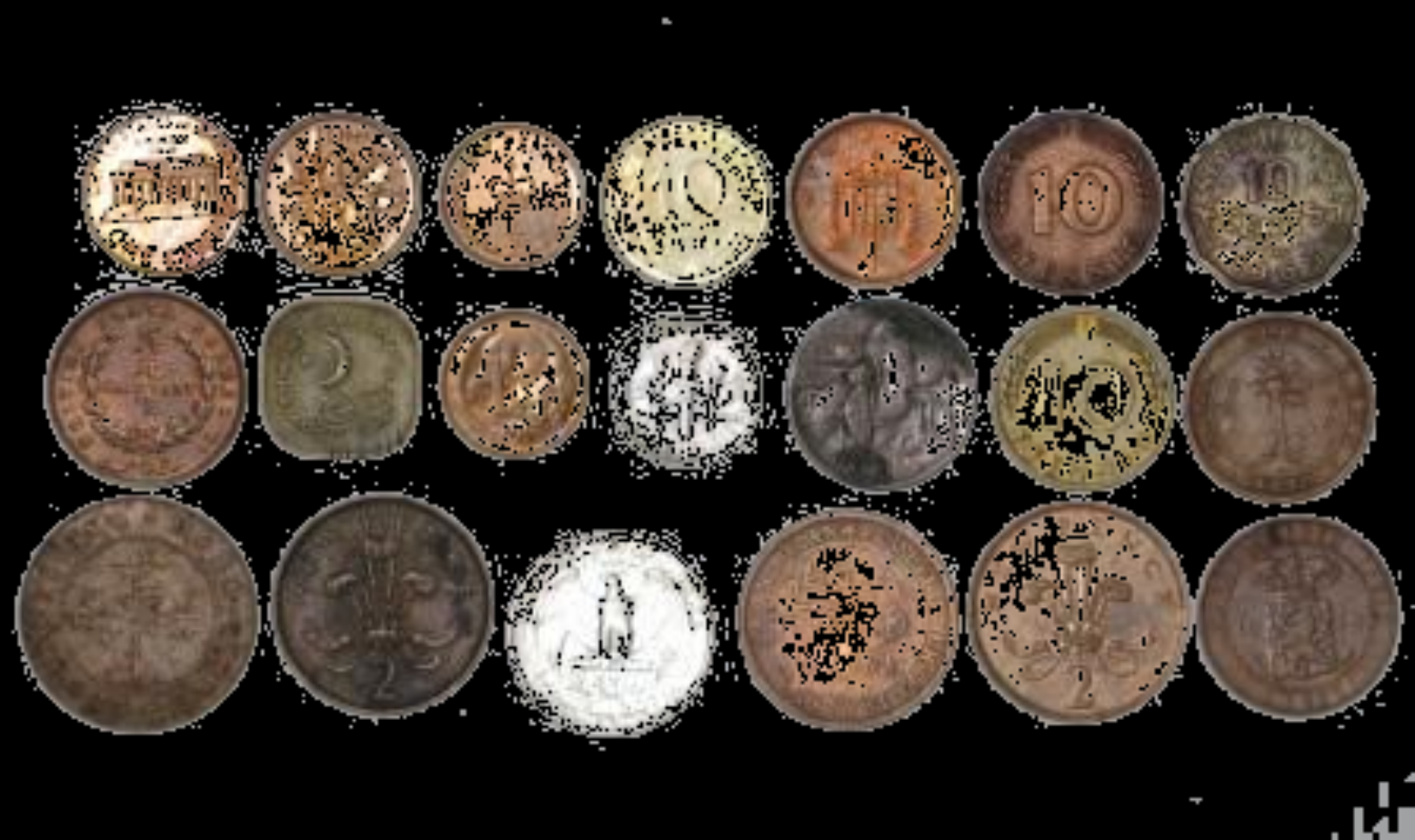}\\

\includegraphics[width=1.80cm]{Coins.pdf}
\includegraphics[width=1.80cm]{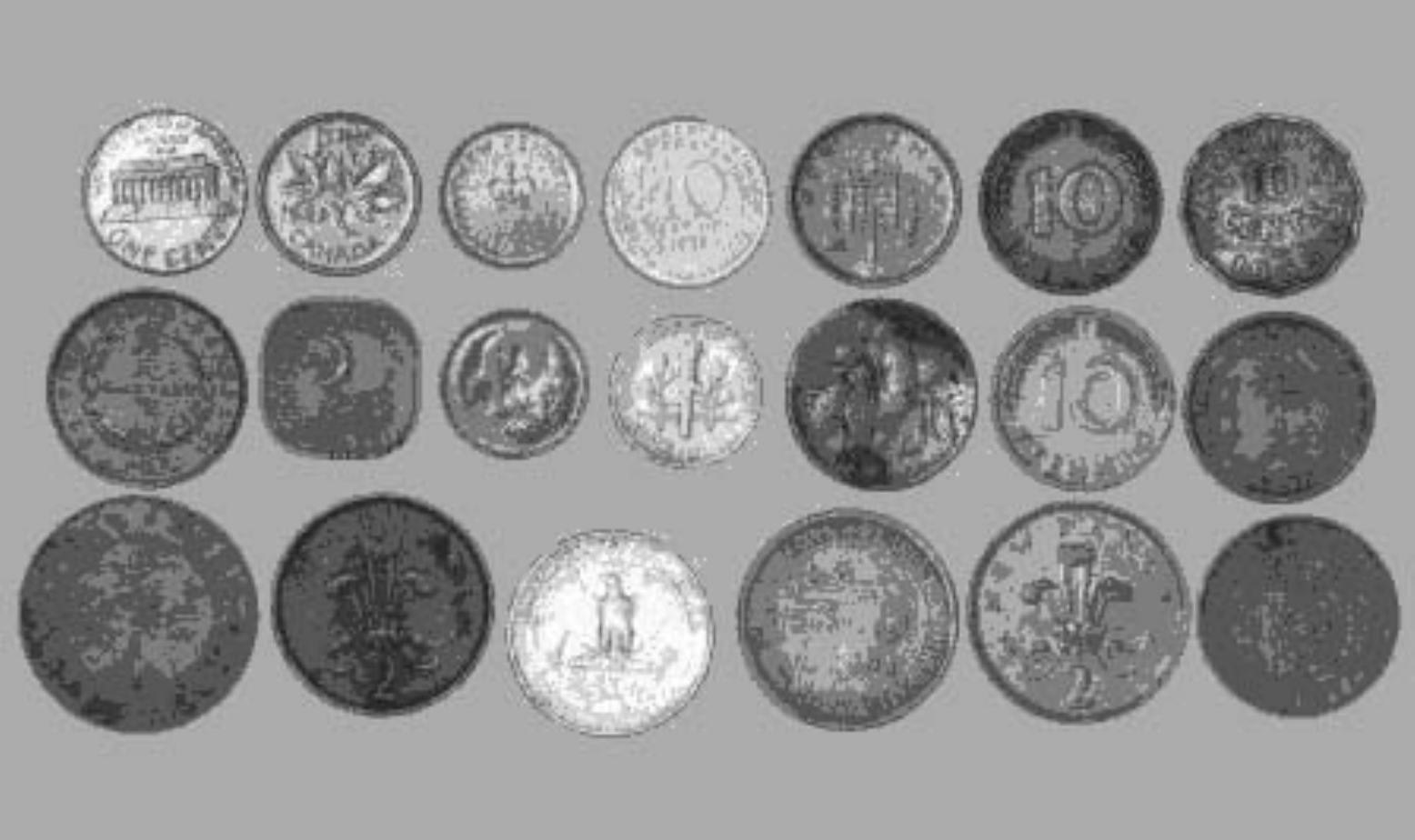}
\includegraphics[width=1.80cm]{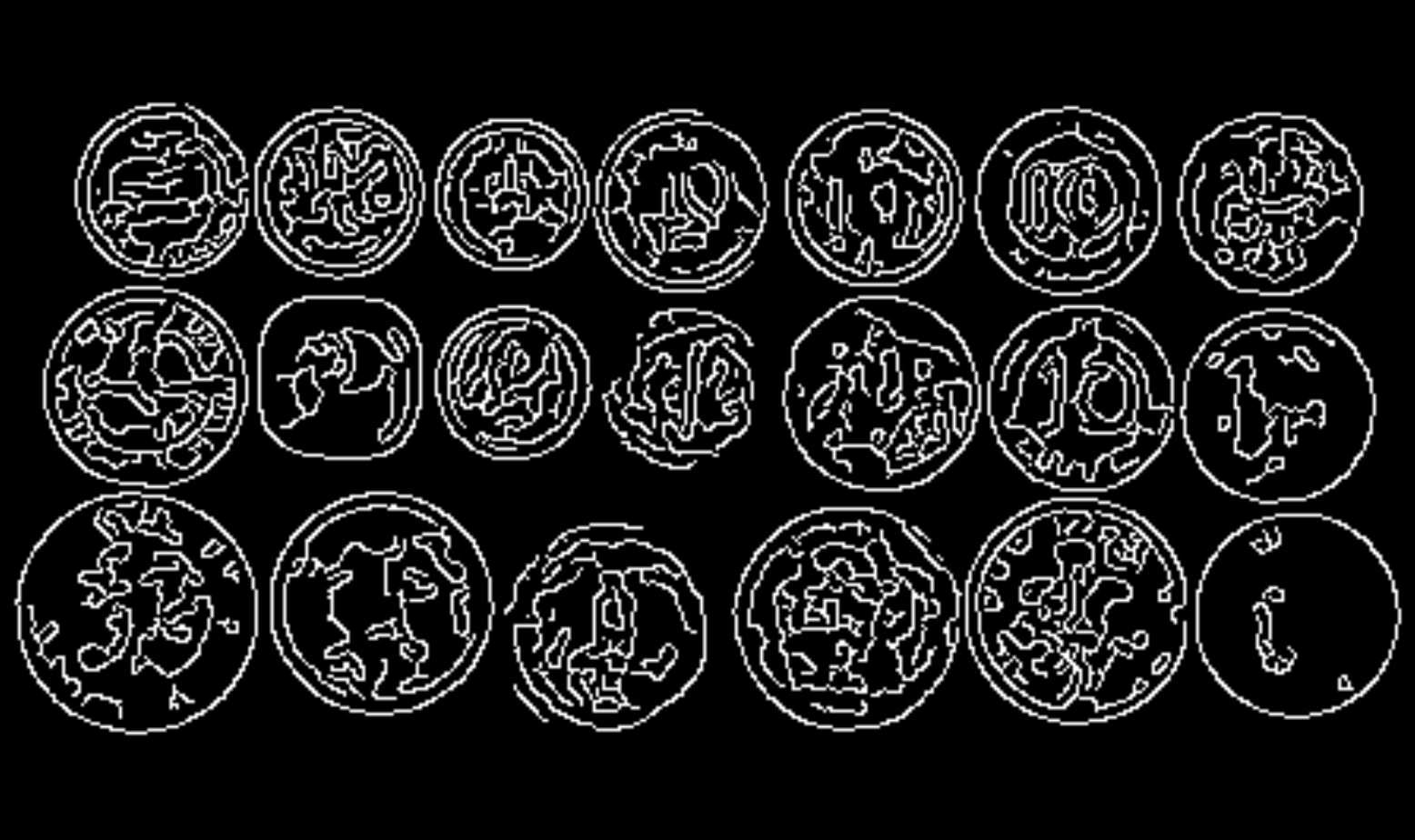}
\includegraphics[width=1.80cm]{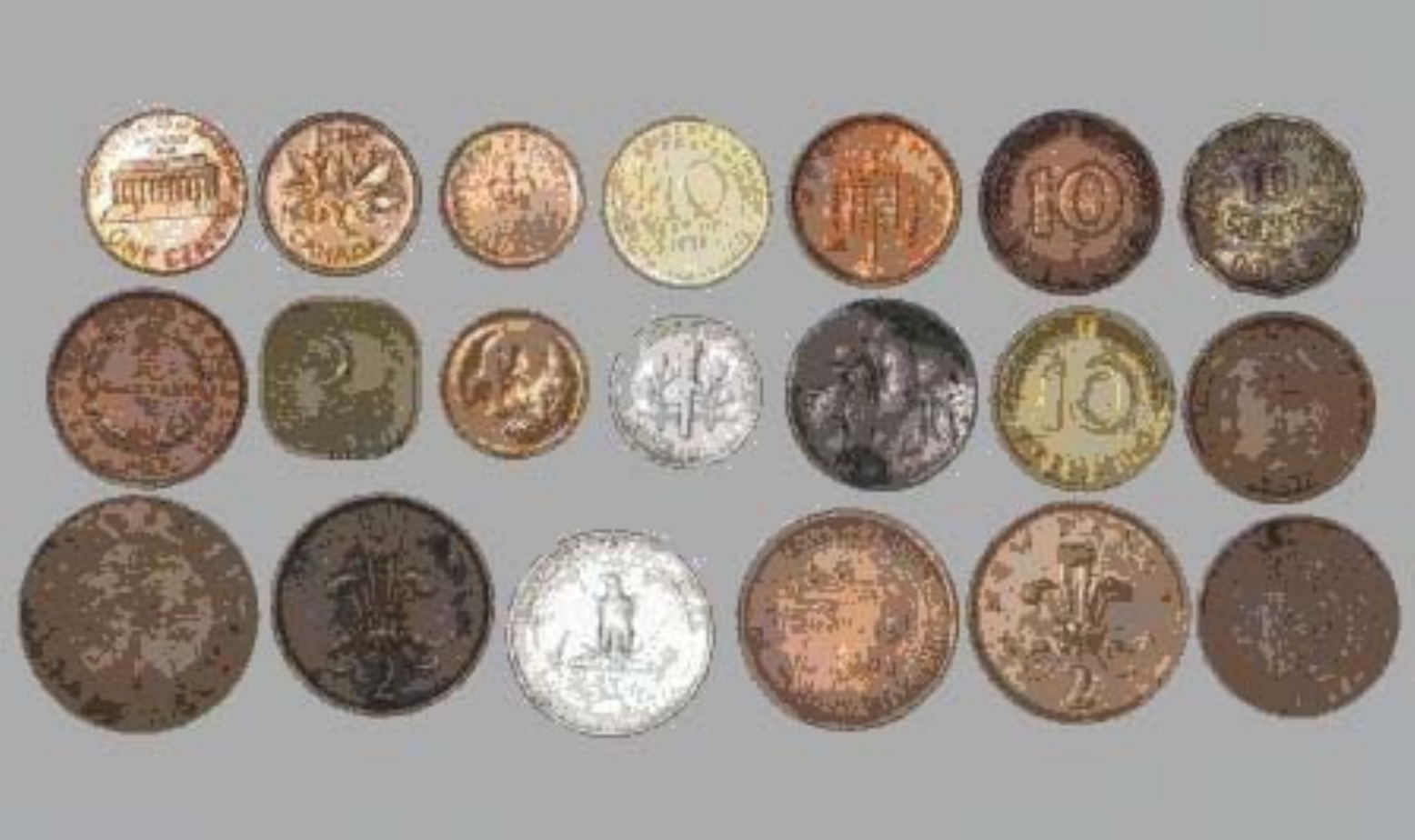}
\includegraphics[width=1.80cm]{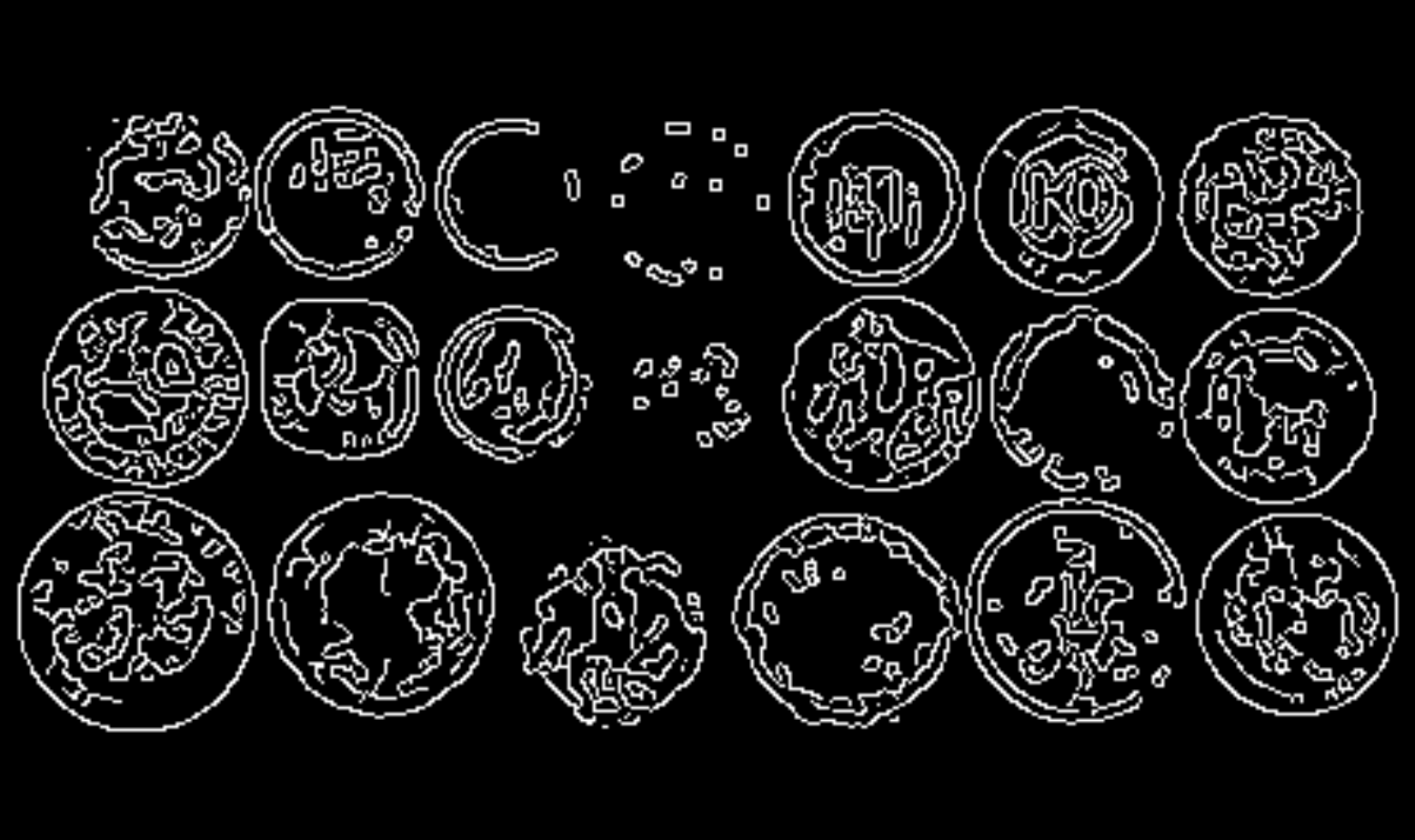}
\includegraphics[width=1.80cm]{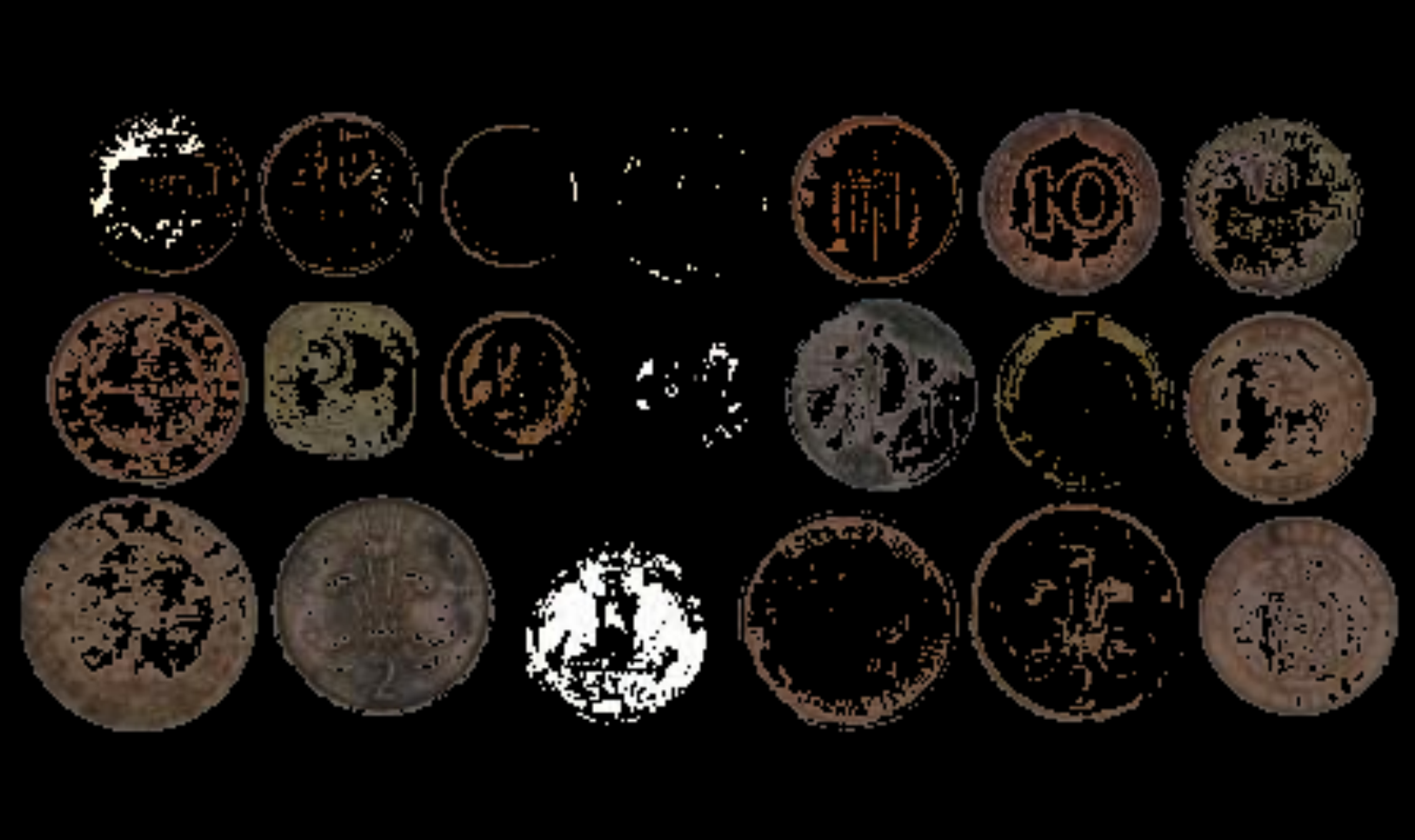}\\

\includegraphics[width=1.80cm]{Coins.pdf}
\includegraphics[width=1.80cm]{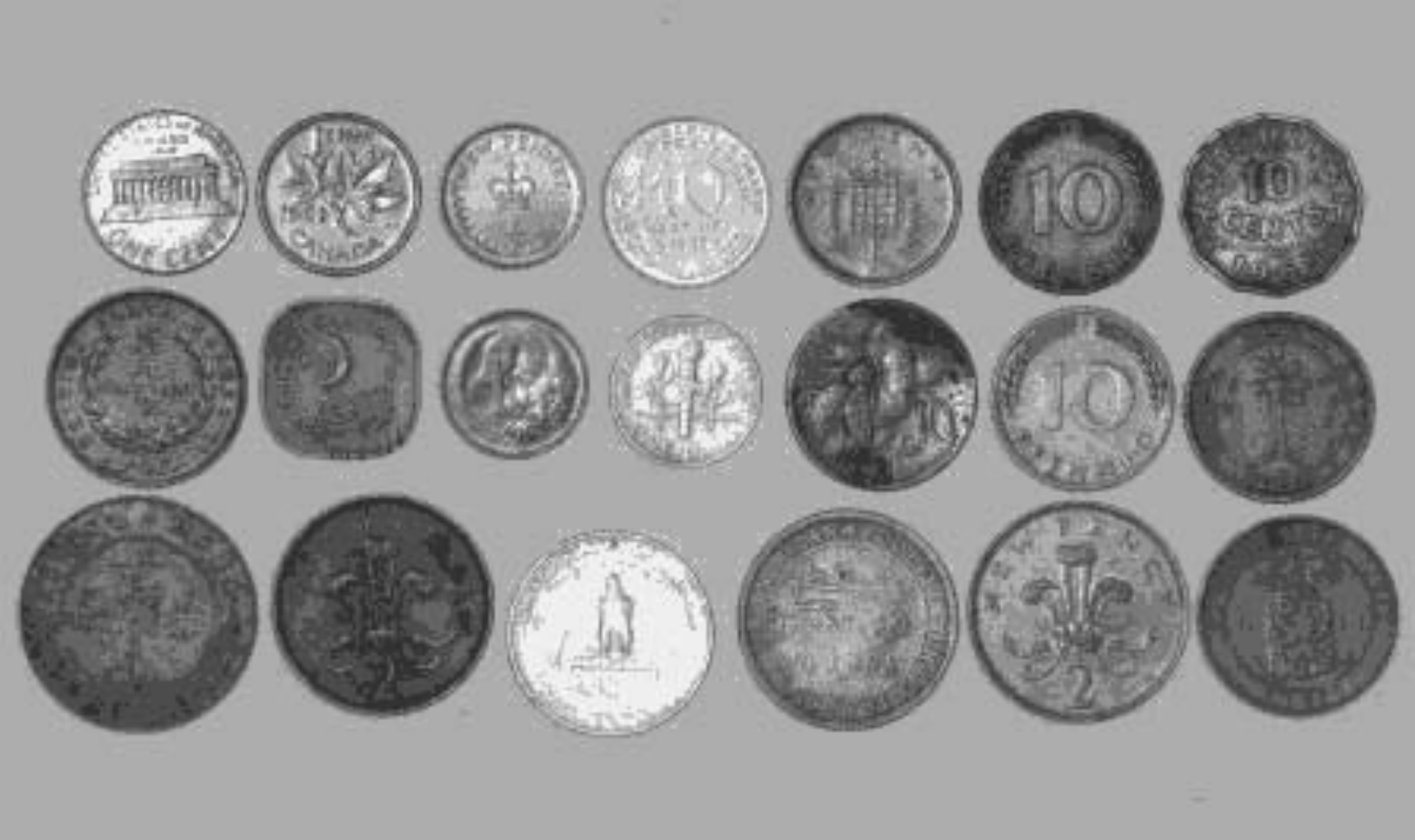}
\includegraphics[width=1.80cm]{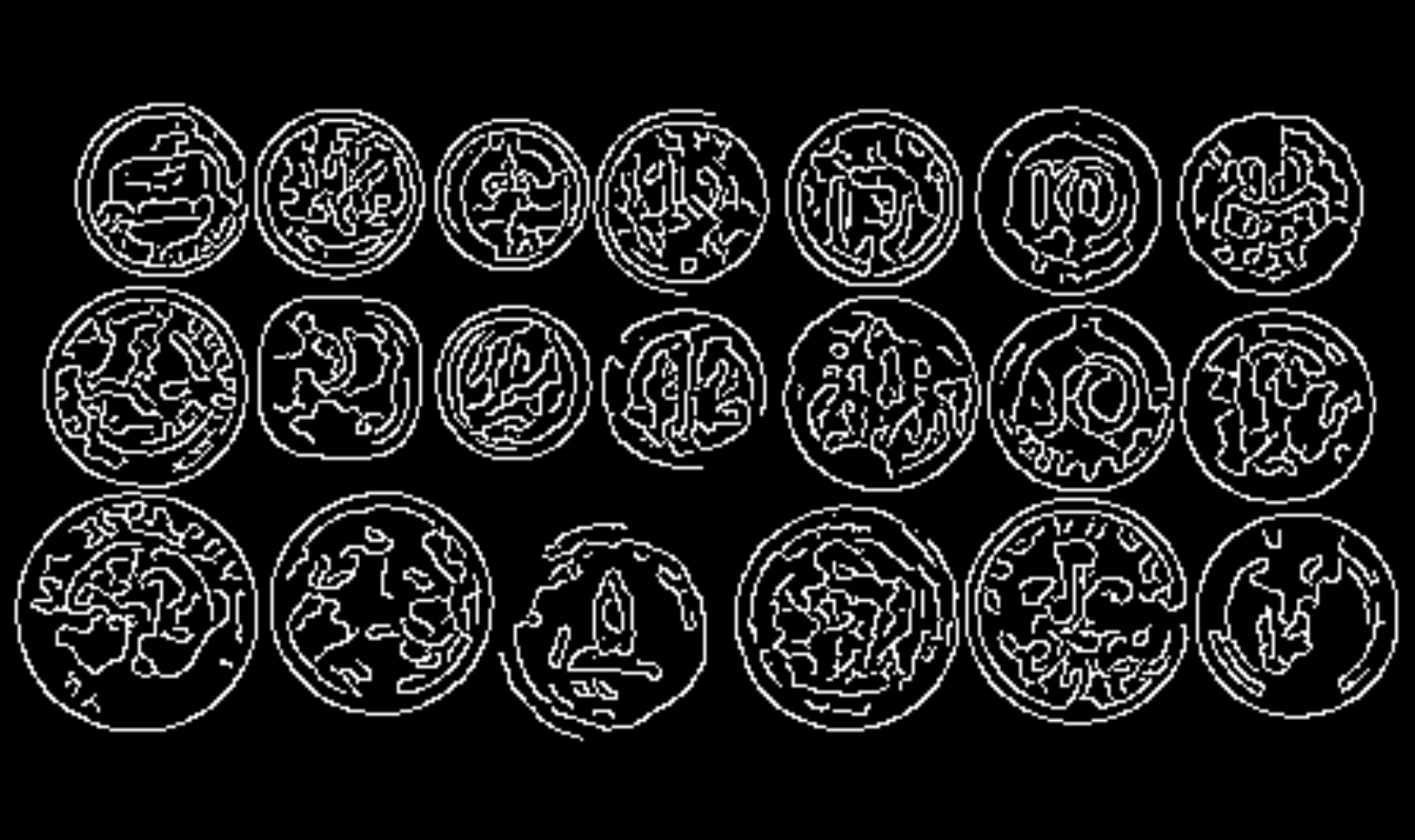}
\includegraphics[width=1.80cm]{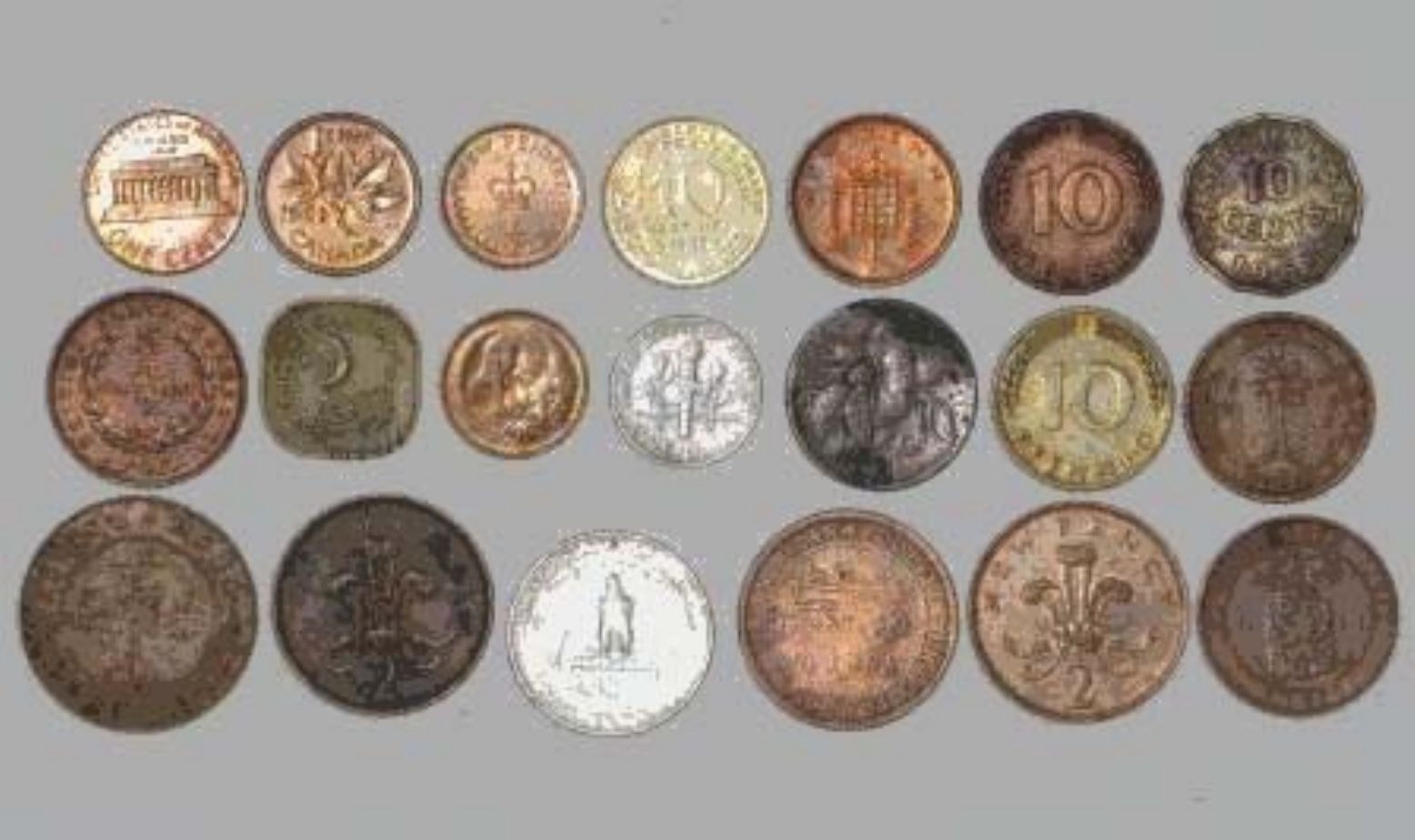}
\includegraphics[width=1.80cm]{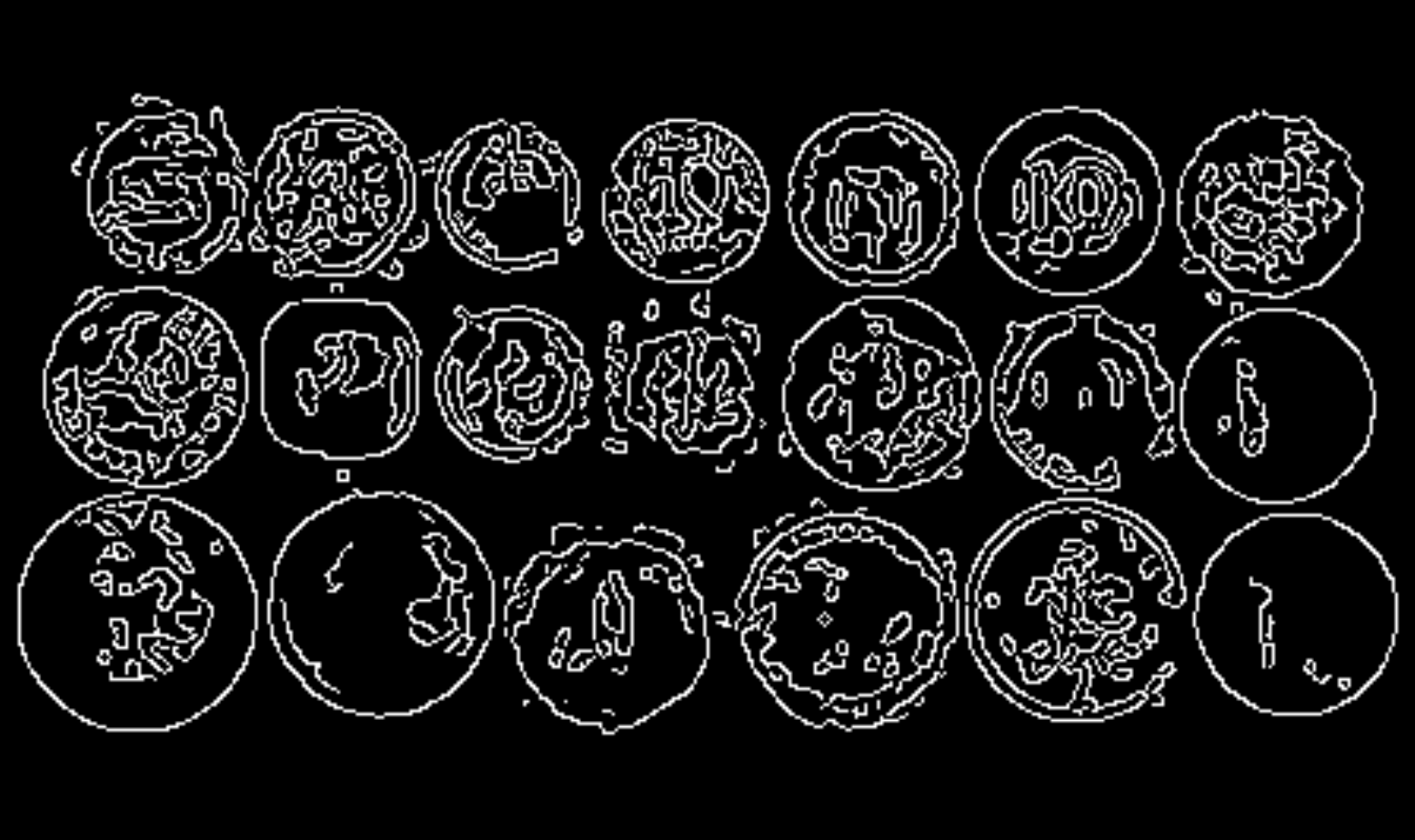}
\includegraphics[width=1.80cm]{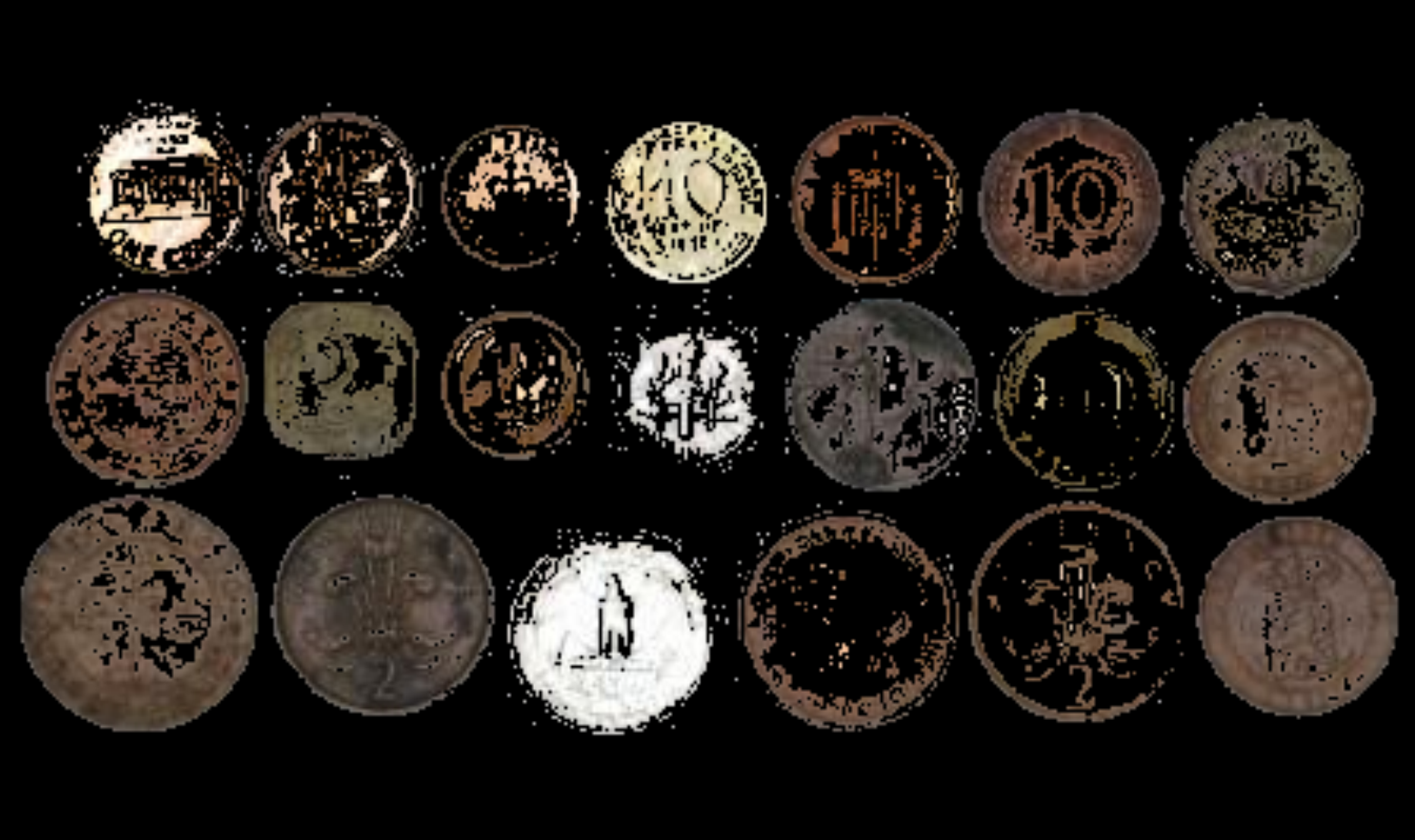}\\
\caption{Simulated output of Coins. Row 1 - Methodology-I, Row 2 - Methodology-II, Row 3 - Otsu's method. Column 1 - Orginal Image, Column 2 - Segmented Y component, Column 3 - Canny edge detection of segmented Y component, Column 4 - Segmented color image, Column 5 - Canny edge detection of extracted Y component, Column 6 - Extracted image.}

\label{fig:6}
\end{center}
\end{figure*}

\noindent $Step\, 4$: In the case of some anomaly in extracting the object precisely, vary the parameters $\kappa_1$ and $\kappa_2$, discreetly.\\
$Step\, 5$: Extract the object by choosing the right combination of segmented values.

However, the above steps are only required to be followed for one image from a particular class. For rest of the images of the same class, the value of the threshold would remain the same. Moreover, the above steps are only suggestive, and with experience user would be able to correctly guess the required threshold, $\kappa$ values, and combination of segments to be extracted to obtain the object. Furthermore, the methodologies can also be chosen by observing the approximate intensity values of the desired object. If they lie towards the weighted mean, then methodology-I should be chosen, otherwise methodology-II should be utilized.
 
The results are summarized in $Fig.\,3-8$. In $Fig.\,3-7$, column 1 represents the original image, while column 6 contains the extracted object. Column 3 and 5 are the edges of the segmented Y component ($shown\,in\,Column\,2$) and Y component of the extracted object using Canny edge detection algorithm for comparison. It can be easily seen that the edges in column 3 are focusing on the overall segmented gray scale image, while on the other hand, column 5 is showing the edges within the object which has been extracted, providing no information about the background. Hence, the separated object can be easily perceived from column 5.

Canny edge detection \cite{32} is optimal compared to other edge detection techniques like Sobel, Prewitt and Roberts – to name a few. Apart from being optimal, Canny edge detection method is also robust against noise which is suitable for noisy test images used to evaluate the algorithms. However, in this paper, the task of edge detection is to visualize the effectiveness of the segmentation process to extract the desired object. It does not have any involvement in thresholding, segmentation or/and object separation.

For each image in $Fig.\,3-7$, different threshold levels are taken depending upon the requirement. If the image background lies towards the weighted mean of the histogram, and overlaps with some part of the object to be extracted, then higher number of threshold levels are required to separate the two, more effectively with the help of methodology-I. On the other hand, if the background pixels lie at the ends of the histogram, mixed with some part of the object, then higher number of thresholds will be required using methodology-II.

\begin{figure*}[ht]
\begin{center}
\includegraphics[width=2.70cm]{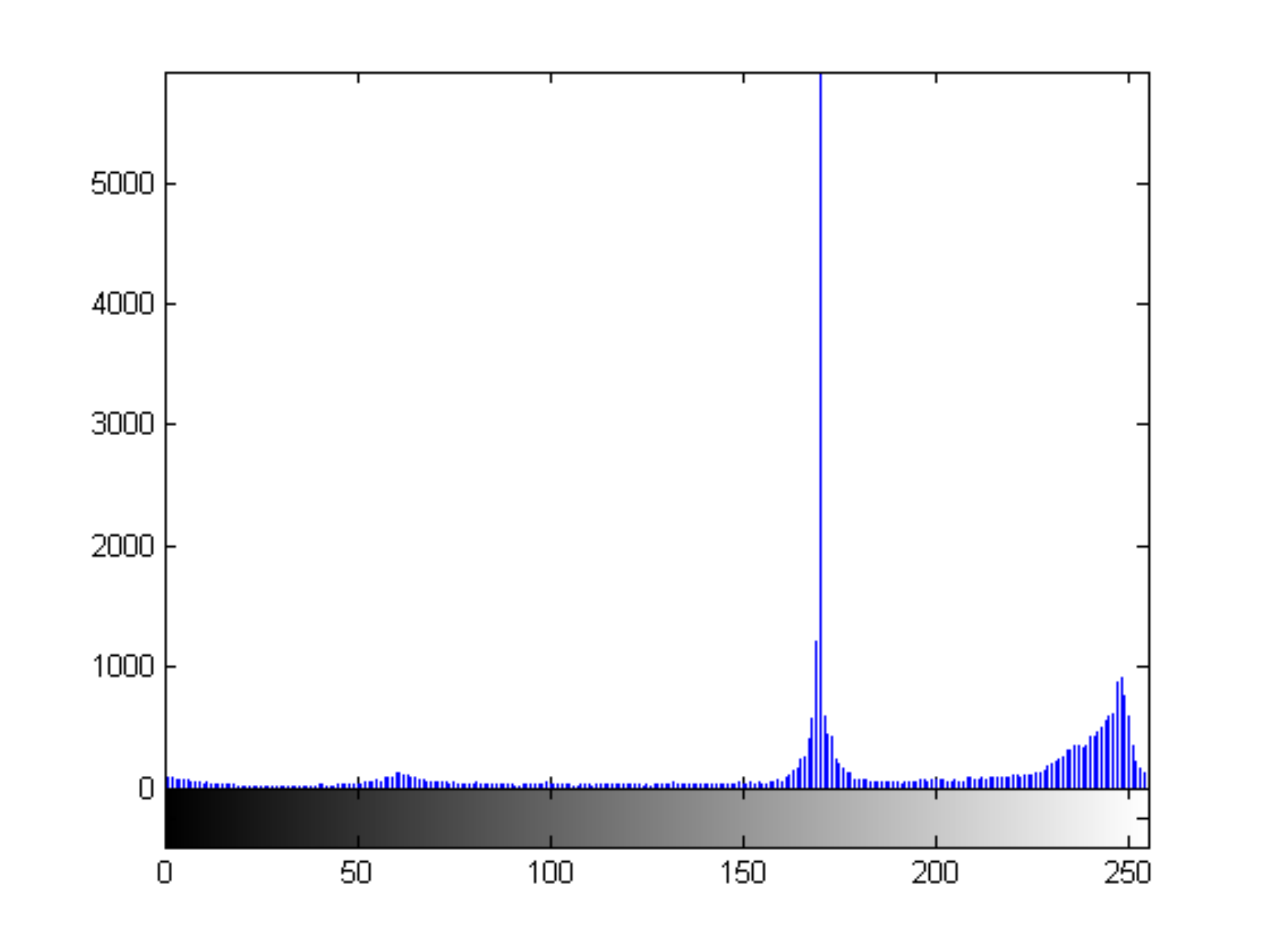}
\includegraphics[width=2.70cm]{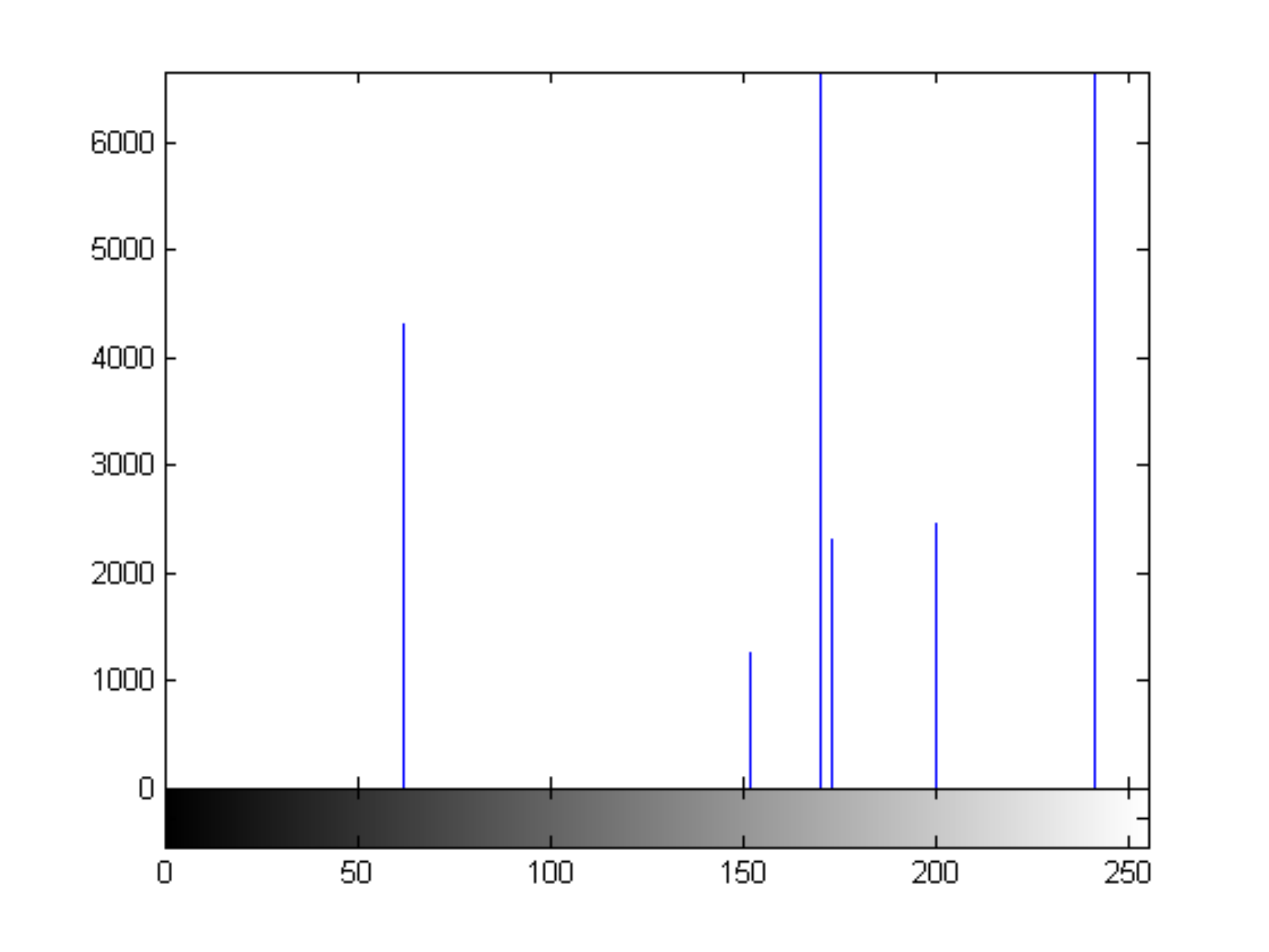}
\includegraphics[width=2.70cm]{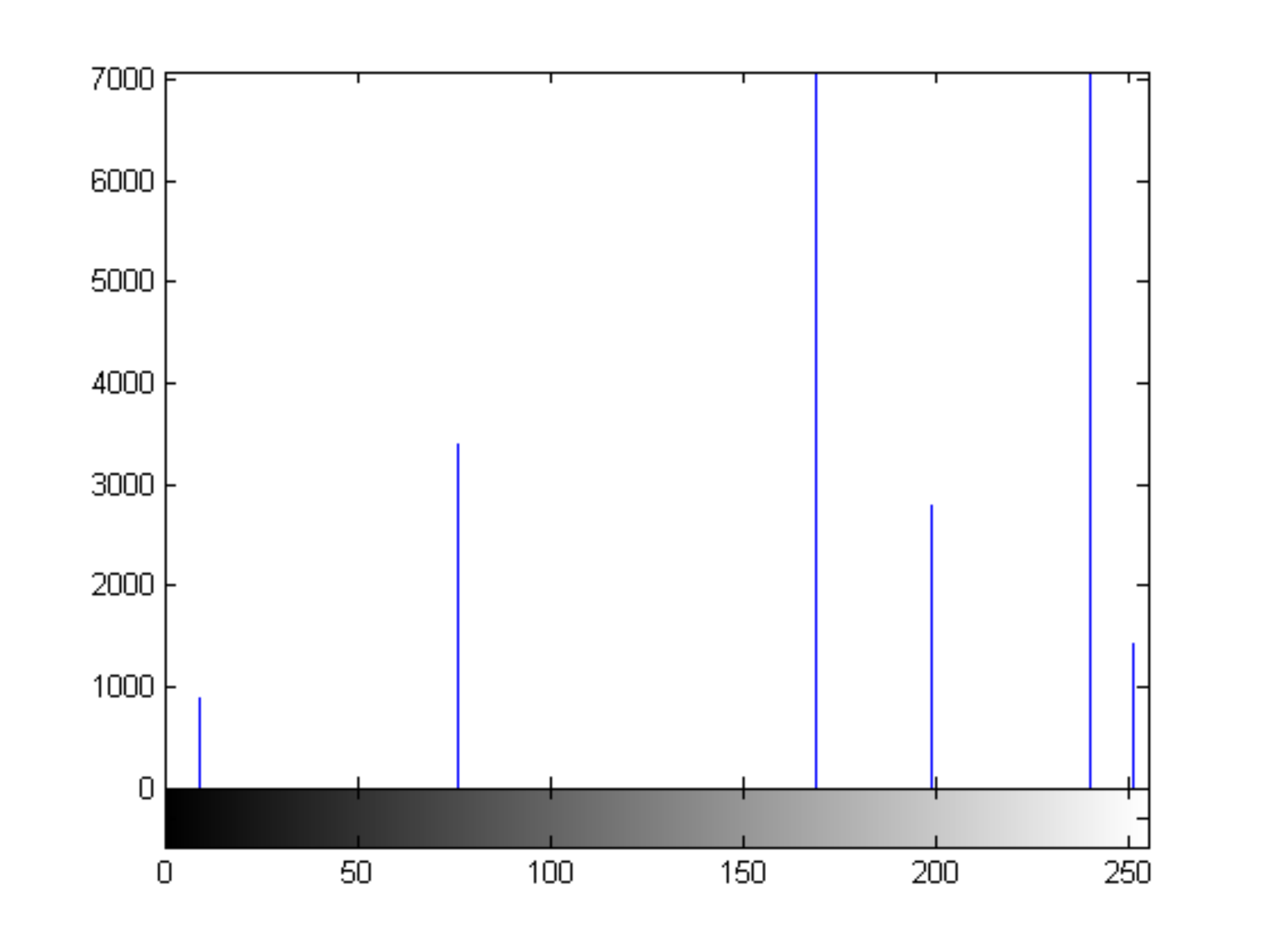}
\includegraphics[width=2.70cm]{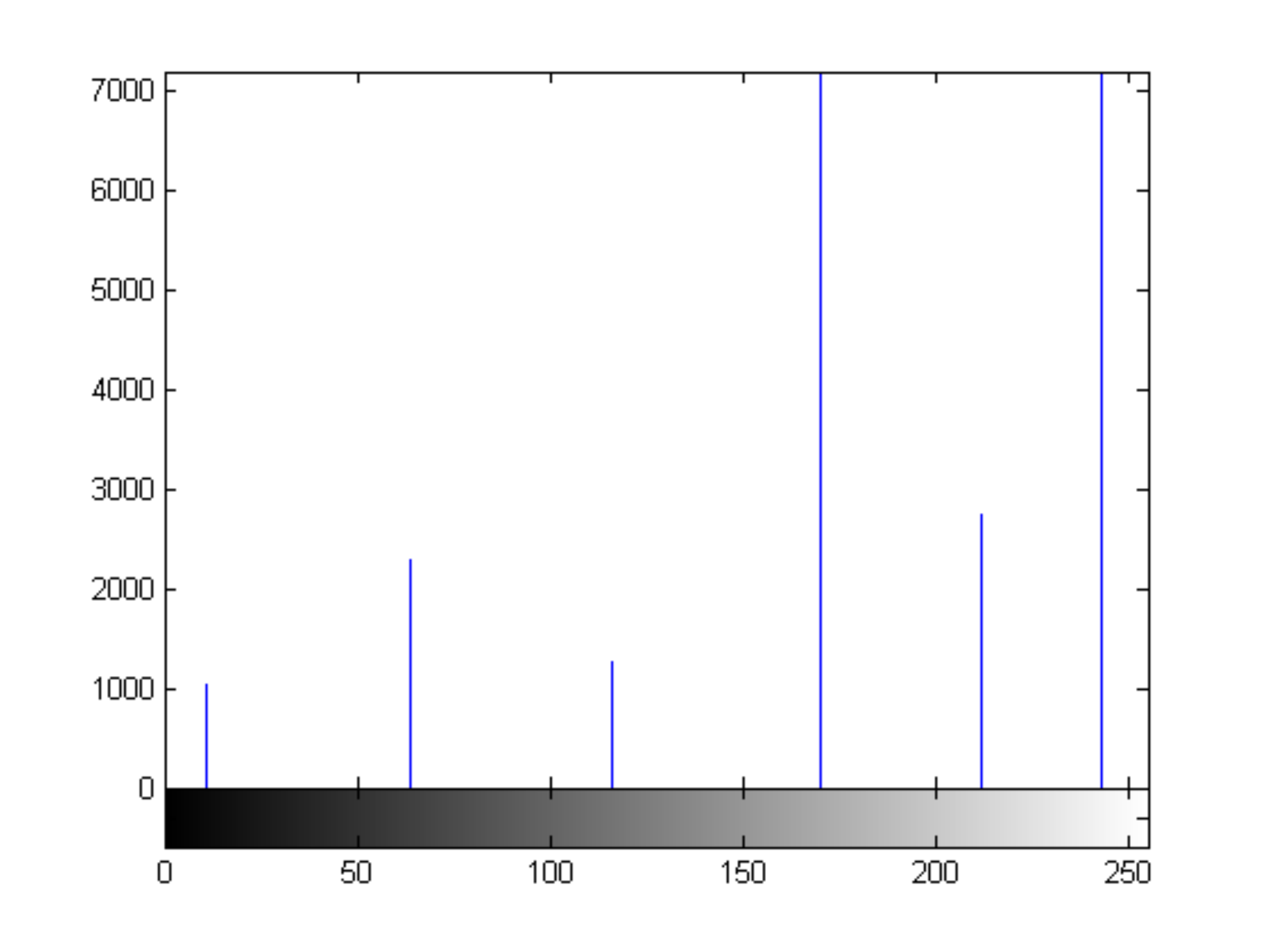}\\

\includegraphics[width=2.70cm]{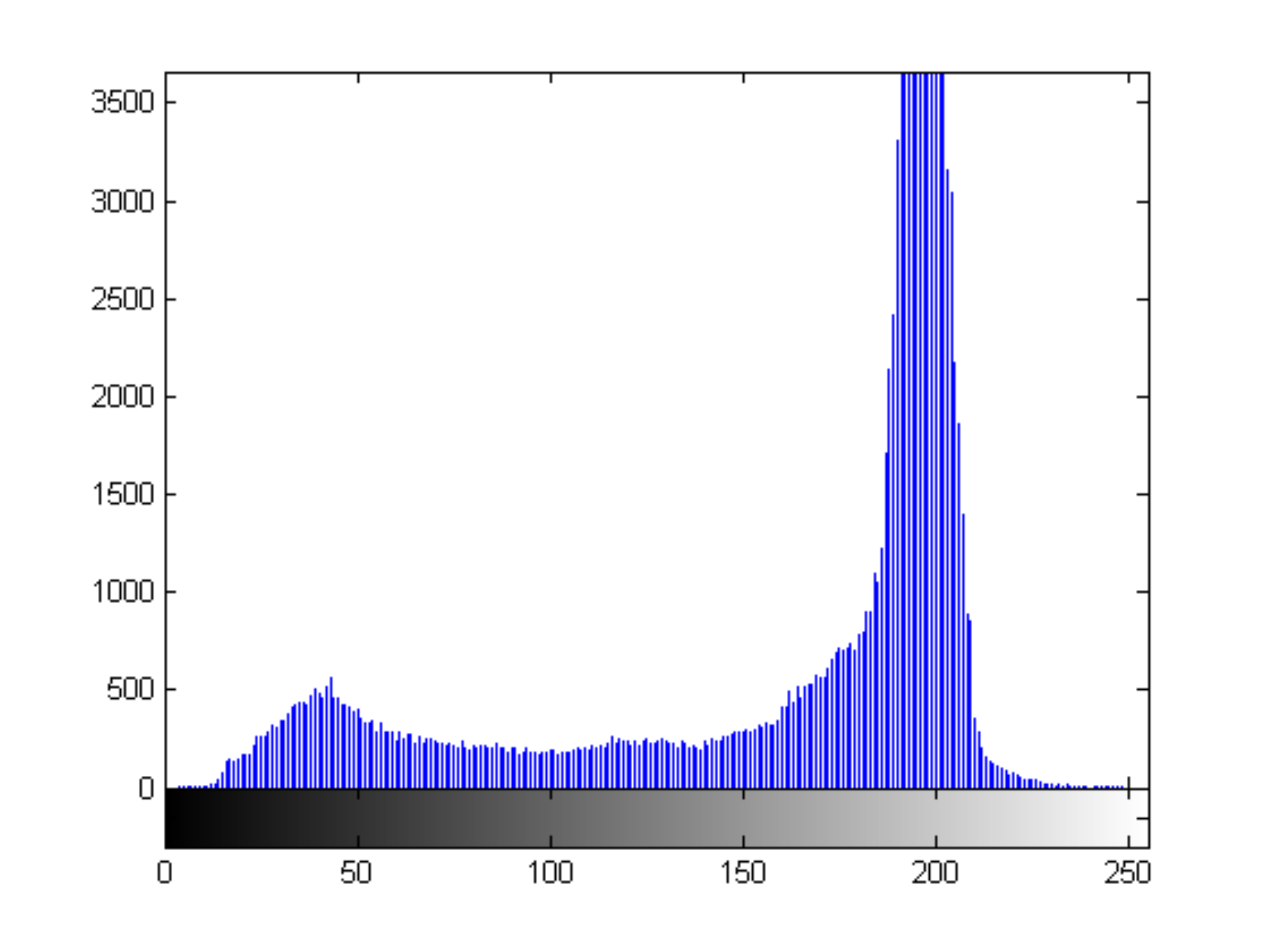}
\includegraphics[width=2.70cm]{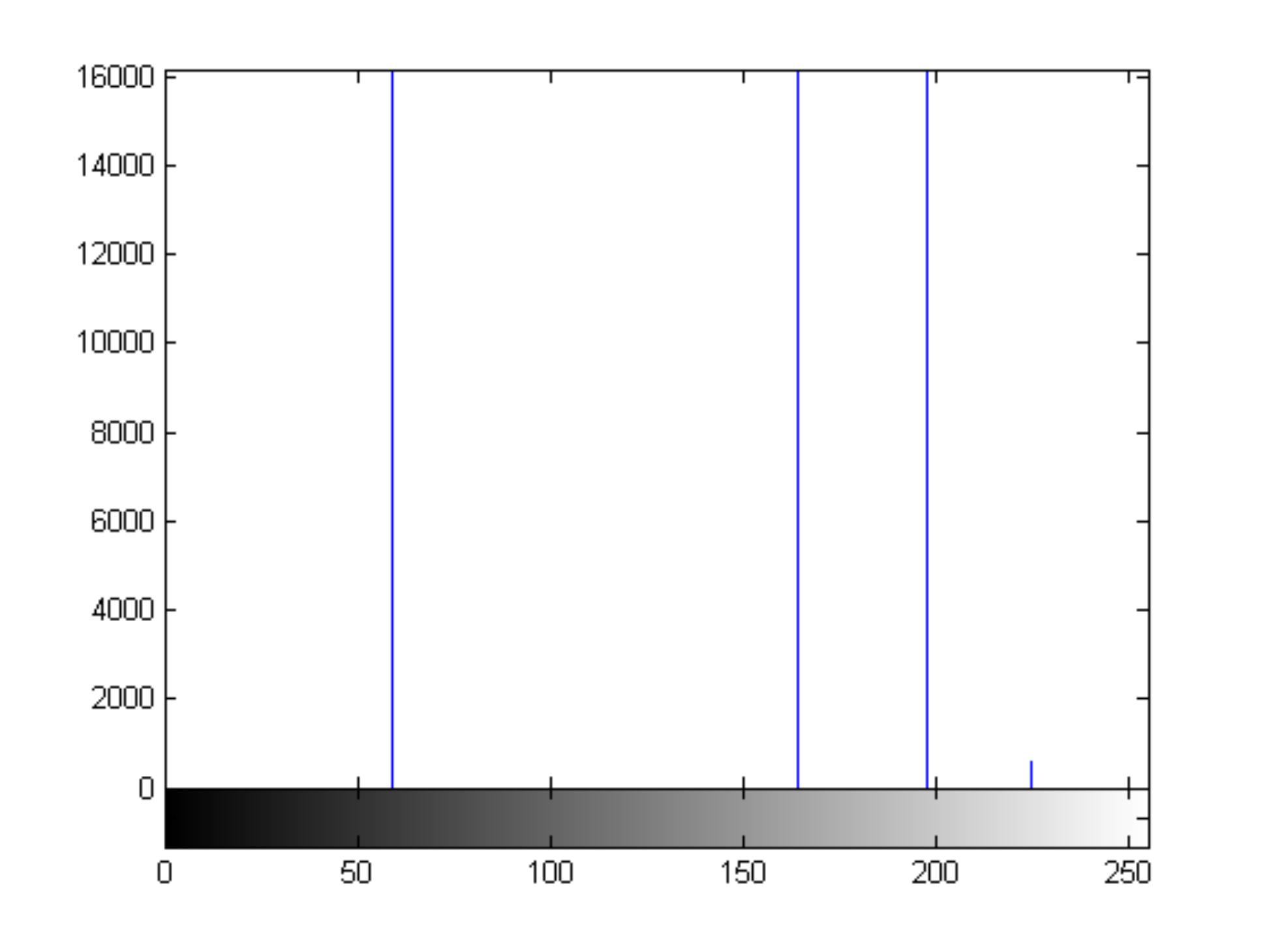}
\includegraphics[width=2.70cm]{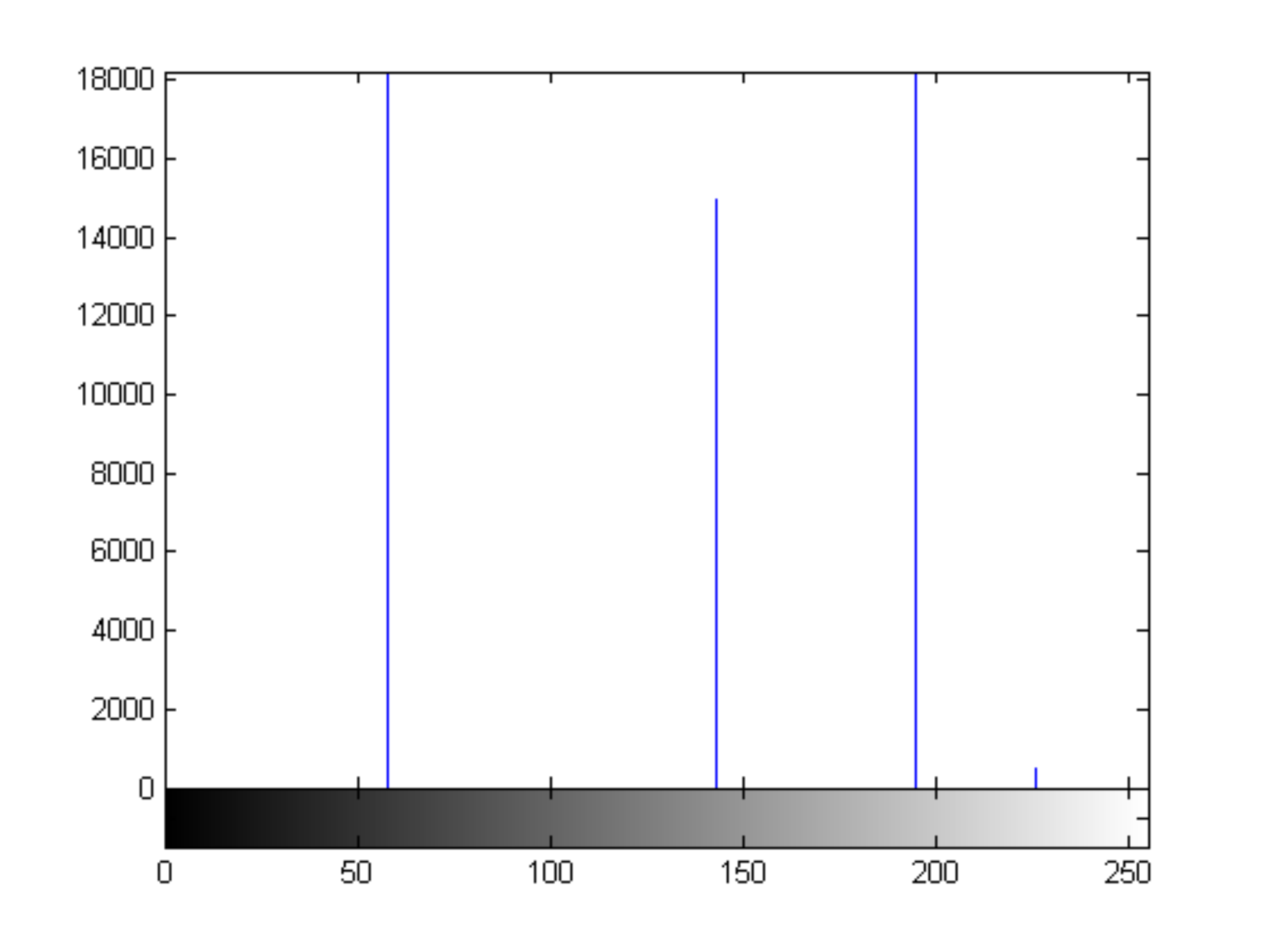}
\includegraphics[width=2.70cm]{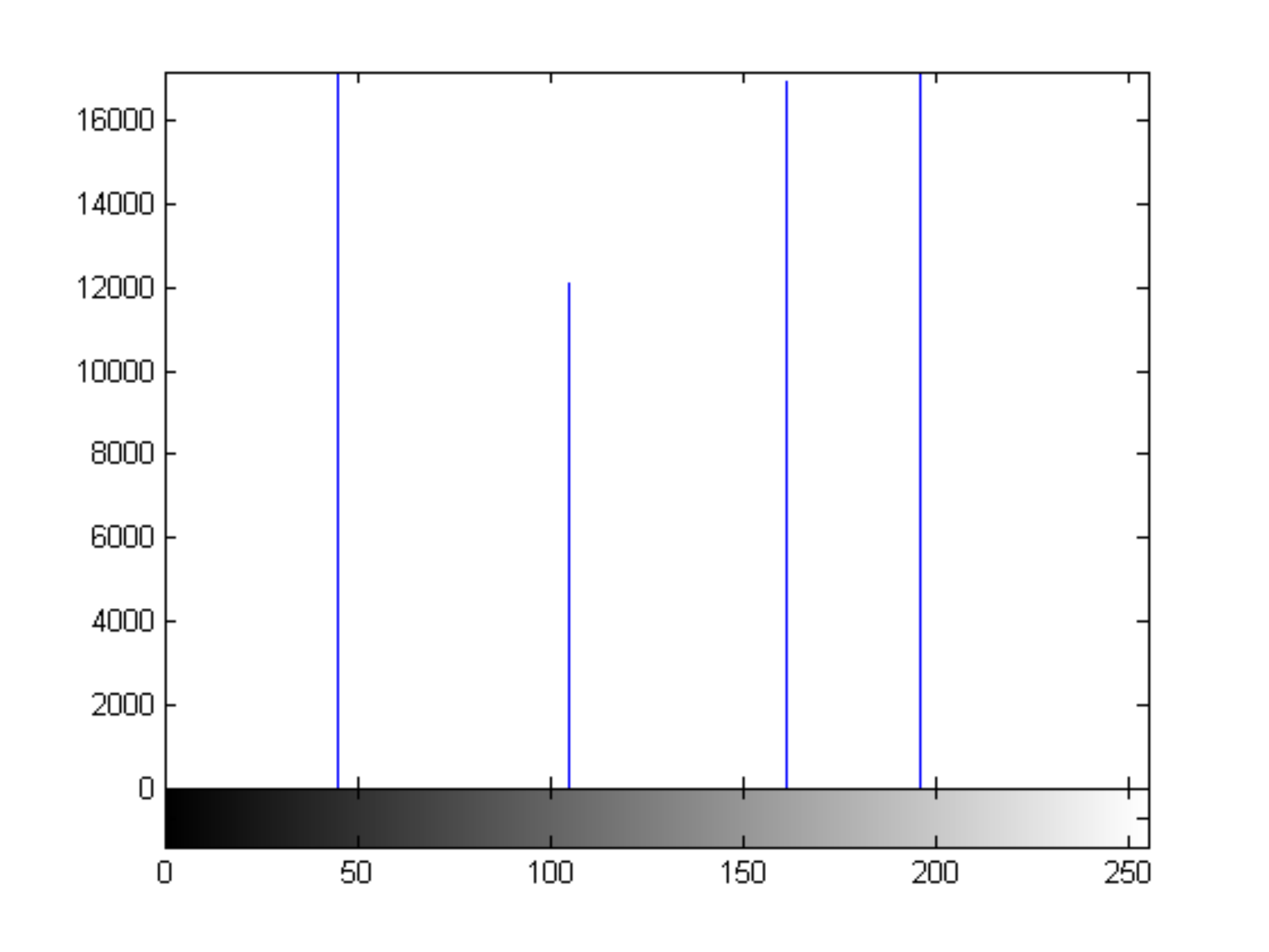}\\

\includegraphics[width=2.70cm]{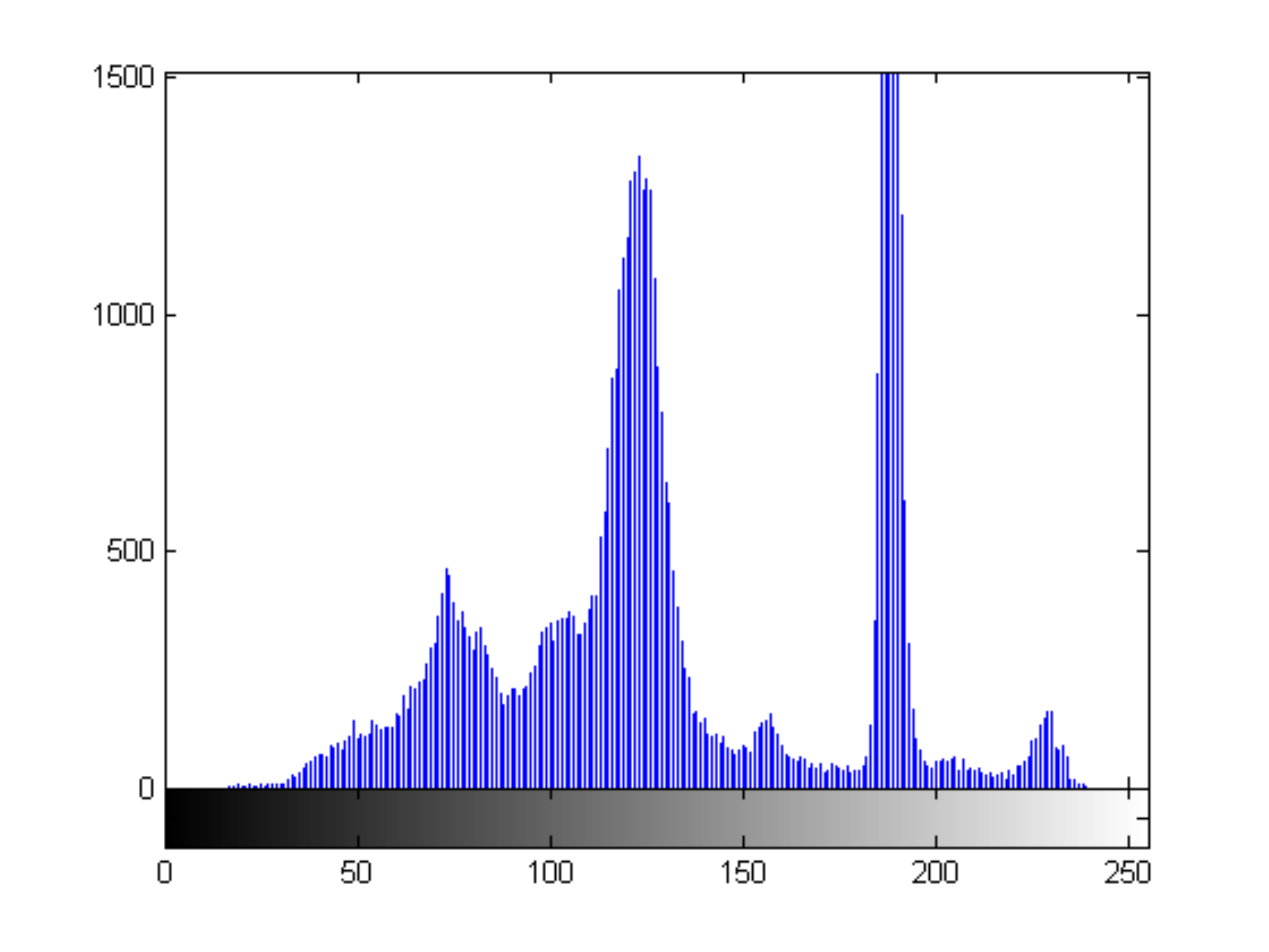}
\includegraphics[width=2.70cm]{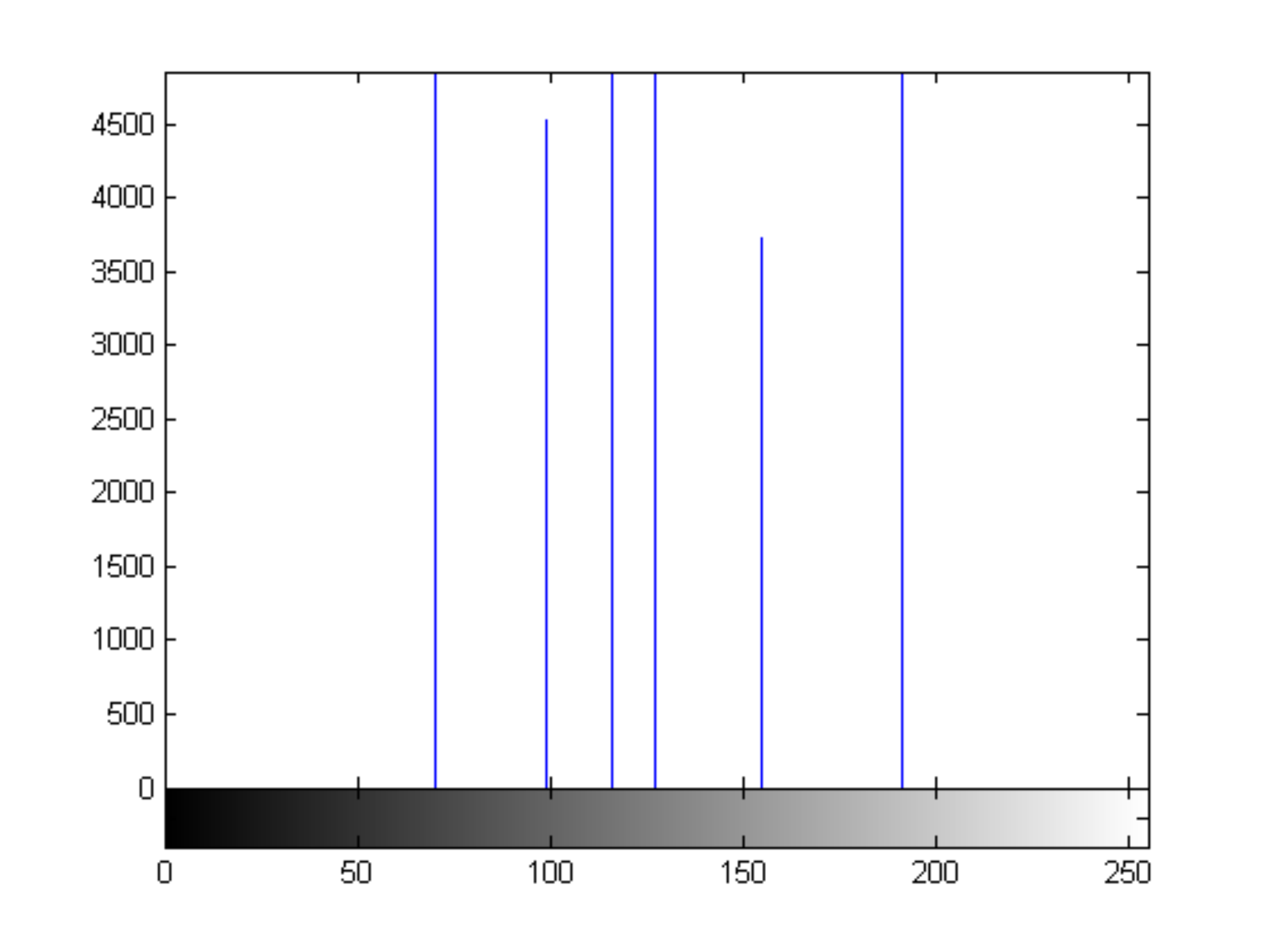}
\includegraphics[width=2.70cm]{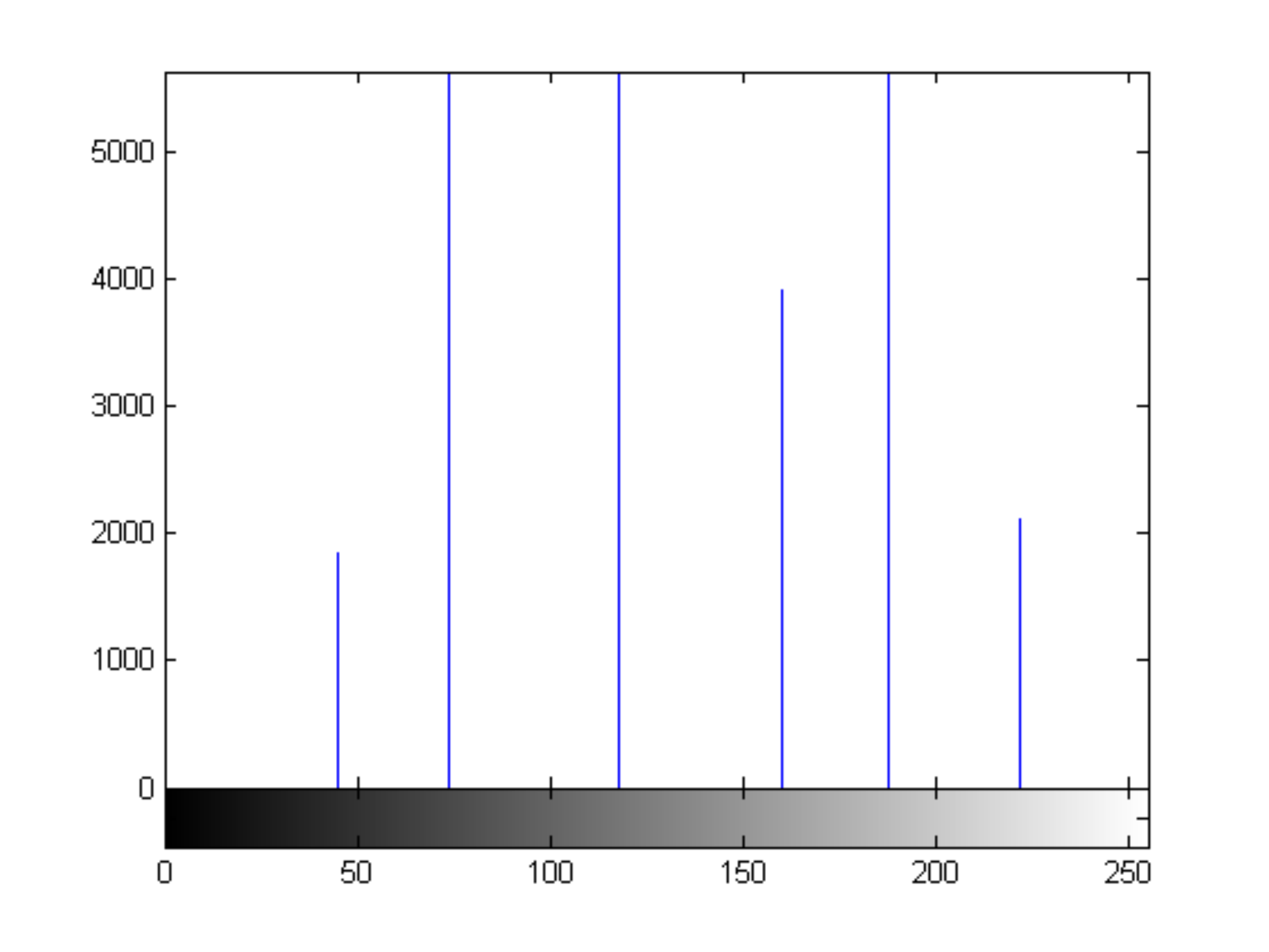}
\includegraphics[width=2.70cm]{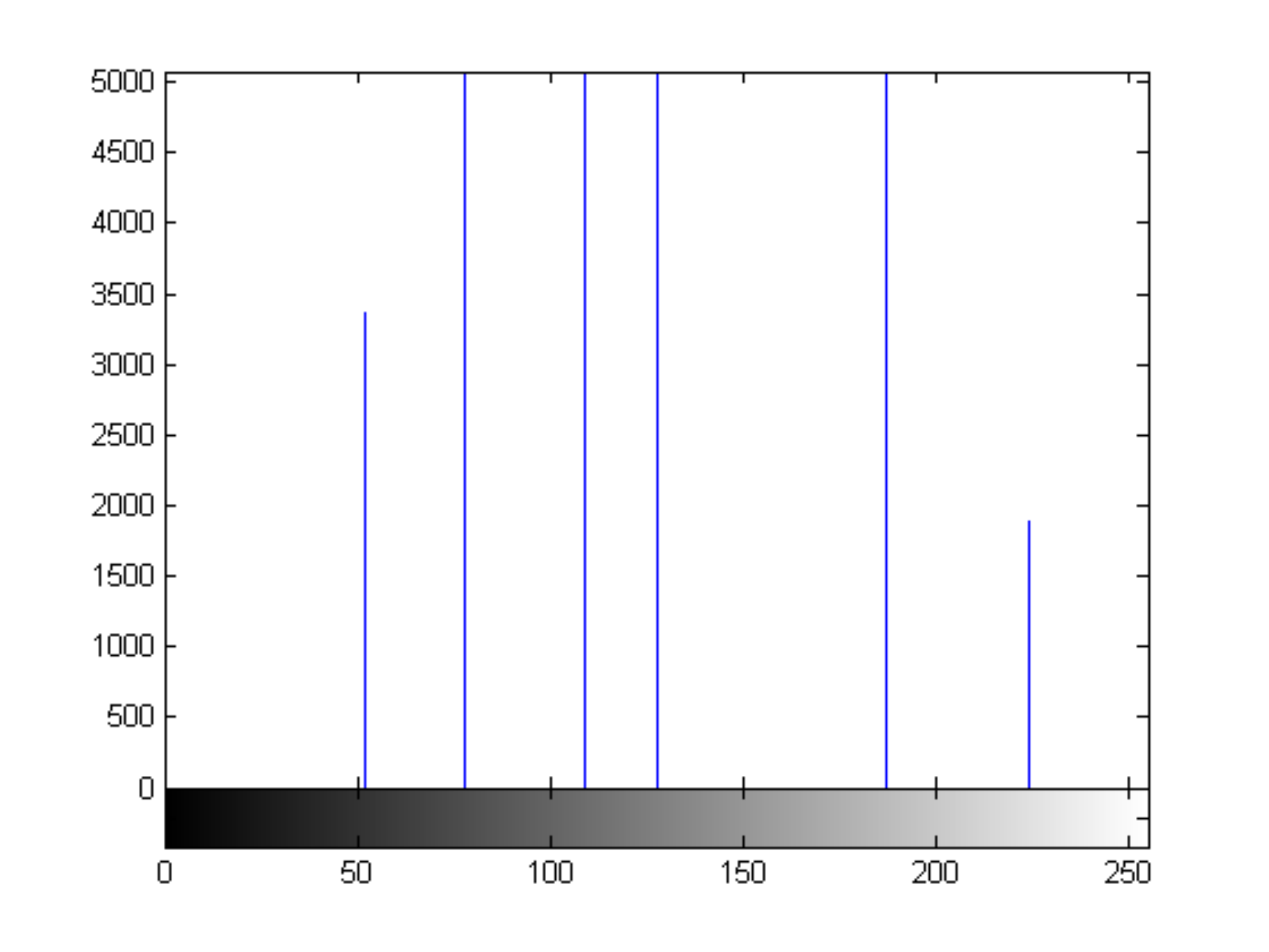}\\

\includegraphics[width=2.70cm]{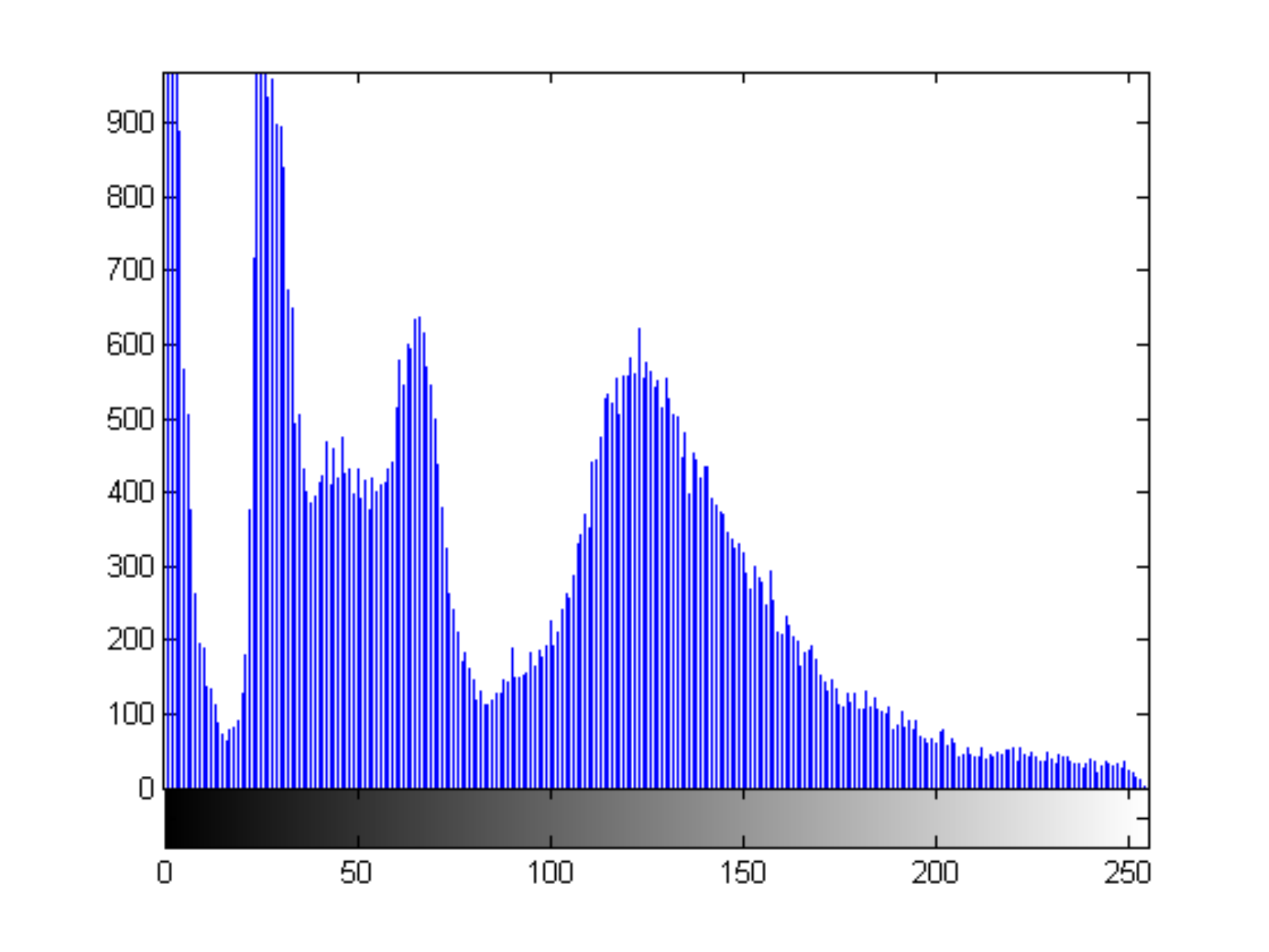}
\includegraphics[width=2.70cm]{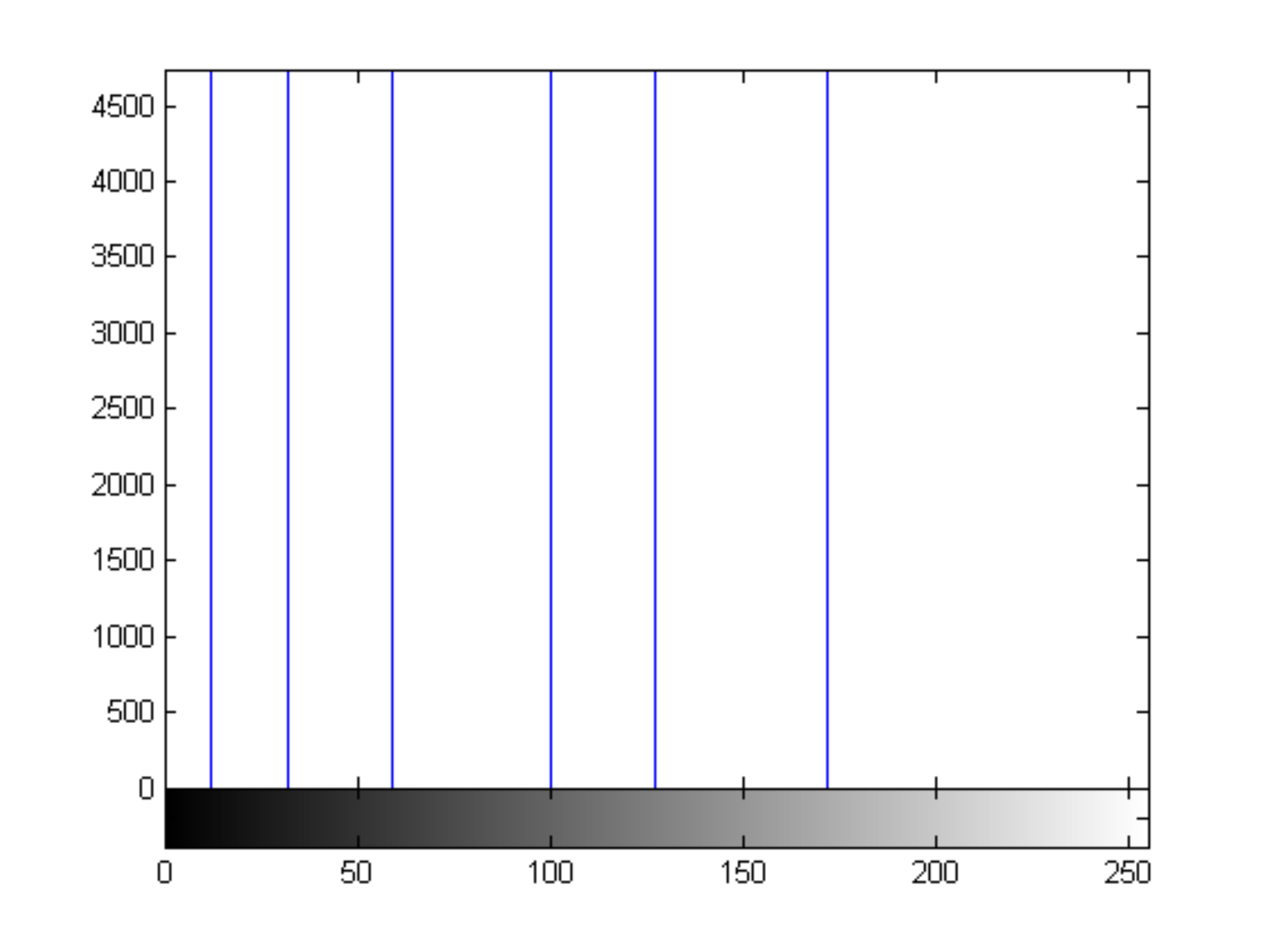}
\includegraphics[width=2.70cm]{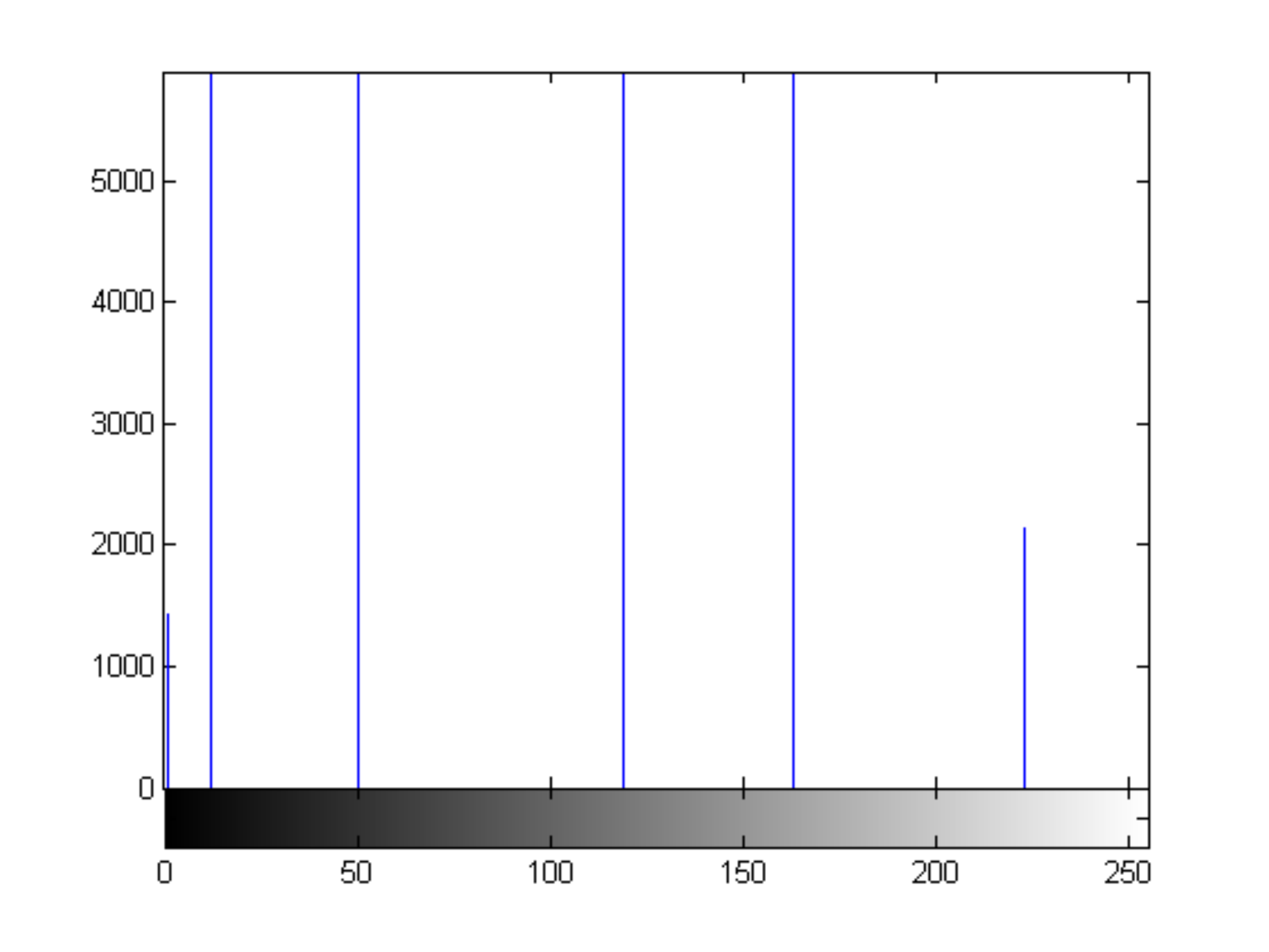}
\includegraphics[width=2.70cm]{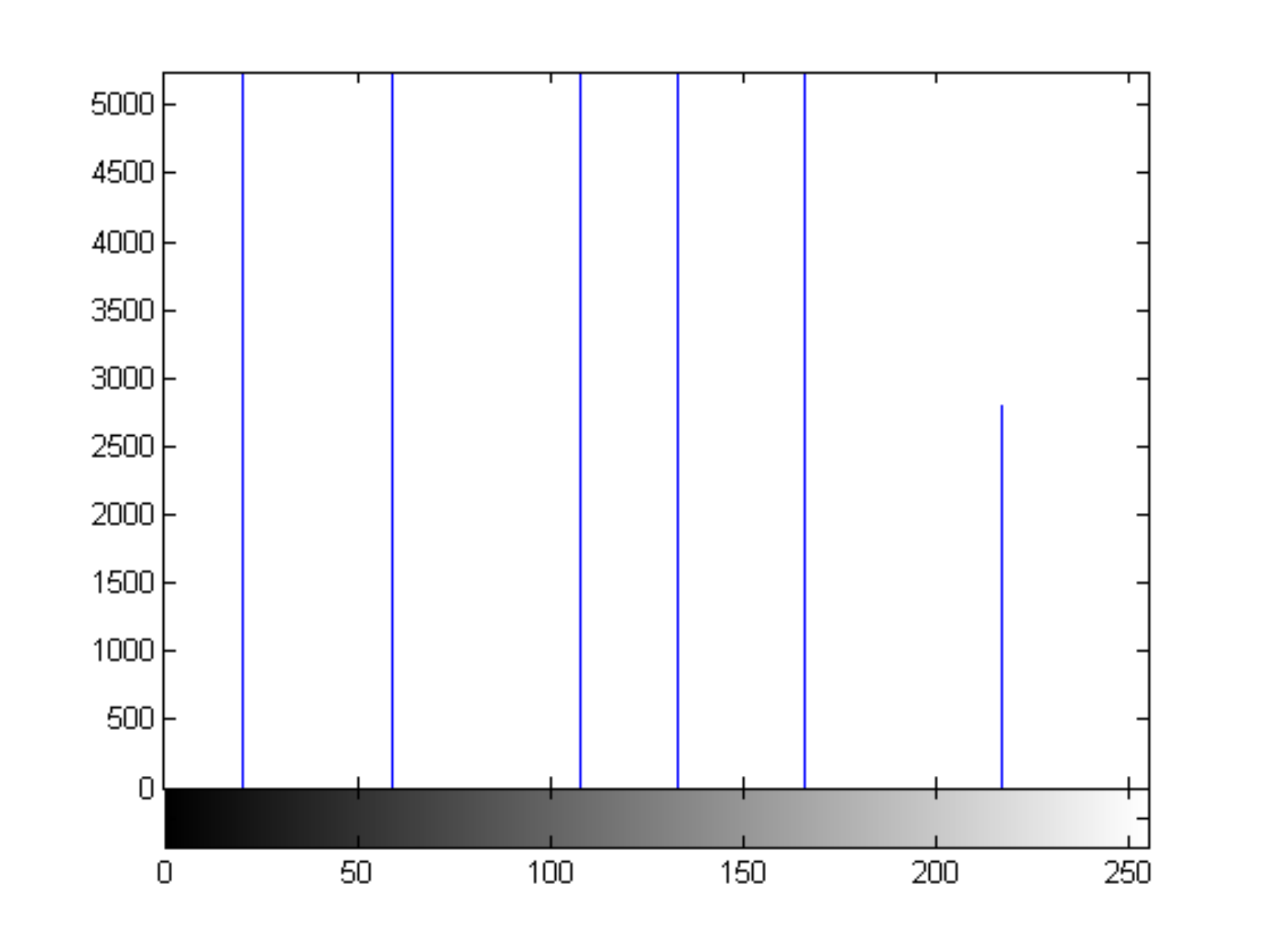}\\

\includegraphics[width=2.70cm]{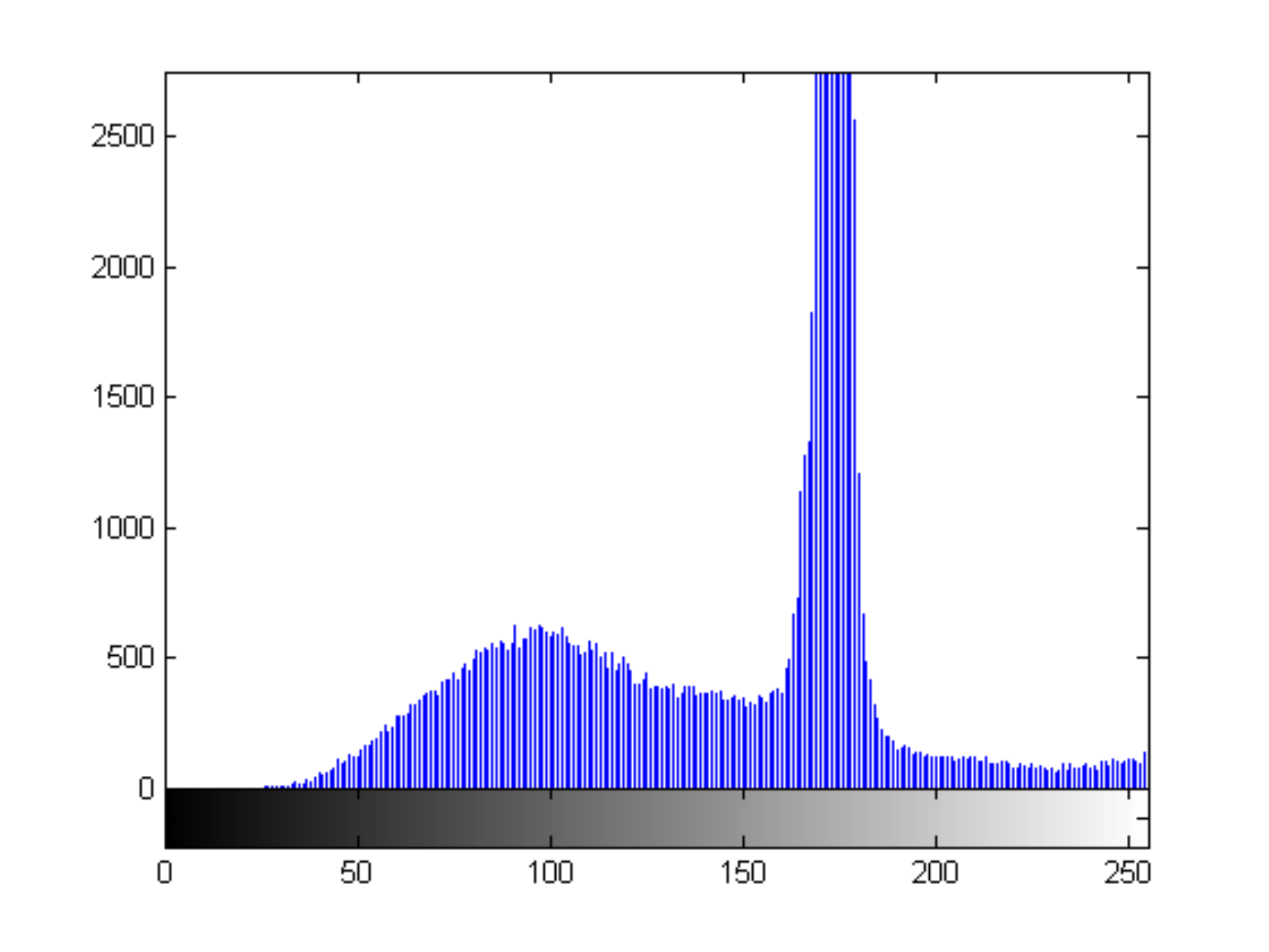}
\includegraphics[width=2.70cm]{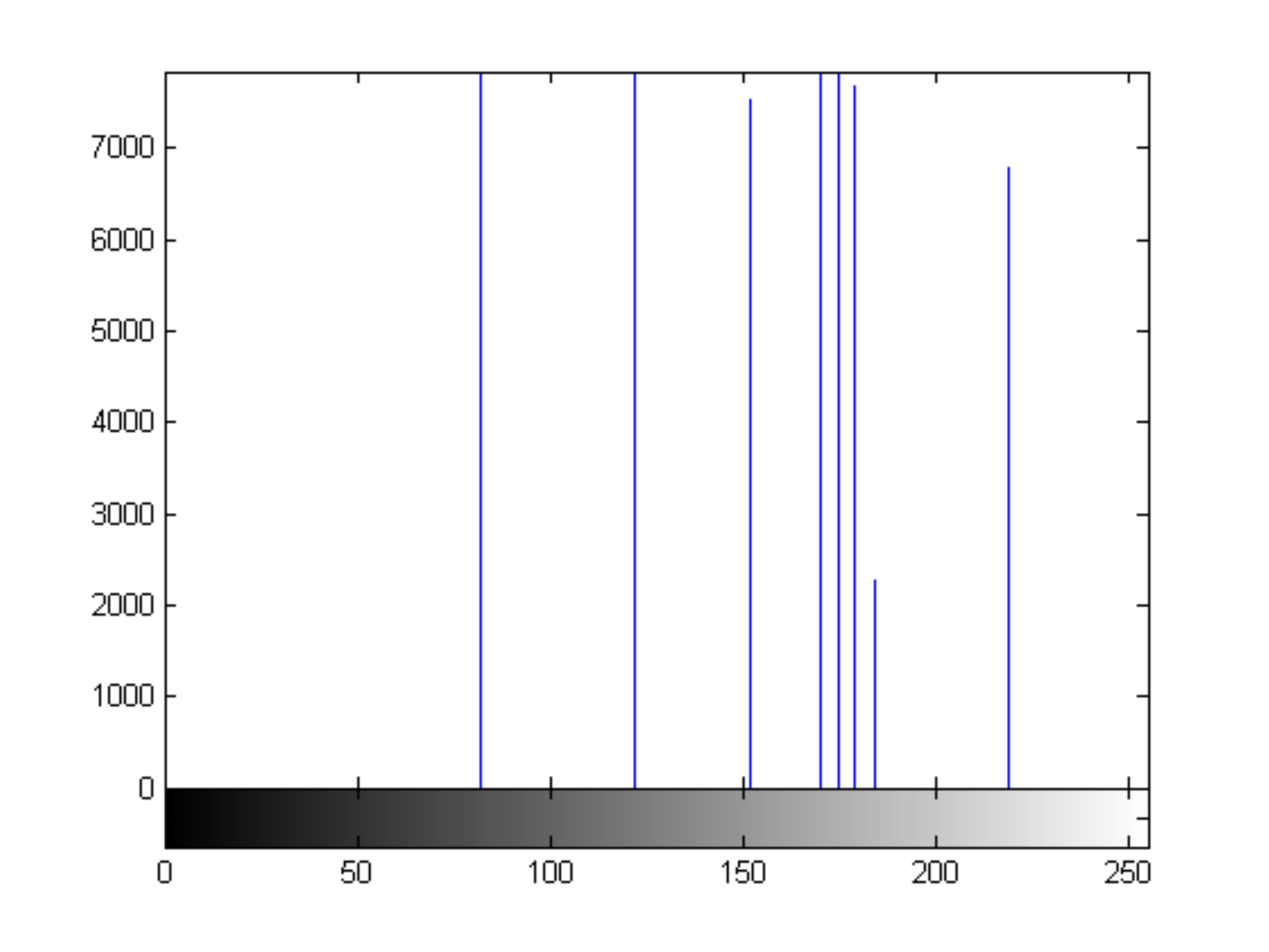}
\includegraphics[width=2.70cm]{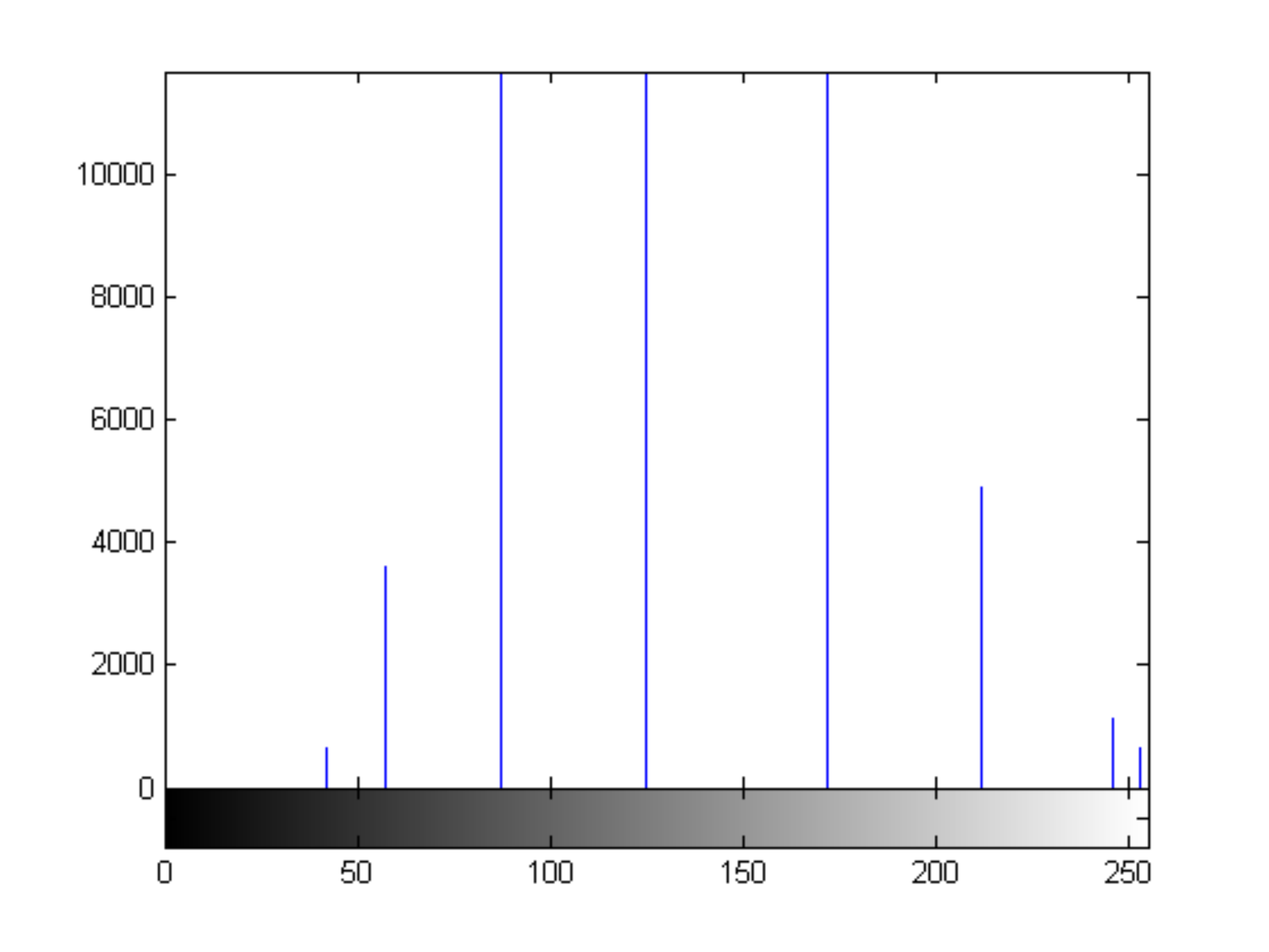}
\includegraphics[width=2.70cm]{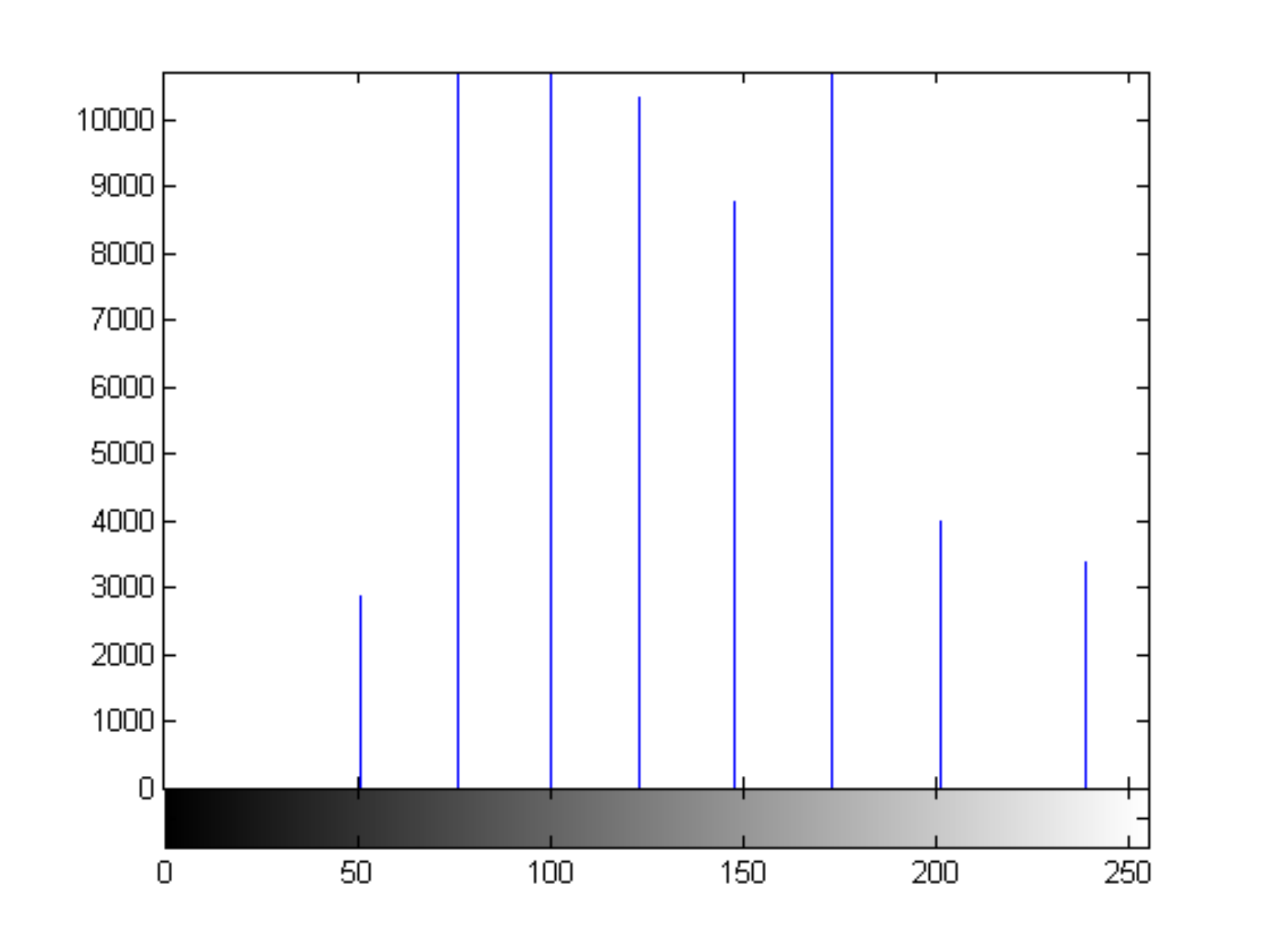}\\

\caption{Histograms of Y component at threshold levels mentioned in Table 2. Row 1 - Aeroplane, Row 2 - Eagle, Row 3 - House, Row 4 - Coin, Row 5 - Coins. Column 1 - Orginal Image, Column 2 - Methodology-I, Column 3 - Methodology-II, Column 4 - Otsu's Method.}
\label{fig:4}
\end{center}
\end{figure*}

\label{sec:5}
\begin{table*}[h]
\caption{The segmented and extracted values of different images at suitable threshold level using methodology-I, methodology-II, and Otsu's method.}
\label{tab:1}
\begin{center}
\begin{tabular}{cccc}
\hline
Image Name & Thresholds & Segmented Values & Extracted Values\\
\hline
\bf{Methodology-I} & & & \\
Airplane & 5 & 62 152 170 173 200 241 & 62 152 200 241\\
Eagle & 3 & 59 164 198 225 & 59\\
House & 5 & 70 99 116 127 155 191 & 70 99 116 127 191\\
Coin & 5 & 12 32 59 100 127 172 & 100 127 172\\
Coins & 7 & 82 122 152 170 175 179  184 219 & 82 122 152 184 219\\
\hline
\bf{Methodology-II} & & & \\
Airplane & 5 & 9 76 169 199 240 251 & 9 76 240 251\\
Eagle & 3 & 58 143 195 226 & 58\\
House & 5 & 45 74 118 160 188 222 & 45 74 118 160 222\\
Coin & 5 & 1 12 50 119 163 223 & 119 163 223\\
Coins & 7 & 42 57 87 125 172 212 246 253 & 42 57 87 246 253\\
\hline
\bf{Otsu's method} & & & \\
Airplane & 5 & 11 64 116 170 212 243 & 11 64 212 243\\
Eagle & 3 & 45 105 161 196 & 45\\
House & 5 & 52 78 109 128 187 224 & 52 78 109 128 224\\
Coin & 5 & 20 59 108 133 166 217 & 133 166 217\\
Coins & 7 & 51 76 100 123 148 173 201 239 & 51 76 100 201 239\\
\hline

\end{tabular}
\end{center}
\end{table*}

$Table\,2$ states the number of thresholds, segmented values and extracted values for the results shown in $Fig.\,3-7$. Looking at the segmented values of the two proposed methodologies and Otsu's method, all the algorithms' inclination and the location towards different parts of histogram can be clearly observed. Comparison of segmented values shows that methodology-I converges towards the weighted mean of histogram, while methodology-II diverges to the ends of the histogram. Otsu's method follows its own approach of maximizing class variance, neither converging nor diverging towards any part of the histogram. Segmented values of $Table\,2$ are displayed in $Fig.\,8$ for each algorithm, along with the original histogram of Y component for the test images. Moreover, the convergence and divergence observed in $Table\,2$ can easily be seen in $Fig.\,8$. 

In addition, $Table\,2$ demonstrates the need for minimal number of threshold required to extract the desired object. It can be seen through the values extracted from the segmented values that the object would not have been separated from the rest of the image if the threshold chosen was less than the current threshold. As an example, for the image 'Airplane', the extracted values are 62, 152, 200, 241 from the segmented values 62, 152, 170, 173, 200, 241 ($Table\,2$). The threshold level 5 ensures that the pixel values (152, 170) and (173, 200), which would be represented by single value at threshold 3 in methodology-I, are making a distinction between the object and the background pixel i.e., segmented pixel values 152 and 200 represent the object, while the values 170 and 173 represent the background. In the image 'Eagle', only segmented pixel value 59 is required to separate the object from the background, and therefore threshold level 3 is sufficient to do the job. The choice of number of threshold levels can be better achieved by the psycho$-$visual discretion of user, hence proving the utility of a semi$-$automated approach. For the case, when methodology-II is adopted for images 'House' and 'Coin', threshold level 5 is performing the similar task as it did for images using methodology-I but in a converse manner (for the end of the histogram pixels). Having a higher number of threshold levels than required will not affect the object separation process, but a smaller number of threshold levels will. This is because higher number of thresholds will result in over segmentation of object or background, which can be negated by taking or rejecting more segment values respectively. In the case of less threshold levels, under segmentation would occur which may not be able to separate background and object pixels.

\subsection{Object Separation Results with Variation of $\kappa$ Values}

In some extracted images, a very small part of the background may appear with the separated object or some part of the desired object may not be extracted. This deficiency can be removed by varying the values of $\kappa_1$ and $\kappa_2$. The increase and decrease of $\kappa_1$ and $\kappa_2$ values will result in finer and broader segment size, respectively. It will help in separating out the background and the foreground pixels, which were overlapping in the initial case. For the results in $Fig.\,9$, $\kappa_1 = \kappa_2 = \kappa$. $Fig.\,9$ shows the original and extracted images at $\kappa$ values of 0.25, 0.5, 1 and 2. It can be seen that pixel values of the background and the object are segregating or merging with a change in $\kappa$ value. Amongst all the results displayed in $Fig.\,9$, objects in the images 'Aeroplane', 'Coin' and 'Coins' are most efficiently extracted at $\kappa=1$, 'Eagle' at $\kappa=0.5$, and 'House' at $\kappa=2$. In the image House, it is seen that some part of the wall is of sky blue color, matching the color of the background. In this case, the background cannot be completely eliminated, but overlap can be minimized by varying the value of $\kappa$ between 1 to 2. The results can be further refined to get the desired object by taking different values of $\kappa_1$ and $\kappa_2$, depending on the skewness of the histogram. It is worth mentioning that skewness parameter $\kappa$ can be successfully tuned by the user, providing freedom to isolate the desired object.

\begin{figure*}[t]
\begin{center}
\includegraphics[width=2cm]{aeroplane.pdf}
\includegraphics[width=2cm]{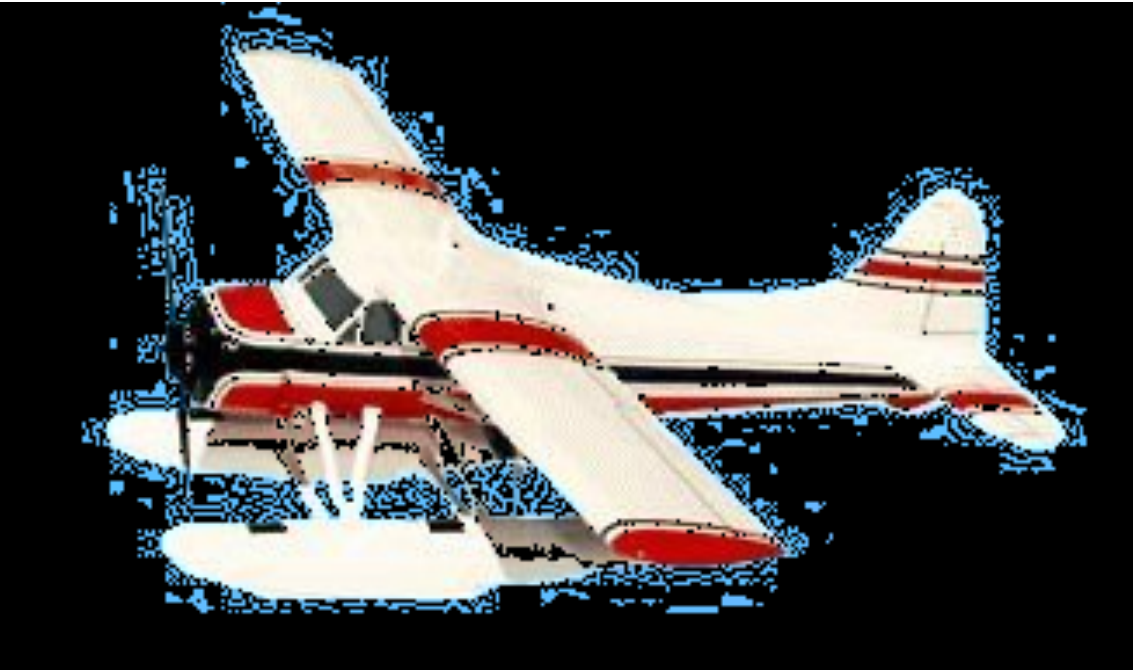}
\includegraphics[width=2cm]{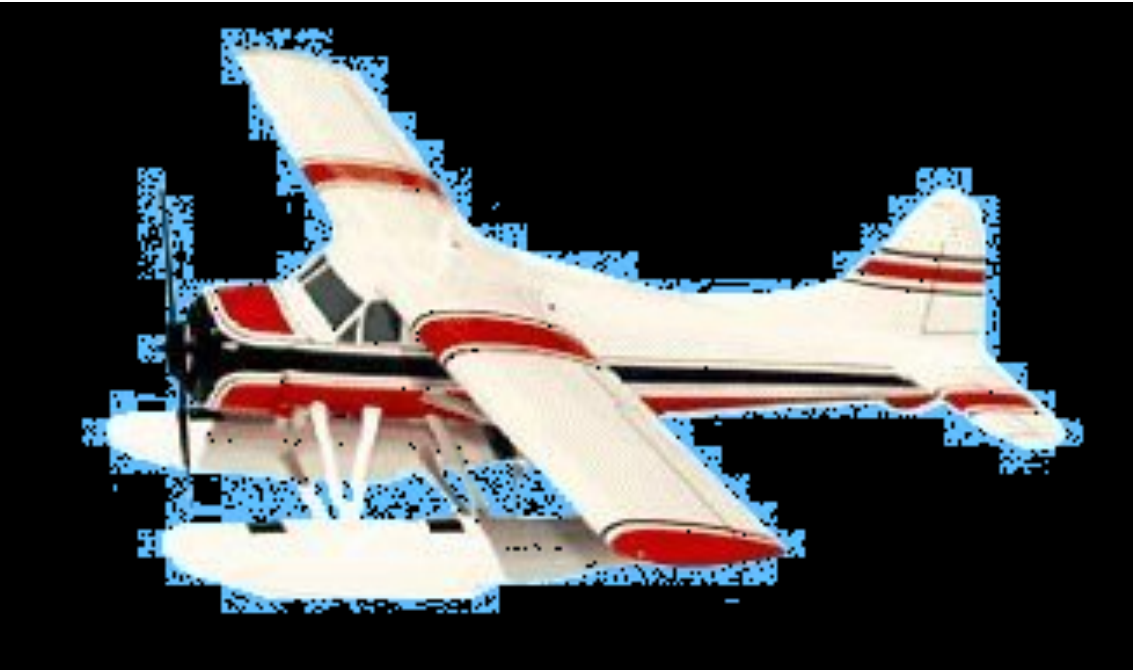}
\includegraphics[width=2cm]{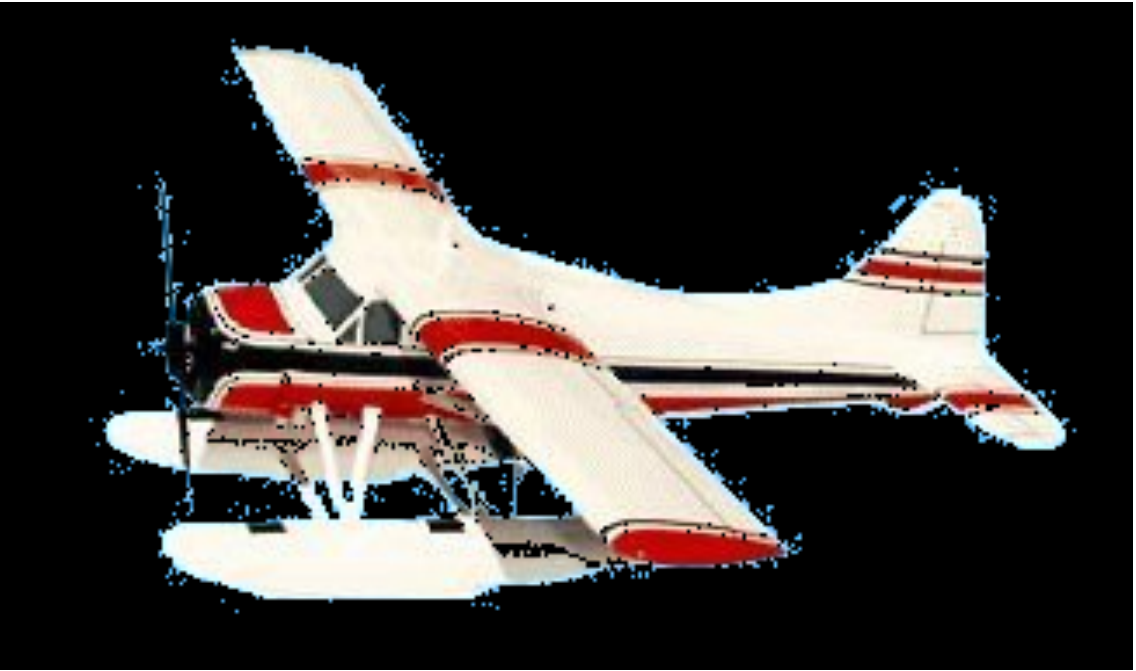}
\includegraphics[width=2cm]{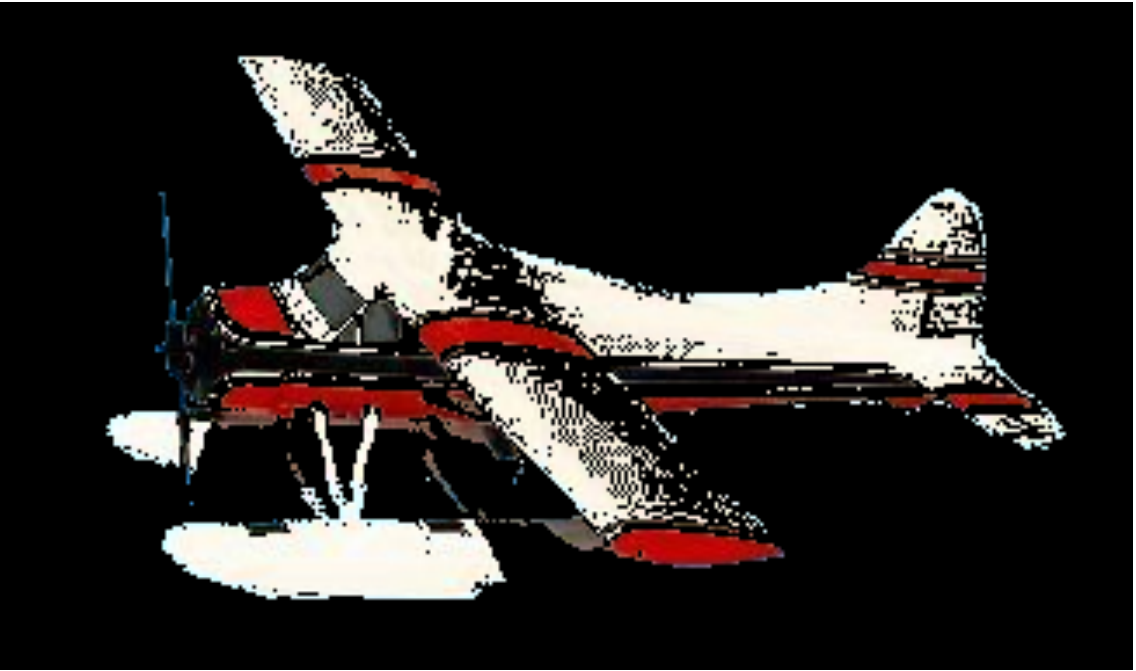}\\

\includegraphics[width=2cm]{Eagle.pdf}
\includegraphics[width=2cm]{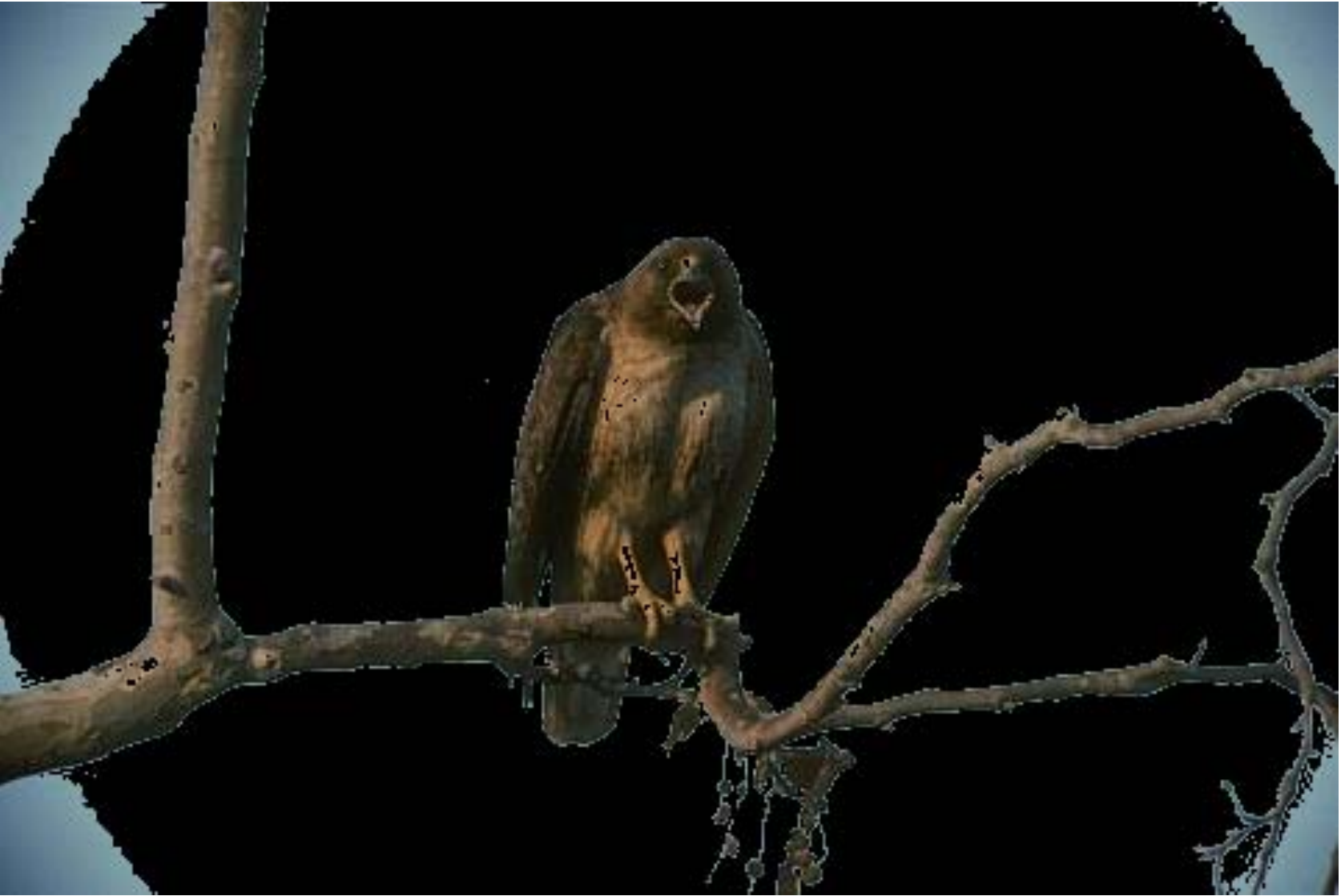}
\includegraphics[width=2cm]{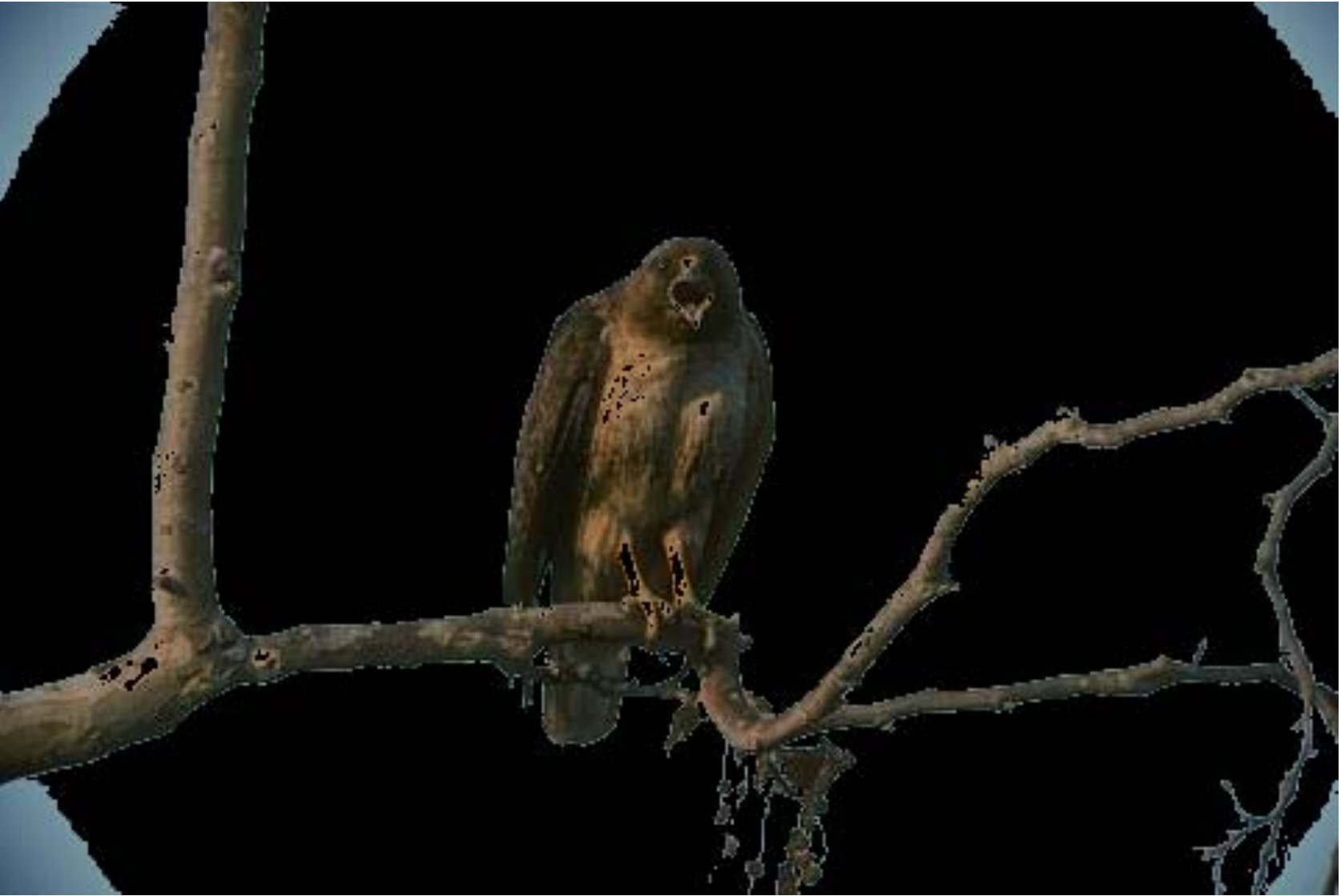}
\includegraphics[width=2cm]{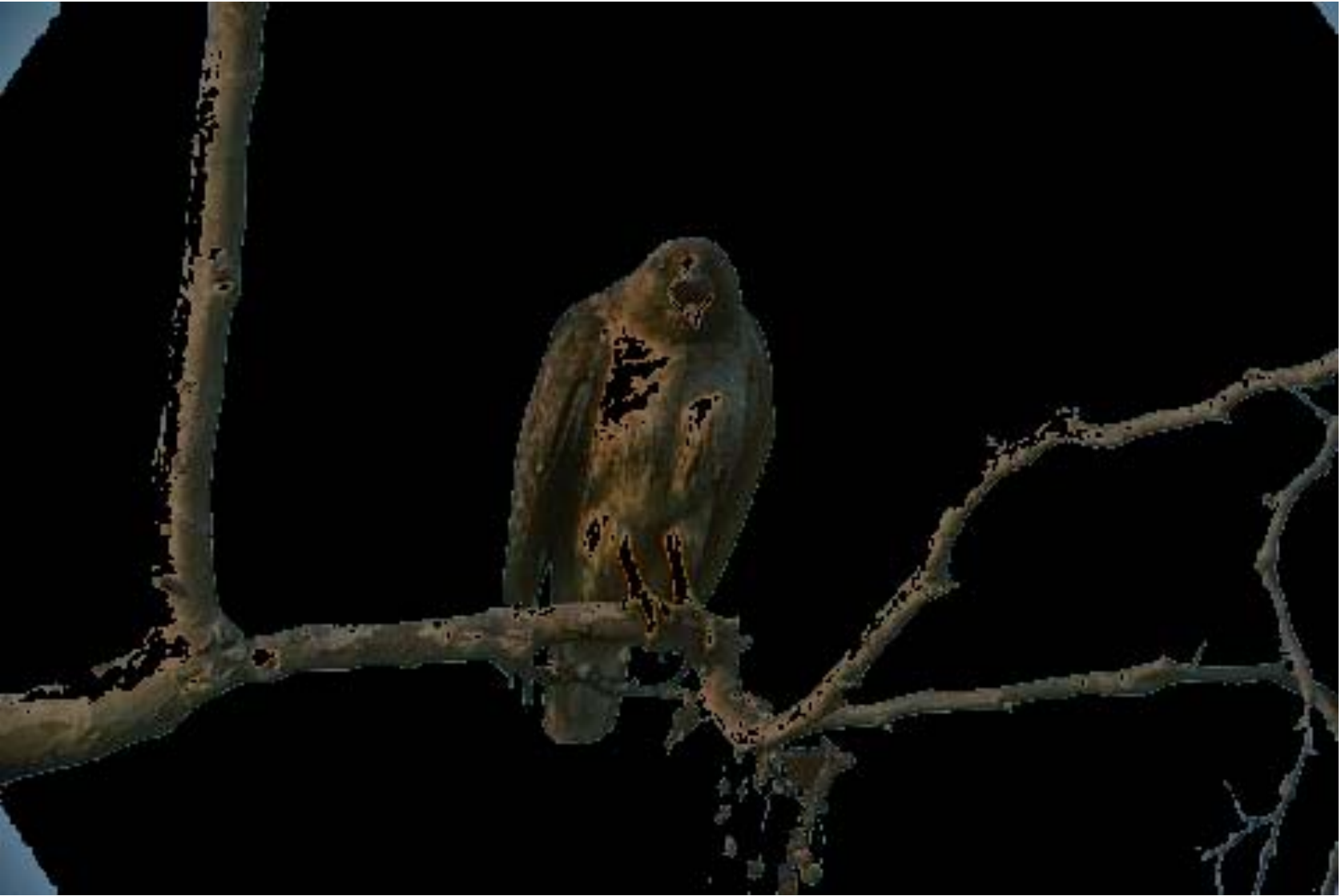}
\includegraphics[width=2cm]{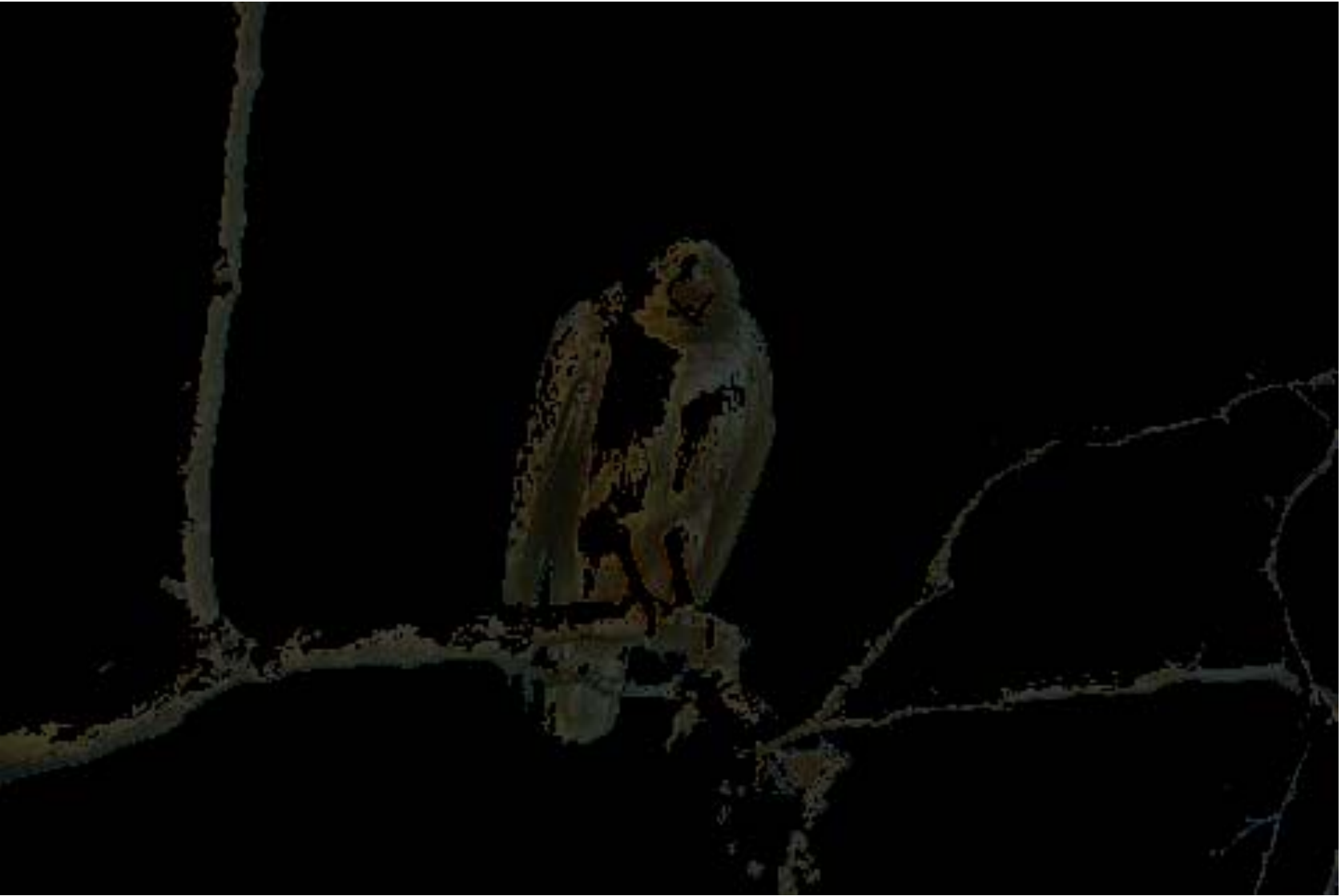}\\

\includegraphics[width=2cm]{Houses.pdf}
\includegraphics[width=2cm]{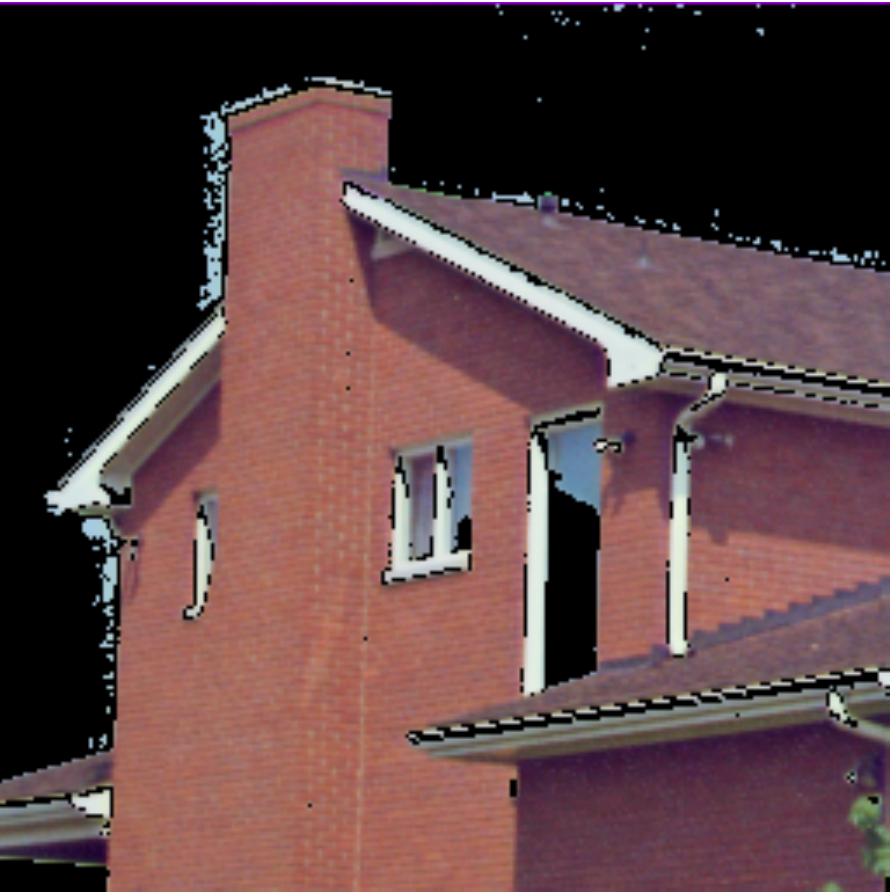}
\includegraphics[width=2cm]{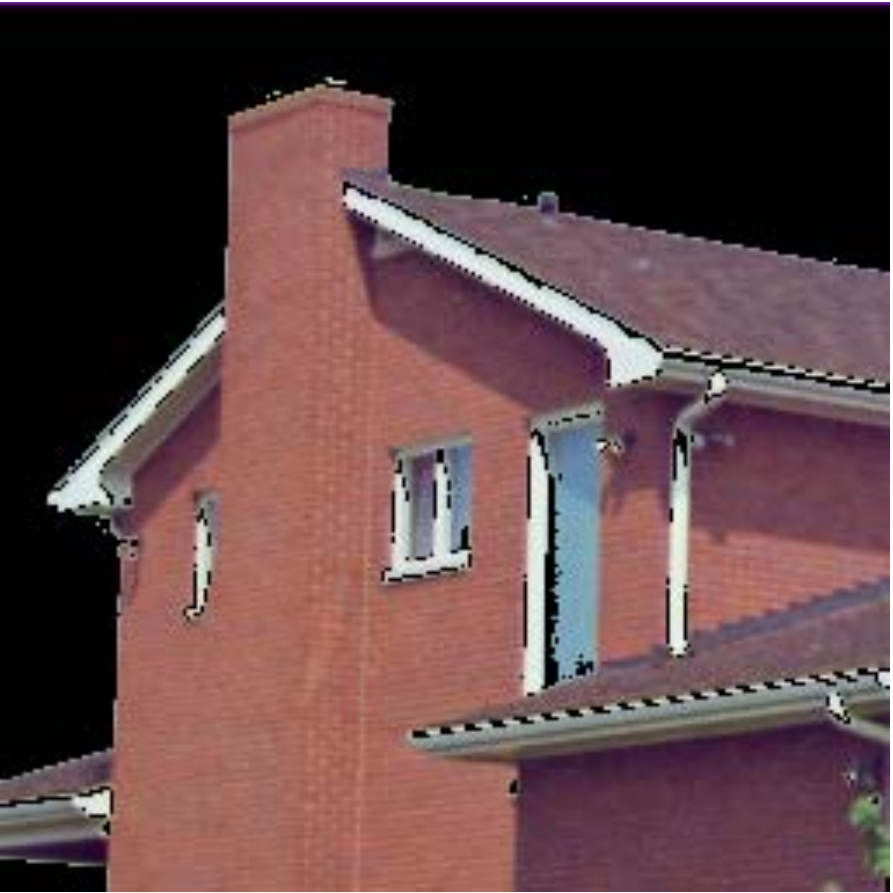}
\includegraphics[width=2cm]{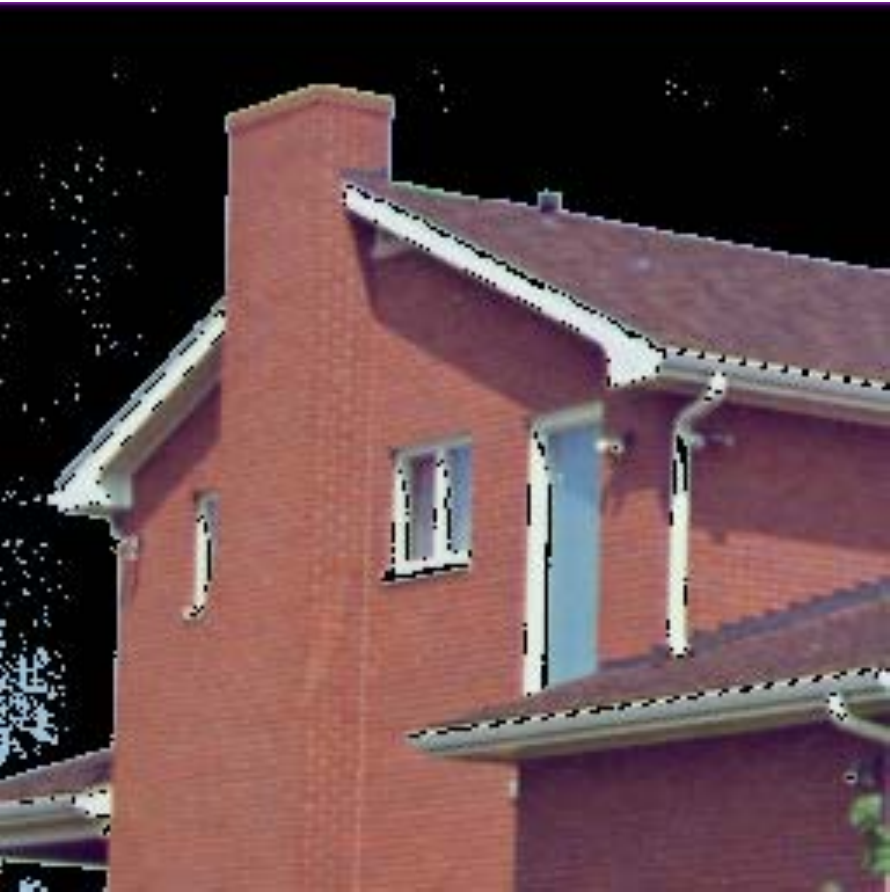}
\includegraphics[width=2cm]{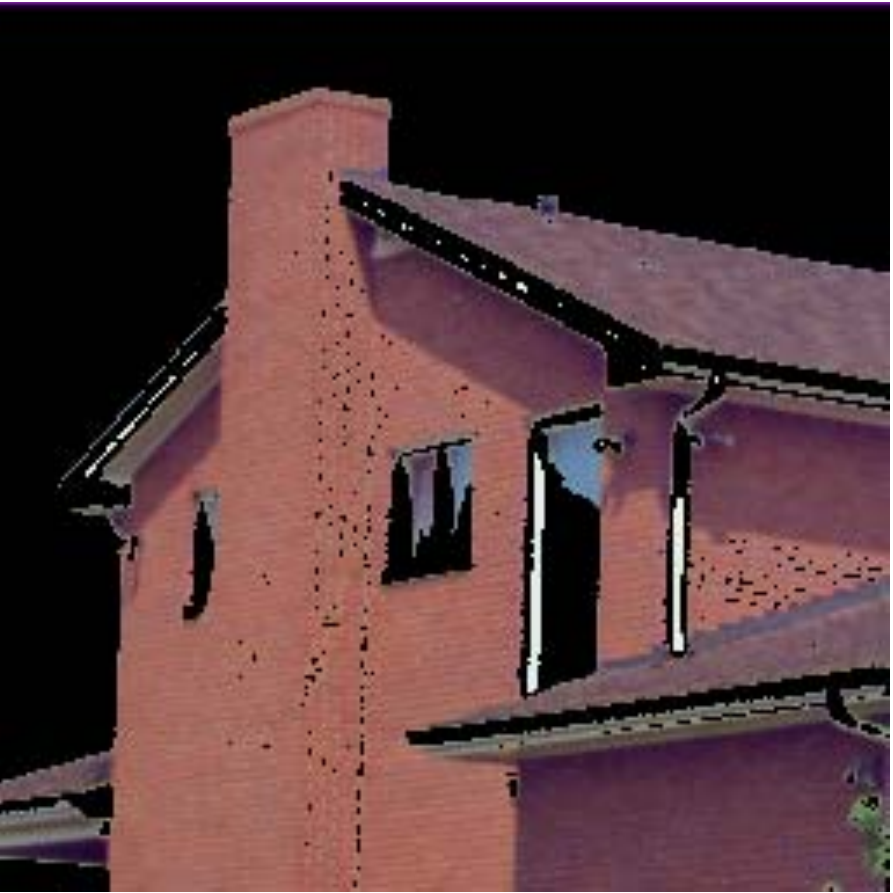}\\

\includegraphics[width=2cm]{Coin.pdf}
\includegraphics[width=2cm]{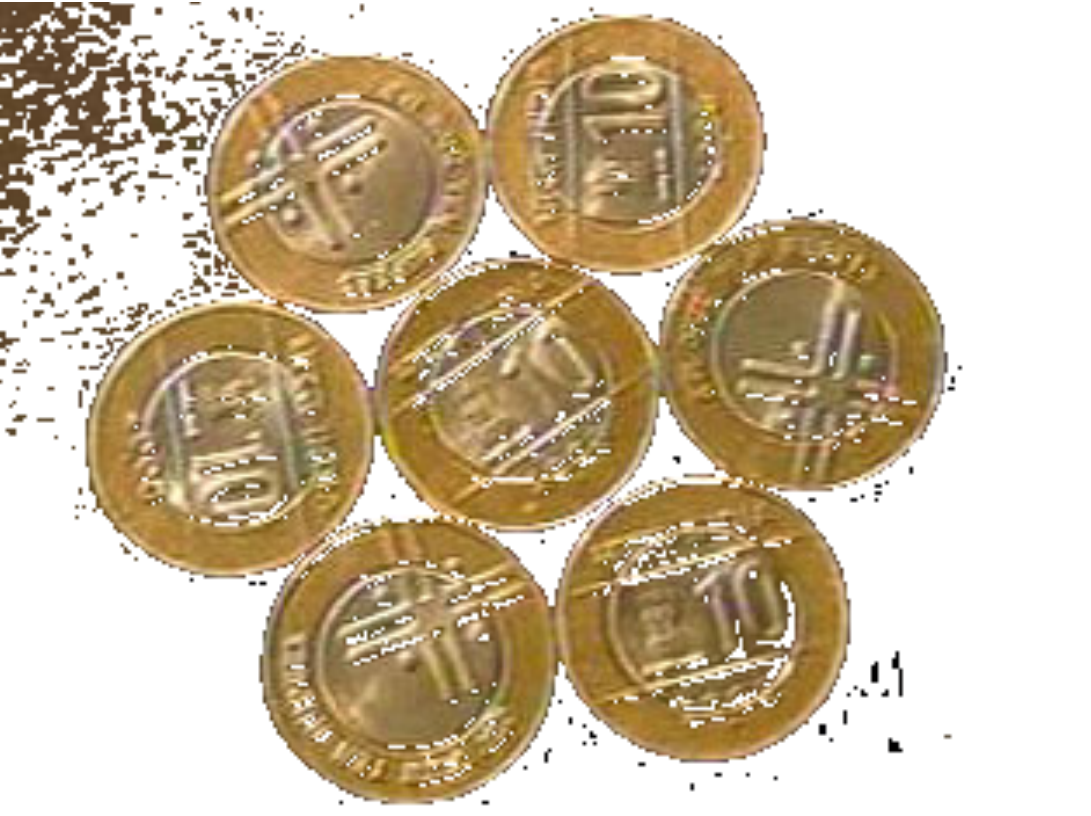}
\includegraphics[width=2cm]{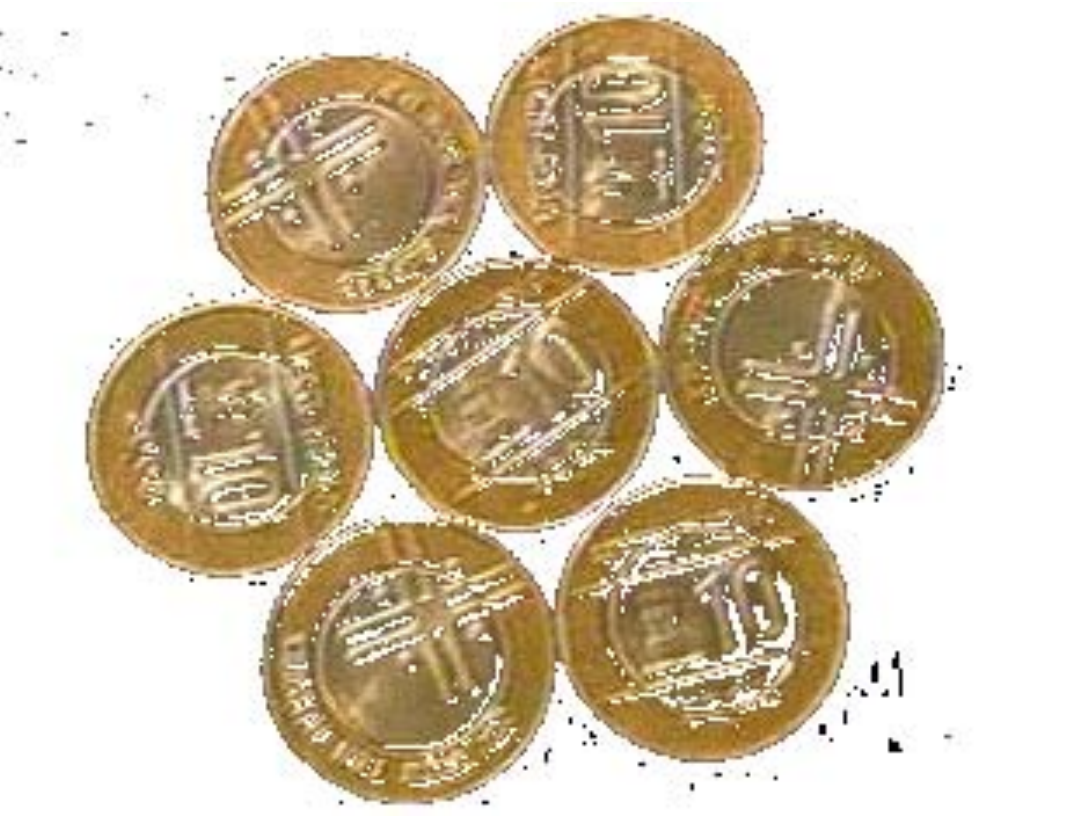}
\includegraphics[width=2cm]{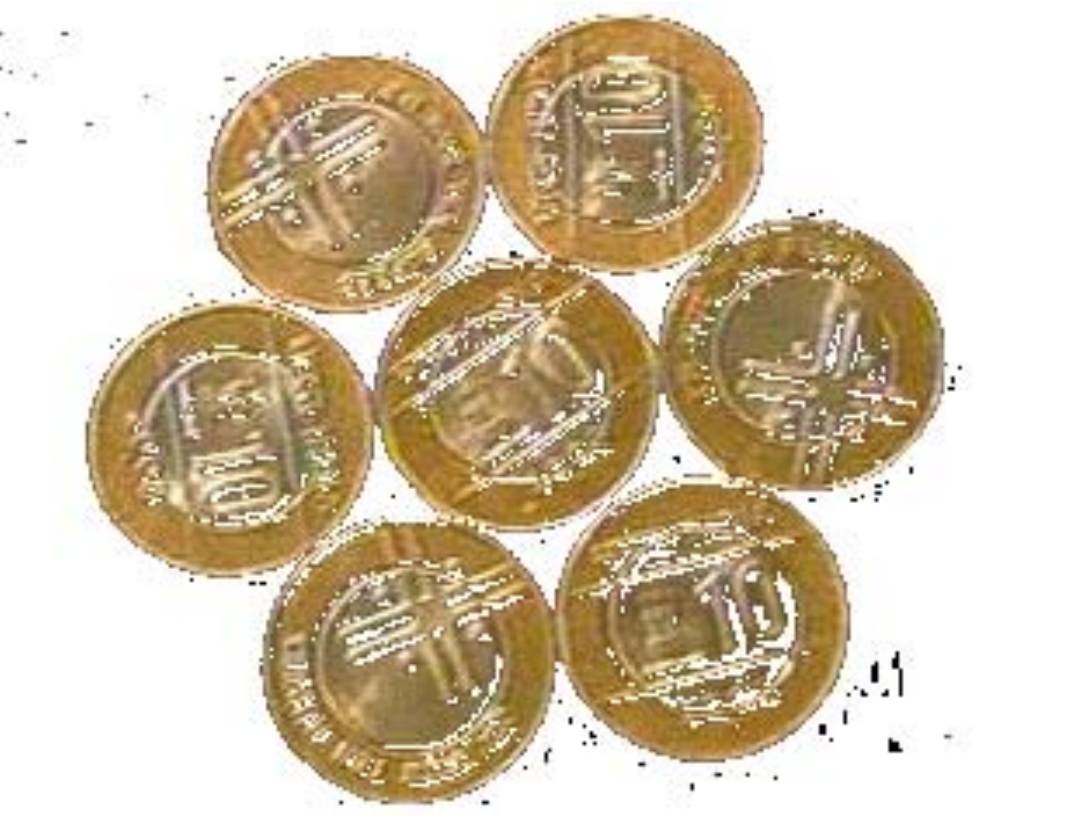}
\includegraphics[width=2cm]{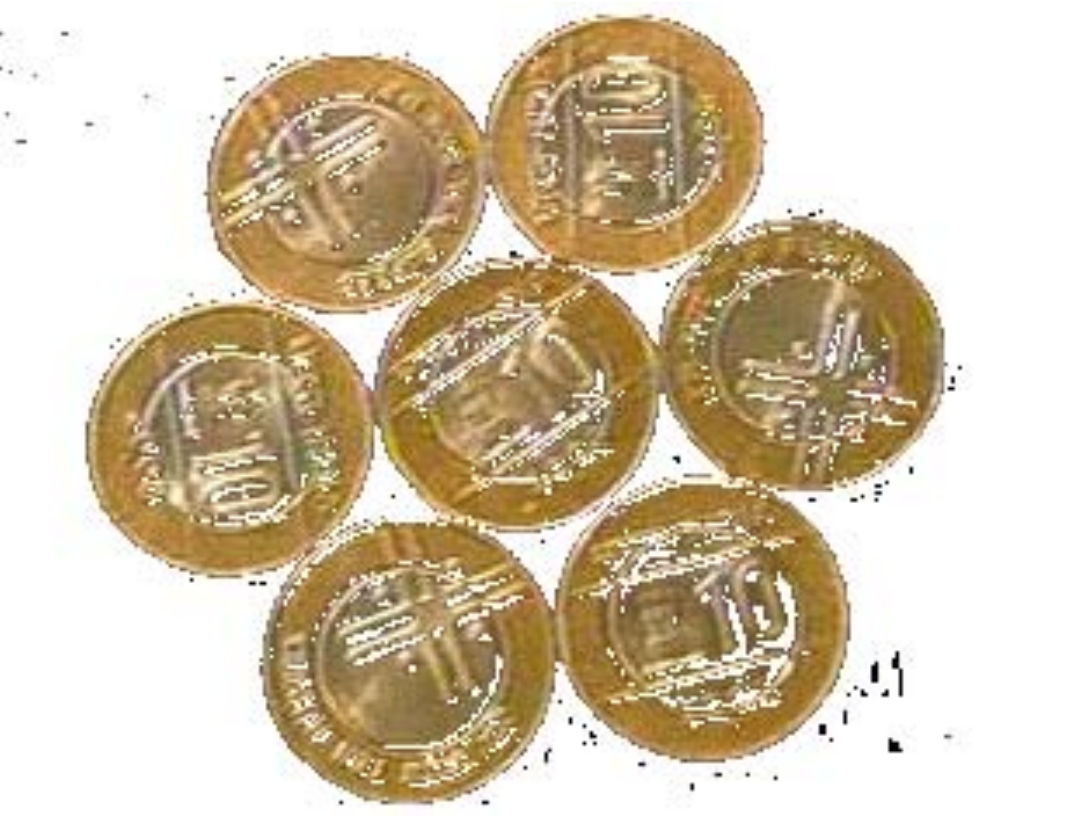}\\

\includegraphics[width=2cm]{Coins.pdf}
\includegraphics[width=2cm]{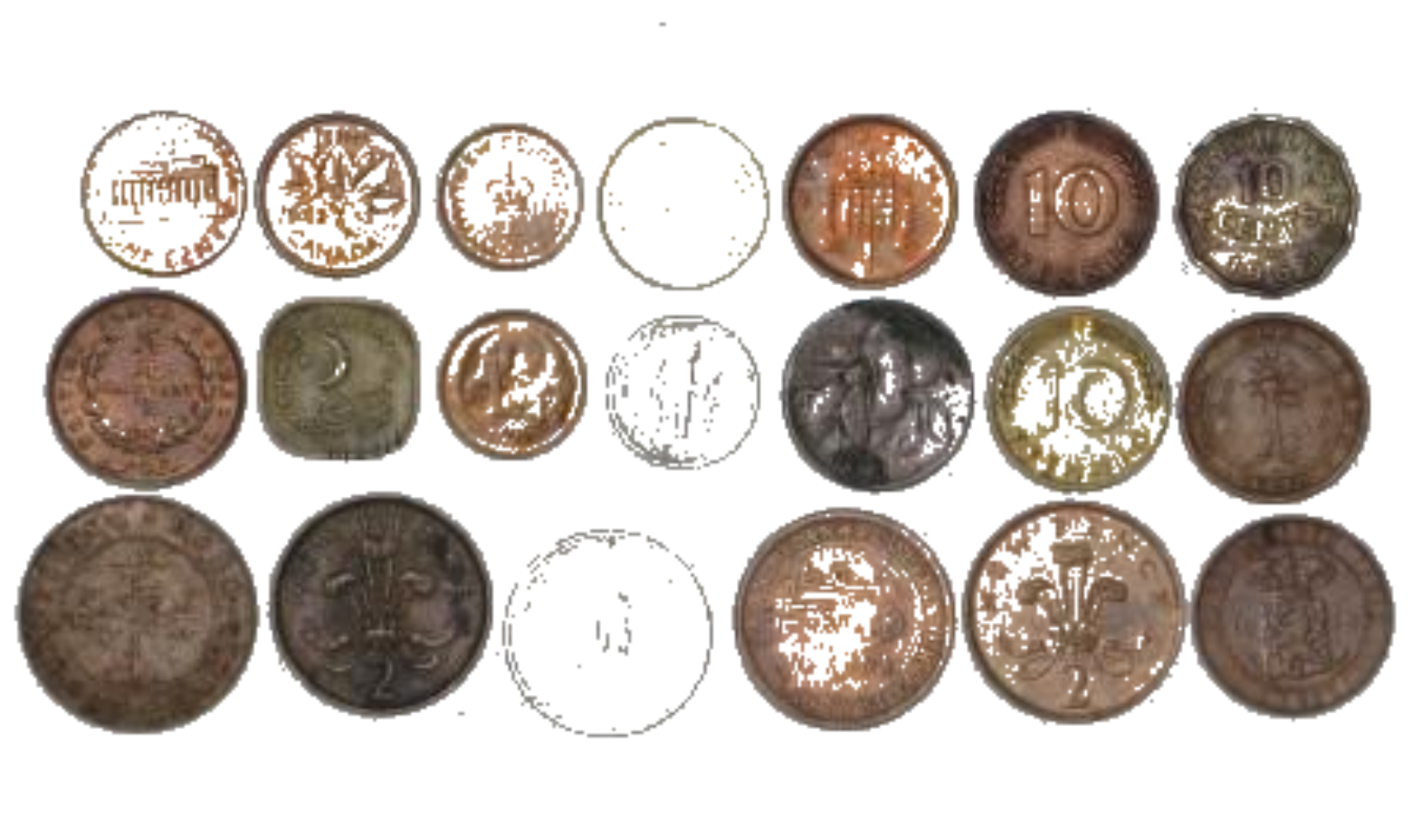}
\includegraphics[width=2cm]{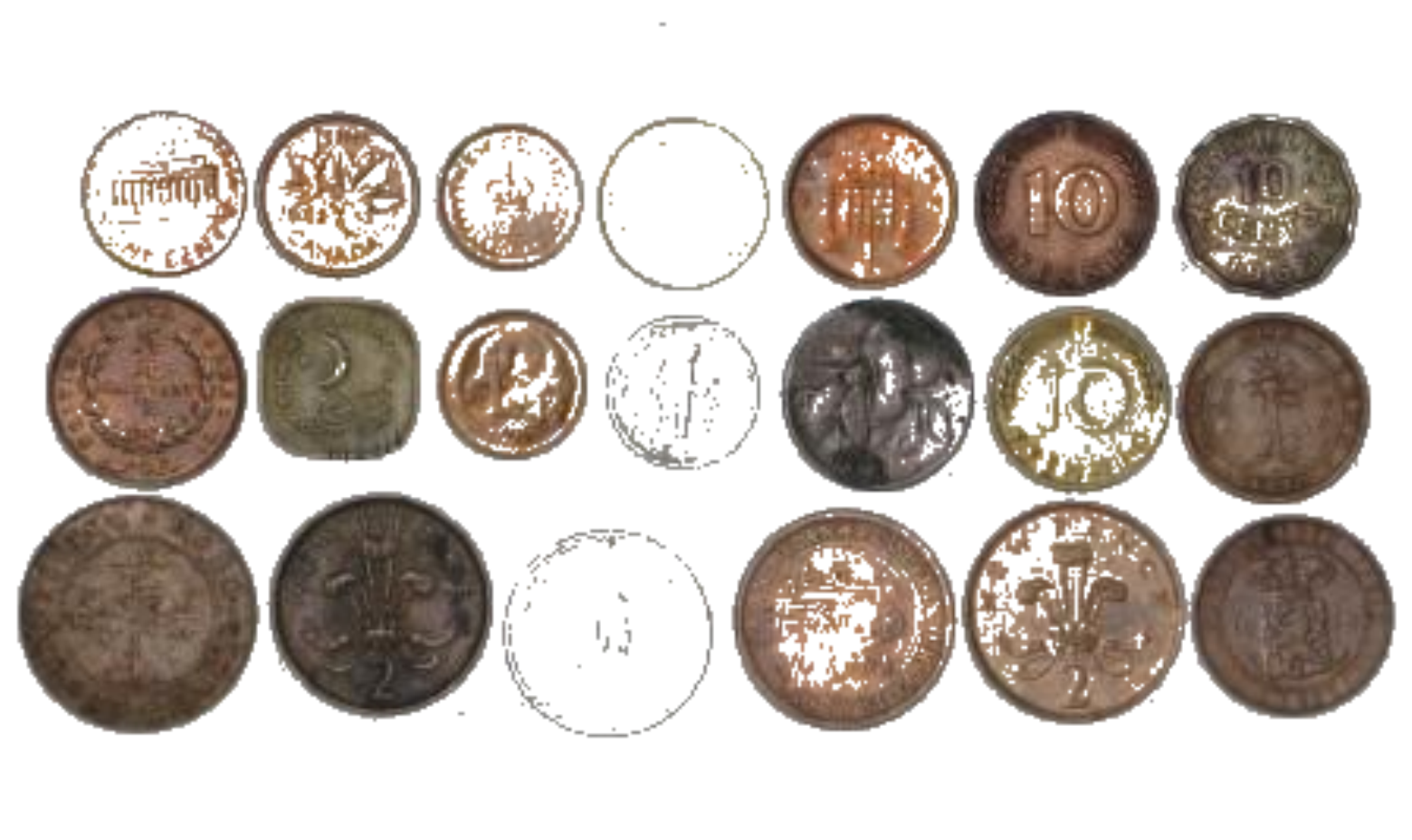}
\includegraphics[width=2cm]{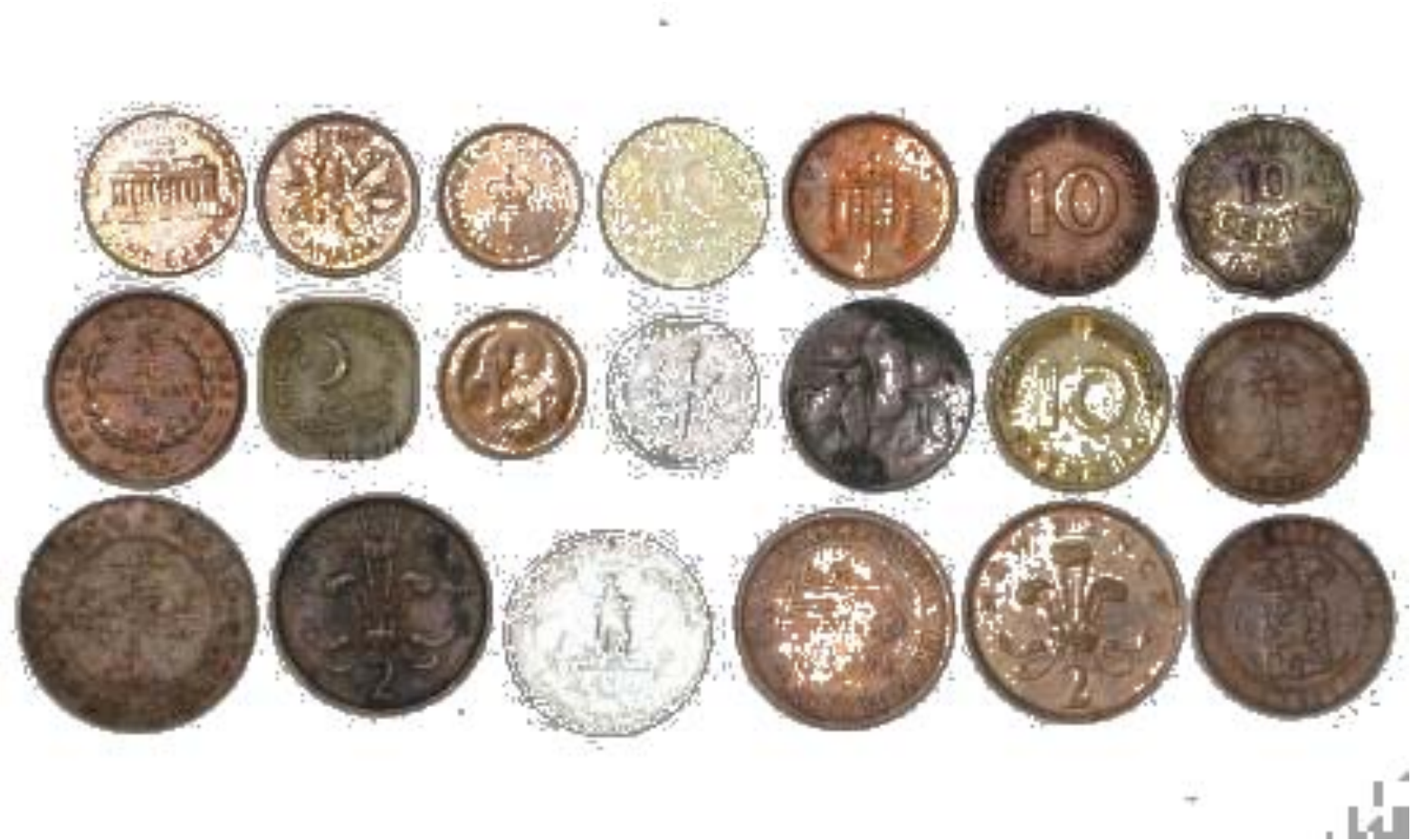}
\includegraphics[width=2cm]{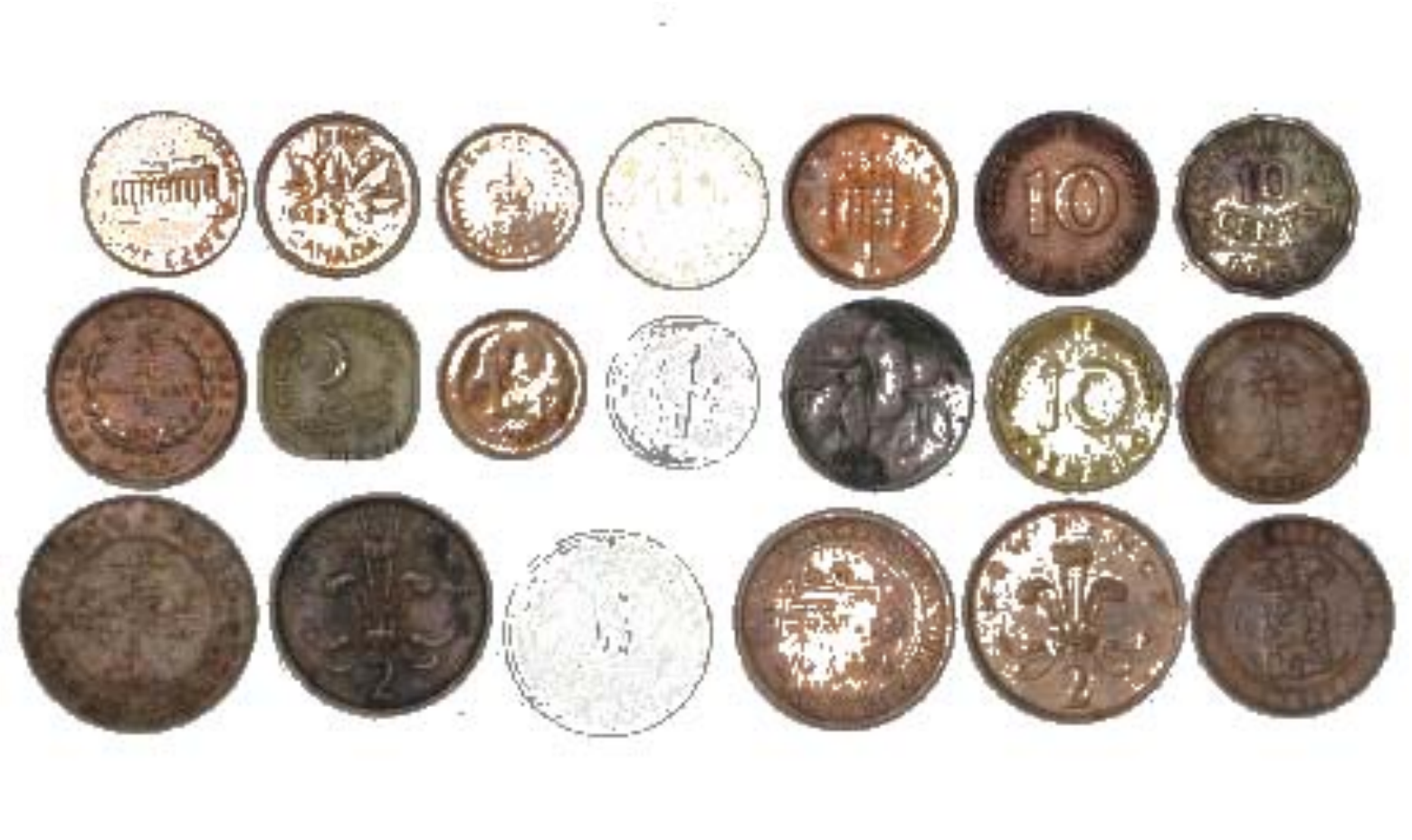}\\

\caption{Simulated output of different images. Row 1 - Aeroplane, Row 2 - Eagle, Row 3 - House, Row 4 - Coin, Row 5 - Coins at different $\kappa$ values. Images Aeroplane, Coin and Coins show best results at $\kappa=1$, Eagle and Infrared at $\kappa=0.5$, and House at $\kappa=2$}
\label{fig:3}
\end{center}
\end{figure*}

\subsection{Comments}

\begin{enumerate}

\item The number of thresholds is always odd which also implies that the number of segmented regions is always even i.e. one more than the number of thresholds. However, there might be a situation where, for desired segmentation, even number of threshold values are required. Although the proposed algorithms are incapable of achieving even number of thresholds, there are two ways in which proposed methodologies overcome the above problem. First, the variation of either $\kappa_1$ or $\kappa_2$, or both, can result in shifting the threshold values, which would bring the desired object values into single or distinct segments, in place of overlapping ones. Second, the number of thresholds can be increased. As the aim of this paper is to separate the specific object from the rest of the image, over-thresholding is not a concern. This is because an object represented by one segment, or more than one, would yield the same result. The only change would be in the number of segmented values extracted to isolate the object. The focus is on extracting the desired object. It may or may not be achieved by optimally segmenting the image.

\item No method to obtain optimized parametric values is suggested in the paper because the statistical characteristics of an object to be extracted vary drastically among the distinct set of images. However, for a particular set of images, the fixed or optimized value of the parameters can be found manually by the user for one image, and then the same values can be utilized for the complete set in an automated manner. Manual intervention to define the best parametric values results in the algorithm being semi-automated. But, the algorithm can be completely automated for images belonging to certain areas, if the parameter values are manually initialized.

\item $\kappa$ is an important parameter which can be used to refine segmentation and adjust thresholds. Its comprehensive role needs further exploration that lies beyond the scope of the paper. The paper provides an enabling framework for object separation requiring an active role of $\kappa$, but it would be difficult to claim that there is a generalized formula to find optimal value $\kappa$ for all classes of images. Hence, a separate study is required to investigate it. However, it is believed that for a specific class of images there can be a method to find an optimal $\kappa$ and therefore it is suggested to different field users to calculate their own optimal $\kappa$ formula. It would be difficult for the authors of this paper to survey the characteristics of every class of images, and find the formula for an optimal $\kappa$. Another argument against finding optimal $\kappa$ is the lack of knowledge whether a single $\kappa$ value is enough for an image, or there can also be different $\kappa$ values at each threshold levels.

\item Although, the proposed methodologies are able to separate out the desired object from the image, there are still some small artifacts. These can be removed using three methods: (1) by segmenting the color spaces, (2) varying the $\kappa$ values, and (3) using some preprocessing or postprocessing operations. None of the previous mentioned methods are used in the results to remove the artifacts because choosing any one would intrinsically mean that others are inferior methods. Also, as stated earlier, choosing the appropriate $\kappa$ would automatically solve this problem. 

\end{enumerate}

\section{Conclusion}
\label{sec:5}

In conclusion, a semi-automated statistical algorithm is proposed, which simultaneously achieves content based image segmentation and controlled object segregation. The fact that visually distinguishable objects in images are characterized by well separated pixel distributions, suggests use of thresholding techniques, to isolate these distributions from each other and hence, separate out the corresponding objects. We implement this approach making use of global image characteristics like mean, variance and skewness to isolate an object, taking into account the fact that human eye is less sensitive to variations in black and white regions as compared to the gray scale. Relying on the fact that normal distributions tend towards sharply peaked $\delta-$function distributions in the limit of zero variance, the present approach progressively isolates objects in the lower and higher pixel values, zooming onto the objects in the visually sensitive gray pixel values or vice$-$versa. It is evident that the present approach can apply equally well in other representations of the image, wherein object characteristics can be captured by localized distributions. It is worth investigating the utility of the present approach in wavelet domain, wherein different objects can be well localized as a function of scale. The use of a control parameter $\kappa$, for achieving optimal clustering for object separation, helps remove ambiguity of closely overlapping distributions representing degenerate cases.
\begin{acknowledgements}
The authors would like to thank anonymous reviewers for providing useful comments and suggestions which has enhance the quality of this paper.

The authors also want to express gratitude towards Aadhar Jain (Sibley School of Mechanical and Aerospace Engineering, Cornell University) and Parinita Nene (School of Electrical and Computer Engineering, Cornell University) for sparing their valuable time to read the paper, and correct the grammatical errors. Additionally, their queries while reading the paper have been useful in improving the quality of paper and the contents covered.

The Berkeley Segmentation Dataset and SIPI Image Database images were used to test the proposed methodologies.
\end{acknowledgements}

\end{document}